\documentclass[conference,compsoc]{IEEEtran}

\usepackage[utf8]{inputenc} 
\usepackage[T1]{fontenc}    
\usepackage{hyperref}       
\usepackage{url}            
\usepackage{booktabs}       
\usepackage{amsfonts}       
\usepackage{nicefrac}       
\usepackage{microtype}      
\usepackage{xcolor}         
\usepackage{graphicx}
\usepackage{subcaption}

\usepackage{amsmath}
\usepackage[capitalize]{cleveref}
\crefname{section}{Appendix}{Appendices}
\usepackage{enumitem}

\usepackage{adjustbox}
\usepackage{threeparttable}
\usepackage{algorithm, algcompatible}
\algnewcommand\INPUT{\item[\textbf{Input:}]}%
\algnewcommand\OUTPUT{\item[\textbf{Output:}]}%

\makeatletter
\renewcommand\paragraph{\@startsection{paragraph}{4}{\z@}%
  {1.5ex \@plus 0.5ex \@minus .2ex}%
  {-1em}%
  {\normalfont\normalsize\bfseries}}
\makeatother

\title{Multi-Task Consistency-based Detection of Adversarial Attacks}

\author{\IEEEauthorblockN{Cong Chen}
\IEEEauthorblockA{Qualcomm Technologies, Inc.\\
San Diego, CA 92121, USA\\
congchen@qti.qualcomm.com}
\and
\IEEEauthorblockN{Jean-Philippe Monteuuis}
\IEEEauthorblockA{Qualcomm Technologies, Inc.\\
San Diego, CA 92121, USA\\
jmonteuu@qti.qualcomm.com}
\and
\IEEEauthorblockN{Jonathan Petit}
\IEEEauthorblockA{Qualcomm Technologies, Inc.\\
San Diego, CA 92121, USA\\
petit@qti.qualcomm.com}
}

\begin{document}
\maketitle
\pagestyle{plain}
\begin{abstract}
Deep Neural Networks (DNNs) have found successful deployment in numerous vision perception systems. However, their susceptibility to adversarial attacks has prompted concerns regarding their practical applications, specifically in the context of autonomous driving. Existing defenses often suffer from cost inefficiency, rendering their deployment impractical for resource-constrained applications. In this work, we propose an efficient and effective adversarial attack detection scheme leveraging the multi-task perception within a complex vision system. Adversarial perturbations are detected by the inconsistencies between the inference outputs of multiple vision tasks, e.g., object detection and instance segmentation. To this end, we developed a consistency score metric to measure the inconsistency between vision tasks. Next, we designed an approach to select the best model pairs for detecting inconsistencies effectively. Finally, we evaluated our defense against PGD attacks across multiple vision models on the BDD100k validation dataset. The experimental results demonstrated that our defense achieved a ROC-AUC performance of 99.9\% detection within the considered attacker model.
\end{abstract}
\section{Introduction}
The camera-based perception system is critical to enable automated driving (AD). Indeed, camera is the only sensor able to read traffic signs, identify lane markings or drivable areas, and see traffic light colors. To perform such perception tasks (e.g., object detection, classification, segmentation), a wide range of machine learning models were developed, each with its own objective and network architecture~\cite{zou2023object}. For example, from an input image, 2D object detection models output bounding boxes, while semantic segmentation models output masks, or multi-object tracking models output track identifiers. The model outputs help to understand the scene and allow the automated vehicle to maneuver appropriately.

However, camera inputs can be maliciously manipulated to affect the performance of perception tasks, or even downstream tasks of automated vehicles (e.g., path planning, motion control). The idea of adversarial inputs (commonly called \textit{adversarial examples}) is to add specially-crafted noise to images such that the underlying machine learning models do not perform as originally intended~\cite{madry2018towards}. Adversarial examples have been demonstrated in the form of full image perturbations or patches, realized digitally or physically, and with some high attack success rate and universality~\cite{chow2020adversarial}. Because of their low level of sophistication and effectiveness, it is key to deploy defenses to protect automated vehicles against such threats. Defenses range from preemptive techniques (e.g., adversarial training~\cite{shafahi2019adversarial}, certified robustness~\cite{xiang2023patchcure}) to reactive techniques (e.g., real-time detection of perturbations~\cite{xiang2022patchcleanser}, image compression~\cite{das2018compression}).

In this paper, we focus on reactive techniques, aiming at real-time detection of perturbations, because it does not require any adversarial data generation or additional training. Especially, we propose to leverage the output of multiple perception tasks to identify perturbations on every image prior to use by downstream tasks. Prior work showed the effectiveness of checking inconsistencies of edge extractions between outputs of semantic segmentation and depth estimation~\cite{klingner2022detecting}, but with some limitations. Their inconsistency check only detects adversarial perturbations on the entire image and might show limited performance on local perturbations. Therefore, we propose a consistency-based detection technique that is effective regardless of the perturbation's location. As long as the perturbation causes inconsistent inference output across models, locally or globally, our defense can capture the inconsistency. Especially, we demonstrate the benefits of cross-model consistency by using 2D object detection and instance segmentation models. Indeed, 2D object detection models are commonly used in AD to detect road objects, and then to convert 2D bounding boxes to 3D bounding boxes~\cite{feng2020deep,arnold2019survey}. Instance segmentation is also used in AD to provide finer object boundaries~\cite{zhou2020joint}. Both models share the objective of detecting objects, and hence, can be used to identify inconsistencies.

Our \textbf{contributions} are as follows:
\begin{itemize}[leftmargin=*,noitemsep,topsep=0pt]
\item We propose a lightweight consistency detector based on outputs from object detection and instance segmentation models. 
\item We develop a technique to select the optimal model pair, deriving requirements w.r.t model architecture.
\item We define a metric to capture the consistency score between two models' output.
\item We generate and publish an adversarial BDD100k dataset to assess the effectiveness of our defense, and allow reproducibility and comparison of future defenses.
\end{itemize}

\section{System Model}
\label{sec:system}
\subsection{Vision Multi-Task System}
Perception systems perform multiple vision tasks such as object detection, segmentation, and depth estimation. Because of its better generalization performance and efficiency~\cite{guo2020multi}, one architecture considered for automated driving is Multi-Task Learning (MTL)~\cite{miraliev2023real}. 
A common approach in MTL is to have a shared feature extractor and multiple task-specific heads~\cite{caruana1997multitask,kokkinos2017ubernet, lu2017fully}. 
In this paper, because our detection method must work with MTL and non-MTL architecture, 
our architecture consists of one model per task. With this flexible approach, we can evaluate the performance of our detector when the tasks share (or not) the same backbone. 
Indeed, the attack success rate strongly correlates with the architecture similarity between tasks as highlighted by~\cite{xie2017adversarial}. Interestingly, from a security perspective, it may be more robust to have an architecture with different backbone per task than a common backbone architecture for all tasks (like in the MTL architecture). 


\subsection{Attacker Model}
We follow the same attacker model as defined by~\cite{xiang2022patchcleanser}, where the attacker performs a white-box attack (i.e., has access to the model's architecture and weights). We assume a model $\mathbb{F}$ with an underlying data distribution $\mathcal{D}$ over pairs consisting of image $\mathbf{x}$ and its corresponding ground truth $y$. $\mathcal{X}$ 
denotes the image space. 
The attacker adds the perturbation $\delta$ to the genuine image $\mathbf{x}$ to create an adversarial image ($\mathbf{x'} = \mathbf{x} + \delta$) (with $||\delta||_p\leq\epsilon$, where $\epsilon$ is the bound on the $L_p$ norm perturbation) such as $ \mathbf{x'} \in \mathcal{A}(\mathbf{x}) \subset \mathcal{X}$, where constraint $\mathcal{A}$ defines the attacker’s capability. The goal of the attacker is to minimize the alteration of the genuine image $\mathbf{x}$ while ensuring the attack succeed, and is formulated as:
\begin{equation}
\min ||\mathbf{x'} - \mathbf{x}||\ s.t.\ \mathbb{F}(\mathbf{x'})\neq\mathbb{F}(\mathbf{x})
\end{equation}
where $\mathbb{F}(\mathbf{x'})\neq\mathbb{F}(\mathbf{x})$ can be the removal or injection of bounding boxes/masks.

To achieve her goal, the attacker uses a projected gradient descent (PGD) attack~\cite{madry2017towards}.
\begin{equation}
    \mathbf{x'}_{t+1} = \Pi_{\mathbf{x}+\mathcal{X}}(\mathbf{x'}_{t} + \alpha sgn(\nabla_{\mathbf{x}}L(\theta,\mathbf{x'},y)))
\end{equation}
where, $L(\theta,\mathbf{x'},y)$ is the global loss function defined as the sum of classification loss and localization loss ($L = L_{cls} + L_{loc}$). Hence, the perturbation targets a misclassification or mislocalization.

\section{Multi-Task Consistency}
\label{sec:consistency}
We first define the multi-task consistency score between model outputs across different vision tasks. In particular, we use object detection (OD) and instance segmentation (SEG) as example vision tasks in this paper. Then, we explain how to use the consistency score to detect adversarial perturbations. 

\begin{figure*}[bt!]
    \centering
    \begin{subfigure}{.4\linewidth}
        \centering
        \textbf{Object Detection}
        \par\medskip
        \includegraphics[width=\linewidth]{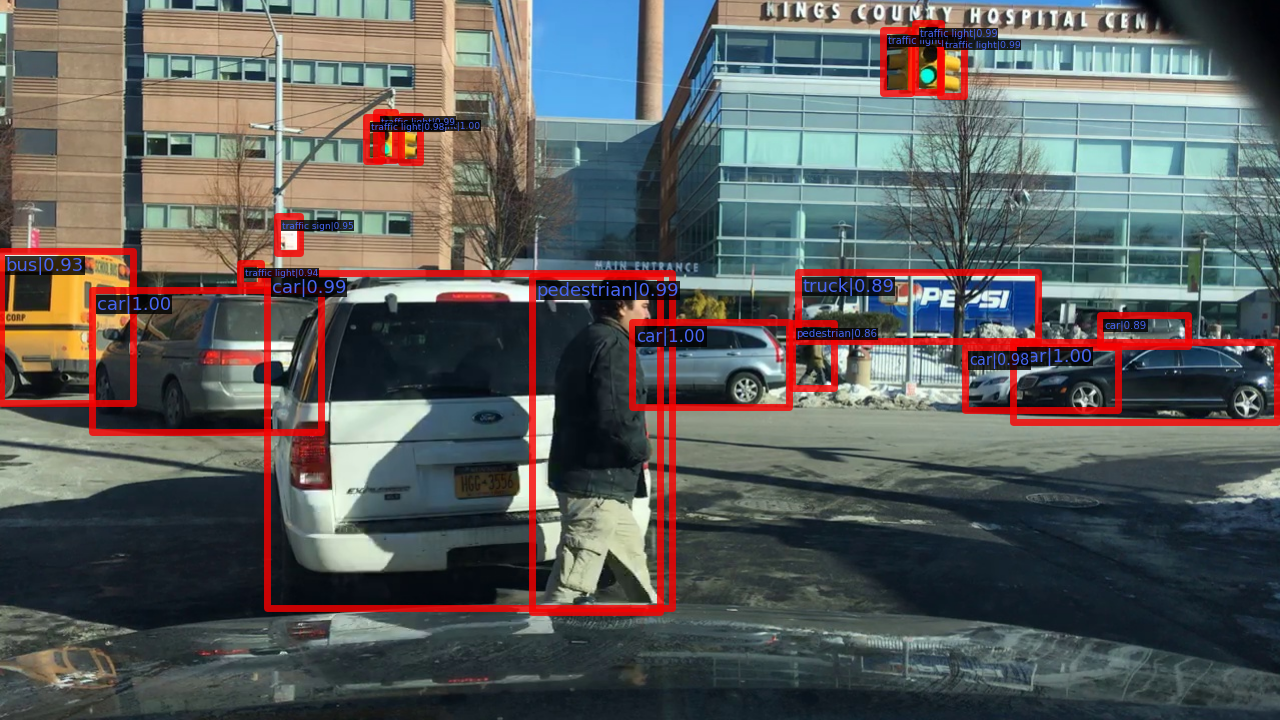}
        \caption{Clean image}
        \label{1-intro-clean_det}
    \end{subfigure}
    \hspace{2em}
    \begin{subfigure}{.4\linewidth}
        \centering
        \textbf{Instance Segmentation}
        \par\medskip
        \includegraphics[width=\linewidth]{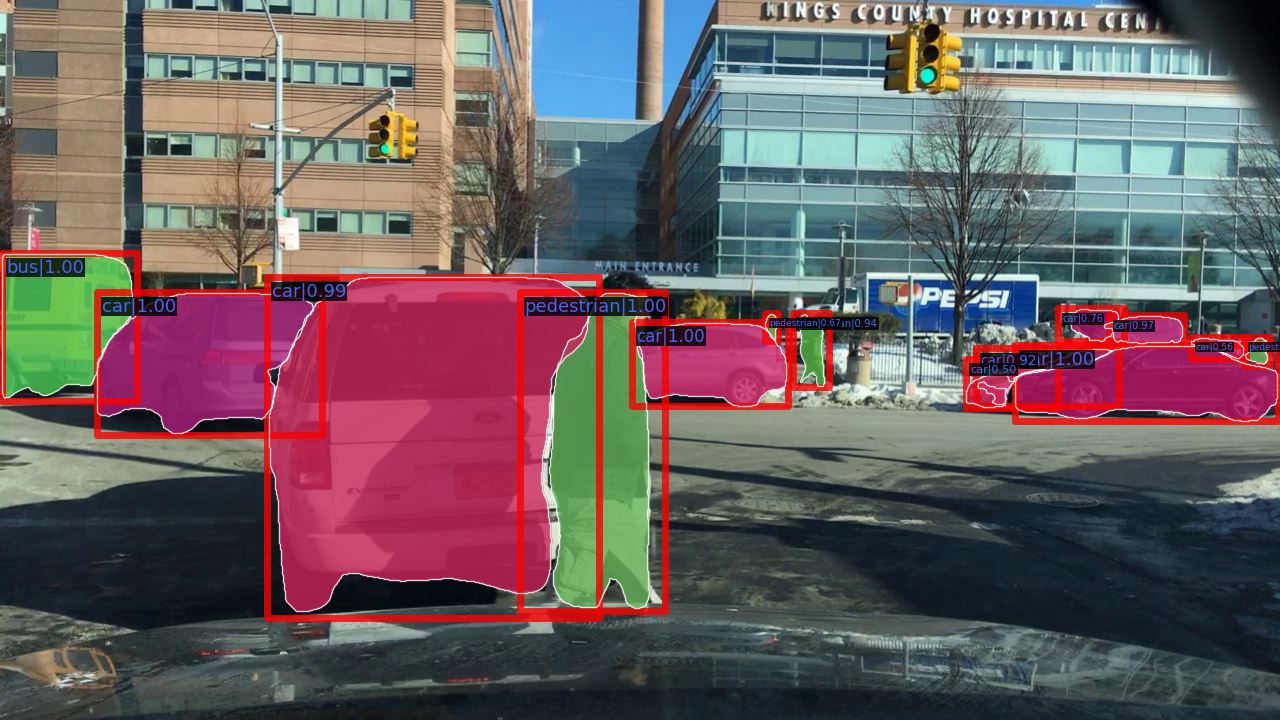}
        \caption{Clean image}
        \label{1-intro-clean_seg}
    \end{subfigure}

    \begin{subfigure}{.4\linewidth}
        \includegraphics[width=\linewidth]{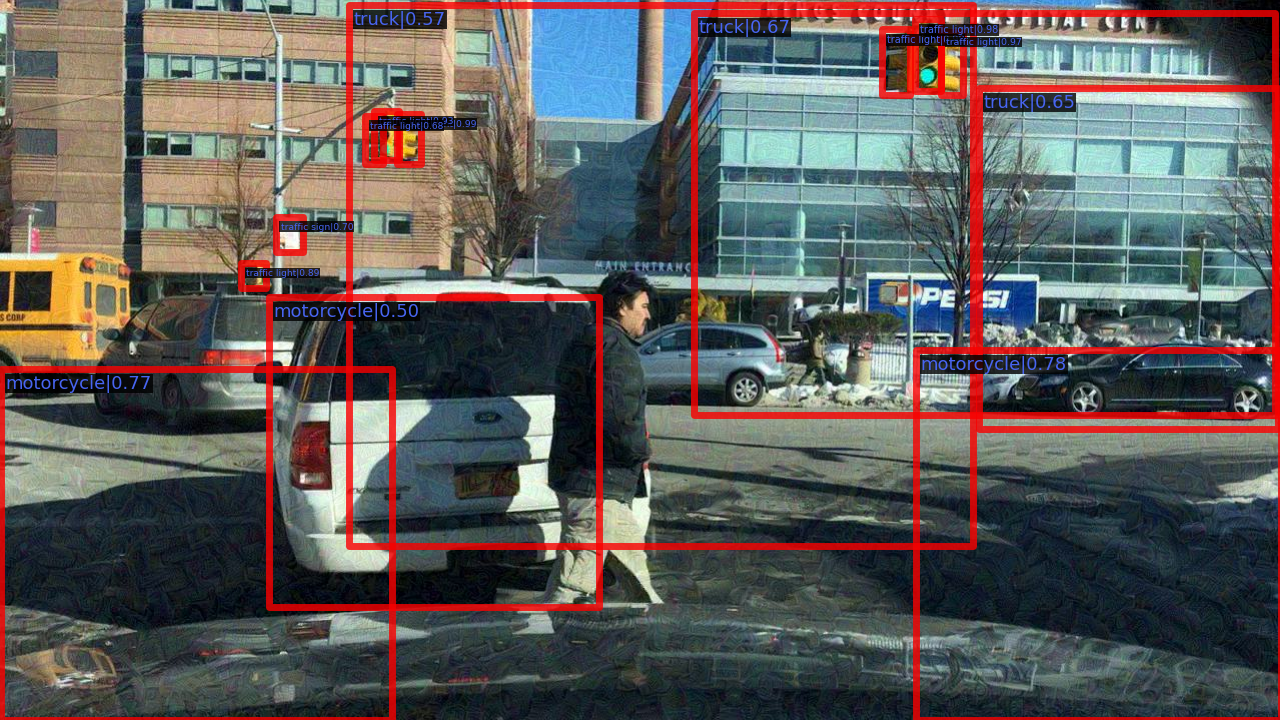}
        \caption{Perturbation optimized for object detection}
        \label{1-intro-ae_det}
    \end{subfigure}
    \hspace{2em}
    \begin{subfigure}{.4\linewidth}
        \includegraphics[width=\linewidth]{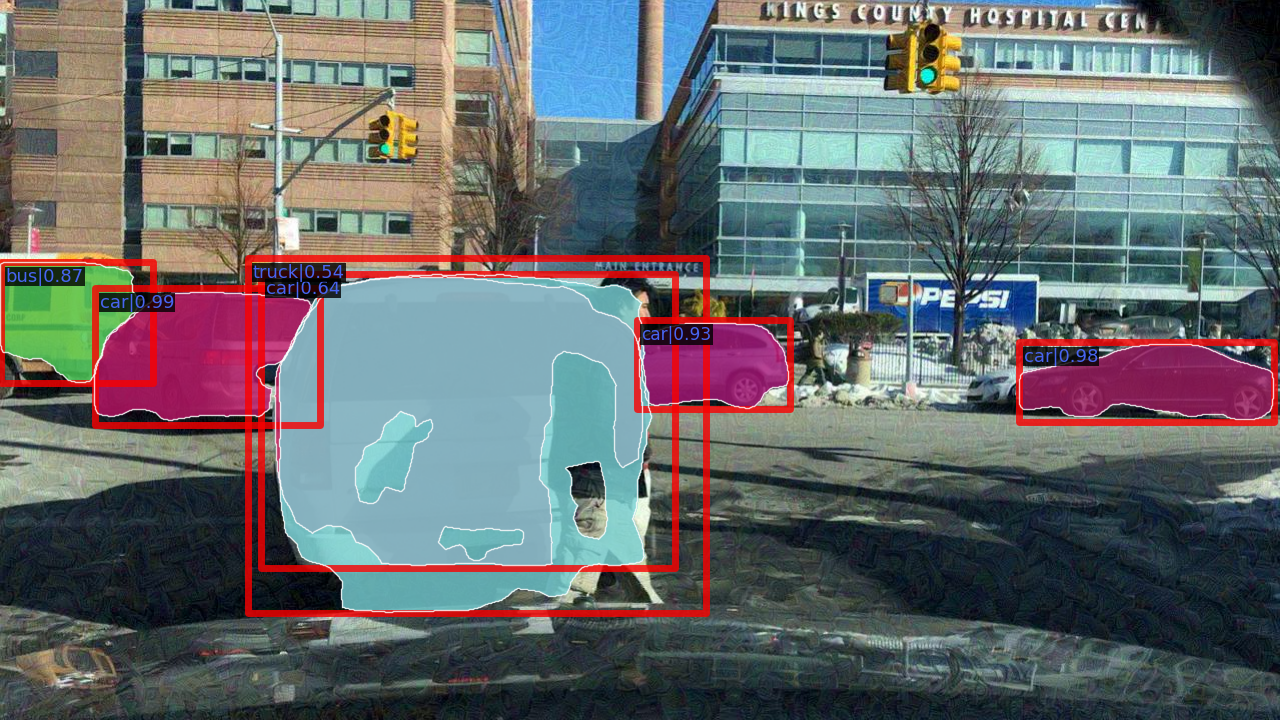}
        \caption{Perturbation optimized for object detection}
        \label{1-intro-ae_det_seg}
    \end{subfigure}

    \begin{subfigure}{.4\linewidth}
        \includegraphics[width=\linewidth]{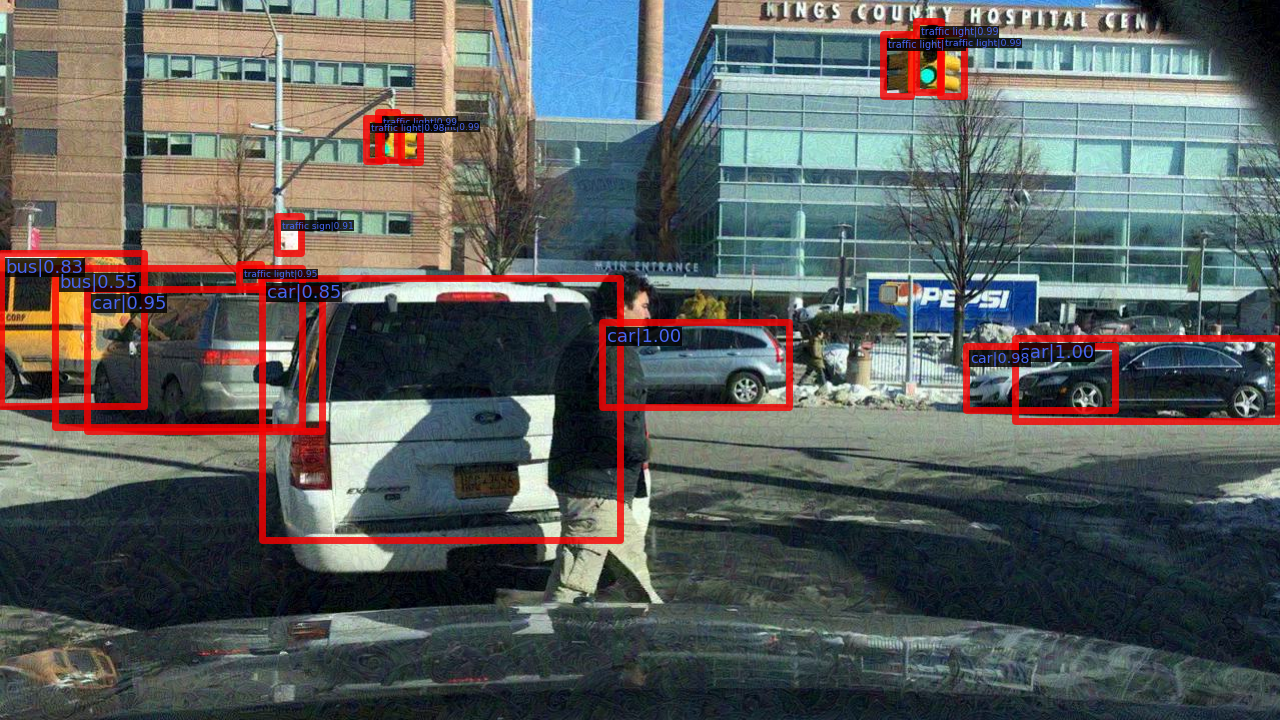}
        \caption{Perturbation optimized for instance segmentation}
        \label{1-intro-ae_seg_det}
    \end{subfigure}
    \hspace{2em}
    \begin{subfigure}{.4\linewidth}
        \includegraphics[width=\linewidth]{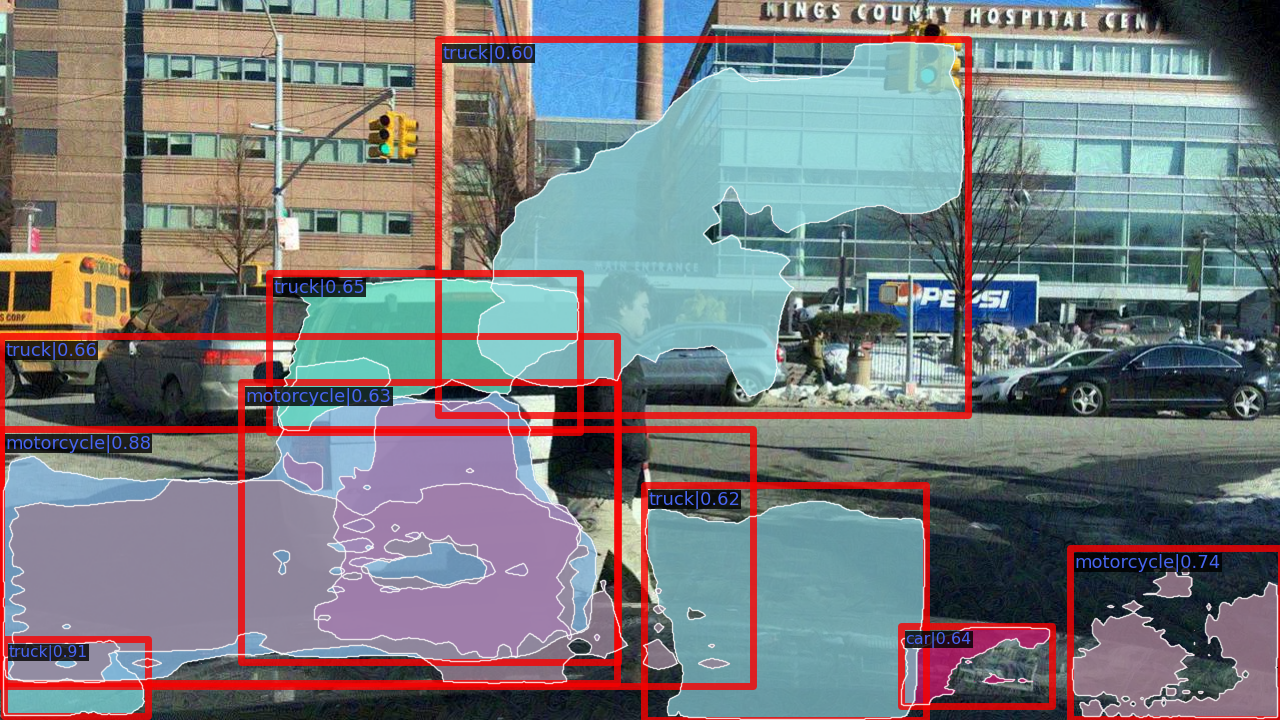}
        \caption{Perturbation optimized for instance segmentation}
        \label{1-intro-ae_seg}
    \end{subfigure}
    \caption{Impact of adversarial perturbation on the vision models}
    \label{fig:1-intro}
\end{figure*}

\subsection{Consistency between Vision Tasks}
As shown in \cref{fig:1-intro}, the inference outputs for object detection and instance segmentation on clean images exhibit overall consistency. Indeed, the object bounding boxes match with the object masks. However, on the perturbed images, discrepancies arise. For instance, in \cref{1-intro-ae_det}-\cref{1-intro-ae_det_seg}, the perturbation optimized for the object detection model successfully deceived the object detector, leading to numerous false positive and false negative predictions. On the other hand, the same perturbation did not fool the instance segmentation model, which accurately predicted the bounding boxes and masks.\footnote{We note a slight impact on the foreground objects' masks.} Similar impact is observed in \cref{1-intro-ae_seg_det}-\cref{1-intro-ae_seg} where the perturbation is optimized for instance segmentation. In fact, we can identify two types of consistency between the model outputs: 
\begin{itemize}[leftmargin=*]
    \item Location Consistency: refers to detecting an object at the same location within an input image using both an object detection model and an instance segmentation model. It involves calculating the Intersection over Union (IoU) between each detected object from both models. If the IoU exceeds a predefined threshold (e.g., 50\%), the object pair is considered \textit{location consistent}.
    \item Semantic Consistency: goes beyond location and ensures that the labels of the object pair are identical as well. In this paper, we consider a detection as \textit{consistent} if both location and semantic consistency are proven.
\end{itemize}
\paragraph{Consistency Score}
In this work, we call \textit{consistent detection} ($CD$) a matching pair of box and mask (location and label wise). In order to measure the overall consistency of a single image, \cref{eq:task_cs} defines the \textit{Consistency Score} $C_\text{task}$ as the ratio of total number of consistent detection over the total number of detection from either model ($N_\text{task}$).

\begin{equation}
    \label{eq:task_cs}
    C_{\text{task}} = \frac{|CD|}{N_{\text{task}}} \quad \text{task}\in\{det,seg\}
\end{equation}




Then, as in \cref{eq:mean_cs}, we define \textbf{consistency score} $C$ as a harmonic mean of $C_{\text{det}}$ and $C_{\text{seg}}$ to measure the overall consistency of the inferences on input images by both models.

\begin{equation}
    \label{eq:mean_cs}
    C = \frac{2 \cdot C_{\text{det}} \cdot C_{\text{seg}}}{C_{\text{det}} + C_{\text{seg}}}
\end{equation}

\begin{figure}[!t]
    \centering
    \includegraphics[scale=0.45]{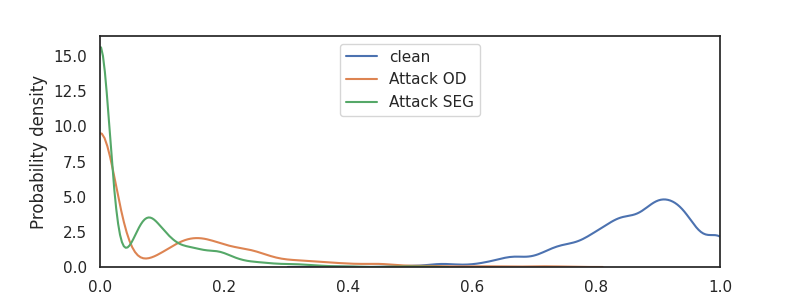}
    \caption{Empirical study of the consistency score distribution of FRCNN R50 on BDD100k dataset. Blue line shows consistency scores for clean images. Orange line shows consistency scores for perturbed images (target OD). Green line shows consistency scores for perturbed images (target SEG). The clear divergence between distributions, confirms the ability of our detector to identify perturbations.}
    \label{fig:4-consistency-dist}
\end{figure}

\paragraph{Empirical Study on BDD100k}
From \cref{fig:4-consistency-dist}, we observe that FRCNN R50 on clean images have higher consistency score, while perturbed images have much lower consistency score. This implies that we can distinguish between clean and perturbed images using the consistency score. We present other consistency score distribution plots of other models in \cref{app_subsec:det_perf}.

\subsection{Consistency Score based Attack Detection}
\begin{figure}[t]
    \centering
    \includegraphics[width=\linewidth]{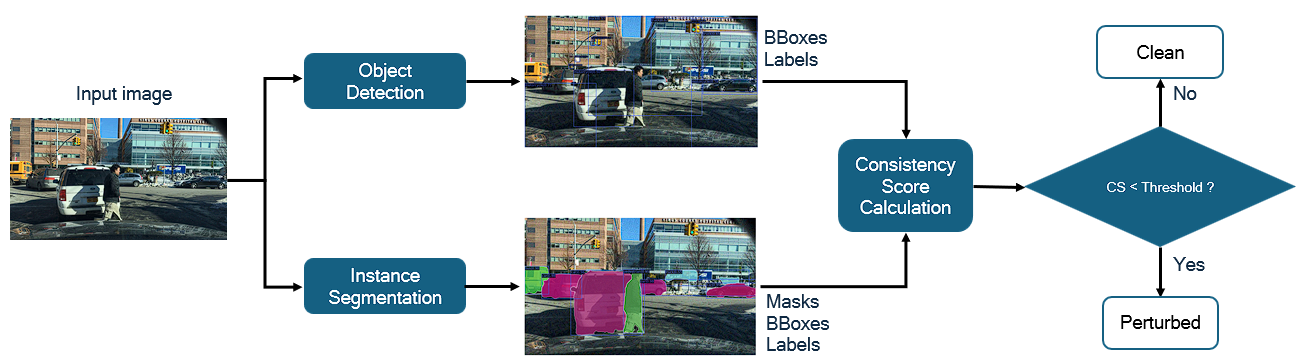}
    \caption{Pipeline of consistency score based adversarial perturbation detection}
    \label{fig:4-consistency-pipeline}
\end{figure}

Inspired by the above observation, we propose a consistency score based adversarial attack detection scheme illustrated in \cref{fig:4-consistency-pipeline}. 

\paragraph{Notations} In order to formulate the problem, we denote the output of the object detection model as a set of annotations of detected objects $S_\text{det}=\{(\text{BBox}_{\text{det},i}, \text{Label}_{\text{det},i})|i=1,...K_\text{det}\}$ where $\text{BBox}_{\text{det},i}$ is the bounding box coordinates for the $i$-th detection, $\text{Label}_{\text{det},i}$ is its corresponding class label, and $K_\text{det}$ is the total number of detection by the object detection model. Similarly, we denote the output of the instance segmentation model as $S_\text{seg}=\{(\text{BBox}_{\text{seg},j}, \text{Label}_{\text{seg},j})|j=1,...K_\text{seg}\}$. 

\paragraph{Step 1: Consistency Score Calculation} 
Following \cref{eq:task_cs}, the consistency score is calculated between the two tasks output. In \cref{app:algo}, we propose \cref{alg:cs_calc} as an implementation of the Consistency Score Calculation module of \cref{fig:4-consistency-pipeline}.


\paragraph{Step 2: Adversarial determination} With the consistency score generated for the input image, the next step is to decide if it is a clean or perturbed image. As shown in \cref{fig:4-consistency-pipeline}, a threshold-based binary classification takes the consistency score as input. If the consistency score is lower than the predefined threshold, the input is labelled as ``perturbed''. As shown in \cref{fig:4-consistency-dist}, setting a high cut-off threshold (e.g., 0.75) would trigger false positives. Conversely, selecting a low threshold (e.g., 0.2) would trigger false negatives. Therefore, there is a trade-off between false positive rate and false negative rate. Implementers would have to pick the appropriate threshold using known techniques~\cite{optimalthres}.

\paragraph{Cross-task model selection}
When designing a multi-task consistency detector, it is important to select the appropriate models used for each task. Indeed, the two models 
could share the same backbone and underlying structure, or only share the same backbone, or share similar backbone but with different layer depth. We aim at answering the question ``What model architectures or parameters affect the ability to detect adversarial inputs via multi-task consistency?''. For example, should the feature extractors be different? if so, to what extent? \cite{ghamizi2022adversarial} hinted that one should carefully select the auxiliary tasks added to reduce model vulnerability. Indeed, the addition of auxiliary tasks can have negative effects (e.g., larger model size, slower convergence of the common encoder layers, deterioration of clean performance). They raised the (still open) question of how to select the combination that yields the lowest vulnerability.
One could think that picking the most adversarially robust backbone would be preferable. For example, when investigating ResNet50 and ResNet101 backbones, the only difference is that ResNet101 has 23 conv4\_x layers while ResNet50 has 6 (so a total of 51 additional convolution layers as the name indicates). This means that ResNet101 has larger receptive fields than ResNet50. As shown by~\cite{xiang2023patchcure}, smaller receptive fields impose a bound on the number of features that can be corrupted, hence more adversarially robust. This could justify the use of ResNet50 backbone over ResNet101. However, in our context, we select models that, even if fooled by the attack, yield to inconsistent outputs. So, having two weak models could be acceptable as long as their outputs are inconsistent. 

\section{Experiments}
\label{sec:experiments}
In this section, we outline the implementation details of the datasets, models, attack parameters, and evaluation metrics used to ensure reproducibility. We then analyze the experimental results of the multi-task consistency-based detector. Additionally, we offer recommendations for a cross-model strategy to select the best model pairs for the detector.

\subsection{Implementation Details}

\paragraph{Datasets} Our evaluation relies on a set of genuine and adversarial datasets based on the BDD100k dataset. Details of the BDD100k dataset can be found in \cref{app_subsec: BDD100k}.

\paragraph{Models} We use 11 existing models from BDD100k model zoo~\cite{bdd100kmodelzoo} and from mmdetection 2.0 framework~\cite{mmdetection}: six models for object detection (OD) and five models for instance segmentation (SEG). All models are fine-tuned  on the BDD100k dataset. 
Our selection of models aims to maximize the diversity of models for a given vision task to understand how it affects the performance of our defense. Indeed, an adversarial attack may transfer from one model to another if their architectures are similar. Therefore, we chose our models based on a set of criteria. The first one is the type of architecture (e.g., transformer or CNN). A second criteria is the depth of the backbone (ResNet50 versus ResNet101). The last criteria is to ensure a diversity of heads among the models (e.g., FRCNN versus RetinaNet).

\paragraph{Attack} 
We utilized 1,000 clean images from the BDD100k instance segmentation validation dataset for our attack. This dataset was selected due to its comprehensive annotations, which include both segmentation masks and bounding boxes, allowing us to fairly assess the impact on both object detection (OD) and segmentation (SEG) models. We then applied the PGD-40 attack (40 iterations with a perturbation strength $\epsilon=16/255$) to each of the eleven models. This resulted in 11 adversarial datasets: six from attacking the OD models and five from attacking the SEG models. We use the clean dataset alongside these 11 adversarial datasets to evaluate the performance of the models and our detection scheme.

\paragraph{Evaluation Metrics} To evaluate the prediction performance of the models on both the clean dataset and the eleven adversarial datasets, we utilize the mean Average Precision (mAP), a widely accepted metric for assessing computer vision models. For evaluating our detection scheme, we employ the receiver operating characteristic (ROC) curve, a popular metric that illustrates the performance of a classification model across all classification thresholds. The area under the curve (AUC) provides a measure of our adversarial attack detection performance.

\subsection{Experimental Evaluation}
First, we study the effectiveness and transferability of the attack. Next, we assess the performance of our detector on detecting perturbations in digital domain.\footnote{To further demonstrate the applicability of our defense against physical adversarial perturbations, we tested a physical patch attack that targets misdetection of traffic signs. The attack was effective and our defense was able to detect it in real-time. Details can be found in \cref{sec:phy_test}.} 

\subsubsection{Prediction Performance Under Attack}

\begin{table*}[t]
\centering
\caption{Impact of the attack on the mAP of vision models} 
\label{tab:perf_drop}
\begin{adjustbox}{width=1\textwidth}
\begin{threeparttable}
\begin{tabular}{|c|c|c|cccccc|ccccc|}
\hline
\multicolumn{1}{|c|}{\textbf{Task}} & \multicolumn{1}{c|}{\textbf{Vision}} & \textbf{Clean} & \multicolumn{6}{c|}{\textbf{Attack Object Detection} ($\downarrow$)} & \multicolumn{5}{c|}{\textbf{Attack Segmentation} ($\downarrow$)} \\ 
\cline{4-14}
 & \multicolumn{1}{c|}{\textbf{Models}} & mAP ($\uparrow$) & \textbf{F R50}  &\textbf{F R101}& \textbf{F SwinT}  & \textbf{R R50}&\textbf{R R101}& \textbf{R PVT} & \textbf{M R50}  &\textbf{M R101}& \textbf{G R50}  &\textbf{G R101}& \textbf{M2F SwinT}\\ 
\hline
 & \textbf{F R50} & 30.2 & \underline{\textbf{0.18}} & 5.7 & 18.4 & \underline{0.34} & \underline{3.6} & 11.8 & 8.8 & 9.2 & 8.8 & 10.0 & 24.7\\
 & \textbf{F R101}& 30.3 & 7.5 & \underline{\textbf{0.17}} & 18.5 & \underline{3.2} & \underline{1.0} & 13.0 & 14.5 & 6.4 & 14.1 & 8.06 & 25.0\\
\textbf{OD} & \textbf{F SwinT} & 31.8 & 17.0 & 16.5 & \underline{\textbf{1.5}} & 11.5 & 12.0 & 14.8 & 20.7 & 18.6 & 20.7 & 19.0 & 22.4 \\
& \textbf{R R50} & 28.7 & \underline{2.2} & \underline{4.4} & 17.0 & \underline{\textbf{0.01}} & \underline{2.7} & 10.2 & 7.1 & 7.4 & 7.0 & 8.9 & 23.1 \\
 & \textbf{R R101}& 29.2 & 7.7 & \underline{2.2} & 17.9 & \underline{2.63} & \underline{\textbf{0.02}} & 11.9 & 14.0 & 5.6 & 13.5 & 7.1 & 24.3 \\
 & \textbf{R PVT} & 29.8 & 12.8 & 13.0 & 18.4 & 7.5 & 8.9 & \underline{\textbf{0.04}} & 18.2 & 15.0 & 17.8 & 15.0 & 24.8 \\
\hline
 & \textbf{M R50} & 19.8 & \underline{1.5} & \underline{2.6} & 10.1 & \underline{0.6} & \underline{2.6} & 7.1 & \underline{\textbf{0.01}} & \underline{1.6} & \underline{0.5} & \underline{2.2} & 13.4 \\
 & \textbf{M R101}& 20.5 & \underline{4.2} & \underline{1.8} &  10.3 & \underline{2.5} & \underline{1.4} & 8.0 & \underline{4.4} &\underline{\textbf{0.01}} & \underline{4.2} & \underline{0.72} & 13.4 \\
 \textbf{SEG} & \textbf{G R50} & 20.1 & \underline{1.7} & \underline{3.1} & 10.7 & \underline{0.62} & \underline{2.9} & 6.9 & \underline{0.73} & \underline{1.8} & \underline{\textbf{0.01}} & \underline{2.1} & 13.3 \\
 & \textbf{G R101} & 20.7 & \underline{4.2} & \underline{2.0} & 10.3 & \underline{2.6} & \underline{1.7} & 7.6 & \underline{4.4} & \underline{0.3} & \underline{4.1} & \underline{\textbf{0.01}} & 13.2 \\ 
 & \textbf{M2F SwinT}& 21.0 & 9.4 & 9.1 & 7.1 & 7.3 & 7.8 & 9.7 & 9.7 & 7.9 & 10.1 & 9.1 & \underline{\textbf{2.8}} \\ 
\hline
\end{tabular}
\begin{tablenotes}[flushleft]
    \item Acronyms: Object Detection (OD), Instance Segmentation (SEG), FRCNN (F), RetinaNet (RN), MRCNN (M), GCNET (G), Mask2Former (M2F)
    \item \textbf{A bold value} is the lowest mAP score among all targeted models for a given adversarial dataset.
    \item \underline{An underlined value} indicates the adversarial dataset successfully dropped the mAP score of the targeted model below 5 mAP. 
\end{tablenotes}
\end{threeparttable}

\end{adjustbox}
\end{table*}

\cref{tab:perf_drop} shows the mAP for each model across twelve test datasets. The table’s diagonal highlights that the attack is most effective on the target model for which the perturbation is optimized. For instance, the attack on the FRCNN R50 model decreases its mAP from 30.2 to 0.18. 

The perturbations demonstrate transferability across models with similar network architectures, regardless of the task. For instance, the adversarial dataset generated by attacking the OD model FRCNN R50 decreases the mAP of the SEG model MRCNN R50 from 19.8 to 1.5. Conversely, the attack on MRCNN R50 reduces the mAP of FRCNN R50 from 30.2 to 8.8. This indicates that the perturbation can transfer to different tasks or models with the same backbone architectures. Transferability is also observed in models that share the same backbone type but differ in depth. As shown in \cref{tab:perf_drop}, attacks on models with an R50 backbone (see columns) can transfer to models with an R101 backbone (see rows), and vice versa. However, for models with the same baseline architecture but different backbones, such as FRCNN R50 and FRCNN SwinT, or RetinaNet R50 and RetinaNet PVT, the transferability is less evident. This indicates that the backbone plays a more crucial role in the transferability of the attack.

\begin{figure}[!t]
    \centering
    \includegraphics[scale=0.45]{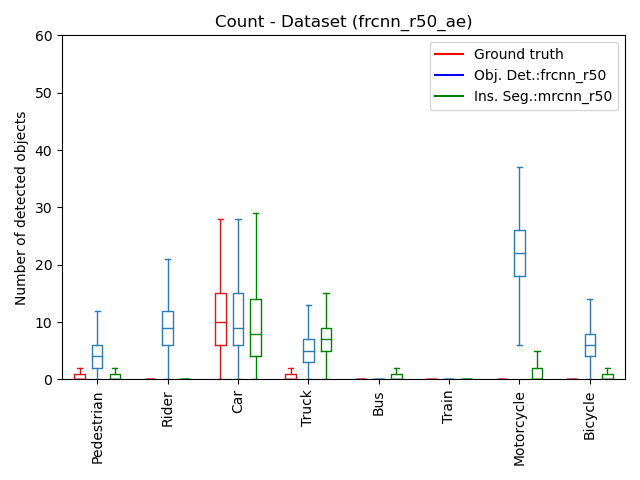}
    \caption{Attack on FRCNN R50: Impact on object counts}
    \label{stat1}
\end{figure}

While perturbations can transfer between different models and tasks, their fine-grained impact varies significantly across models. This variation is evident in several aspects, such as the number of objects detected. \cref{stat1} illustrates this using one model pairs: (FRCNN R50, MRCNN R50). It shows the distribution of the number of objects for each category given the adversarial dataset optimized on FRCNN R50. The attack generates significantly more objects on FRCNN R50 than on MRCNN R50, especially for categories like rider and motorcycle. 
This demonstrates that even when perturbations transfer, they can lead to inconsistent impacts on different models.\footnote{See \cref{app_subsec:transferability} for full transferability analysis across all model pairs considered.}

\subsubsection{Perturbation Detection Performance}
\label{sec:det_performance}
\begin{figure*}[!t]
    \centering
    \includegraphics[scale=0.75]{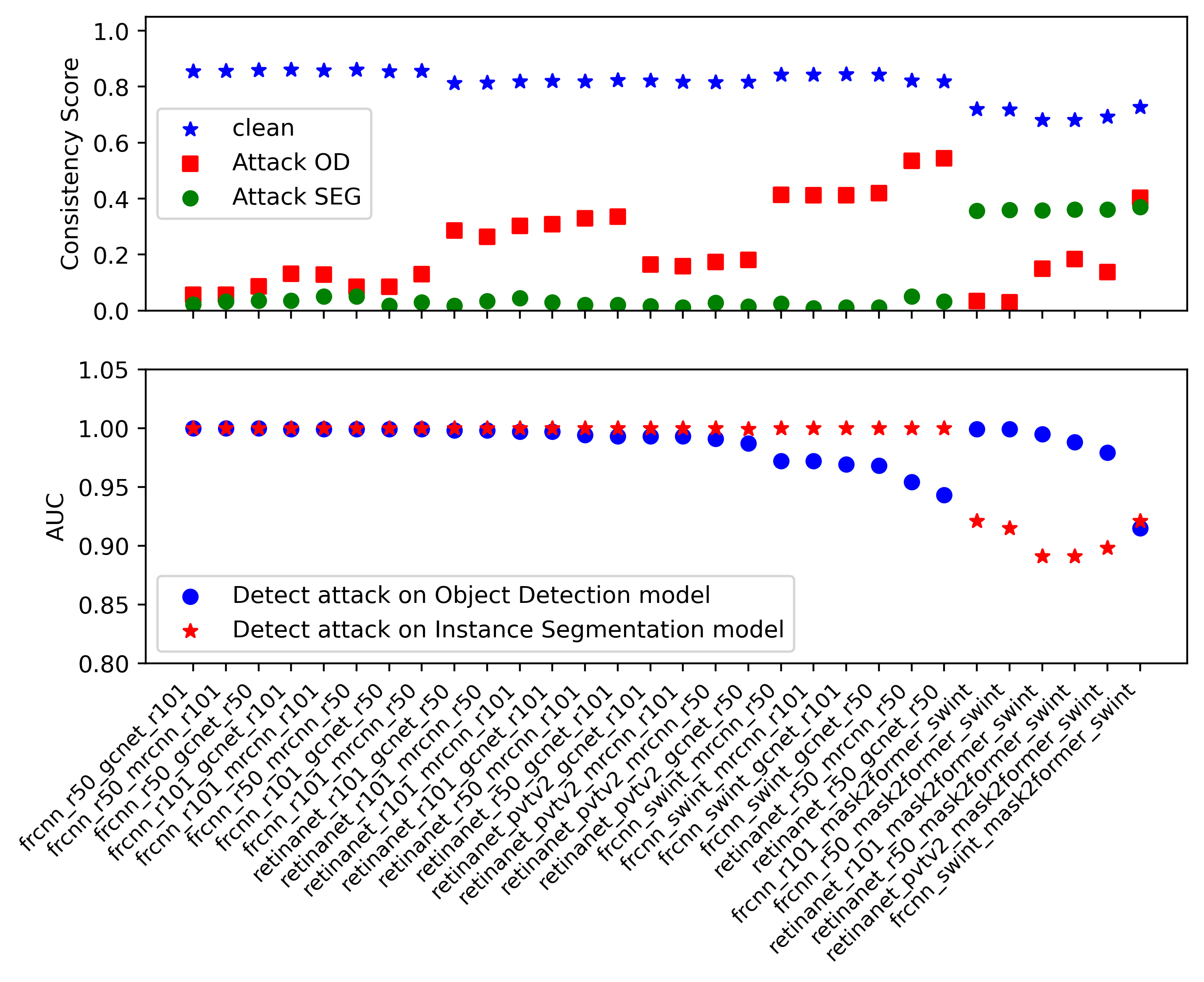}
    \caption{Top: Consistency Score for all model pairs. The lower the better the pair is for our detector. Bottom: The AUC for all model pairs. The higher the better the pair is our detector.}
    \label{fig:AUC_CS}
\end{figure*}

We evaluated the performance of our detector across 30 (6 OD $\times$ 5 SEG) model pairs. 
As previously noted in \cref{fig:4-consistency-dist}, we aim for a model pair that exhibits a high consistency score on clean inputs while a lower score on adversarial inputs, facilitating the identification of perturbations. 
\cref{fig:AUC_CS} (top) illustrates the average consistency score of model pairs across the three datasets (clean, attack OD, attack SEG). The blue stars represent the average consistency score for clean inputs. Generally, the consistency score for clean inputs is high, especially for model pairs with similar baseline architectures (RCNN) and backbones (ResNet), which can extract consistent features from the clean inputs, resulting in consistent outputs. Model pairs with different architectures or backbones exhibit slightly lower consistency score due to their varying feature extraction capabilities, leading to inconsistent outputs. The red squares represent the consistency score for adversarial datasets optimized on OD models. As discussed in the previous section, similar architectures (RCNN and ResNet) result in high transferability but also high inconsistency, causing consistency score to drop as low as 0 for those model pairs. For model pairs with different architectures, the attack shows less transferability, and thus, higher consistency. Similar findings are observed when attacking SEG models (green circles). The full analysis can be found in \cref{app_subsec:transferability}.

The AUC curves in \cref{fig:AUC_CS} (bottom) demonstrate that all model pairs achieve an AUC greater than 85\%, with most exceeding 95\%, when either model of the pairs is attacked. This highlights the exceptional performance of our consistency-based detector in identifying perturbations. Model pairs with similar backbone types (ResNet) and baseline architecture (RCNN) exhibit the highest performance, achieving an AUC of 99.9\%. Again, it shows that, although the attack can easily transfer between these models, this transferability leads to distinct variations in the number, label, and size of the detected objects. These variations result in a higher level of inconsistency, which our detector can effectively identify. In contrast, model pairs with different backbones or baseline architectures exhibit low transferability and low inconsistency, resulting in a relatively lower AUC.

Next, we are interested to learn how the perturbation strength of the attack can impact the prediction performance of the models and the detection performance of our detector. We evaluate the robustness of the models against attack size $\epsilon \in \{1/255, 2/255, 4/255, 8/255, 16/255\}$. 
Results of all model pairs can be found in the \cref{app_subsec:det_perf}.
We observe that the mAP of both models decreases as the perturbation strength increases, which is expected. Conversely, 
the detection performance in terms of AUC increases. This is the desired behavior because stronger perturbation leads to greater inconsistency (lower consistency score) between the outputs of model pairs, resulting in higher detection performance for our detector.

\paragraph{Takeaway on multi-task architecture}
The empirical analysis indicates that our detector performs optimally when model pairs exhibit high inconsistency in their outputs. Under our attacker model, the most effective model pairs are those with similar architectures and backbones, as they demonstrate high adversarial transferability but also high inconsistency. 

\subsection{Adaptive Attack}
\label{sec:adaptive-attack}
The PGD attack discussed in the previous section targets only one of the two tasks. However, an adaptive attacker may attempt to optimize the adversarial perturbation to deceive both tasks simultaneously, hoping to evade our defense. A straightforward approach is a \textbf{Joint Attack} that can fool both models (more details can be found in \cref{joint_attack}). Nevertheless, the impact of the perturbation on each task is independent, and the outputs of both tasks can vary significantly in terms of object location, size, and labels. This variability allows our detector to still capture the inconsistency and detect the perturbation.

Therefore, we create an \textbf{Adaptive Attack}, which aims at (i) fooling the task output, and (ii) creating consistent outputs. To ensure that the outputs of both tasks are consistent, we introduce a consistency loss term $L_{con}$ that measures the cosine similarity ($CS$) between the object proposals of two models before NMS. Specifically, $L_{con} = CS(pred_{det}, pred_{seg})$. 
The total loss for the consistent adaptive attack is the weighted sum of the consistency loss and the losses of the two models, given by $L_{total} = \beta \times L_{con} + (1-\beta) \times (L_{det} + L_{seg})$, where the value of weight $\beta$ is in range $[0, 1]$. 

This formulation is crucial for balancing the two competing objectives of the attack. 
First, the weight $\beta$ is used to balance the loss scales, as the consistency loss and the models' losses may have different magnitudes. Without appropriate weighting, the larger-scale loss may dominate the gradients. 
Second, the weight is used to balance the task priorities, i.e., maximizing the prediction errors and aligning the model outputs. A high $\beta$ will enforce strict alignment but risks reducing the attack performance, while a low weight could lead to inconsistency between model outputs.

\cref{tab:Joint_attack} shows the comparison of the attacks' impact on model mAP and consistency score $C$ (as a reminder, a low $C$ means easier attack detection). When only one task is targeted, the mAP of the targeted model is significantly reduced, while the other task remains relatively unaffected, therefore leading to a low consistency score between the model outputs. In contrast, the Joint Attack results in degradation of both tasks' performance, highlighting the increased vulnerability when both models are attacked simultaneously. Similarly, the Adaptive Attack also causes a significant drop in mAP for both models but also exhibits an increase of the consistency between the model outputs (due to the prioritization of consistency loss during optimization).

\begin{table}[t]
\centering
\caption{Impact of single model attacks, Joint Attack, and Adaptive Attack on mAP and consistency score $C$}
\label{tab:Joint_attack}
\small
\begin{tabular}{l|l|l|l}
\hline
   & $\text{mAP}_{det}$  & $\text{mAP}_{seg}$ & $C$  \\
  \hline
  \textbf{Clean} & 29.8 & 19.8 & 0.91 \\
  \hline
  \textbf{Attack Det} & 0.04 & 7.1 & 0.1\\
  \hline
  \textbf{Attack Seg} & 18.2 & 0.01 & 0.06\\
  \hline
  \textbf{Joint Attack} & 0.12 & 0.37 & 0.34 \\
  \hline
  \textbf{Adaptive Attack} & 1.2 & 2.7 & 0.52 \\
  \hline
\end{tabular}
\end{table}

\subsection{Adaptive Attacker: Impact of \texorpdfstring{$\beta$}{beta}}
\label{app:adaptive}
\begin{figure}[!t]
    \centering
    \includegraphics[scale=0.45]{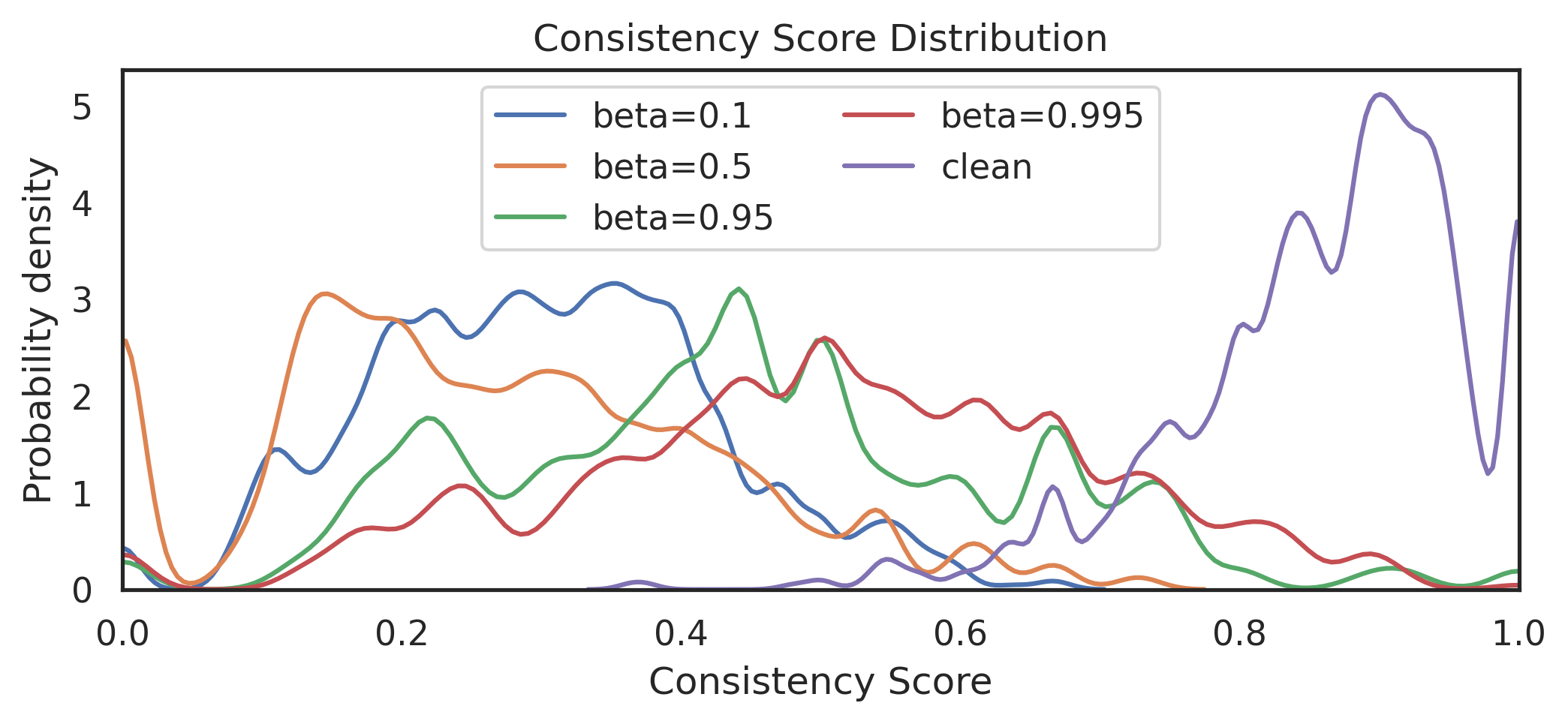}
    \caption{When prioritizing consistency loss with higher $\beta$, the model outputs are more consistent.}
    \label{fig:adaptive-consistency-dist}
\end{figure}

Figure~\ref{fig:adaptive-consistency-dist} shows the effect of $\beta$ on the consistency score.
As shown in \cref{fig:adaptive-consistency-dist}, when $\beta$ is low, the consistency scores for the perturbed images are low, indicating highly inconsistent outputs between models. As $\beta$ increases, the distribution shifts to the right, closer to the clean distribution, demonstrating that the model outputs are becoming more consistent. This indicates that by prioritizing the consistency loss, the attack results in more consistent outputs.

\begin{figure}[t]
    \centering
    \includegraphics[scale=0.45]{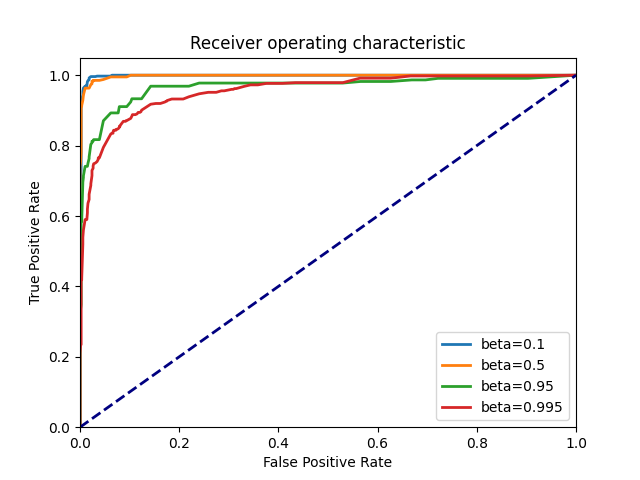}
    \caption{As $\beta$ increases, the AUC of ROC decreases showing that the prioritizing consistency loss causes less inconsistent outputs. However, the AUC is still high as 0.95 indicating space to improve our loss function design.}
    \label{fig:5-beta-values}
\end{figure}

As shown in \cref{fig:5-beta-values}, the ROC curves indicate the performance of our detector on detecting adaptive attacks under different values of $\beta$. A low AUC indicates that the adaptive attack can evade our detector because there is less inconsistency between model outputs. As the $\beta$ increases, the AUC becomes lower showing the outputs of both models become more consistent, which is caused by prioritizing the consistency loss. However the AUC is still very high (0.95 when $\beta$ is 0.995), indicating that our defense is effective against the adaptive attack.

\section{Comparison with other Defenses}
\label{sec:defense}
To evaluate the effectiveness of our multi-task consistency detector, we compare its performance against established defense mechanisms. A broader discussion of defense strategies is provided in \cref{defense}. Notably, Dong et al. \cite{dong2022adversarially} introduced RobustDet, a defense approach that modifies the model architecture as an alternative to adversarial training. We include RobustDet in our comparative analysis to highlight the advantages of our proposed method.

\subsection{Application of RobustDet on Faster RCNN}

\begin{table*}[!h]
\centering
\caption{Impact of the attack on the mAP of regular and Robust FRCNN R50}
\label{tab:robust_perf}
\small
\begin{tabular}{l|cccc|cccc}
  \hline
 \multicolumn{1}{c|}{\textbf{Model}}  & \multicolumn{4}{c|}{\textbf{Clean}}  & \multicolumn{4}{c}{\textbf{Attack}} \\
 \cline{2-9}
        & $\text{mAP}$ & $\text{mAP}_{small}$ & $\text{mAP}_{medium}$ & $\text{mAP}_{large}$ & $\text{mAP}$ & $\text{mAP}_{small}$ & $\text{mAP}_{medium}$ & $\text{mAP}_{large}$ \\
  \hline
  \textbf{FRCNN R50} & 30.2 & 12.4 & 34.6 & 54.4 & 0.18 & 0.07 & 0.24 & 0.33 \\
  \hline
  \textbf{Robust FRCNN} & 19.8 & 8.1 & 22.4 & 36.7 & 6.2 & 2.4 & 7.3 & 11.9 \\
  \hline
\end{tabular}
\end{table*}

\cite{dong2022adversarially} evaluated RobustDet on the object detection model SSD with a VGG16 backbone, which was trained on the COCO dataset. However, in our paper, the models evaluated are trained on BDD100k dataset and we do not use SSD. To fairly compare the performance of our consistency-based detector with RobustDet, we applied the \textit{AAconv} technique to one of the models, namely Faster RCNN with RestNet50 backbone (FRCNN R50 in short). 

As previously explained, the core idea of \textit{AAconv} is to replace any regular convolution kernel in a model network with a weighted sum of a set of dynamic convolution kernels, expressed as:

\begin{equation}
\dot{\theta}^{AA_{\textit{conv}}} = \sum_{i=1}^{M} {\theta_i}^{AA_{\textit{conv}}} \cdot \pi_i
\end{equation}

where ${\theta_i}^{AA_{\textit{conv}}}$ is the $i$-th kernel in the set of $M$ dynamic kernels, and $\pi_i$ is its corresponding weight generated by \textit{AID}. To clarify, each convolution kernel in the original network will be replaced by a unique set of dynamic kernels whose parameters are determined during the training phase. Thus, for each convolution layer in original Faster RCNN, we replace it with a dynamic convolution layer as defined by \cite{dong2022adversarially}. Regarding \textit{AID}, we use the same network architecture Resnet18 as in the original paper. 

\subsection{Performance of Robust FRCNN}

We evaluate the performance of two models based on FRCNN R50: the standard model and a robust model. \cref{tab:robust_perf} presents the performance of both models under clean and adversarial datasets. For the standard model, we used the same adversarial dataset as previously mentioned. For the robust model, we applied PGD attack using same attack parameters. \cref{tab:robust_perf} shows that the standard model experiences a significant performance drop due to the attack, compared to the robust model. Specifically, its mAP decreases from 30.2 to 0.18, while for the Robust FRCNN R50, it decreases from 19.8 to 6.2. Hence, RobustDet technique enhances the model’s adversarial robustness. It is worth noting that the clean mAP for the robust model is not high, indicating there is still room to adjust the training parameters to improve both its clean and adversarial performance.

\subsection{Comparison to RobustDet}
\label{sec:comp-RobustDet}

In this section, we compare our detector to the adversarial training method RobustDet proposed by~\cite{dong2022adversarially}. For a fair comparison, we applied RobustDet to FRCNN R50 which resulted in a robust model named Robust FRCNN. More details about our implementation and performance results can be found in \cref{app_subsec:robustdet}. Our detector functions as a binary classifier, determining whether an input is adversarial or not. In contrast, RobustDet, similar to adversarial training, enhances the model’s robustness. To ensure a fair comparison, we introduce the metric \textit{Detection Rate}, which represents the true positive rate for a given adversarial dataset. Specifically, we utilize one of our best model pairs (FRCNN R50, MRCNN R50) for our detector and assess its detection rate on the adversarial dataset for FRCNN R50. For Robust FRCNN R50, we evaluate its performance by calculating the consistency score between its output and the ground truth annotations under adversarial conditions. A high consistency score indicates that Robust FRCNN successfully mitigates the perturbation, whereas a low score signifies failure. Therefore, the detection rate is the ratio of adversarial inputs with a consistency score above the consistency threshold.

\begin{table}[t!]
\centering
\caption{Robust FRCNN R50 vs our detector: comparison of attack detection rate, model size, and frame per second.}
\label{tab:perf_comparison}
\small
\begin{tabular}{l|p{1.5cm}|p{1.5cm}|p{1.5cm}}
\hline
  \textbf{Defense} & Detection \newline Rate & Model Size \newline (MB) & Speed \newline (FPS) \\
  \hline
  \textbf{RobustDet} & 19 & 643 & 11 \\
  \hline
  \textbf{Our detector} & \textbf{99.9} & \textbf{350} & \textbf{20} \\
  \hline
\end{tabular}
\end{table}

As shown in \cref{tab:perf_comparison}, our consistency-based detector successfully identifies all adversarial inputs (100\%) in the adversarial datasets, thanks to the high inconsistency between the outputs of the model pair. In contrast, Robust FRCNN R50 performs poorly ($\text{mAP}=6.2$), failing to ensure prediction outputs align with the ground truth, resulting in a very low detection rate (19\%). On top of being less able to detect adversarial inputs, Robust FRCNN R50 employs a dynamic convolution kernel that is four times the size of a regular convolution kernel, significantly increasing its model size. In comparison, our detector has a combined weight size of only 350MB for both OD and SEG. Finally, our detector achieves faster inference speeds on the same hardware due to its lightweight architecture. In summary, our detector demonstrates stronger performance than RobustDet.

\section{Open Challenges}
\label{sec:discussion}
In this section we describe open challenges that will serve as future work.


\paragraph{Generalization} In this paper, we investigated object detection and instance segmentation models, finding the best model pairs to use in a multi-task consistency detector. We are interested in generalizing the approach to any combination of tasks. For example, depth estimation can be combined with semantic segmentation to detect inconsistency. Moreover, we would like to understand if the recommendations (about the model architectures) generalized across tasks.

\paragraph{Tuple multi-task consistency} We propose to extend the detector with more than two tasks and investigate how the detection rate correlates to the number of tasks. Though, \cite{ghamizi2022adversarial} demonstrated that what matters the most is not the number of tasks or how they correlate, but how much the tasks individually impact the vulnerability of the model. Indeed, the more vulnerable the tasks in the model are, the less likely adding new tasks increases the robustness of the model; and adding a vulnerable task may actually decrease the robustness of the whole model. Thus, a comprehensive analysis is required to answer this challenge.

\section{Related Work}
\label{sec:related-work}

In recent years, many defenses were created to detect~\cite{hendrycks2016baseline, liu2019detection, tian2021detecting, sperl2020dla} or to improve the robustness of vision systems against adversarial perturbations~\cite{hendrycks2019using, mkadry2018towards}. Especially, a strong emphasis has been put on the security of image classification task. Examples of defenses used in image classification include: use of additional detection networks~\cite{liu2019detection}, analysis of network output~\cite{hendrycks2016baseline, tian2021detecting}, or use of certain activation patterns within the hidden layers~\cite{sperl2020dla}. These detection methods focus on the output structure or network topology of an image classifier and are thereby not transferable to more complex vision tasks. 

As described earlier, multi-task learning (MTL)~\cite{kendall2018multi} tackles a wide range of vision tasks efficiently. \cite{mao2020multitask} showed that MTL increases the adversarial robustness due to the increased difficulty of successfully attacking several tasks. Thus, subsequent work explored other task combinations~\cite{xie2017adversarial, klingner2020improved, wang2020defending, kumar2021syndistnet}, or compared the effectiveness of adding different auxiliary tasks~\cite{ghamizi2022adversarial, gurulingan2021uninet, haleta2021multitask}. While the positive effects of MTL on adversarial robustness are quite well-explored, we are the first to check the consistency between outputs from object detection and instance segmentation models, deriving recommendations to select best model pairs. 


\section{Conclusion and Future Work}
\label{sec:conclusion}
Vision models are paramount to many applications such as autonomous driving. Their robustness have been shown to be brittle under adversarial setting. From the observation that adversarial inputs yield different effects when fed to different models, we propose an adversarial perturbation detection method based on multi-task perception. We showed an example of our lightweight defense using instance segmentation and object detection tasks. We generated adversarial BDD100k datasets and demonstrated our consistency score can effectively detect perturbations. Then, we empirically identified the optimal model pairs, demonstrating that even if sharing the same backbone, the attack can be detected because of uncoordinated perturbations on both models. The optimal models pair had a 99.9\% detection rate. 

Future work will focus on exploring other combinations of vision task, and continue investigating consistent joint multi-task perturbations.
Indeed, in this paper, we investigated object detection and instance segmentation models, finding the best model pairs to use in a multi-task consistency detector. We are interested in generalizing the approach to any combination of tasks. For example, motion estimation can be combined with semantic segmentation to detect inconsistency. However, one must define the features and adapt the consistency score metric. Moreover, we would like to understand if the recommendations (about the model architectures) generalized across tasks.



{
    \small
    \bibliography{arxiv}
    \bibliographystyle{IEEEtran}
}

\clearpage
\setcounter{page}{1}
\appendices

\section{Algorithm}
\label{app:algo}
\cref{alg:cs_calc} describes the high level steps involved in the computation of the consistency score. 
\begin{algorithm}
\caption{Consistency Score Calculation}
\label{alg:cs_calc}
  \begin{algorithmic}
    \INPUT $\mathcal{S}_{det}$ //A set of pairs of bounding boxes and labels from object detection
    \INPUT $\mathcal{S}_{seg}$ //A set of pairs of bounding boxes and labels from instance segmentation
    \OUTPUT Consistency Score ($C$)
    \STATE $|CD| = 0$ // Number of pairs (box and mask) 
    \STATE $IoU = calc\_iou(\mathcal{S}_{det}, \mathcal{S}_{seg})$ // IoU score and label similarity for between pairs of 
    $\mathcal{S}_{det}$ and $\mathcal{S}_{seg})$
    \STATE $IoU = prune(IoU, threshold)$ // Prune each box with all IoU scores below threshold
    \STATE $n\_box_{seg}, n\_box_{det} = get\_number(IoU)$ // Get remaining number of boxes for each task
    \STATE $|CD| = compute\_n\_pairs(n\_box_{seg}, n\_box_{det})$ // Get total number of pairs
    \STATE $C = compute\_c(|CD|, len(\mathcal{S}_{det}), len(\mathcal{S}_{seg})$ 
  \end{algorithmic}
\end{algorithm}

\section{Dataset: BDD100k}
\label{app_subsec: BDD100k}
The BDD100K dataset is a public dataset of driving scenes, which contains 100k frames
and annotations for 10 vision tasks. Compared with other driving datasets, the BDD100k dataset has a diversity of geography, environment, and weather. Therefore, we use the BDD100k as the benchmark dataset to train the models and evaluate our detection. In particular, we use the 100k subfolder for object detection task, which is split to 70k training, 10k validation and 20k testing images. We also use the 10k subfolder for instance segmentation task, which is split to 7k training, 1k validation and
2k testing images.

\section{Models}
\label{app_subsec: model}
In our experiment, we use the following models for object detection: frcnn\_r50, frcnn\_r101, retinanet\_r50, retinanet\_r101, retinanet\_pvtv2, frcnn\_swint.
For instance segmentation, we use:  gcnet\_r50, gcnet\_r101, mrcnn\_r50, mrcnn\_r101, mask2former\_swint.

\section{Adversarial Transferability}
\label{app_subsec:transferability}
\begin{figure}[h!]
    \centering
    \begin{subfigure}{.24\linewidth}
        \centering
        \includegraphics[width=\linewidth]{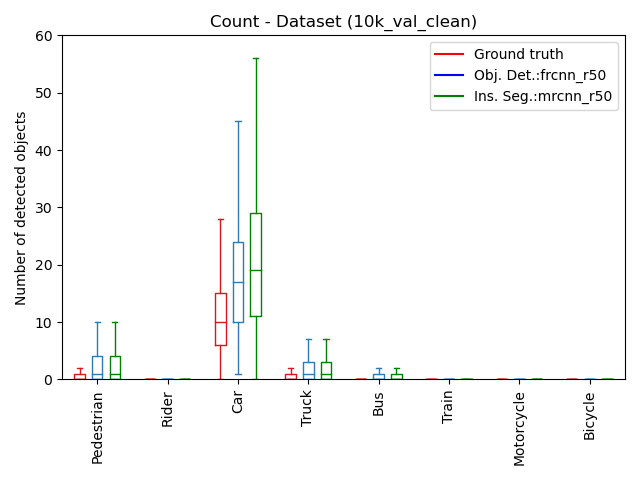}
    \end{subfigure}
    \begin{subfigure}{.24\linewidth}
        \centering
        \includegraphics[width=\linewidth]{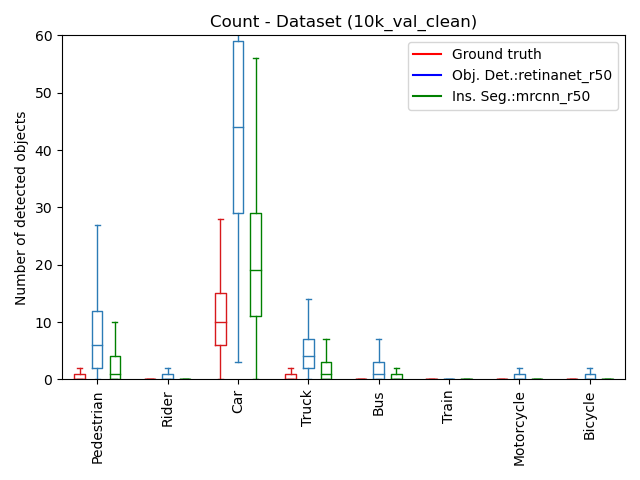}
    \end{subfigure}
    \begin{subfigure}{.24\linewidth}
        \centering
        \includegraphics[width=\linewidth]{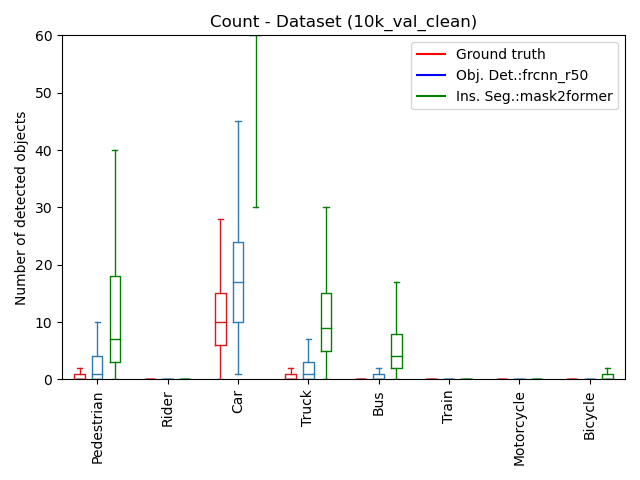}
    \end{subfigure}
    \begin{subfigure}{.24\linewidth}
        \centering
        \includegraphics[width=\linewidth]{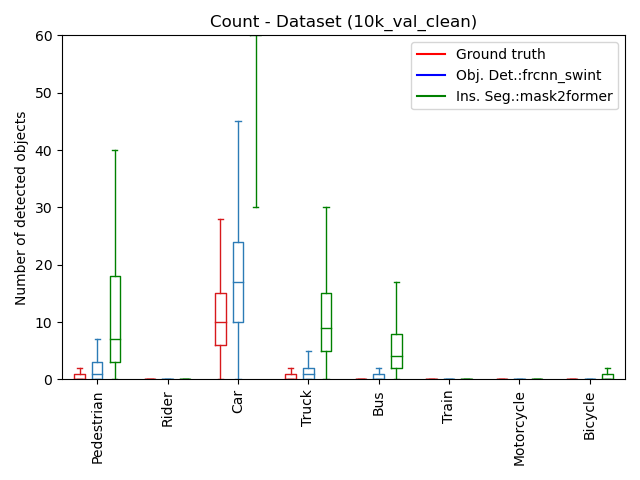}
    \end{subfigure}
    \caption{Clean images on model pairs}
    \label{fig:clean-on-model-pair}
\end{figure}

\begin{figure}[h!]
    \centering
    \begin{subfigure}{.24\linewidth}
        \centering
        \includegraphics[width=\linewidth]{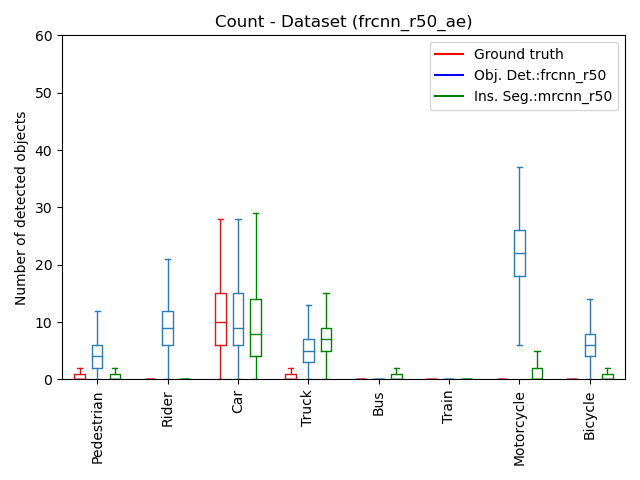}
    \end{subfigure}
    \begin{subfigure}{.24\linewidth}
        \centering
        \includegraphics[width=\linewidth]{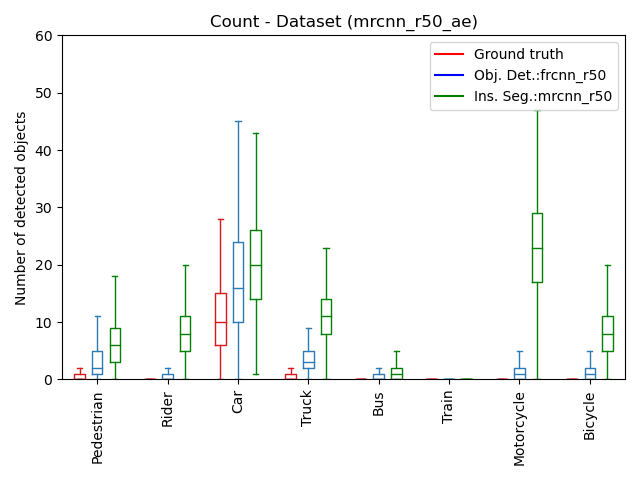}
    \end{subfigure}
    \begin{subfigure}{.24\linewidth}
        \centering
        \includegraphics[width=\linewidth]{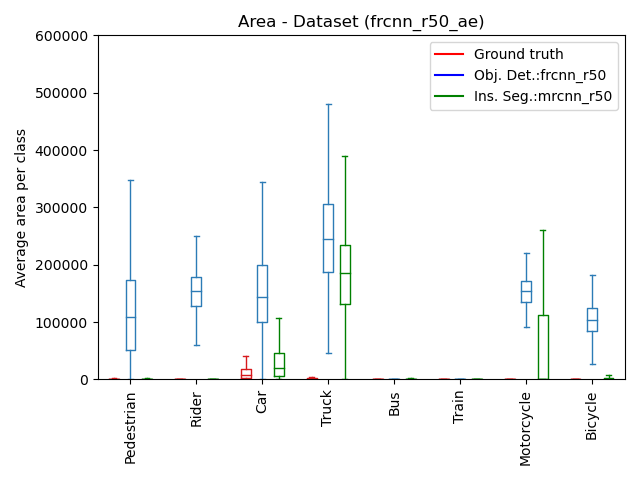}
    \end{subfigure}
    \begin{subfigure}{.24\linewidth}
        \centering
        \includegraphics[width=\linewidth]{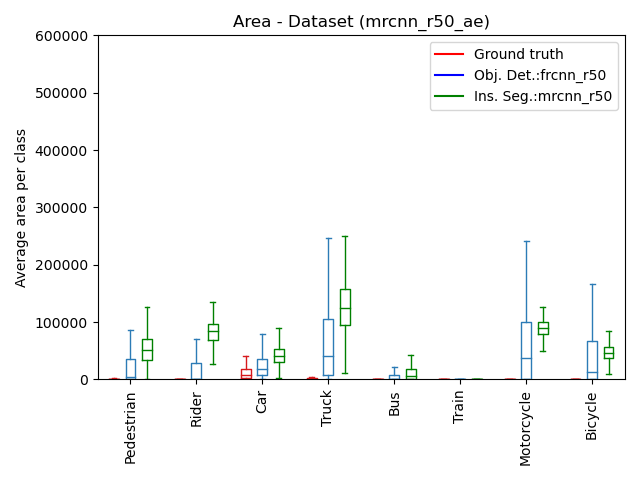}
    \end{subfigure}
    \caption{OD\_frcnn\_r50\_SEG\_mrcnn\_r50}
    \label{fig:same_arch}
\end{figure}

\begin{figure}[h!]
    \centering
    \begin{subfigure}{.24\linewidth}
        \centering
        \includegraphics[width=\linewidth]{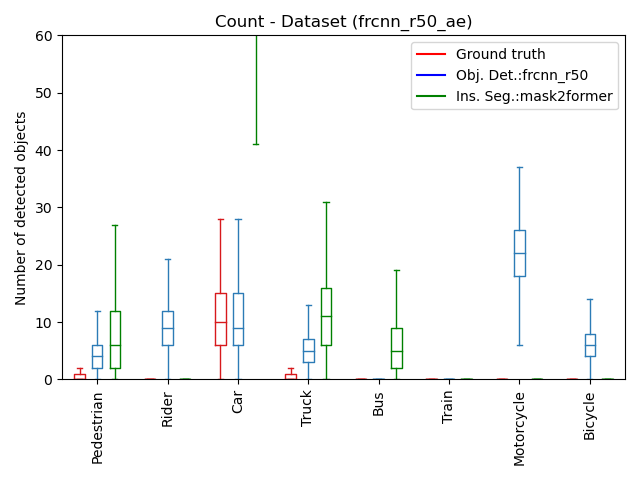}
    \end{subfigure}
    \begin{subfigure}{.24\linewidth}
        \centering
        \includegraphics[width=\linewidth]{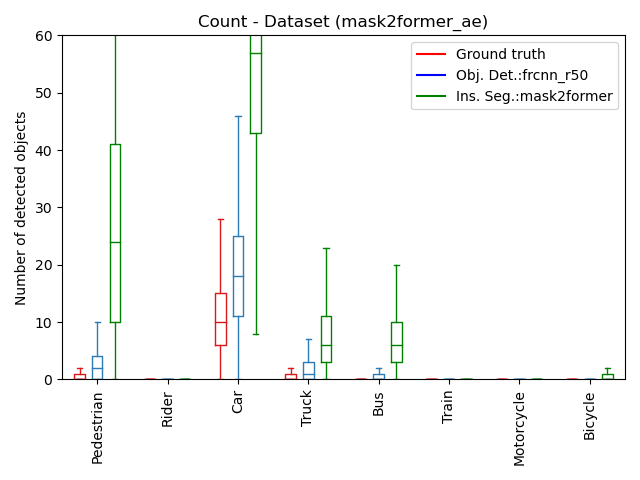}
    \end{subfigure}
    \begin{subfigure}{.24\linewidth}
        \centering
        \includegraphics[width=\linewidth]{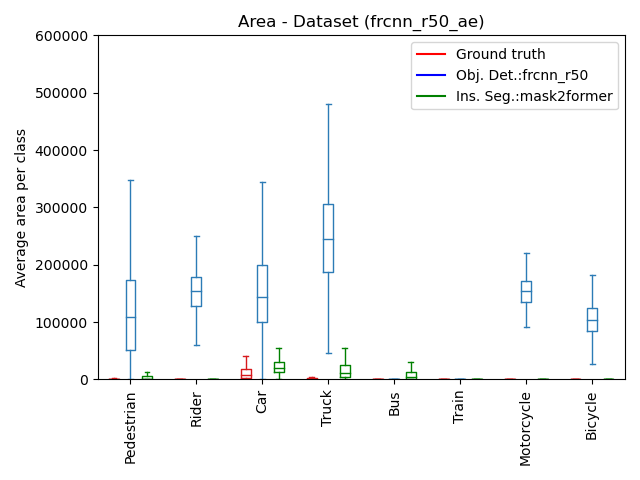}
    \end{subfigure}
    \begin{subfigure}{.24\linewidth}
        \centering
        \includegraphics[width=\linewidth]{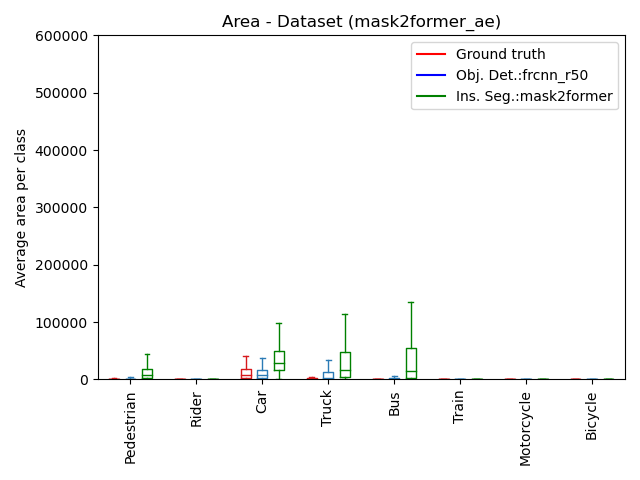}
    \end{subfigure}
    \caption{OD\_frcnn\_r50\_SEG\_mask2former}
    \label{fig:diff_arch}
\end{figure}

This section presents the distinct impact of the attack on OD and SEG models across all model pairs, focusing on the number and size of the detected objects. As previously mentioned, the attack exhibits high transferability between models with similar architectures and backbones, but it also leads to significant inconsistencies in the model outputs. For instance, \cref{fig:same_arch} shows the attack transfer between FRCNN R50 and MRCNN R50, but the number and size of hallucinated objects across the categories vary. For model pairs with different baseline architectures or backbones, such as FRCNN R50 and MASK2FORMER SwinT in \cref{fig:diff_arch}, the adversarial dataset optimized on MASK2FORMER does not fool FRCNN R50, whose outputs remain close to the ground truth in terms of number and size. Similarly, the adversarial dataset optimized on the FRCNN model does not fool MASK2FORMER whose object areas are close to ground truth. Although the number of objects is very large, this is due to the poor performance of MASK2FORMER, which predicts a large number of objects even on clean inputs, as shown in \cref{fig:clean-on-model-pair}. 

This observation also supports the conclusion in \cref{sec:det_performance}. The consistency scores for model pairs with similar architectures are low because the attack transfers between them, resulting in distinct impacts, and thus, high inconsistency. For model pairs with different architectures, the attack does not transfer well, leading to low inconsistency. However, when one model in these pairs performs very poorly even on a clean dataset, it will output many hallucinated objects despite the attack not transferring to it, still resulting in high inconsistency, as seen with FRCNN R50 and MASK2FORMER SwinT.

\begin{figure}[h!]
    \centering
    \begin{subfigure}{.24\linewidth}
        \centering
        \includegraphics[width=\linewidth]{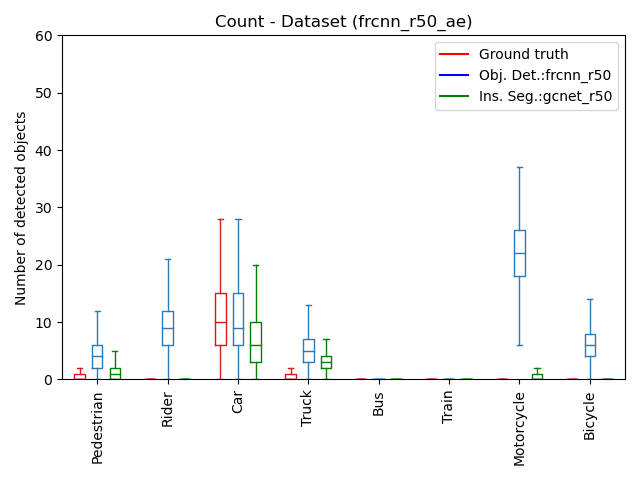}
    \end{subfigure}
    \begin{subfigure}{.24\linewidth}
        \centering
        \includegraphics[width=\linewidth]{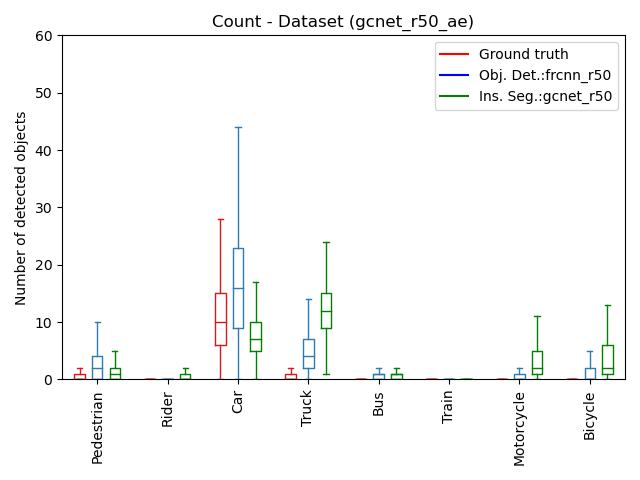}
    \end{subfigure}
    \begin{subfigure}{.24\linewidth}
        \centering
        \includegraphics[width=\linewidth]{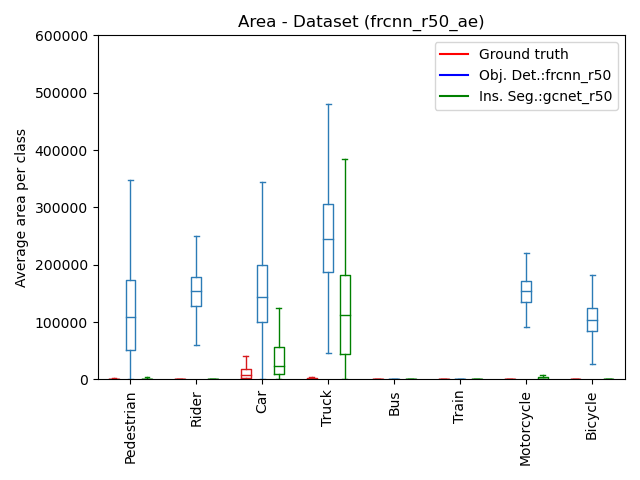}
    \end{subfigure}
    \begin{subfigure}{.24\linewidth}
        \centering
        \includegraphics[width=\linewidth]{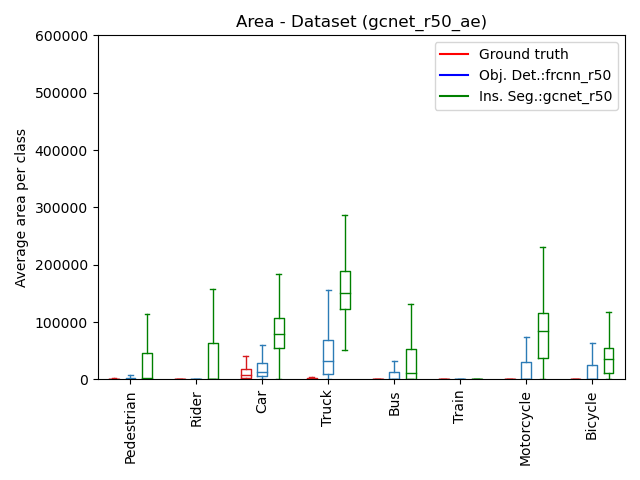}
    \end{subfigure}
    \caption{OD\_frcnn\_r50\_SEG\_gcnet\_r50}
\end{figure}

\begin{figure}[h!]
    \centering
    \begin{subfigure}{.24\linewidth}
        \centering
        \includegraphics[width=\linewidth]{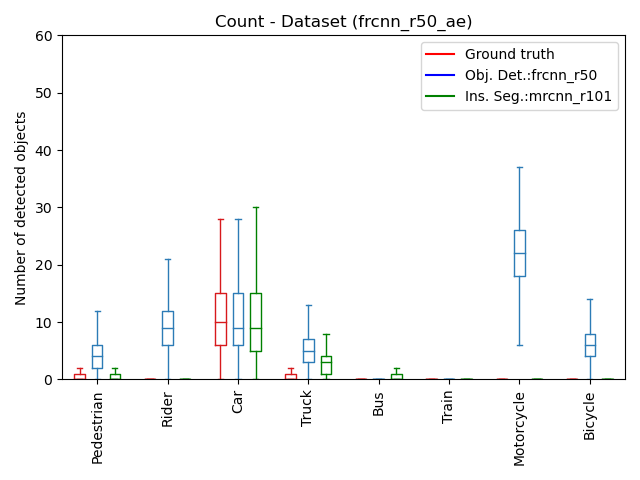}
    \end{subfigure}
    \begin{subfigure}{.24\linewidth}
        \centering
        \includegraphics[width=\linewidth]{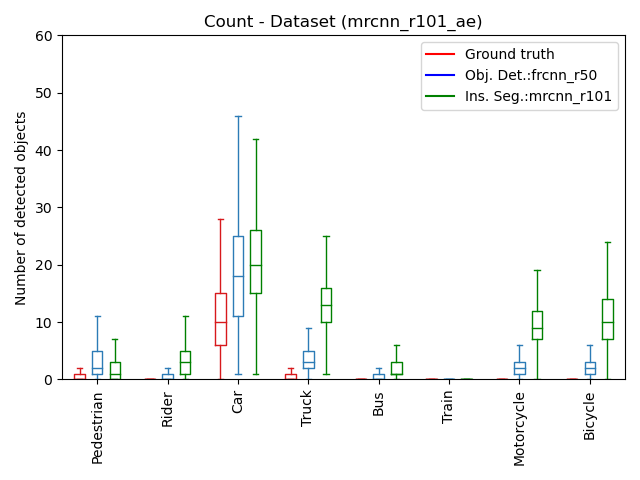}
    \end{subfigure}
    \begin{subfigure}{.24\linewidth}
        \centering
        \includegraphics[width=\linewidth]{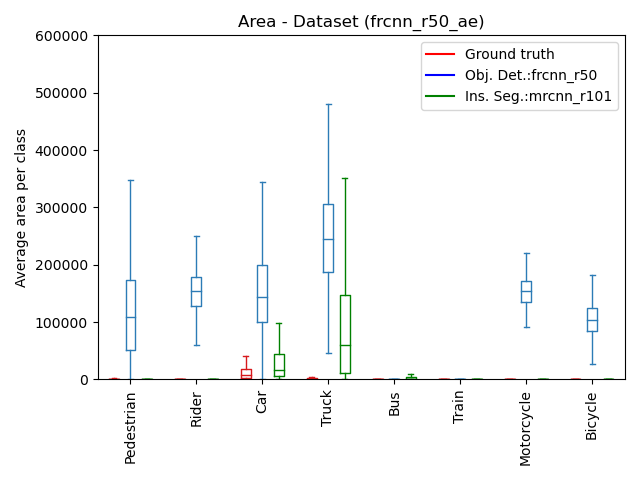}
    \end{subfigure}
    \begin{subfigure}{.24\linewidth}
        \centering
        \includegraphics[width=\linewidth]{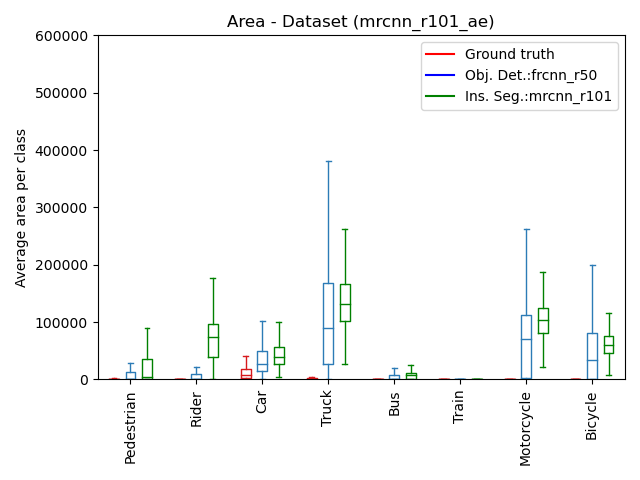}
    \end{subfigure}
    \caption{OD\_frcnn\_r50\_SEG\_mrcnn\_r101}
\end{figure}

\begin{figure}[h!]
    \centering
    \begin{subfigure}{.24\linewidth}
        \centering
        \includegraphics[width=\linewidth]{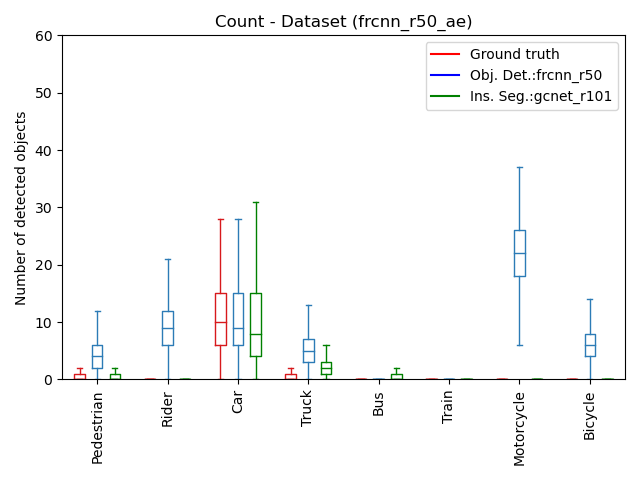}
    \end{subfigure}
    \begin{subfigure}{.24\linewidth}
        \centering
        \includegraphics[width=\linewidth]{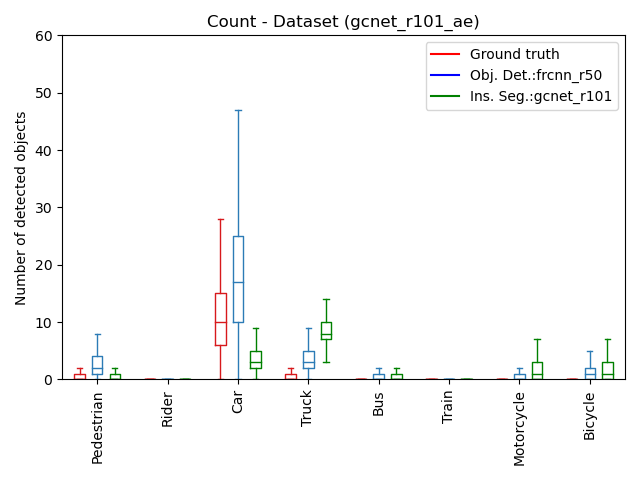}
    \end{subfigure}
    \begin{subfigure}{.24\linewidth}
        \centering
        \includegraphics[width=\linewidth]{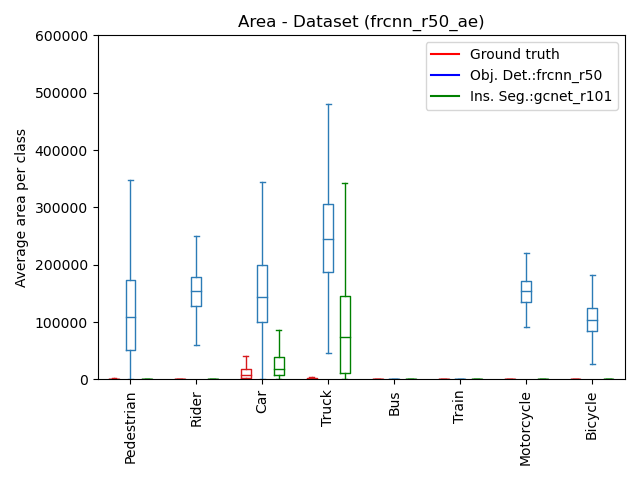}
    \end{subfigure}
    \begin{subfigure}{.24\linewidth}
        \centering
        \includegraphics[width=\linewidth]{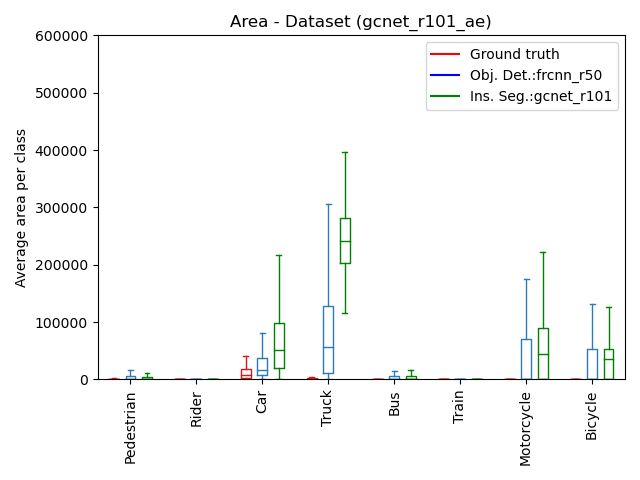}
    \end{subfigure}
    \caption{OD\_frcnn\_r50\_SEG\_gcnet\_r101}
\end{figure}

\begin{figure}[h!]
    \centering
    \begin{subfigure}{.24\linewidth}
        \centering
        \includegraphics[width=\linewidth]{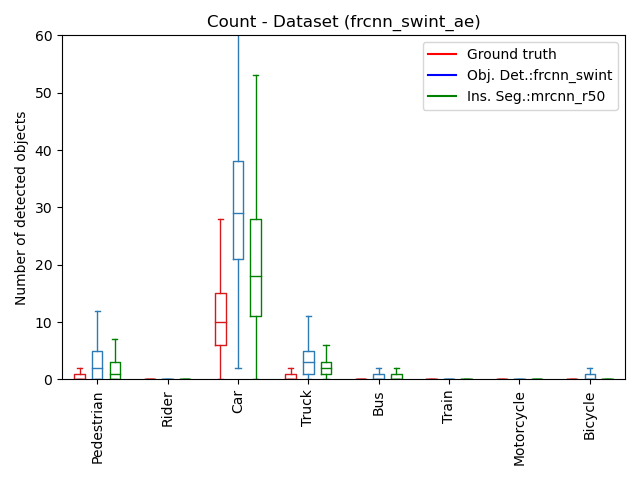}
    \end{subfigure}
    \begin{subfigure}{.24\linewidth}
        \centering
        \includegraphics[width=\linewidth]{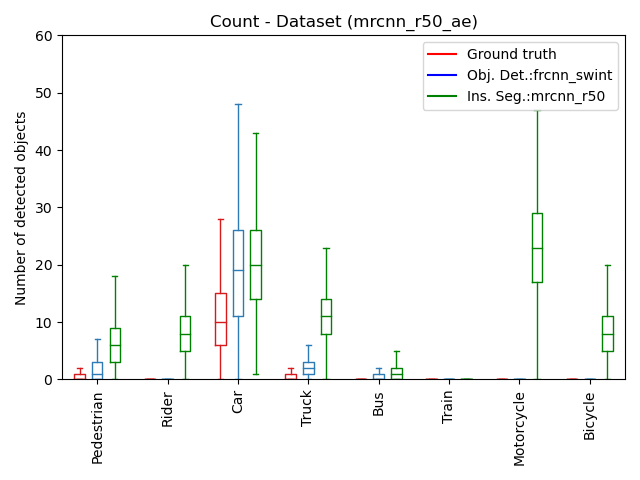}
    \end{subfigure}
    \begin{subfigure}{.24\linewidth}
        \centering
        \includegraphics[width=\linewidth]{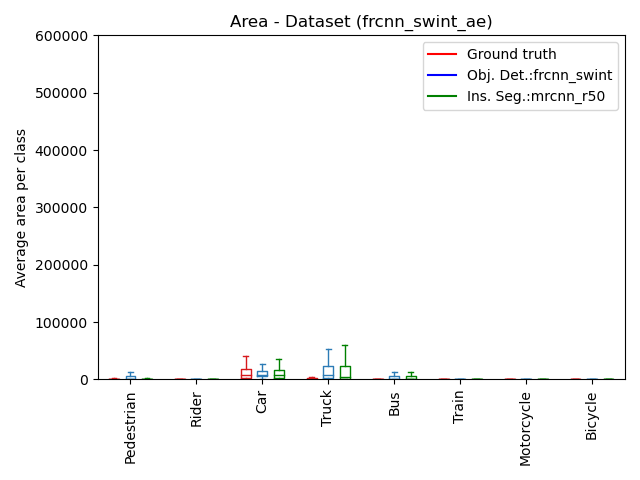}
    \end{subfigure}
    \begin{subfigure}{.24\linewidth}
        \centering
        \includegraphics[width=\linewidth]{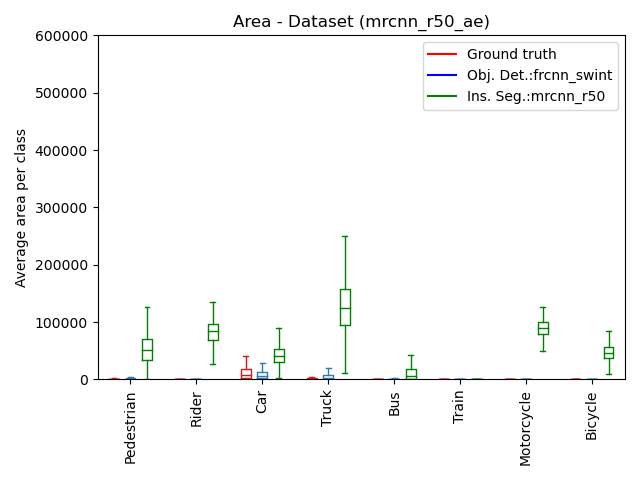}
    \end{subfigure}
    \caption{OD\_frcnn\_swint\_SEG\_mrcnn\_r50}
\end{figure}

\begin{figure}[h!]
    \centering
    \begin{subfigure}{.24\linewidth}
        \centering
        \includegraphics[width=\linewidth]{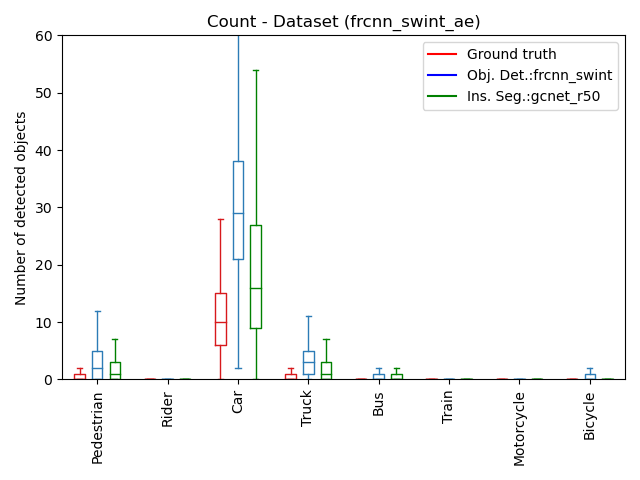}
    \end{subfigure}
    \begin{subfigure}{.24\linewidth}
        \centering
        \includegraphics[width=\linewidth]{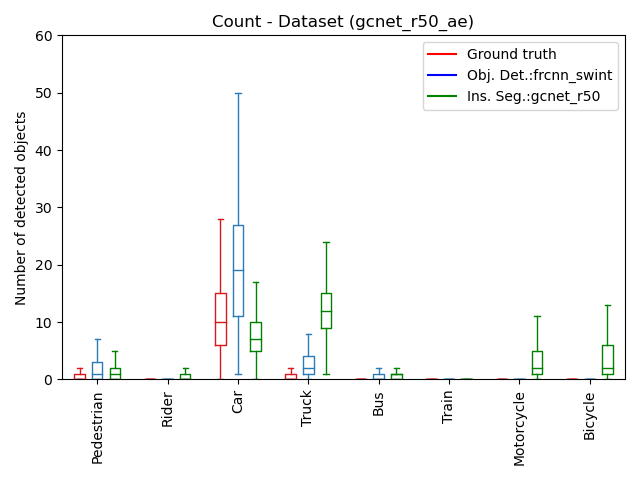}
    \end{subfigure}
    \begin{subfigure}{.24\linewidth}
        \centering
        \includegraphics[width=\linewidth]{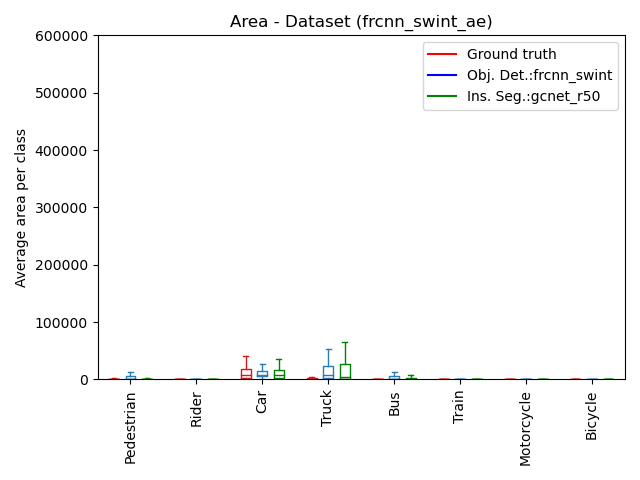}
    \end{subfigure}
    \begin{subfigure}{.24\linewidth}
        \centering
        \includegraphics[width=\linewidth]{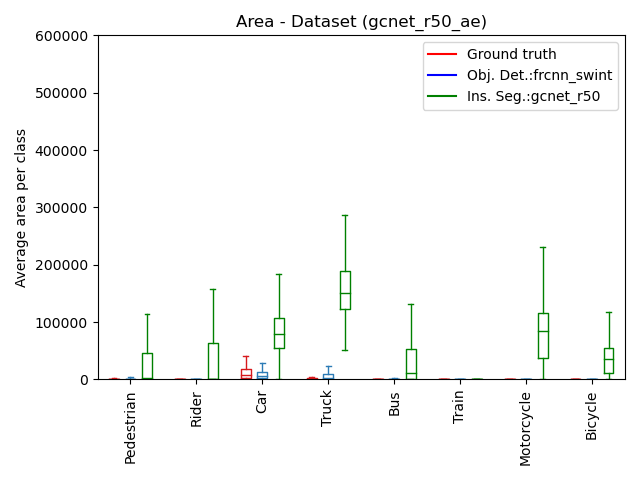}
    \end{subfigure}
    \caption{OD\_frcnn\_swint\_SEG\_gcnet\_r50}
\end{figure}

\begin{figure}[h!]
    \centering
    \begin{subfigure}{.24\linewidth}
        \centering
        \includegraphics[width=\linewidth]{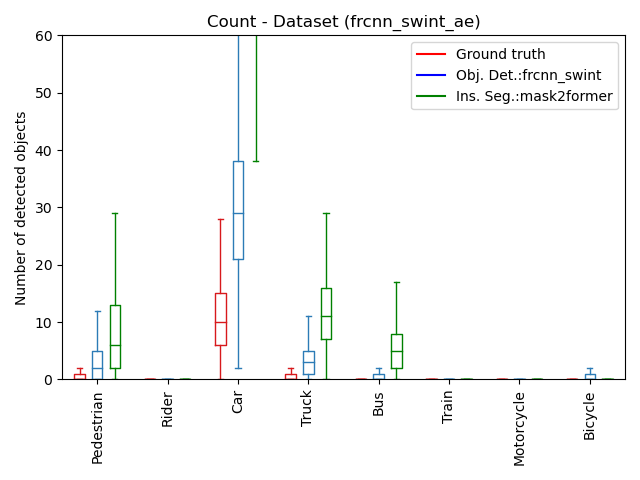}
    \end{subfigure}
    \begin{subfigure}{.24\linewidth}
        \centering
        \includegraphics[width=\linewidth]{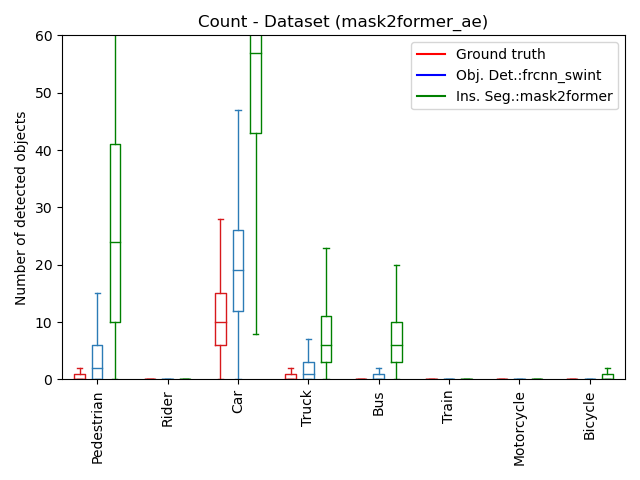}
    \end{subfigure}
    \begin{subfigure}{.24\linewidth}
        \centering
        \includegraphics[width=\linewidth]{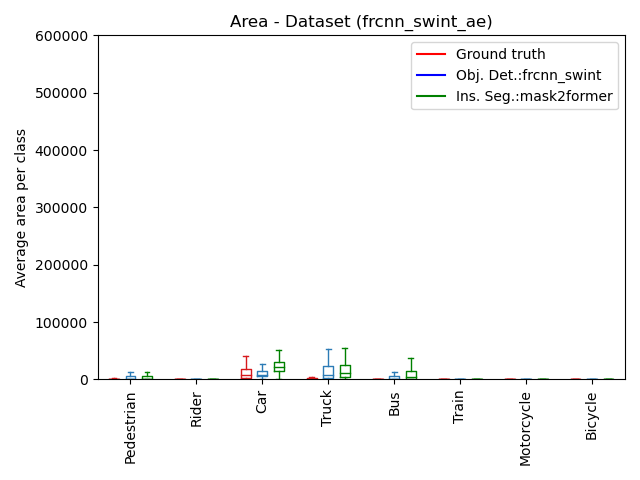}
    \end{subfigure}
    \begin{subfigure}{.24\linewidth}
        \centering
        \includegraphics[width=\linewidth]{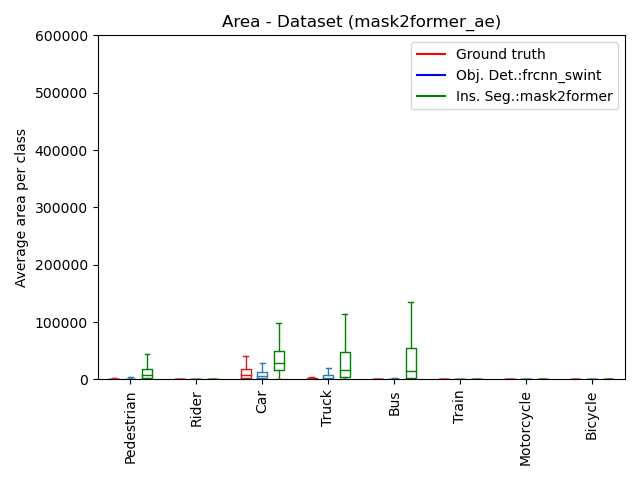}
    \end{subfigure}
    \caption{OD\_frcnn\_swint\_SEG\_mask2former}
\end{figure}

\begin{figure}[h!]
    \centering
    \begin{subfigure}{.24\linewidth}
        \centering
        \includegraphics[width=\linewidth]{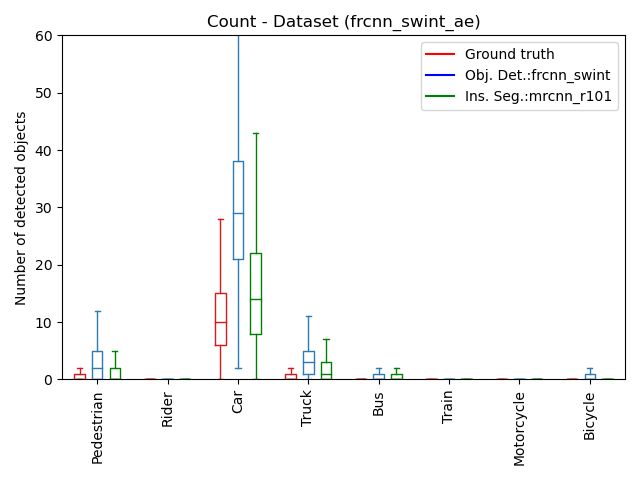}
    \end{subfigure}
    \begin{subfigure}{.24\linewidth}
        \centering
        \includegraphics[width=\linewidth]{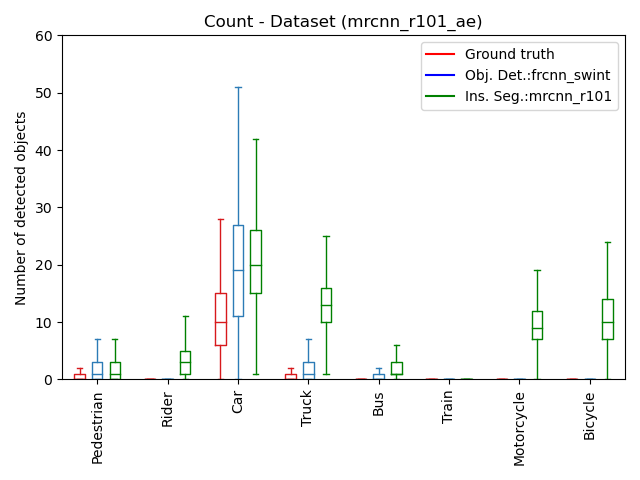}
    \end{subfigure}
    \begin{subfigure}{.24\linewidth}
        \centering
        \includegraphics[width=\linewidth]{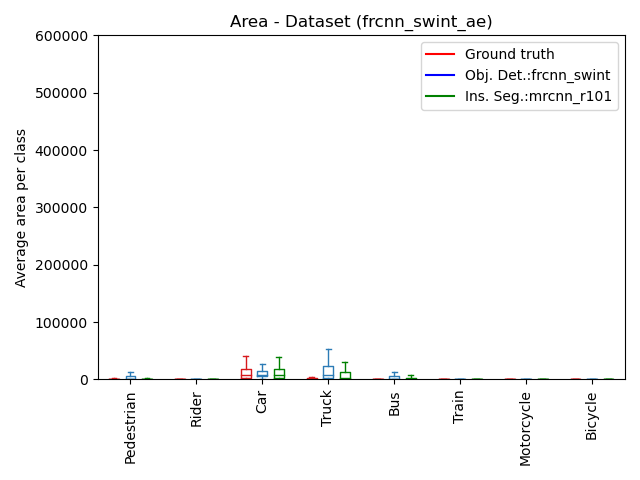}
    \end{subfigure}
    \begin{subfigure}{.24\linewidth}
        \centering
        \includegraphics[width=\linewidth]{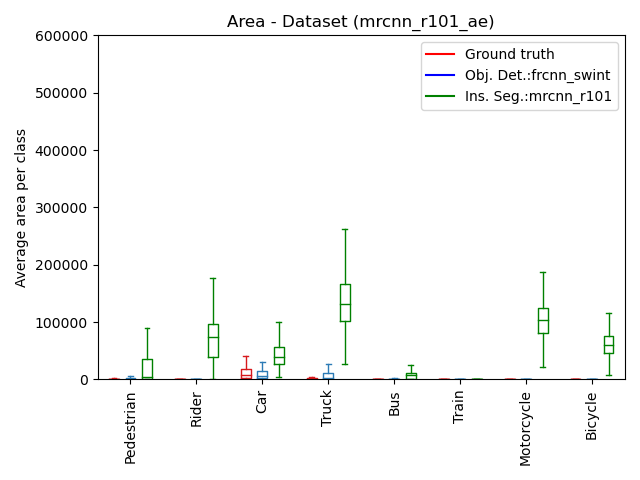}
    \end{subfigure}
    \caption{OD\_frcnn\_swint\_SEG\_mrcnn\_r101}
\end{figure}

\begin{figure}[h!]
    \centering
    \begin{subfigure}{.24\linewidth}
        \centering
        \includegraphics[width=\linewidth]{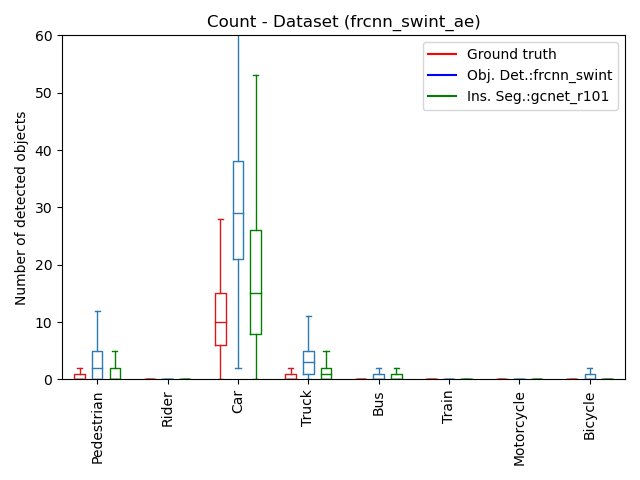}
    \end{subfigure}
    \begin{subfigure}{.24\linewidth}
        \centering
        \includegraphics[width=\linewidth]{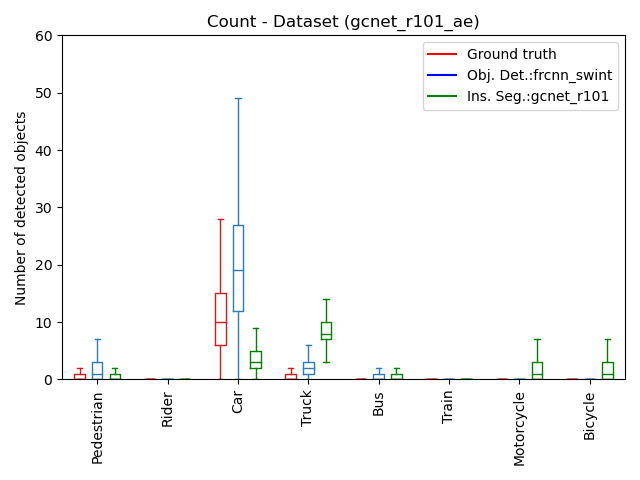}
    \end{subfigure}
    \begin{subfigure}{.24\linewidth}
        \centering
        \includegraphics[width=\linewidth]{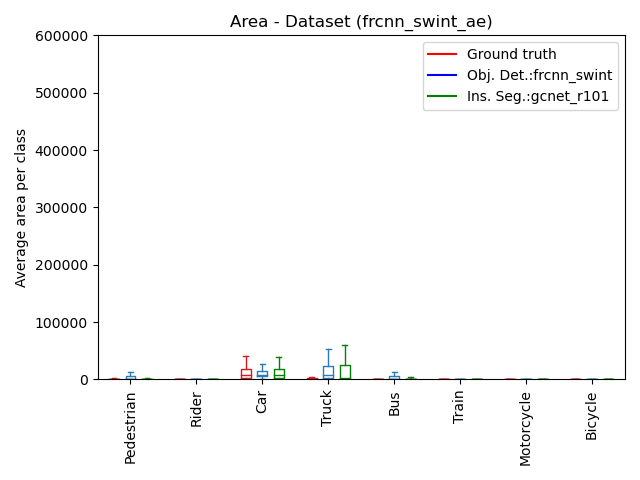}
    \end{subfigure}
    \begin{subfigure}{.24\linewidth}
        \centering
        \includegraphics[width=\linewidth]{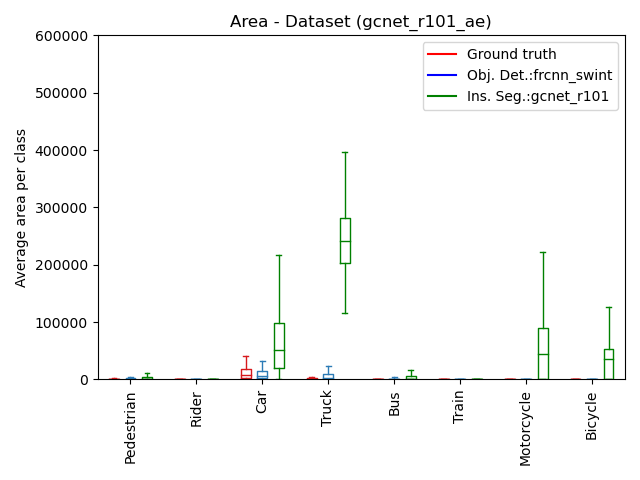}
    \end{subfigure}
    \caption{OD\_frcnn\_swint\_SEG\_gcnet\_r101}
\end{figure}

\begin{figure}[h!]
    \centering
    \begin{subfigure}{.24\linewidth}
        \centering
        \includegraphics[width=\linewidth]{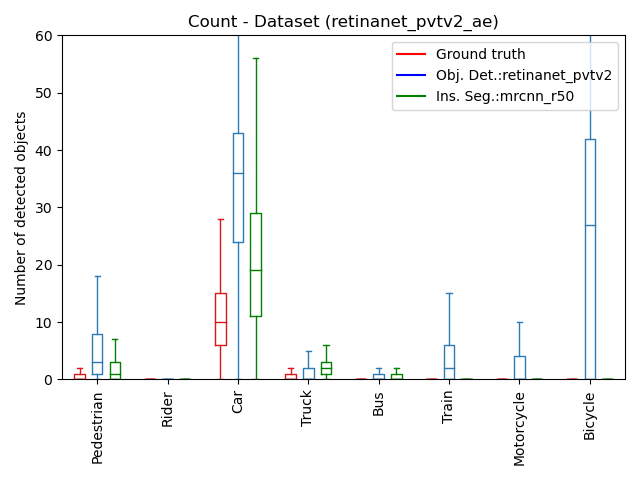}
    \end{subfigure}
    \begin{subfigure}{.24\linewidth}
        \centering
        \includegraphics[width=\linewidth]{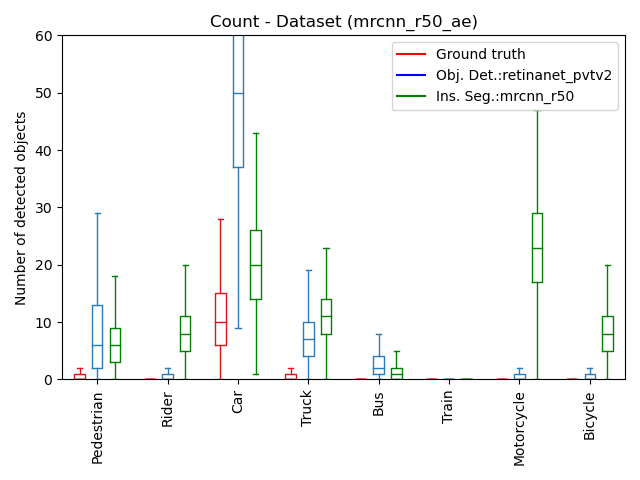}
    \end{subfigure}
    \begin{subfigure}{.24\linewidth}
        \centering
        \includegraphics[width=\linewidth]{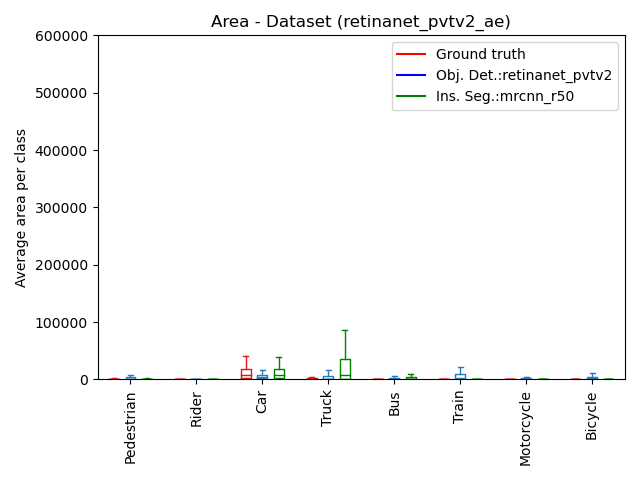}
    \end{subfigure}
    \begin{subfigure}{.24\linewidth}
        \centering
        \includegraphics[width=\linewidth]{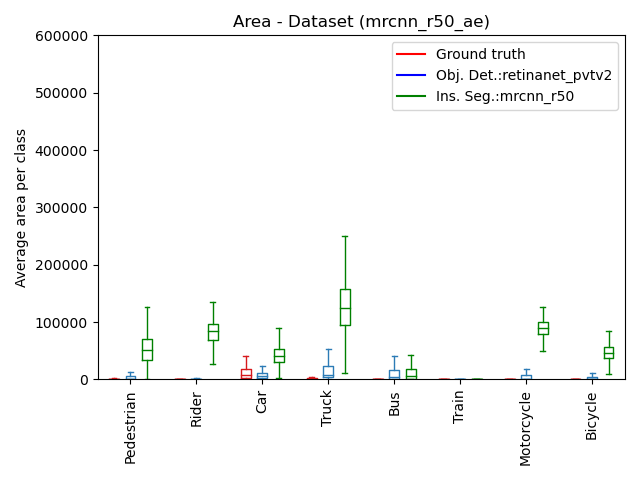}
    \end{subfigure}
    \caption{OD\_retinanet\_pvtv2\_SEG\_mrcnn\_r50}
\end{figure}

\begin{figure}[h!]
    \centering
    \begin{subfigure}{.24\linewidth}
        \centering
        \includegraphics[width=\linewidth]{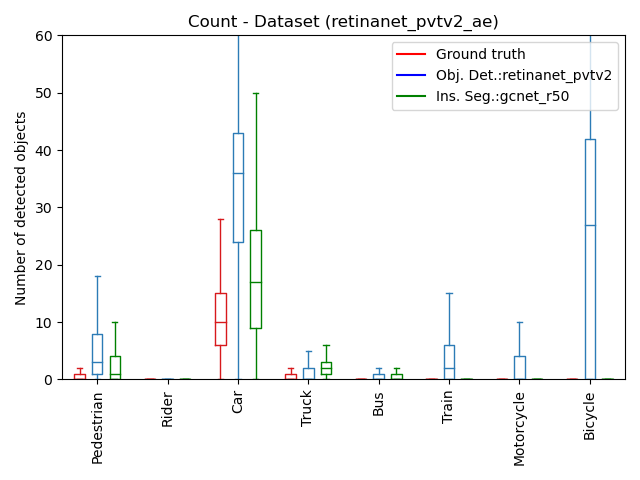}
    \end{subfigure}
    \begin{subfigure}{.24\linewidth}
        \centering
        \includegraphics[width=\linewidth]{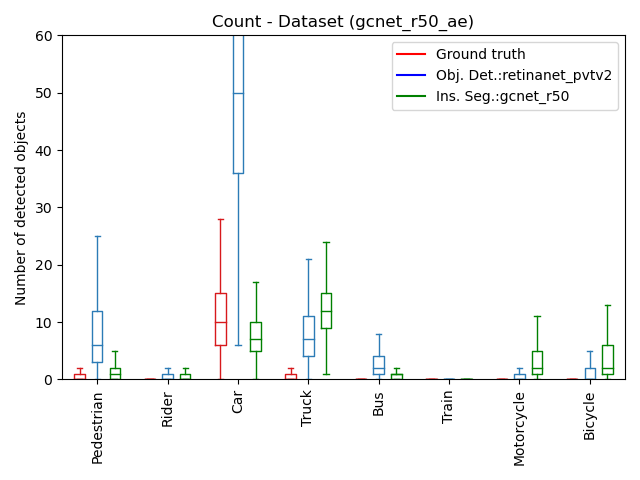}
    \end{subfigure}
    \begin{subfigure}{.24\linewidth}
        \centering
        \includegraphics[width=\linewidth]{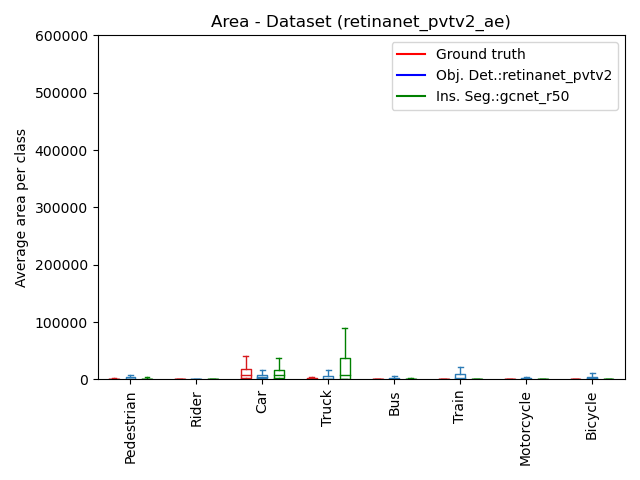}
    \end{subfigure}
    \begin{subfigure}{.24\linewidth}
        \centering
        \includegraphics[width=\linewidth]{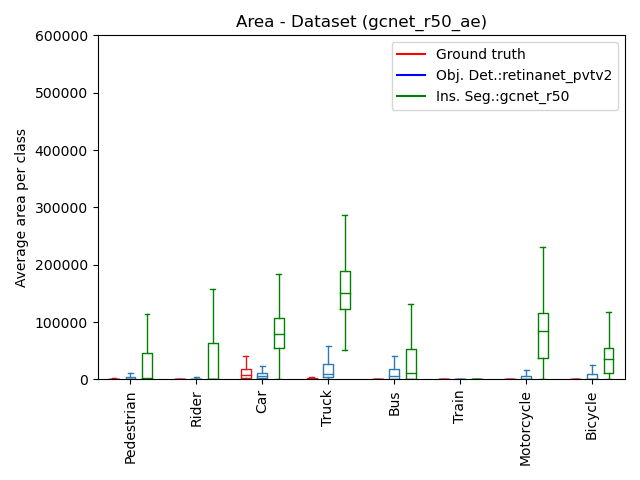}
    \end{subfigure}
    \caption{OD\_retinanet\_pvtv2\_SEG\_gcnet\_r50}
\end{figure}

\begin{figure}[h!]
    \centering
    \begin{subfigure}{.24\linewidth}
        \centering
        \includegraphics[width=\linewidth]{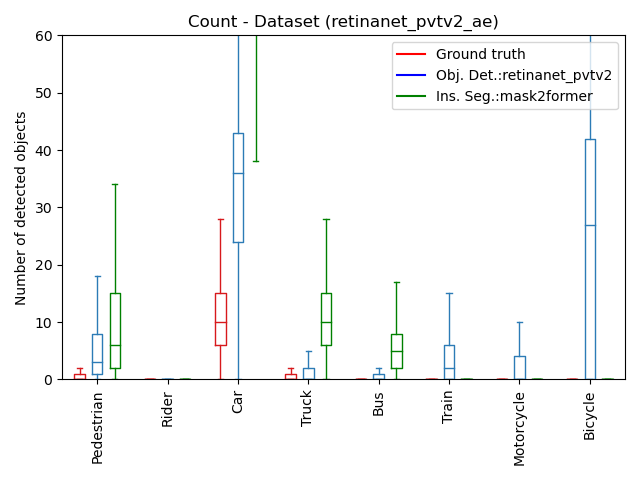}
    \end{subfigure}
    \begin{subfigure}{.24\linewidth}
        \centering
        \includegraphics[width=\linewidth]{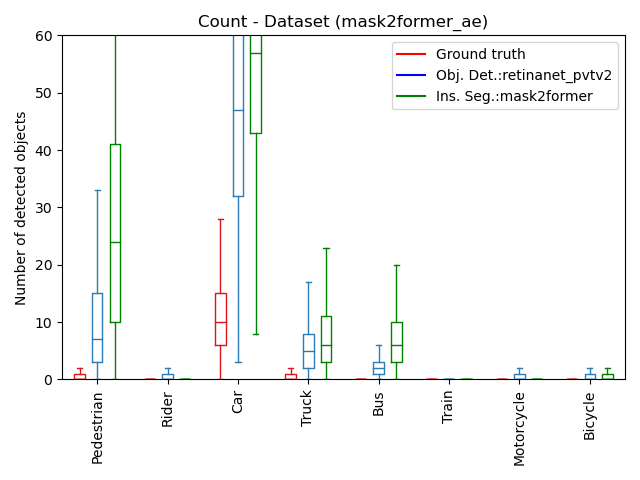}
    \end{subfigure}
    \begin{subfigure}{.24\linewidth}
        \centering
        \includegraphics[width=\linewidth]{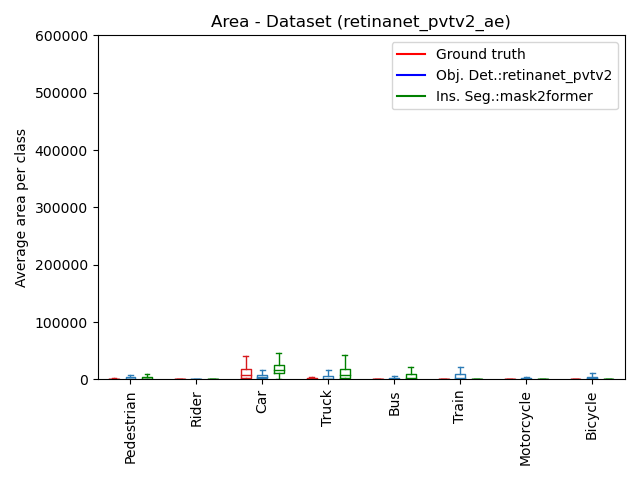}
    \end{subfigure}
    \begin{subfigure}{.24\linewidth}
        \centering
        \includegraphics[width=\linewidth]{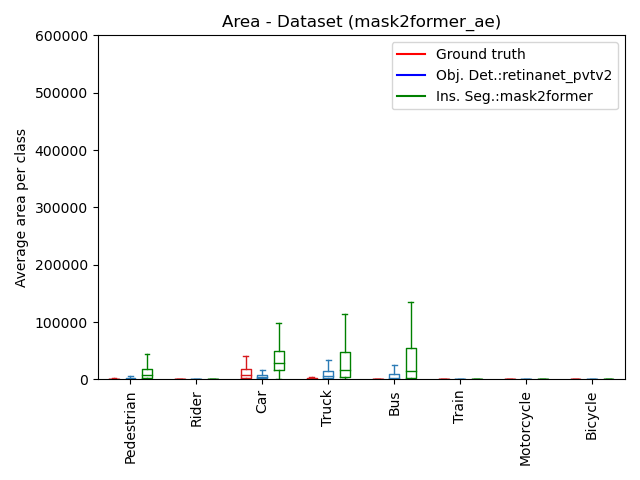}
    \end{subfigure}
    \caption{OD\_retinanet\_pvtv2\_SEG\_mask2former}
\end{figure}

\begin{figure}[h!]
    \centering
    \begin{subfigure}{.24\linewidth}
        \centering
        \includegraphics[width=\linewidth]{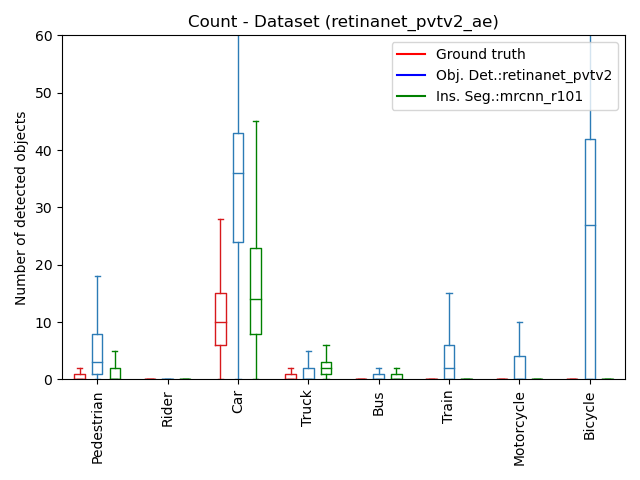}
    \end{subfigure}
    \begin{subfigure}{.24\linewidth}
        \centering
        \includegraphics[width=\linewidth]{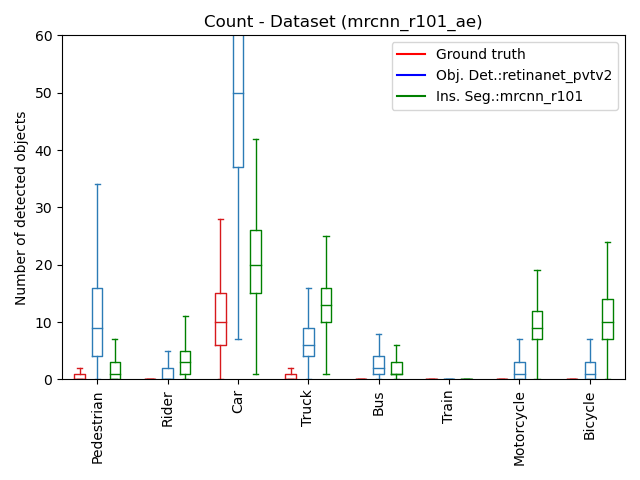}
    \end{subfigure}
    \begin{subfigure}{.24\linewidth}
        \centering
        \includegraphics[width=\linewidth]{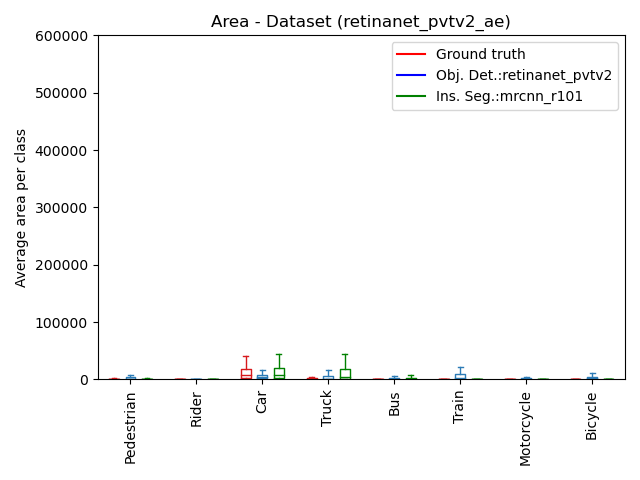}
    \end{subfigure}
    \begin{subfigure}{.24\linewidth}
        \centering
        \includegraphics[width=\linewidth]{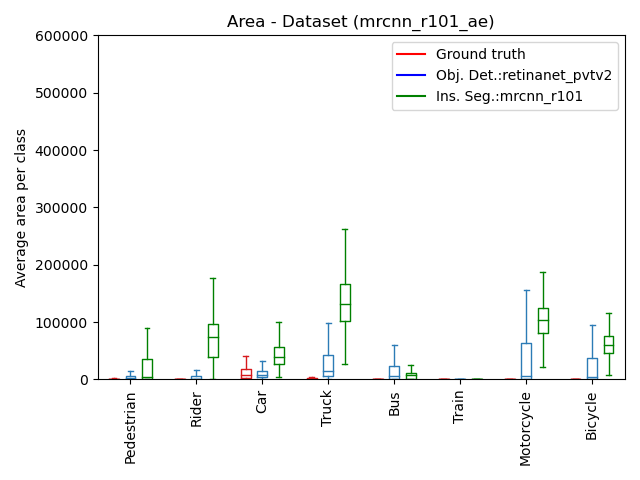}
    \end{subfigure}
    \caption{OD\_retinanet\_pvtv2\_SEG\_mrcnn\_r101}
\end{figure}

\begin{figure}[h!]
    \centering
    \begin{subfigure}{.24\linewidth}
        \centering
        \includegraphics[width=\linewidth]{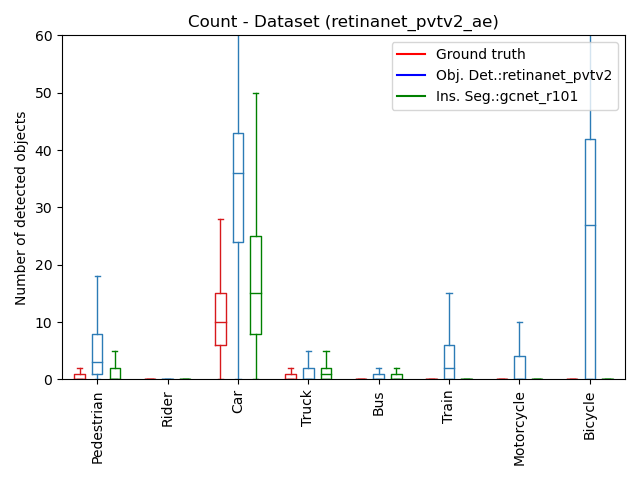}
    \end{subfigure}
    \begin{subfigure}{.24\linewidth}
        \centering
        \includegraphics[width=\linewidth]{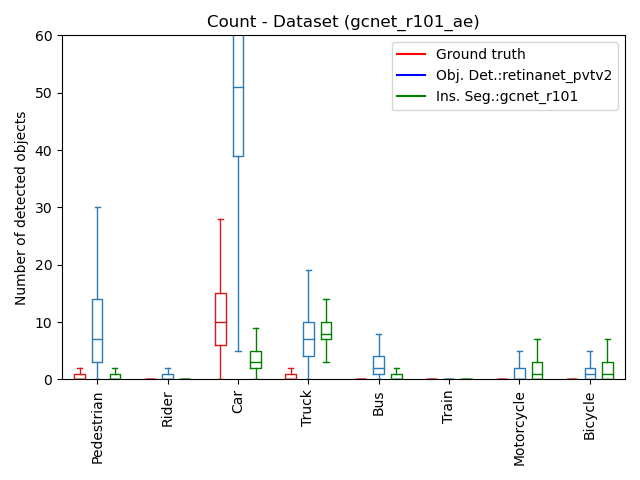}
    \end{subfigure}
    \begin{subfigure}{.24\linewidth}
        \centering
        \includegraphics[width=\linewidth]{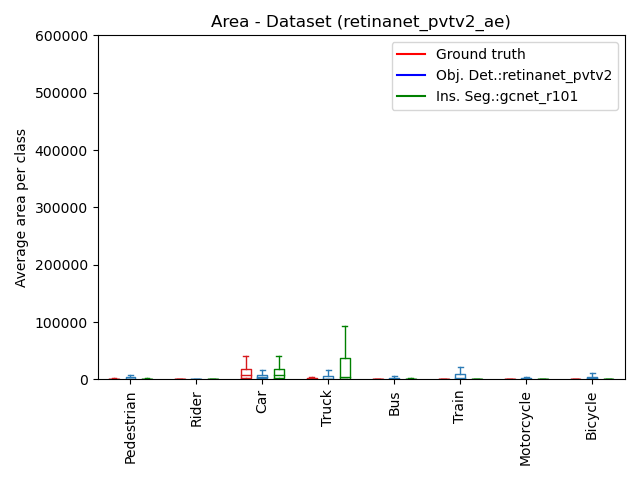}
    \end{subfigure}
    \begin{subfigure}{.24\linewidth}
        \centering
        \includegraphics[width=\linewidth]{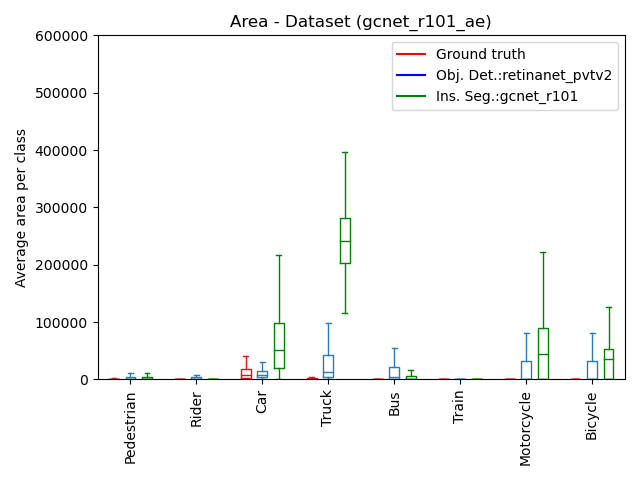}
    \end{subfigure}
    \caption{OD\_retinanet\_pvtv2\_SEG\_gcnet\_r101}
\end{figure}

\begin{figure}[h!]
    \centering
    \begin{subfigure}{.24\linewidth}
        \centering
        \includegraphics[width=\linewidth]{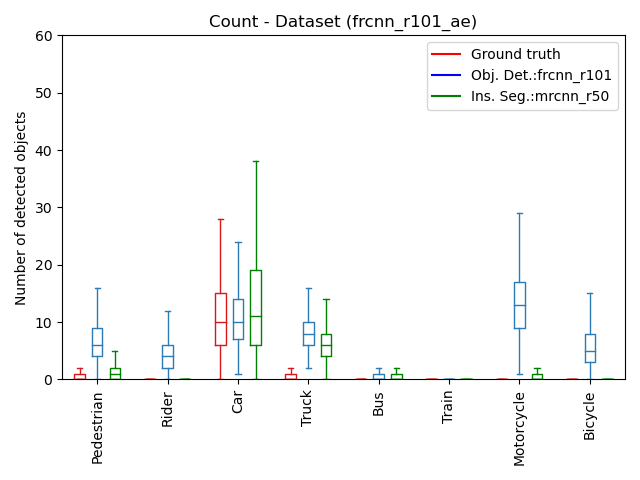}
    \end{subfigure}
    \begin{subfigure}{.24\linewidth}
        \centering
        \includegraphics[width=\linewidth]{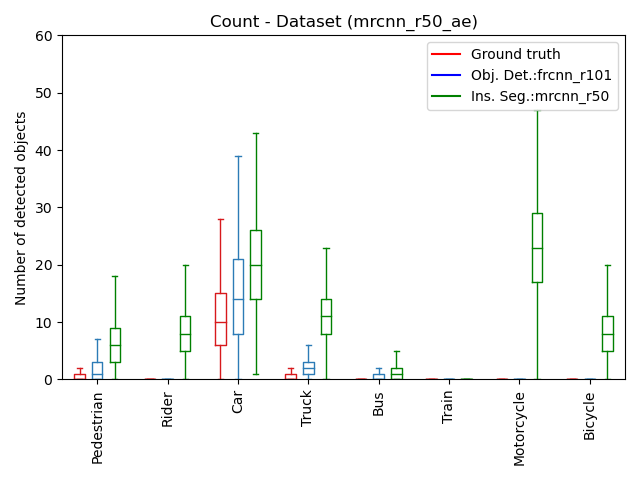}
    \end{subfigure}
    \begin{subfigure}{.24\linewidth}
        \centering
        \includegraphics[width=\linewidth]{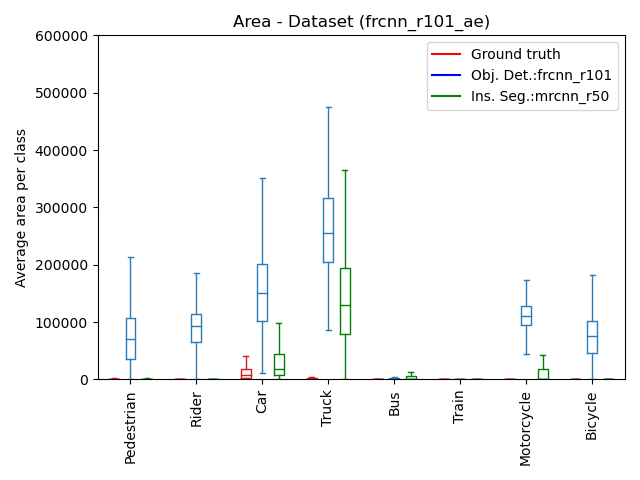}
    \end{subfigure}
    \begin{subfigure}{.24\linewidth}
        \centering
        \includegraphics[width=\linewidth]{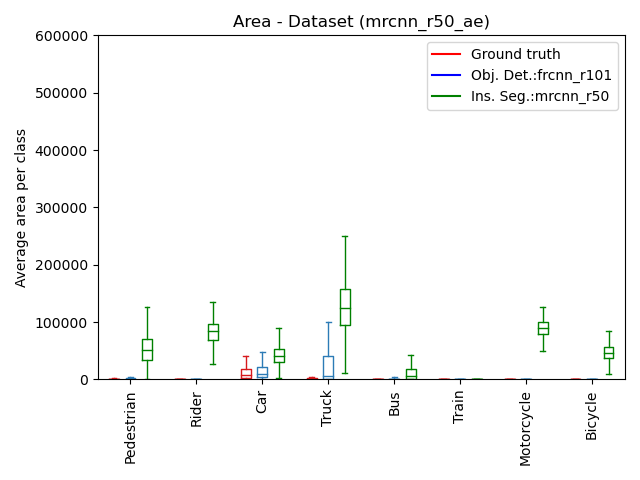}
    \end{subfigure}
    \caption{OD\_frcnn\_r101\_SEG\_mrcnn\_r50}
\end{figure}

\begin{figure}[h!]
    \centering
    \begin{subfigure}{.24\linewidth}
        \centering
        \includegraphics[width=\linewidth]{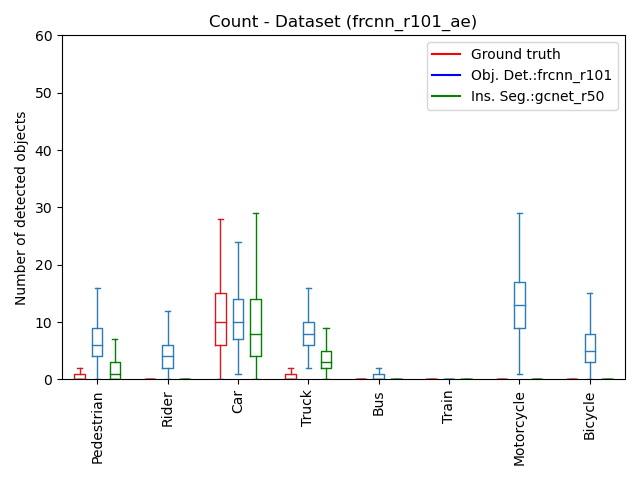}
    \end{subfigure}
    \begin{subfigure}{.24\linewidth}
        \centering
        \includegraphics[width=\linewidth]{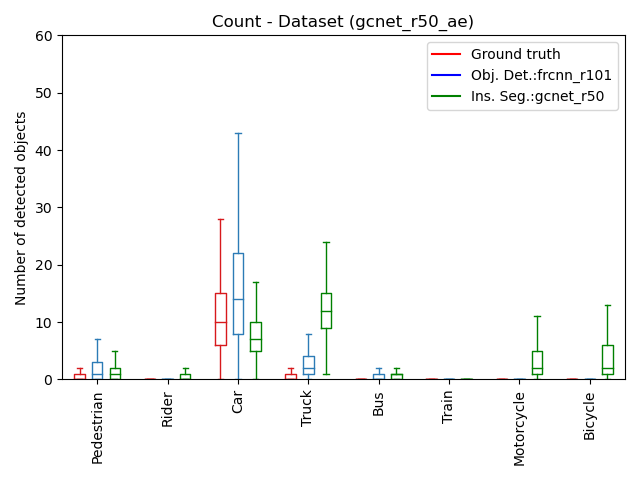}
    \end{subfigure}
    \begin{subfigure}{.24\linewidth}
        \centering
        \includegraphics[width=\linewidth]{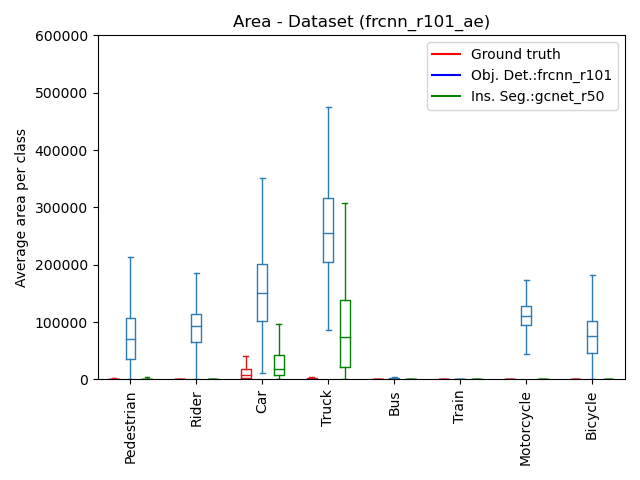}
    \end{subfigure}
    \begin{subfigure}{.24\linewidth}
        \centering
        \includegraphics[width=\linewidth]{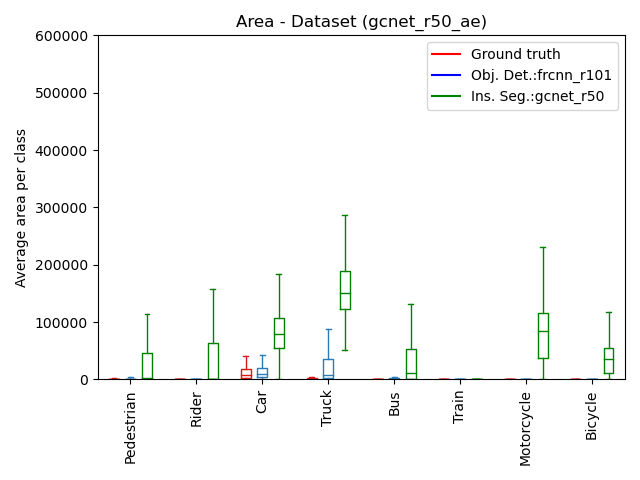}
    \end{subfigure}
    \caption{OD\_frcnn\_r101\_SEG\_gcnet\_r50}
\end{figure}

\begin{figure}[h!]
    \centering
    \begin{subfigure}{.24\linewidth}
        \centering
        \includegraphics[width=\linewidth]{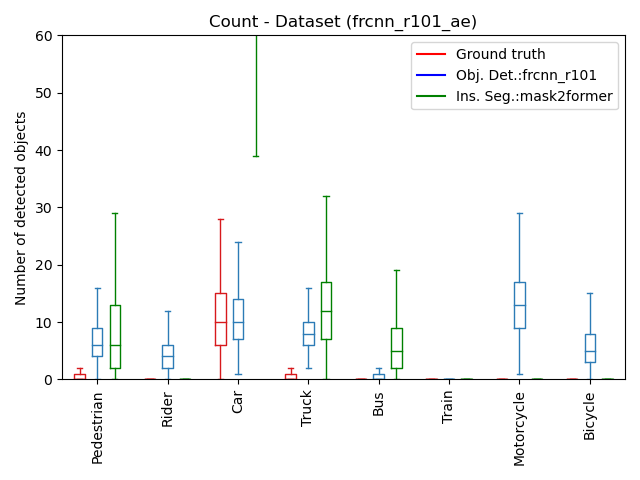}
    \end{subfigure}
    \begin{subfigure}{.24\linewidth}
        \centering
        \includegraphics[width=\linewidth]{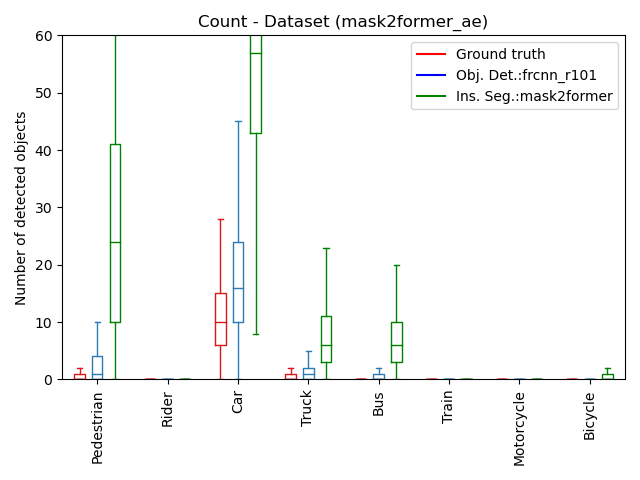}
    \end{subfigure}
    \begin{subfigure}{.24\linewidth}
        \centering
        \includegraphics[width=\linewidth]{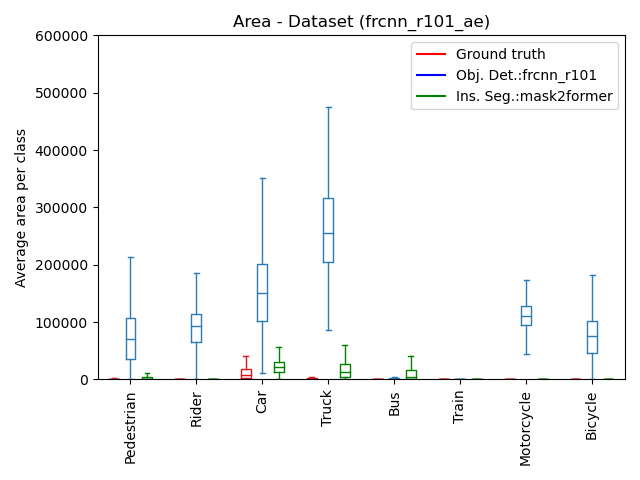}
    \end{subfigure}
    \begin{subfigure}{.24\linewidth}
        \centering
        \includegraphics[width=\linewidth]{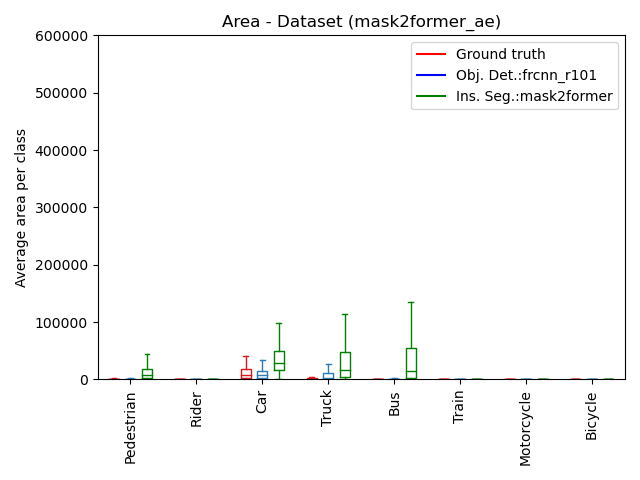}
    \end{subfigure}
    \caption{OD\_frcnn\_r101\_SEG\_mask2former}
\end{figure}

\begin{figure}[h!]
    \centering
    \begin{subfigure}{.24\linewidth}
        \centering
        \includegraphics[width=\linewidth]{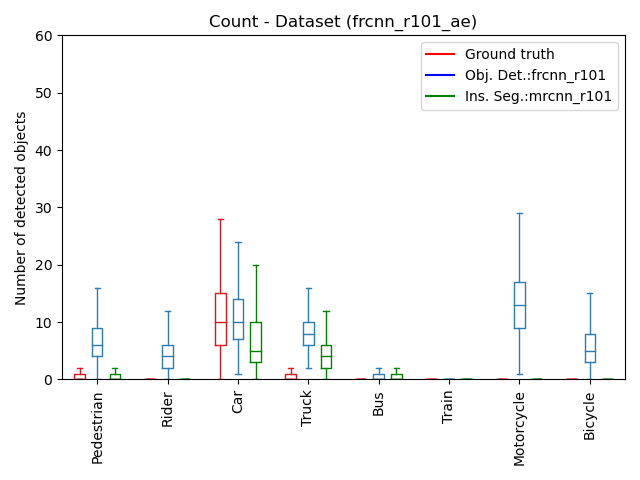}
    \end{subfigure}
    \begin{subfigure}{.24\linewidth}
        \centering
        \includegraphics[width=\linewidth]{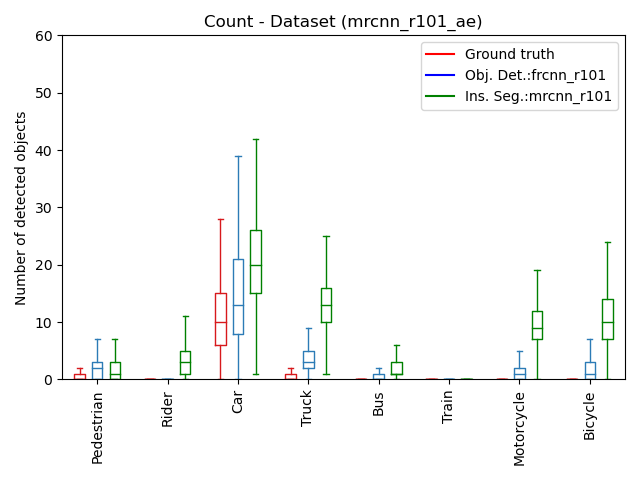}
    \end{subfigure}
    \begin{subfigure}{.24\linewidth}
        \centering
        \includegraphics[width=\linewidth]{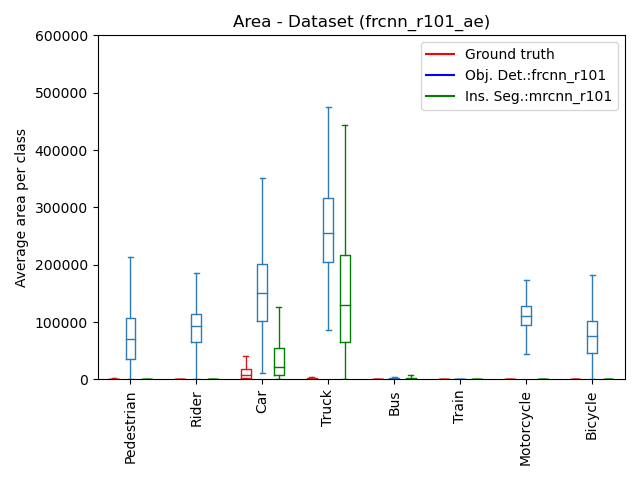}
    \end{subfigure}
    \begin{subfigure}{.24\linewidth}
        \centering
        \includegraphics[width=\linewidth]{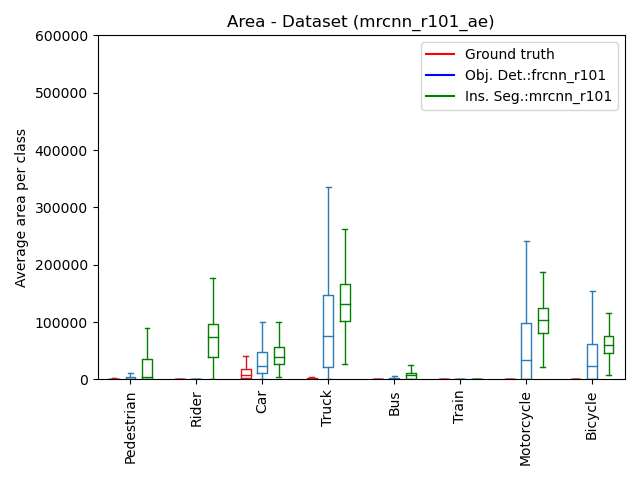}
    \end{subfigure}
    \caption{OD\_frcnn\_r101\_SEG\_mrcnn\_r101}
\end{figure}

\begin{figure}[h!]
    \centering
    \begin{subfigure}{.24\linewidth}
        \centering
        \includegraphics[width=\linewidth]{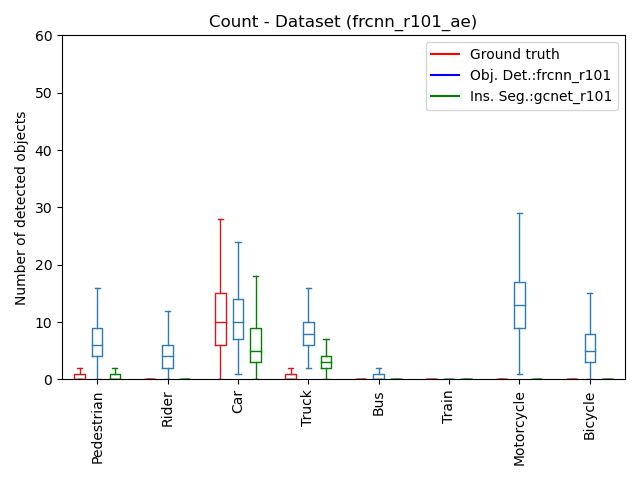}
    \end{subfigure}
    \begin{subfigure}{.24\linewidth}
        \centering
        \includegraphics[width=\linewidth]{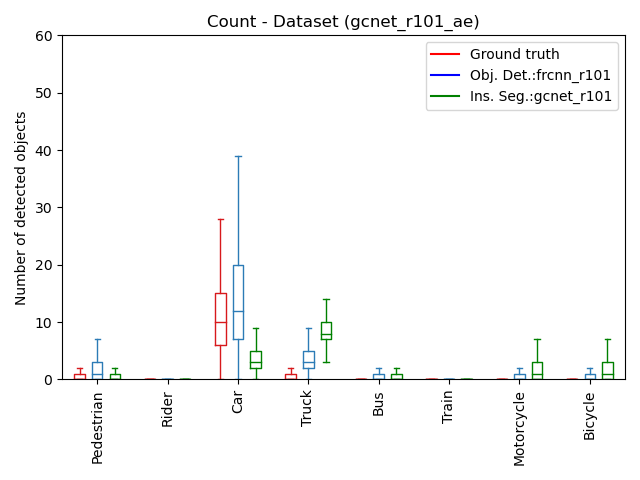}
    \end{subfigure}
    \begin{subfigure}{.24\linewidth}
        \centering
        \includegraphics[width=\linewidth]{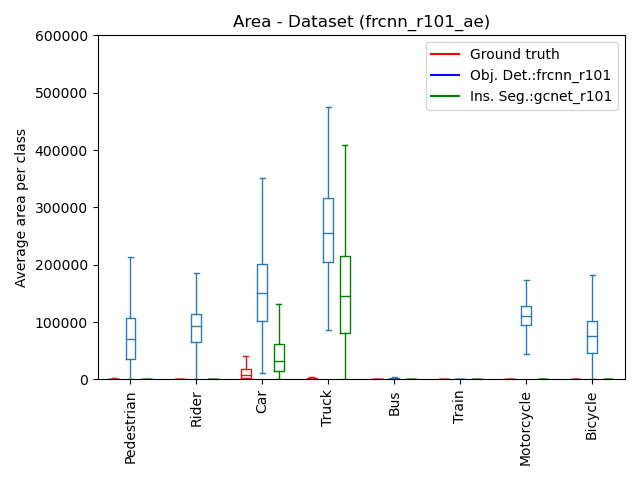}
    \end{subfigure}
    \begin{subfigure}{.24\linewidth}
        \centering
        \includegraphics[width=\linewidth]{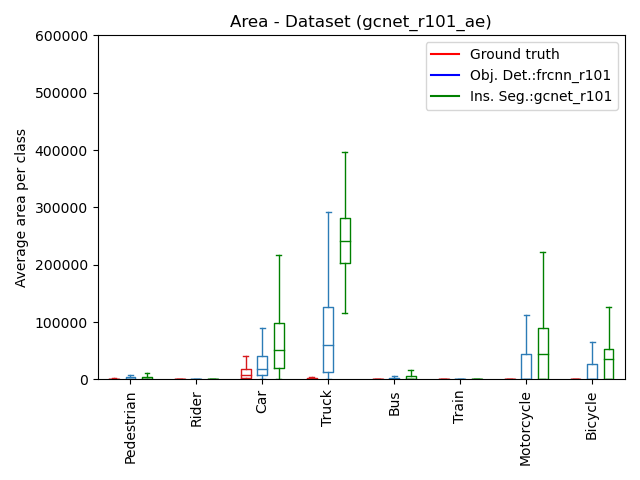}
    \end{subfigure}
    \caption{OD\_frcnn\_r101\_SEG\_gcnet\_r101}
\end{figure}

\begin{figure}[h!]
    \centering
    \begin{subfigure}{.24\linewidth}
        \centering
        \includegraphics[width=\linewidth]{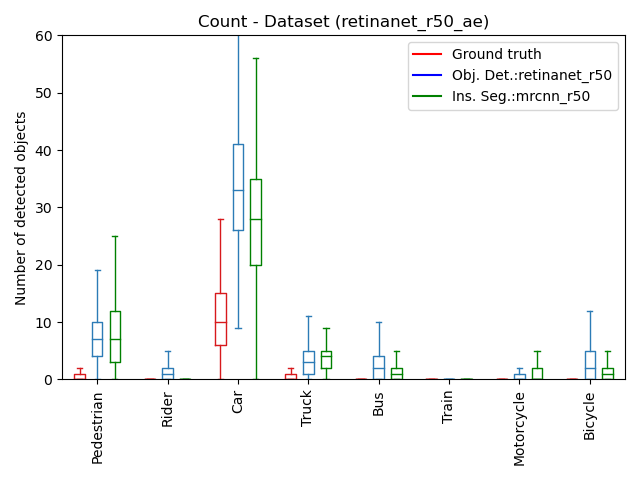}
    \end{subfigure}
    \begin{subfigure}{.24\linewidth}
        \centering
        \includegraphics[width=\linewidth]{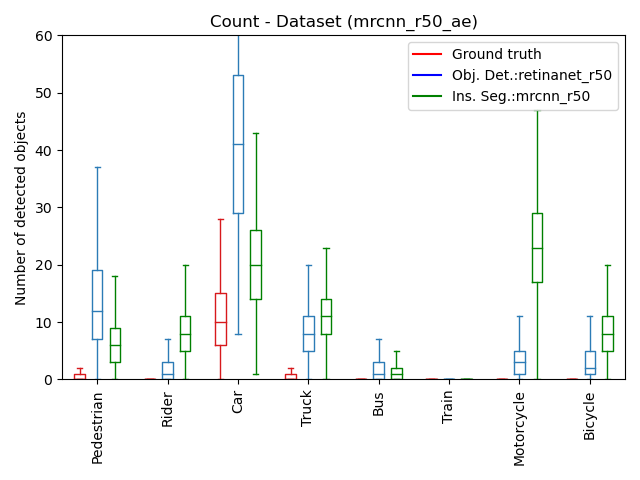}
    \end{subfigure}
    \begin{subfigure}{.24\linewidth}
        \centering
        \includegraphics[width=\linewidth]{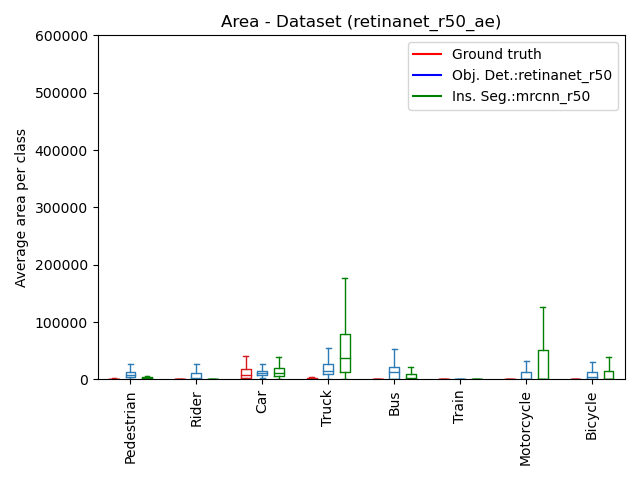}
    \end{subfigure}
    \begin{subfigure}{.24\linewidth}
        \centering
        \includegraphics[width=\linewidth]{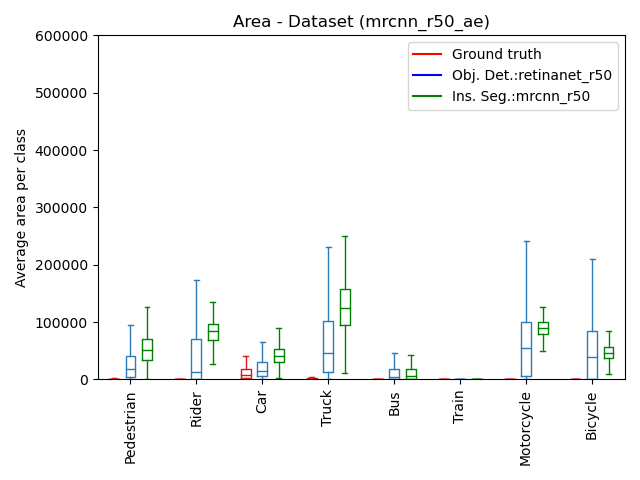}
    \end{subfigure}
    \caption{OD\_retinanet\_r50\_SEG\_mrcnn\_r50}
\end{figure}

\begin{figure}[h!]
    \centering
    \begin{subfigure}{.24\linewidth}
        \centering
        \includegraphics[width=\linewidth]{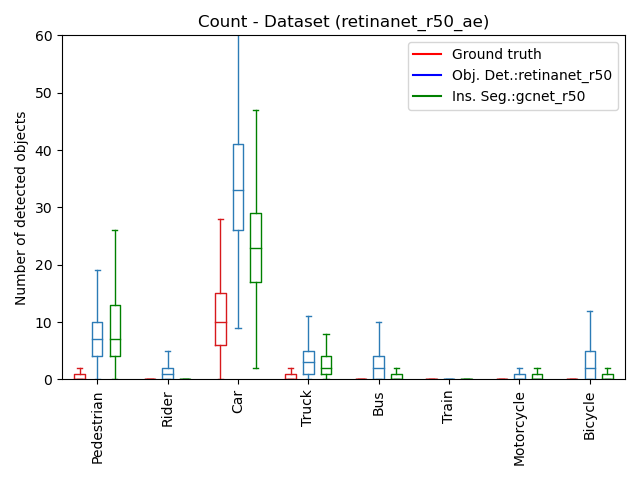}
    \end{subfigure}
    \begin{subfigure}{.24\linewidth}
        \centering
        \includegraphics[width=\linewidth]{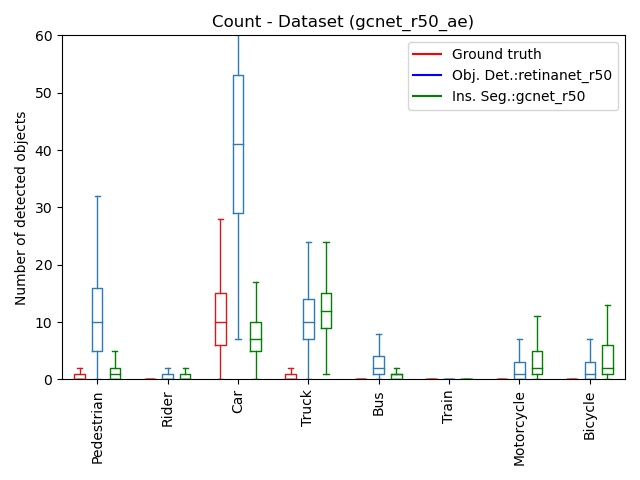}
    \end{subfigure}
    \begin{subfigure}{.24\linewidth}
        \centering
        \includegraphics[width=\linewidth]{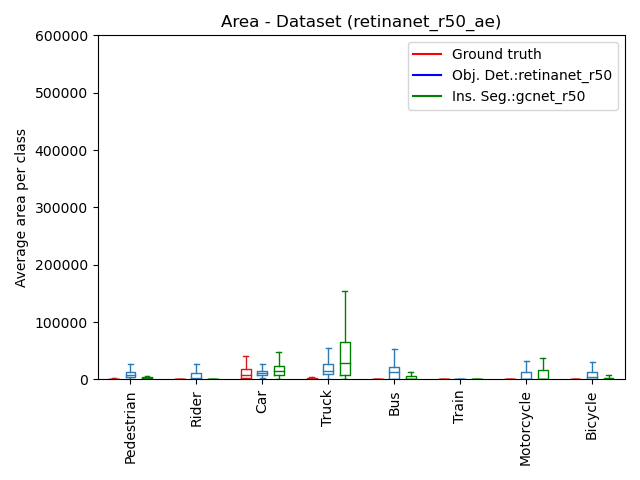}
    \end{subfigure}
    \begin{subfigure}{.24\linewidth}
        \centering
        \includegraphics[width=\linewidth]{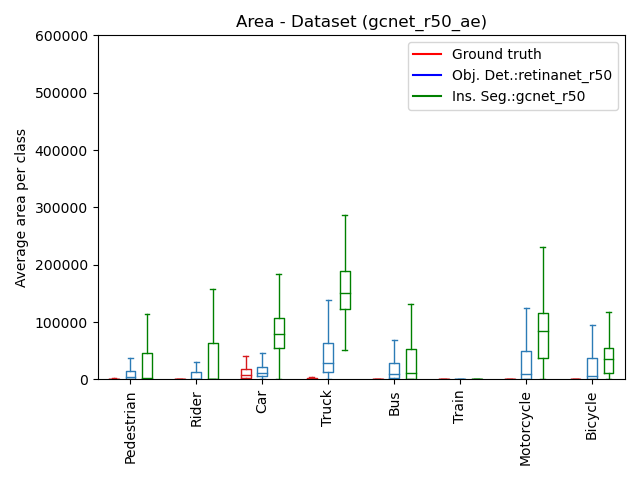}
    \end{subfigure}
    \caption{OD\_retinanet\_r50\_SEG\_gcnet\_r50}
\end{figure}

\begin{figure}[h!]
    \centering
    \begin{subfigure}{.24\linewidth}
        \centering
        \includegraphics[width=\linewidth]{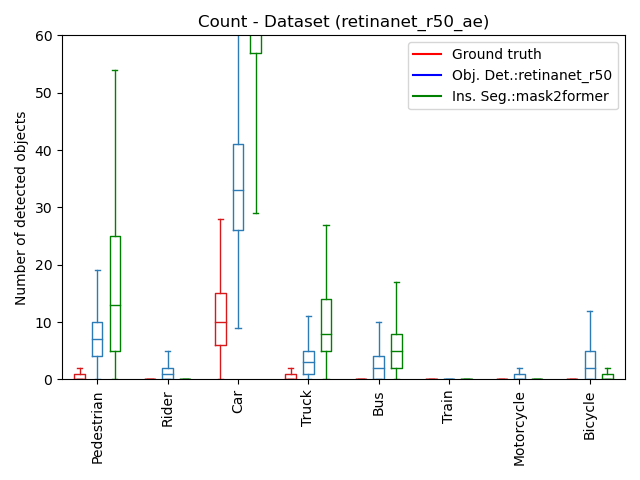}
    \end{subfigure}
    \begin{subfigure}{.24\linewidth}
        \centering
        \includegraphics[width=\linewidth]{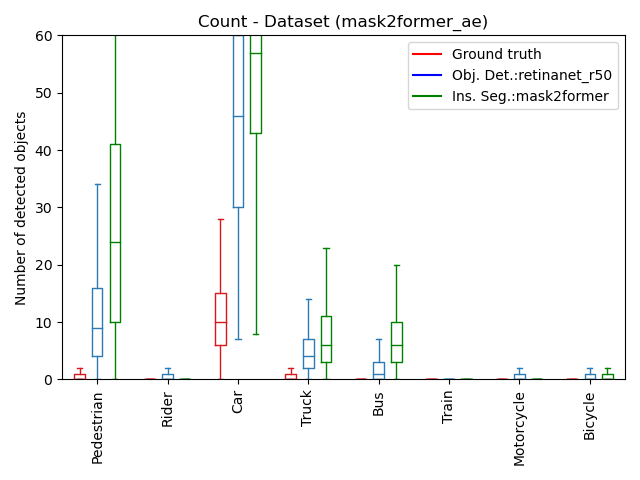}
    \end{subfigure}
    \begin{subfigure}{.24\linewidth}
        \centering
        \includegraphics[width=\linewidth]{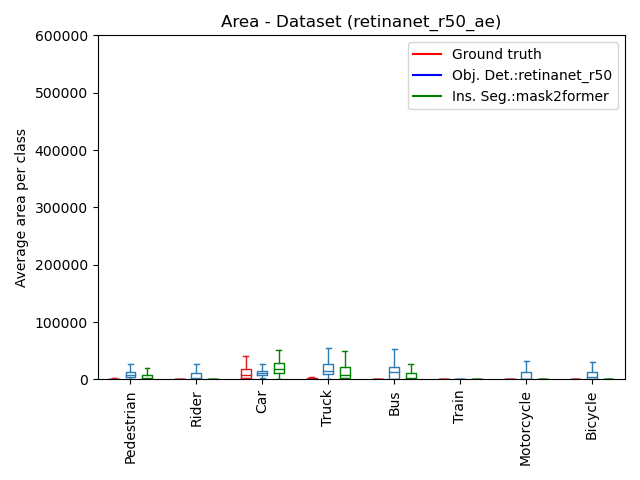}
    \end{subfigure}
    \begin{subfigure}{.24\linewidth}
        \centering
        \includegraphics[width=\linewidth]{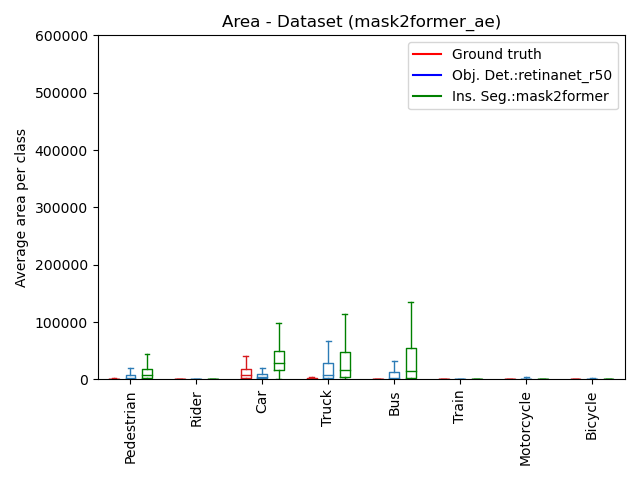}
    \end{subfigure}
    \caption{OD\_retinanet\_r50\_SEG\_mask2former}
\end{figure}

\begin{figure}[h!]
    \centering
    \begin{subfigure}{.24\linewidth}
        \centering
        \includegraphics[width=\linewidth]{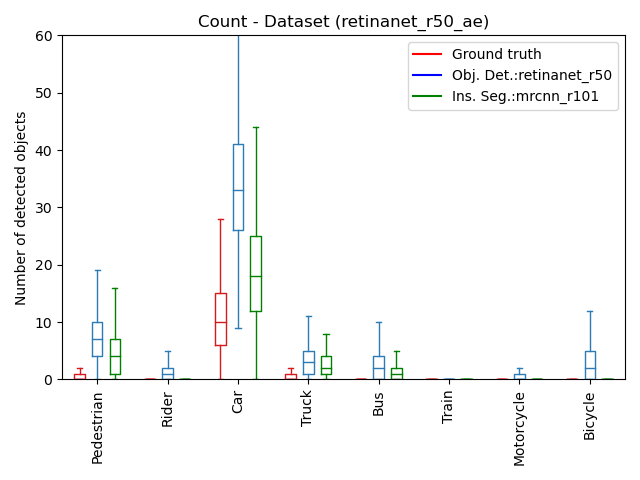}
    \end{subfigure}
    \begin{subfigure}{.24\linewidth}
        \centering
        \includegraphics[width=\linewidth]{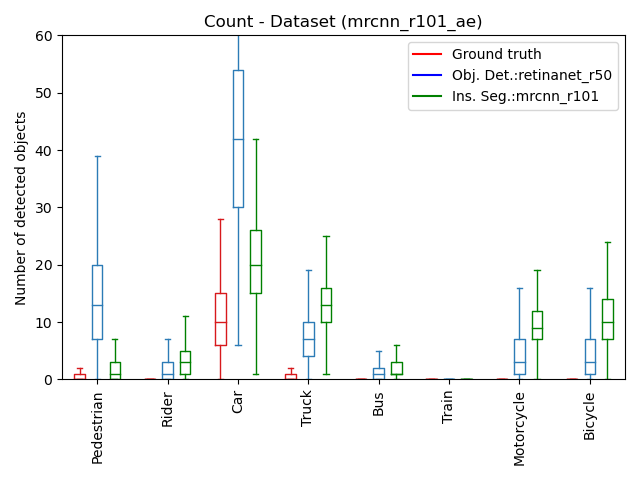}
    \end{subfigure}
    \begin{subfigure}{.24\linewidth}
        \centering
        \includegraphics[width=\linewidth]{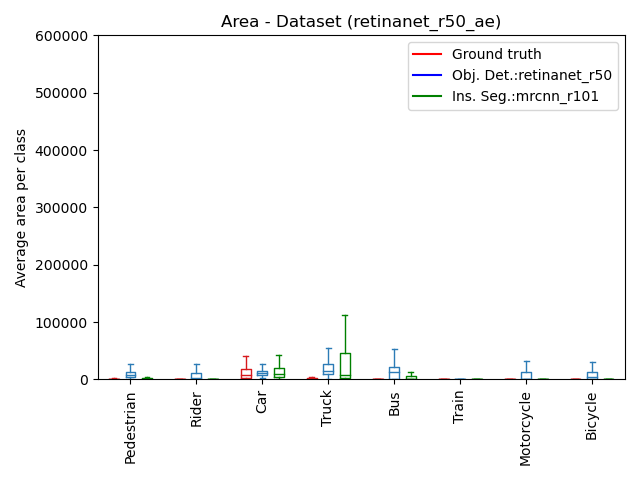}
    \end{subfigure}
    \begin{subfigure}{.24\linewidth}
        \centering
        \includegraphics[width=\linewidth]{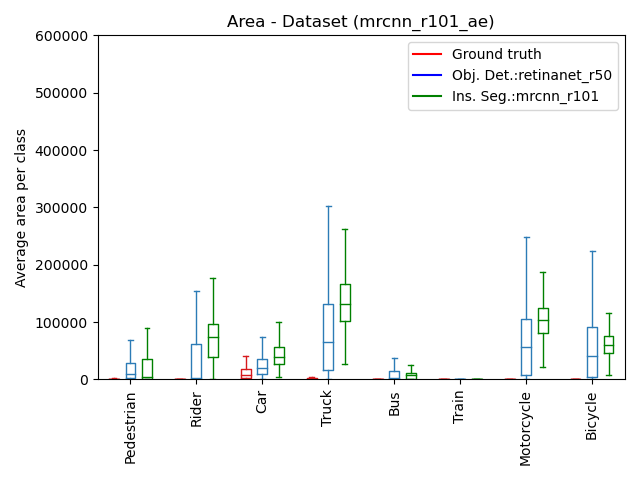}
    \end{subfigure}
    \caption{OD\_retinanet\_r50\_SEG\_mrcnn\_r101}
\end{figure}

\begin{figure}[h!]
    \centering
    \begin{subfigure}{.24\linewidth}
        \centering
        \includegraphics[width=\linewidth]{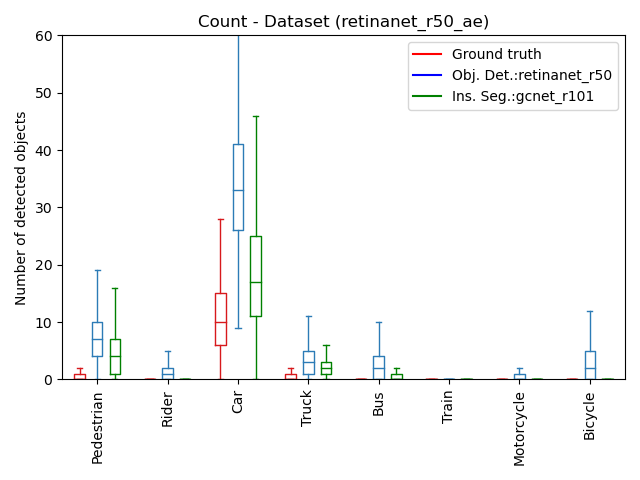}
    \end{subfigure}
    \begin{subfigure}{.24\linewidth}
        \centering
        \includegraphics[width=\linewidth]{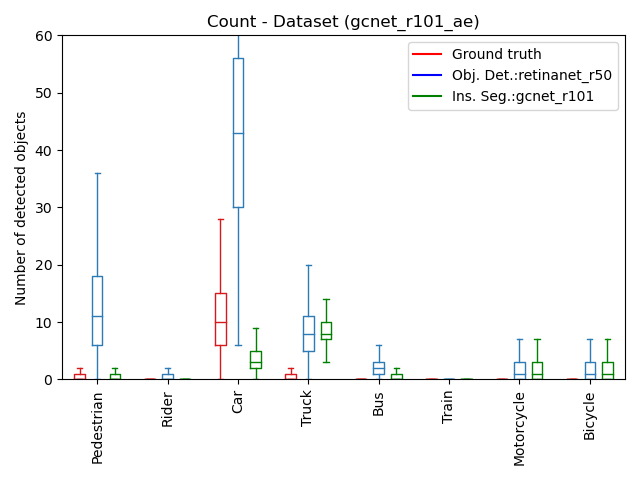}
    \end{subfigure}
    \begin{subfigure}{.24\linewidth}
        \centering
        \includegraphics[width=\linewidth]{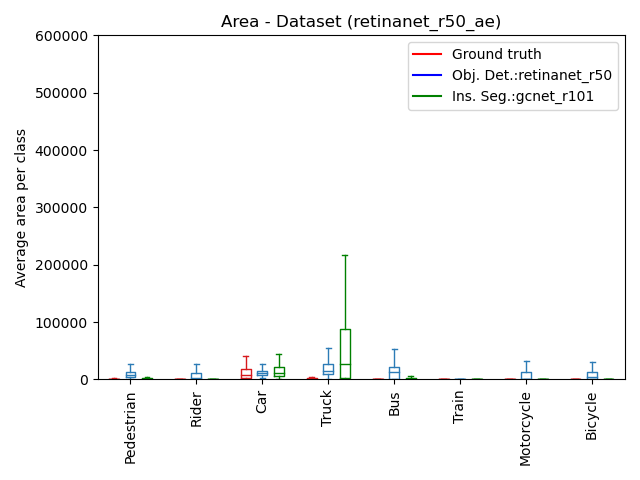}
    \end{subfigure}
    \begin{subfigure}{.24\linewidth}
        \centering
        \includegraphics[width=\linewidth]{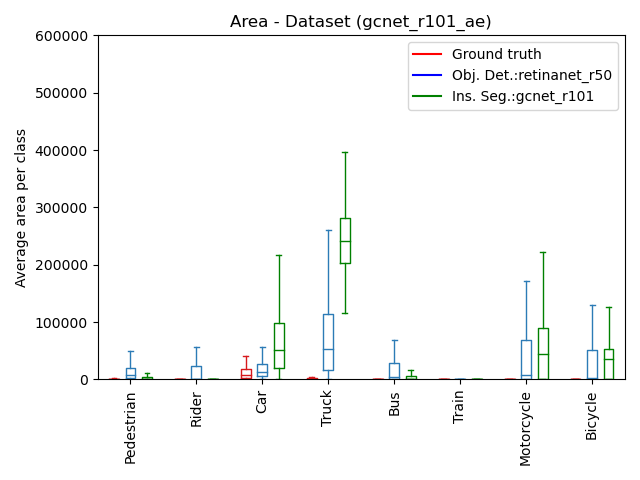}
    \end{subfigure}
    \caption{OD\_retinanet\_r50\_SEG\_gcnet\_r101}
\end{figure}

\begin{figure}[h!]
    \centering
    \begin{subfigure}{.24\linewidth}
        \centering
        \includegraphics[width=\linewidth]{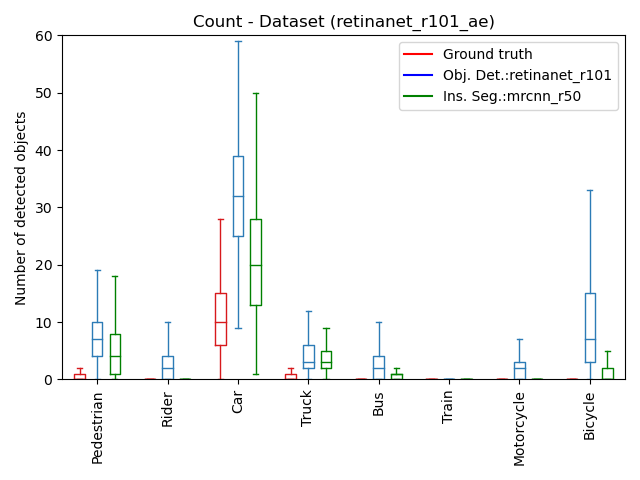}
    \end{subfigure}
    \begin{subfigure}{.24\linewidth}
        \centering
        \includegraphics[width=\linewidth]{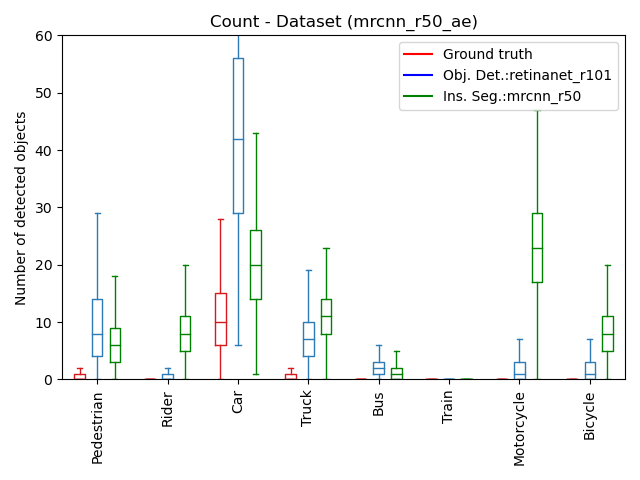}
    \end{subfigure}
    \begin{subfigure}{.24\linewidth}
        \centering
        \includegraphics[width=\linewidth]{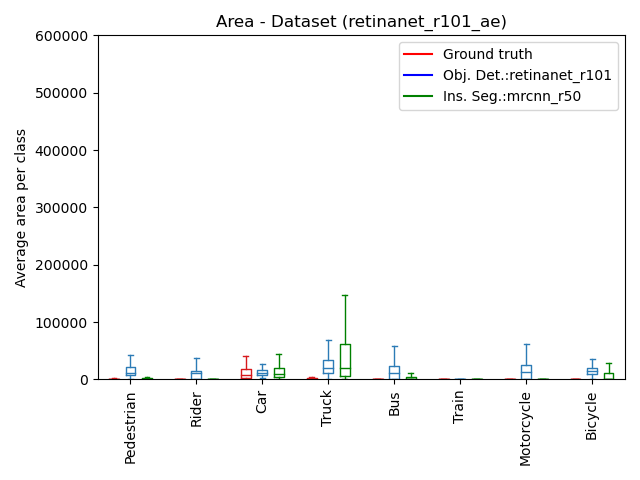}
    \end{subfigure}
    \begin{subfigure}{.24\linewidth}
        \centering
        \includegraphics[width=\linewidth]{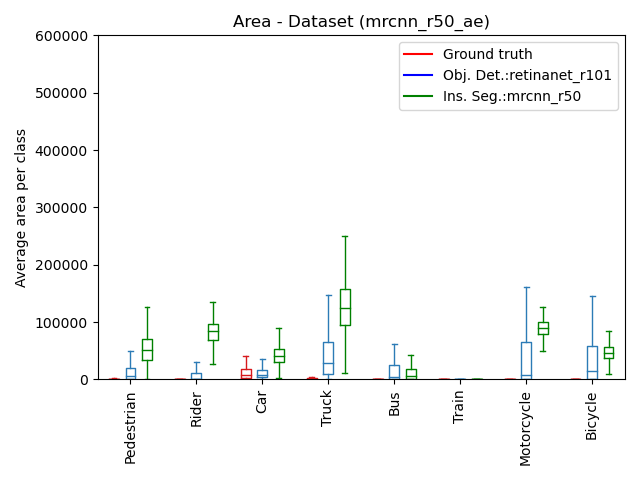}
    \end{subfigure}
    \caption{OD\_retinanet\_r101\_SEG\_mrcnn\_r50}
\end{figure}

\begin{figure}[h!]
    \centering
    \begin{subfigure}{.24\linewidth}
        \centering
        \includegraphics[width=\linewidth]{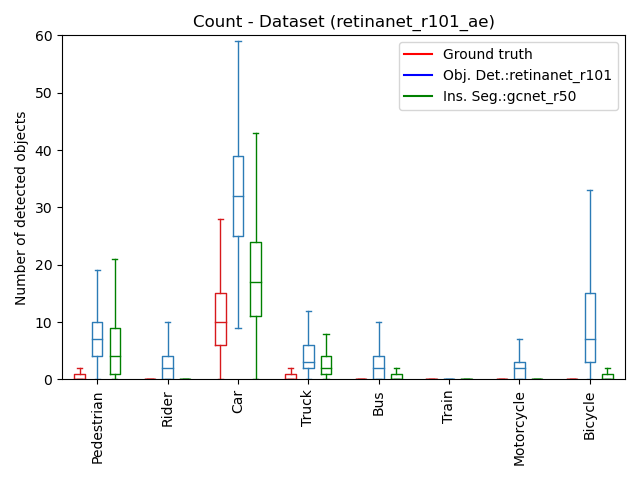}
    \end{subfigure}
    \begin{subfigure}{.24\linewidth}
        \centering
        \includegraphics[width=\linewidth]{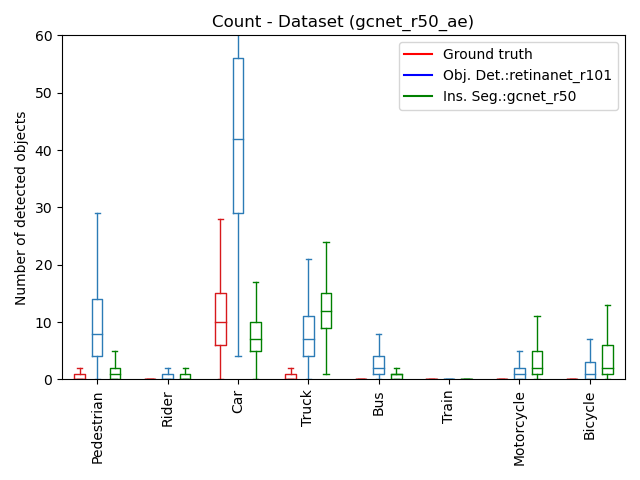}
    \end{subfigure}
    \begin{subfigure}{.24\linewidth}
        \centering
        \includegraphics[width=\linewidth]{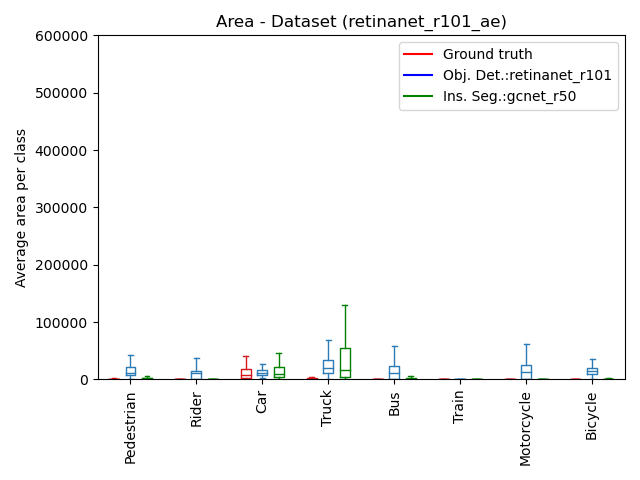}
    \end{subfigure}
    \begin{subfigure}{.24\linewidth}
        \centering
        \includegraphics[width=\linewidth]{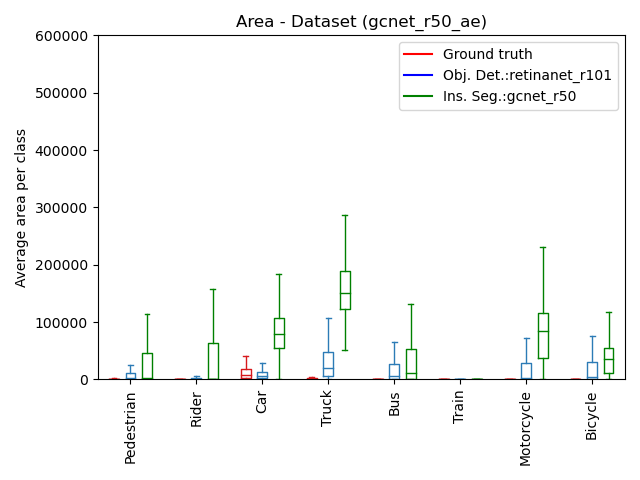}
    \end{subfigure}
    \caption{OD\_retinanet\_r101\_SEG\_gcnet\_r50}
\end{figure}

\begin{figure}[h!]
    \centering
    \begin{subfigure}{.24\linewidth}
        \centering
        \includegraphics[width=\linewidth]{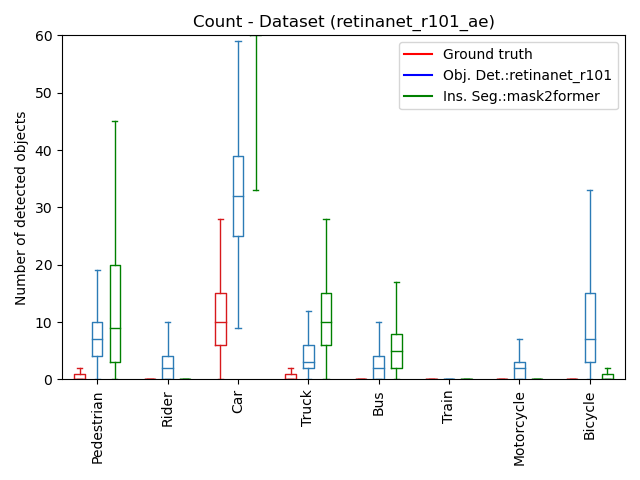}
    \end{subfigure}
    \begin{subfigure}{.24\linewidth}
        \centering
        \includegraphics[width=\linewidth]{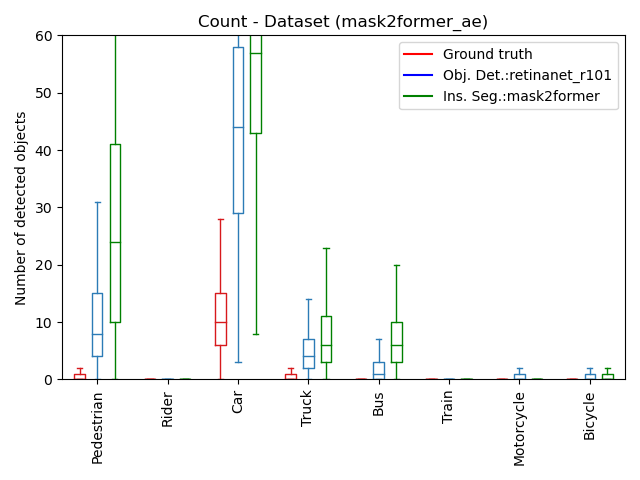}
    \end{subfigure}
    \begin{subfigure}{.24\linewidth}
        \centering
        \includegraphics[width=\linewidth]{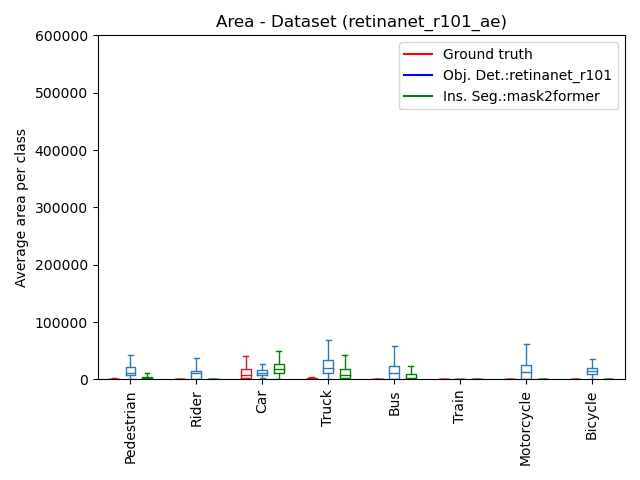}
    \end{subfigure}
    \begin{subfigure}{.24\linewidth}
        \centering
        \includegraphics[width=\linewidth]{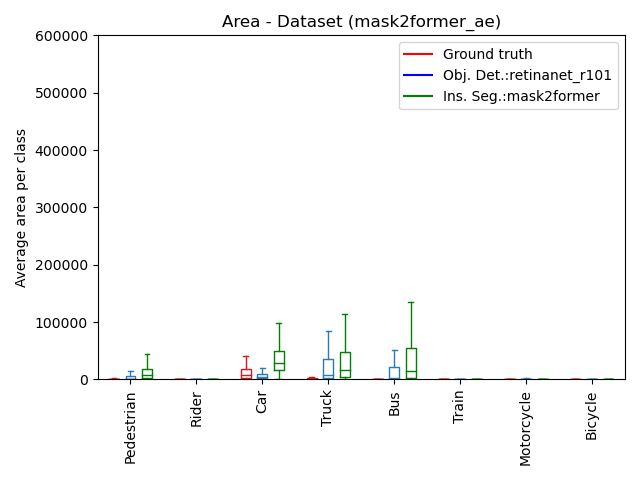}
    \end{subfigure}
    \caption{OD\_retinanet\_r101\_SEG\_mask2former}
\end{figure}

\begin{figure}[h!]
    \centering
    \begin{subfigure}{.24\linewidth}
        \centering
        \includegraphics[width=\linewidth]{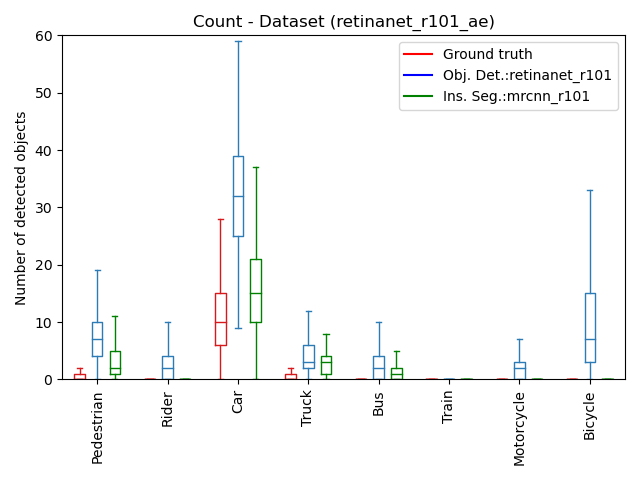}
    \end{subfigure}
    \begin{subfigure}{.24\linewidth}
        \centering
        \includegraphics[width=\linewidth]{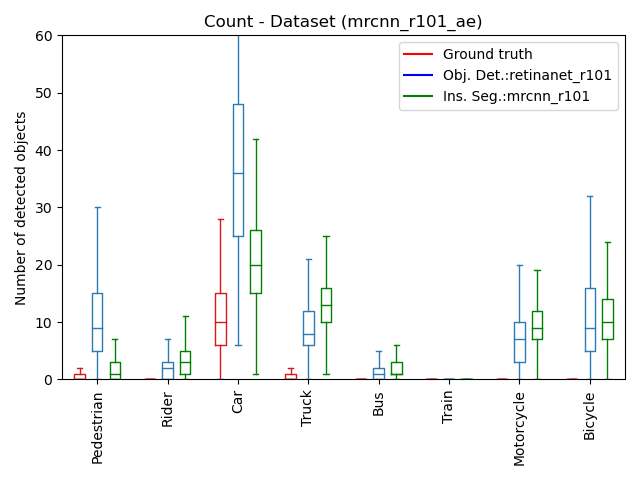}
    \end{subfigure}
    \begin{subfigure}{.24\linewidth}
        \centering
        \includegraphics[width=\linewidth]{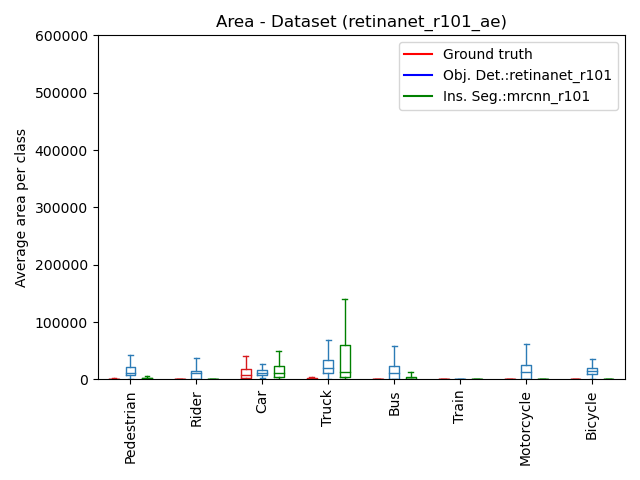}
    \end{subfigure}
    \begin{subfigure}{.24\linewidth}
        \centering
        \includegraphics[width=\linewidth]{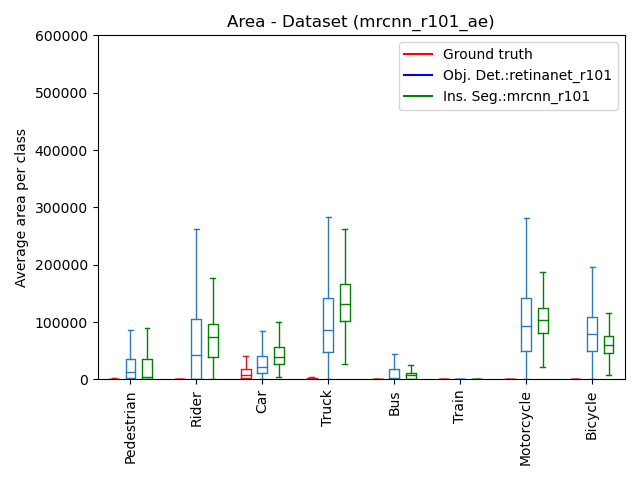}
    \end{subfigure}
    \caption{OD\_retinanet\_r101\_SEG\_mrcnn\_r101}
\end{figure}


\section{Defense}
\label{defense}
Defenses usually fall into three categories: data-based, model-based, detection-based. Data-based defenses use data augmentation at training to improve adversarial robustness. Adversarial training is a common data-based defense~\cite{qian2022survey}. Model-based focuses on selecting a specific network architecture that provides intrinsic adversarial robustness~\cite{ye2019adversarial, guo2020meets, dong2022adversarially}. Finally, detection-based defenses do not require data augmentation or model changes, but add a processing (on the input or the output of the model) in order to detect adversarial inputs. PatchCleanser~\cite{xiang2022patchcleanser} is one example of a double masking technique used to detect presence of adversarial patches in images.

\subsection{Adversarial Training}
The concept of adversarial training (AT) is to train a model on a dataset containing both genuine and adversarial examples in order to build resilience against perturbations. Previous work such as MTD~\cite{zhang2019towards} and Class-Wise Adversarial Training (CWAT)~\cite{chen2021class} defined loss functions to train the model to accurately localize and classify objects in an image despite the presence of adversarial noise. Unfortunately, all AT schemes demonstrated a drop in model accuracy, which is not desirable.

\subsection{Model-based Defense: Robust Network}
\label{app_subsec:robustdet}

\subsubsection{Adversarially-Aware Robust Object Detector (RobustDet)}
\cite{dong2022adversarially} proposed a counter-proposal to adversarial training by modifying the model architecture. The proposal, named RobustDet, aims to modify an existing backbone (e.g., SSD) by adding three security components: an adversarial image discriminator (AID), an "adversarially-aware convolution" (AAconv), and a consistent features with reconstruction (CFR). The AID is a discriminator that outputs a probability vector based on the category of the image. For instance, if the AID discriminates the image as genuine, then the AID will output the probability vector for a genuine image. Otherwise, if the image is adversarial, then the AID will output the probability vector for an adversarial image. For the training phase, the author formulated a dedicated loss function for the AID to generate a probability vector specific to the category of the image (genuine or adversarial). This probability vector will serve as an input for the next module: \textit{AAconv}. Unlike in adversarial training, \textit{AAconv} aims to use specific weights for the model based on the category of the image. To achieve this goal, \textit{AAconv} uses the concept of dynamic convolution to generate different convolution kernels based on the category of the image. The generation of those convolution kernels is possible thanks to the (genuine or adversarial) probability vector provided by the \textit{AID}. The probability vector serves as the weights to generate convolution kernels. This approach allows to have dedicated weights for genuine images and adversarial images instead of having a single set of weights for both categories of images (like in AT). Lastly, the CFR reconstructs the adversarial image into a clean image. 

Looking at their mAP evaluation, RobustDet has higher mAP scores than adversarial training methods such as MTD and CWAT on both genuine and adversarial datasets. However, RobustDet still has at best a 20 mAP score difference between the genuine dataset and the adversarial dataset. This issue means RobustDet do not completely mitigate the mAP loss caused by adversarial examples.

\section{Joint attack}
\label{joint_attack}
A joint attack aims to create perturbations that deceive both tasks simultaneously. However, the impact of the perturbation on each task is independent, and the outputs of both tasks can vary significantly in terms of object location, size, and labels. To craft such perturbations, we applied a PGD attack on both models simultaneously by maximizing the sum of their losses, i.e., $L_{det}+L_{seg}$. 

The results presented in \cref{tab:Joint_attack} demonstrate the effectiveness of the joint attack. When only one task is targeted, the mAP of the targeted model is significantly reduced, but the other task remains relatively unaffected. In contrast, the joint attack results in a substantial degradation of both tasks' performance, highlighting the increased vulnerability when both models are attacked simultaneously.

\begin{figure}[!h]
    \centering
    \includegraphics[scale=0.45]{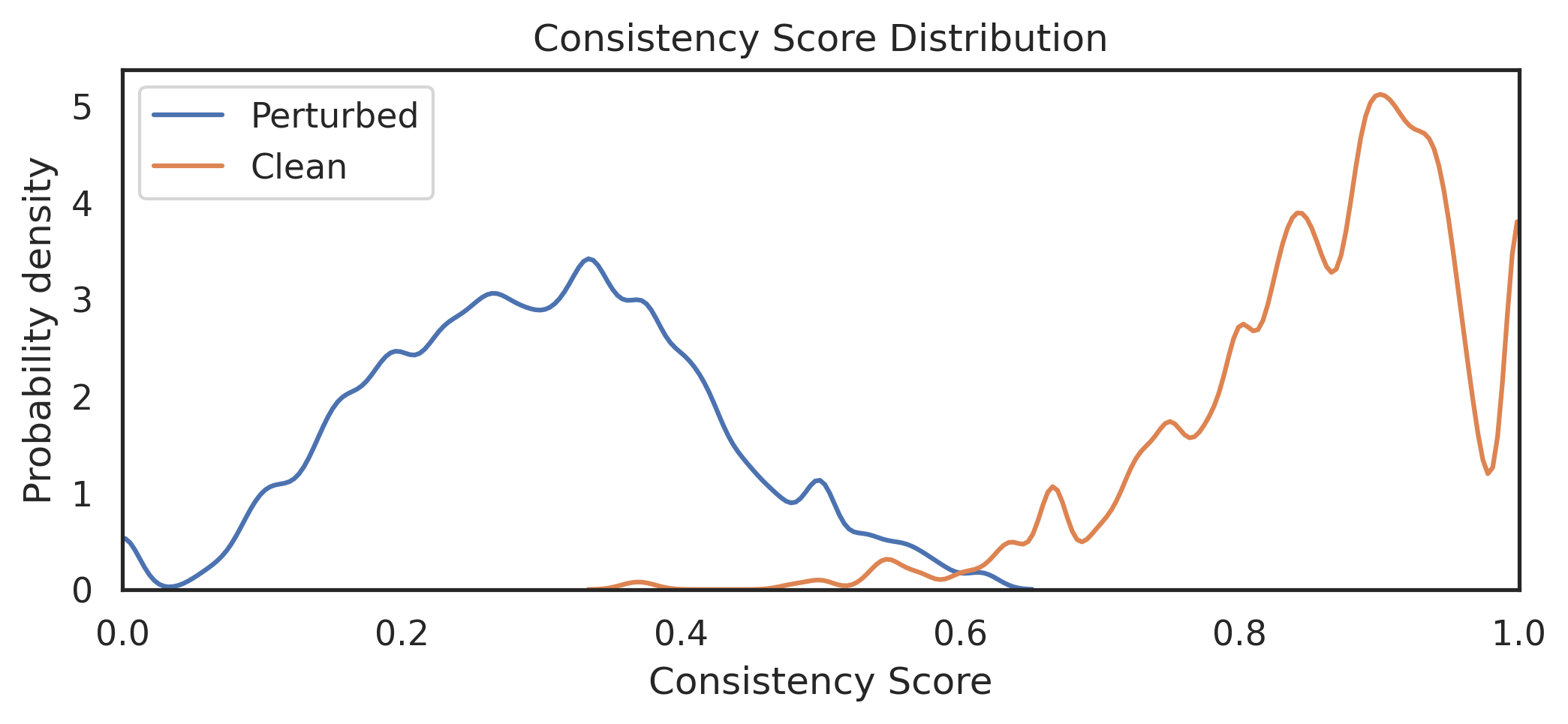}
    \caption{The joint attack still leads to inconsistent outputs between tasks thus exhibiting low consistency score}
    \label{fig:joint-consistency-dist}
\end{figure}

Despite the varying impact of the attack on the models, the output inconsistency between the models persists and can be identified by our consistency-based detector. As shown in \cref{fig:joint-consistency-dist}, the joint attacks still exhibit high inconsistency between model outputs (low consistency score) which can be easily distinguished from the clean images. 

\begin{figure}[!h]
    \centering
    \begin{subfigure}{.7\linewidth}
        \centering
        \par\medskip
        \includegraphics[width=\linewidth]{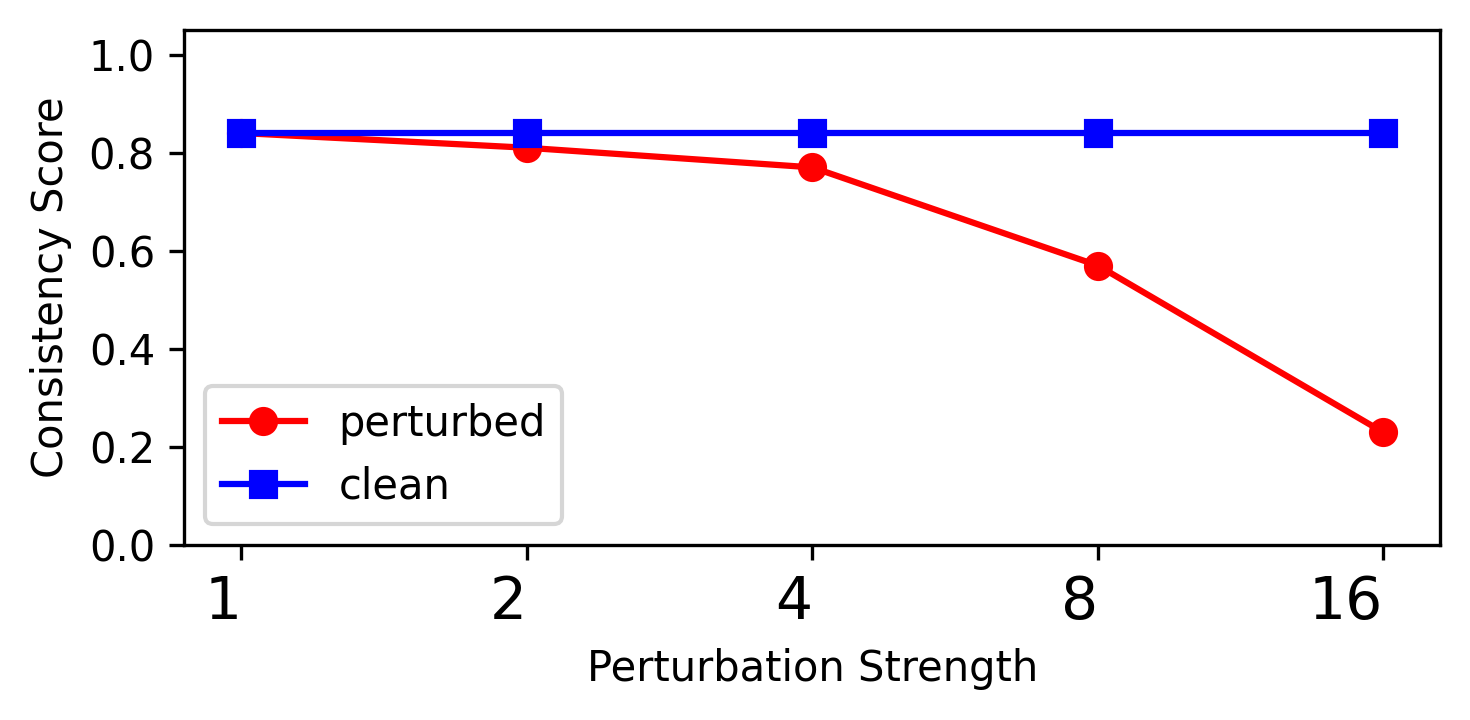}
        \caption{Consistency Score}
        \label{fig:joint_cs_pert}
    \end{subfigure}
    \vfill
    \begin{subfigure}{.7\linewidth}
        \centering
        \par\medskip
        \includegraphics[width=\linewidth]{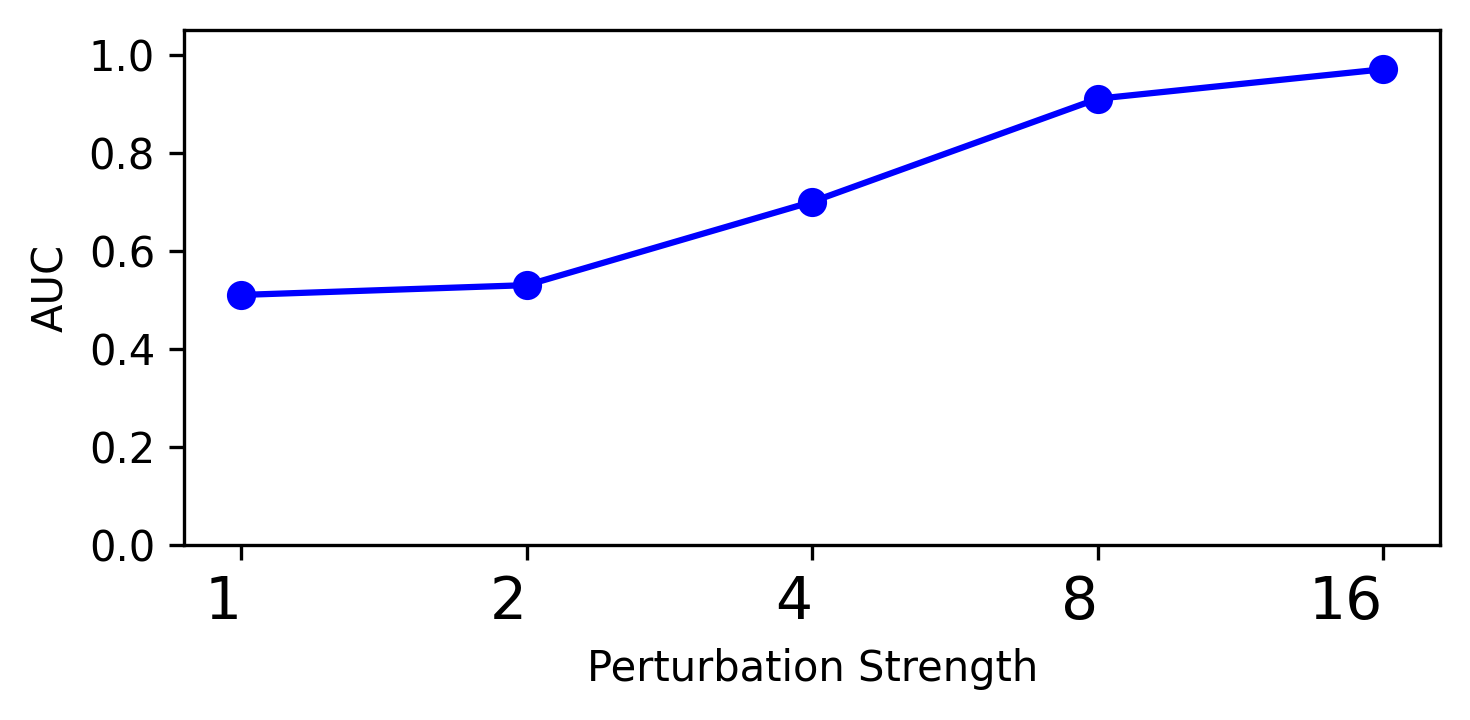}
        \caption{AUC}
        \label{fig:joint_auc_pert}
    \end{subfigure}
    \caption{Detection performance against Joint attack for different perturbation strength. (a) shows consistency score. (b) shows the detector AUC.}
    \label{fig:joint_auc_cs_pert}
\end{figure}

As illustrated in \cref{fig:joint_auc_cs_pert}, an increase in perturbation strength results in a lower consistency score, indicating greater inconsistency in the model outputs. This leads to a higher AUC for our detector, hence better ability in identifying these inconsistencies.

\section{Physical Patch Attack}
\label{sec:phy_test}
We conducted a real-world experiment to demonstrate that the inconsistency can also be efficiently identified by our detector in the physical world. We applied the ShapeShifter attack~\cite{chen2019shapeshifter} to create a perturbed stop sign optimized to fool the object detection model (maximizing objectness loss). We then printed the stop sign and approached it with a test vehicle equipped with a front camera from 50 meters away. The test vehicle was equipped with a medium-performance application processor for perception tasks and our consistency detector.

\begin{figure*}[bt!]
    \centering
    \begin{subfigure}{.35\linewidth}
        \centering
        \textbf{Object Detection}
        \par\medskip
        \includegraphics[width=\linewidth, clip = true, trim=0 0 70mm 20mm]{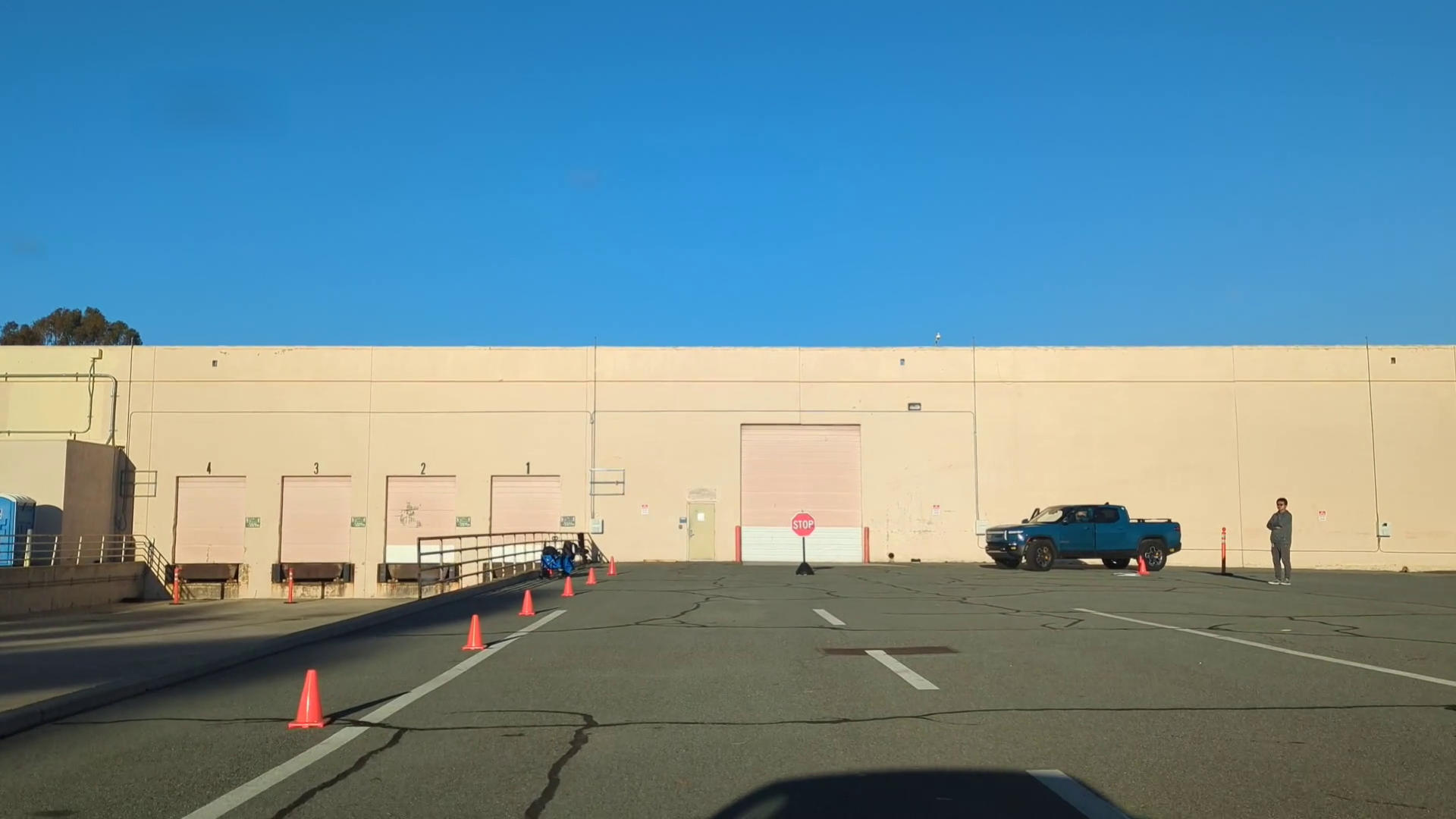}
        \caption{Distance 30m}
        \label{phy_det_30m}
    \end{subfigure}
    \hspace{2em}
    \begin{subfigure}{.35\linewidth}
        \centering
        \textbf{Instance Segmentation}
        \par\medskip
        \includegraphics[width=\linewidth, clip = true, trim=0 0 70mm 20mm]{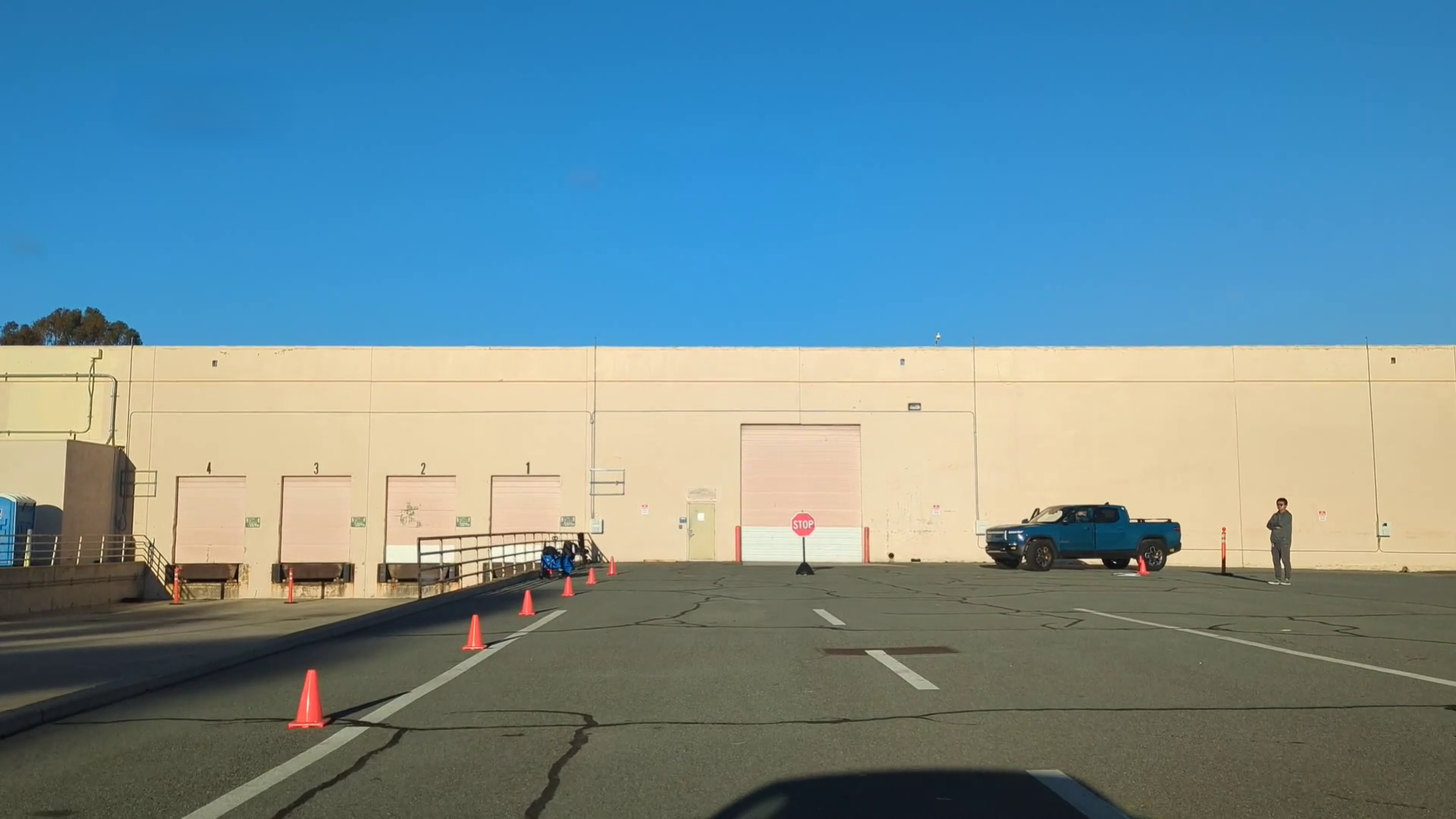}
        \caption{Distance 30m}
        \label{phy_ins_seg_30m}
    \end{subfigure}

    \begin{subfigure}{.35\linewidth}
        \includegraphics[width=\linewidth, clip = true, trim=0 0 70mm 20mm]{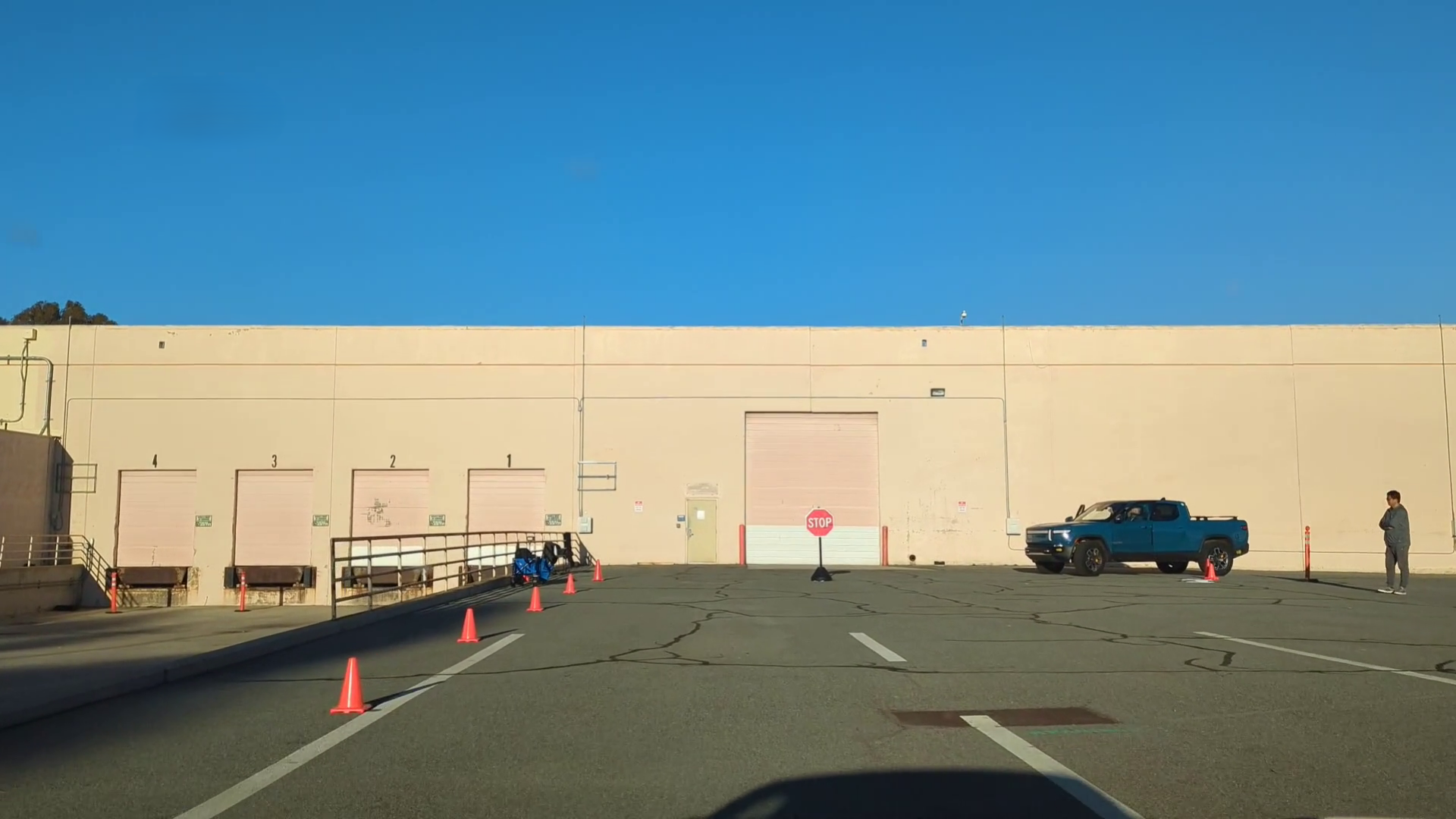}
        \caption{Distance 25m}
        \label{phy_det_25m}
    \end{subfigure}
    \hspace{2em}
    \begin{subfigure}{.35\linewidth}
        \includegraphics[width=\linewidth, clip = true, trim=0 0 70mm 20mm]{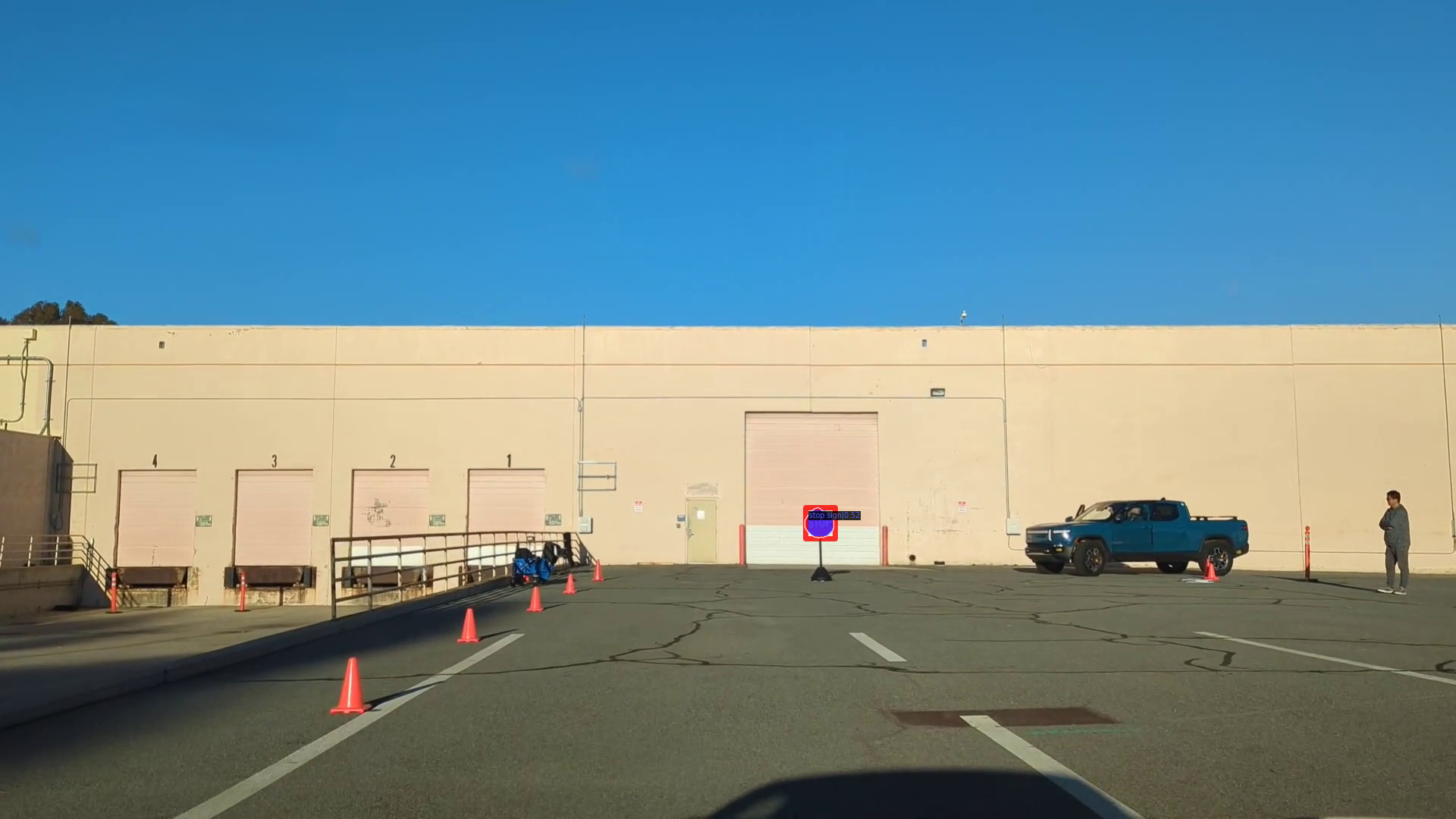}
        \caption{Distance 25m}
        \label{phy_ins_seg_25m}
    \end{subfigure}

    \begin{subfigure}{.35\linewidth}
        \includegraphics[width=\linewidth, clip = true, trim=0 0 70mm 20mm]{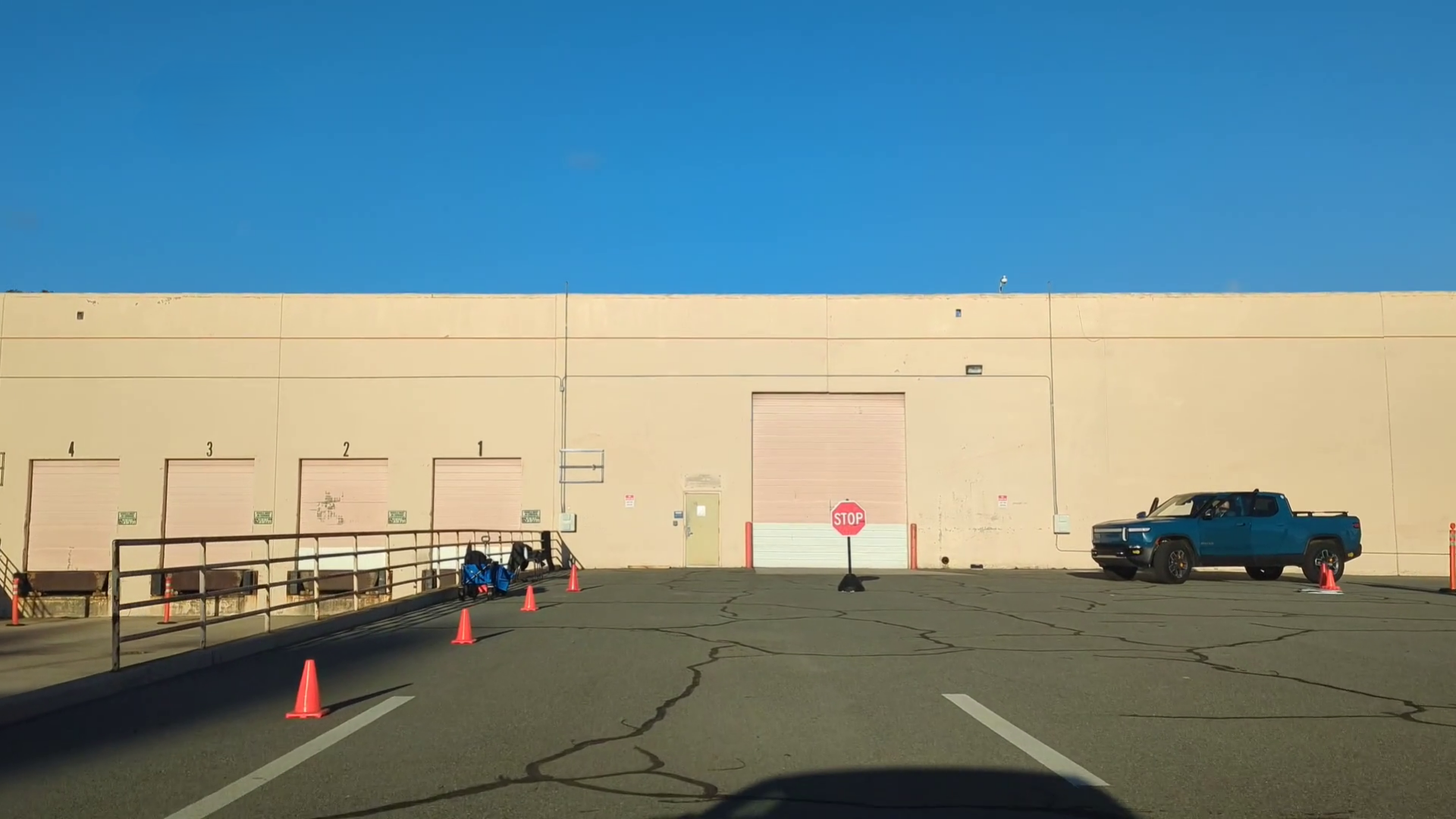}
        \caption{Distance 20m}
        \label{phy_ins_det_20m}
    \end{subfigure}
    \hspace{2em}
    \begin{subfigure}{.35\linewidth}
        \includegraphics[width=\linewidth, clip = true, trim=0 0 70mm 20mm]{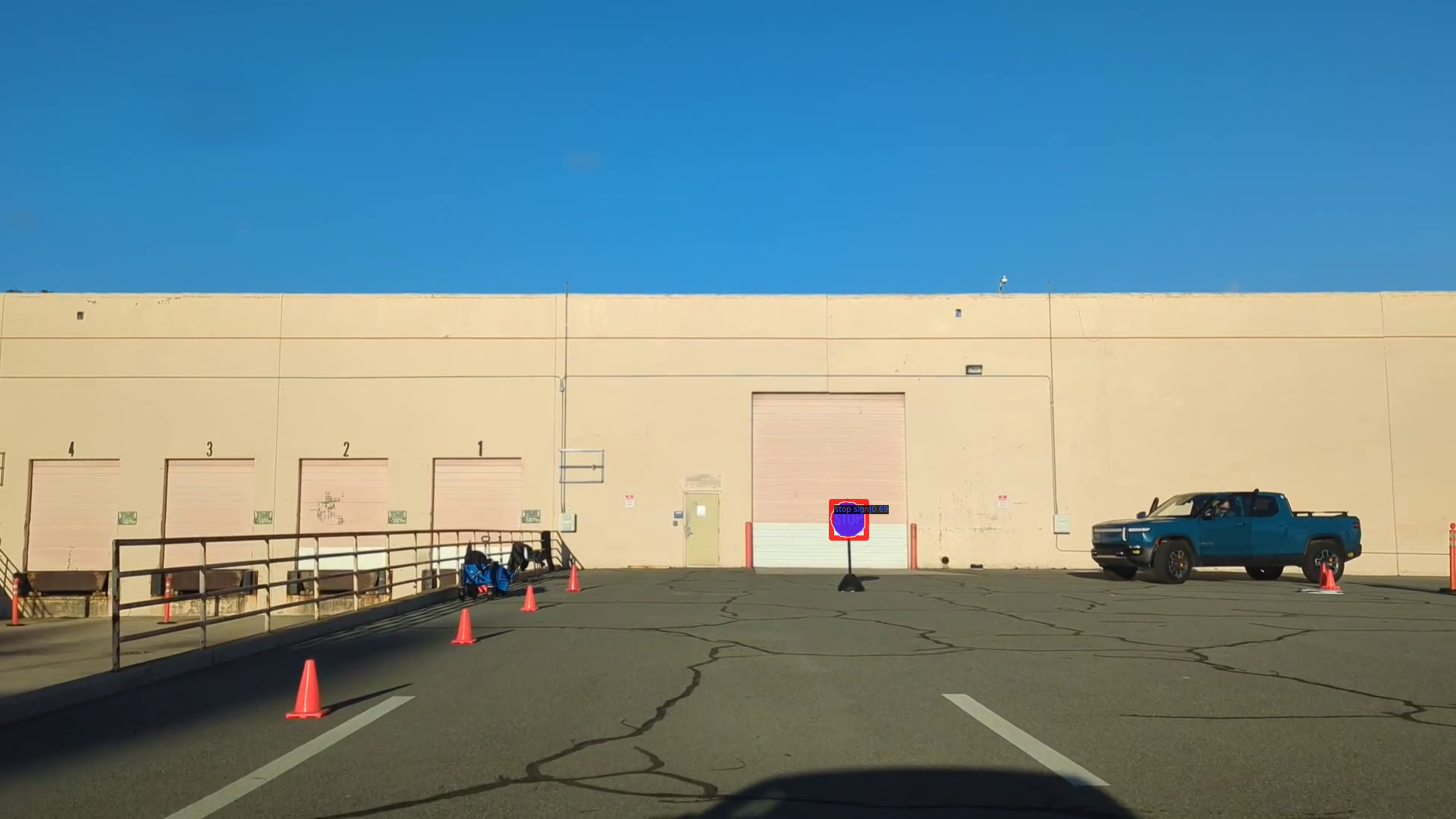}
        \caption{Distance 20m}
        \label{phy_ins_seg_20m}
    \end{subfigure}

    \begin{subfigure}{.35\linewidth}
        \includegraphics[width=\linewidth, clip = true, trim=0 0 70mm 20mm]{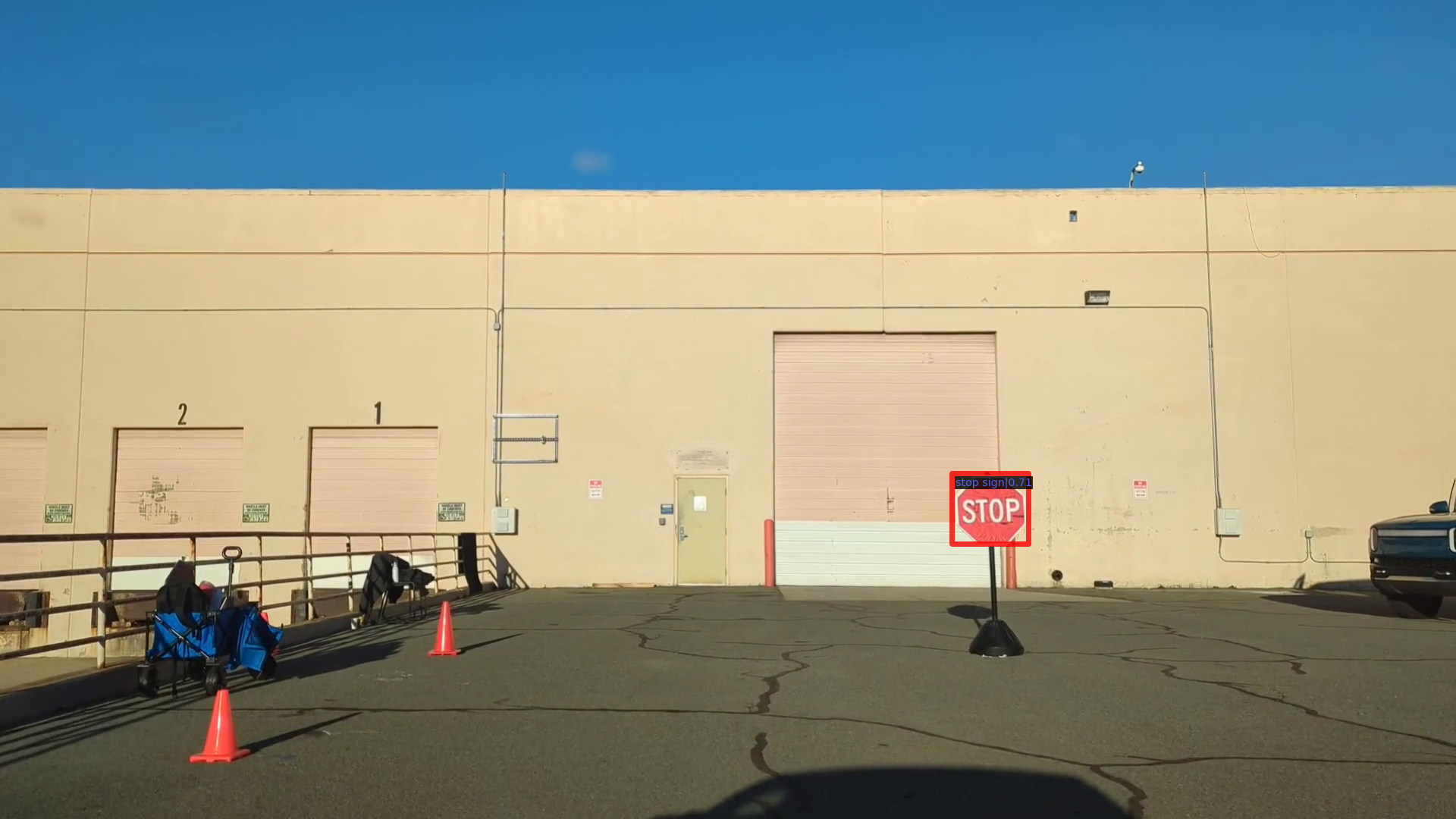}
        \caption{Distance 10m}
        \label{phy_det_10m}
    \end{subfigure}
    \hspace{2em}
    \begin{subfigure}{.35\linewidth}
        \includegraphics[width=\linewidth, clip = true, trim=0 0 70mm 20mm]{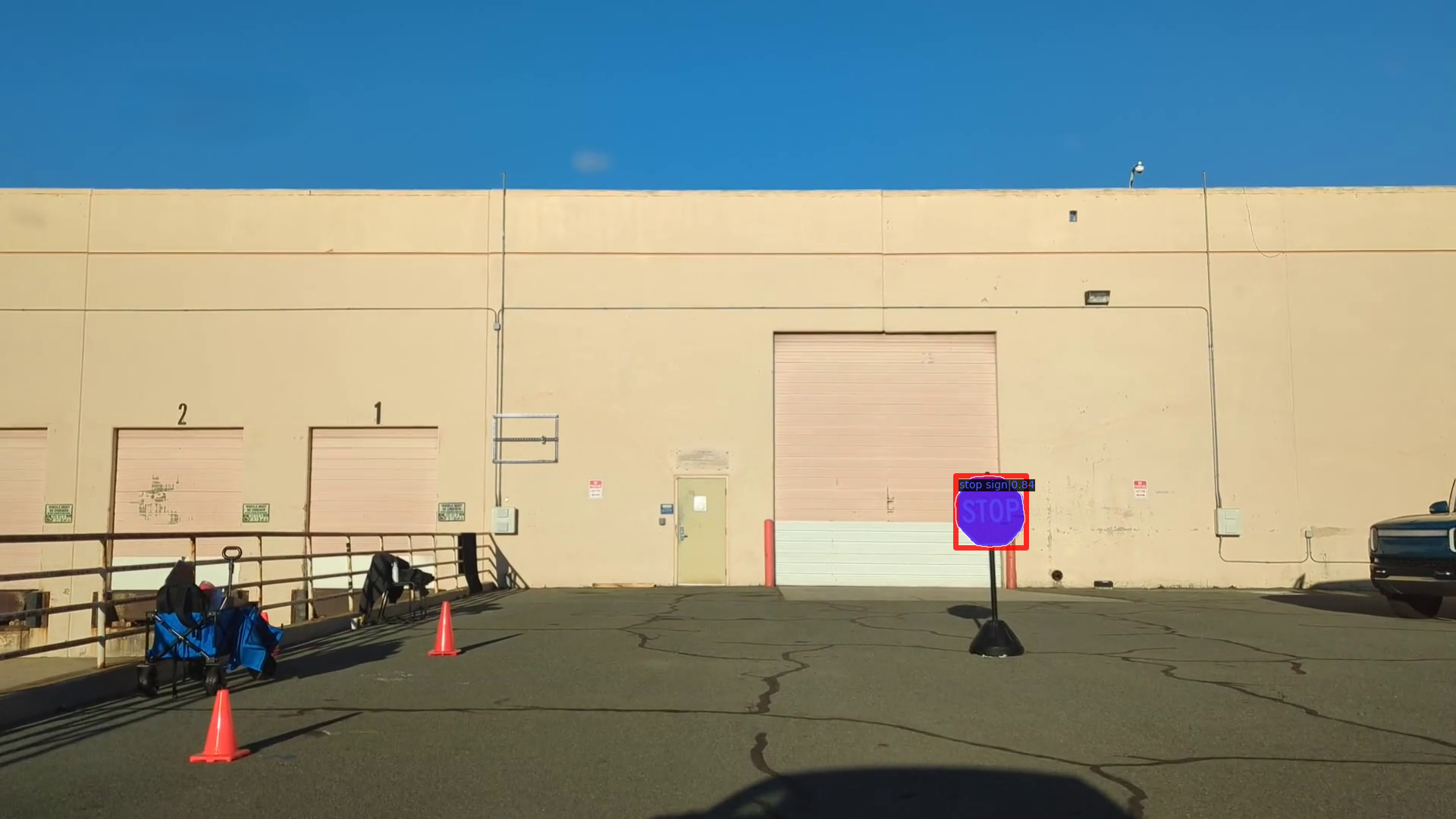}
        \caption{Distance 10m}
        \label{phy_ins_seg_10m}
    \end{subfigure}
    \caption{Physical test of an perturbed stop sign that attacks the OD model}
    \label{fig:phy_test}
\end{figure*}

As illustrated in \cref{fig:phy_test}, the perturbed stop sign successfully deceives the OD model at distances of 25 meters and 20 meters from the camera. However, despite the perturbations, the instance segmentation model remains capable of detecting the stop sign at these distances, as the perturbation is specifically optimized for the OD model. This highlights that inconsistencies between outputs of multi-task models remain even in physical attack. Our defense was able to detect the attack as demonstrated in the log\footnote{At the time of testing, the alert raised by our defense was not propagated to the downstream automated driving tasks, but only logged}.

\section{Detector Performance}
\label{app_subsec:det_perf} 

\begin{figure}[!h]
    \centering
        \includegraphics[scale=0.4]{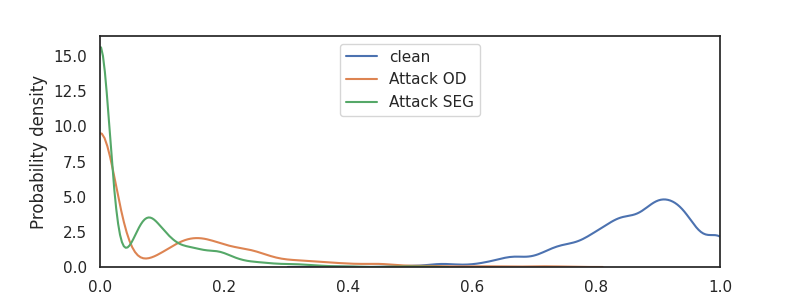}
    \caption{OD\_frcnn\_r50\_SEG\_mrcnn\_r50}
    \label{fig:same_arch_cs}
\end{figure}

\paragraph{Distribution of the consistency score.} This part contains additional results for CS distribution for all model pairs. As previous results showed, the model pairs with similar architecture results in distinct CS distributions between clean inputs and adversarial inputs, e.g., in \cref{fig:same_arch_cs}. This is desired for our detector to identify the perturbation. In contrast, the CS distributions for FRCNN R50 and MASK2FORMER SwinT is more difficult to distinguish, particularly when attacking MASK2FORMER SwinT model. This results in a relatively low AUC for this model pair as seen in \cref{fig:AUC_CS}.

\paragraph{Perturbation strength.} Perturbation strength affects the performance of detector using any model pair. As the perturbation strength increases, it results in stronger impact on the target model and causes higher inconsistency between model pairs.

Using (FRCNN R50, MRCNN R50) pair, we observe in Table~\cref{tab:pert_strength} that the mAP drops when $\epsilon$ increases. This shows the attack works as expected.

\clearpage

\begin{figure}[!h]
    \centering
        \includegraphics[scale=0.4]{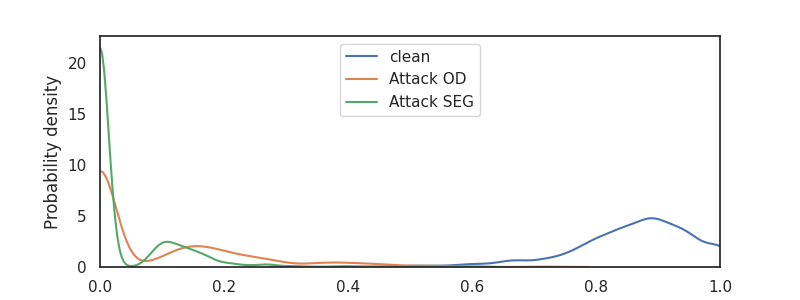}
    \caption{OD\_frcnn\_r50\_SEG\_gcnet\_r50}
\end{figure}

\begin{figure}[!h]
    \centering
        \includegraphics[scale=0.4]{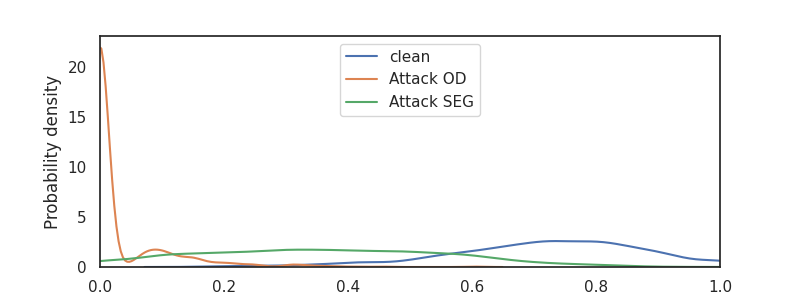}
    \caption{OD\_frcnn\_r50\_SEG\_mask2former}
\end{figure}

\begin{figure}[!h]
    \centering
        \includegraphics[scale=0.4]{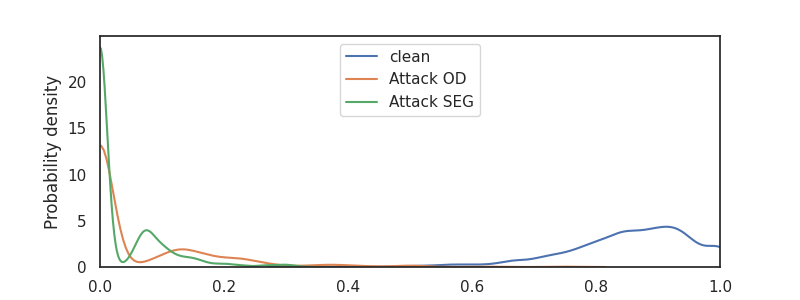}
    \caption{OD\_frcnn\_r50\_SEG\_mrcnn\_r101}
\end{figure}

\begin{figure}[!h]
    \centering
        \includegraphics[scale=0.4]{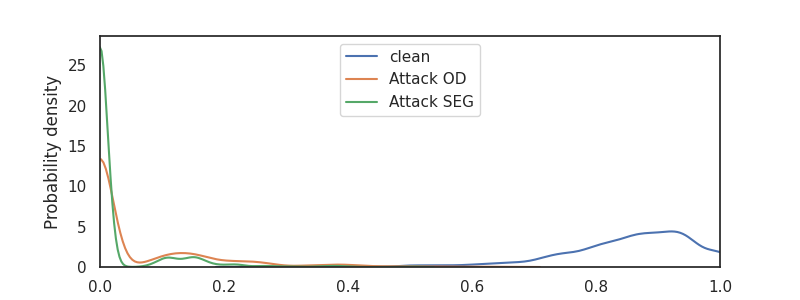}
    \caption{OD\_frcnn\_r50\_SEG\_gcnet\_r101}
\end{figure}

\begin{figure}[!h]
    \centering
        \includegraphics[scale=0.4]{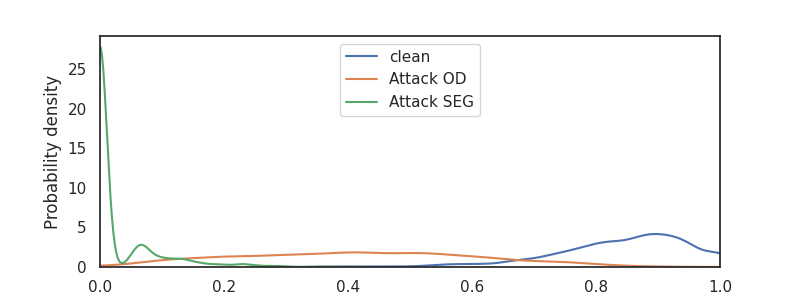}
    \caption{OD\_frcnn\_swint\_SEG\_mrcnn\_r50}
\end{figure}

\begin{figure}[!h]
    \centering
        \includegraphics[scale=0.4]{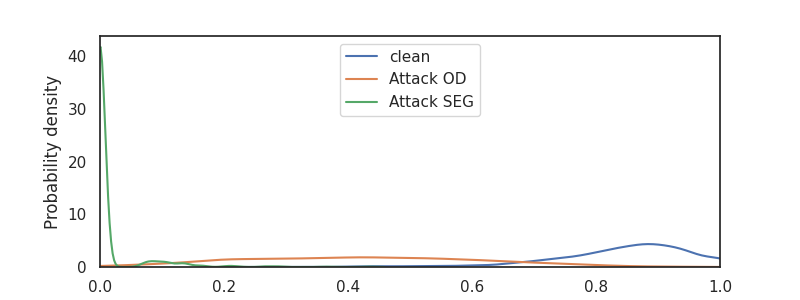}
    \caption{OD\_frcnn\_swint\_SEG\_gcnet\_r50}
\end{figure}

\begin{figure}[!h]
    \centering
        \includegraphics[scale=0.4]{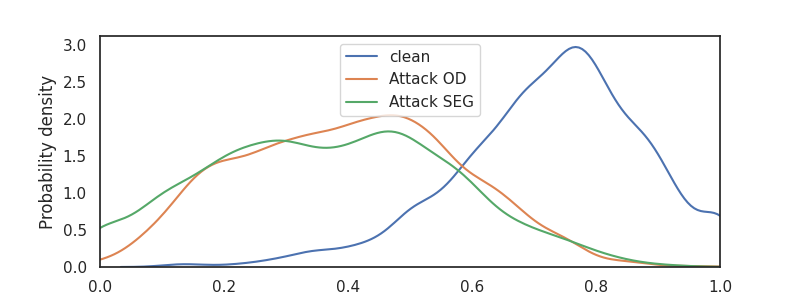}
    \caption{OD\_frcnn\_swint\_SEG\_mask2former}
\end{figure}

\begin{figure}[!h]
    \centering
        \includegraphics[scale=0.4]{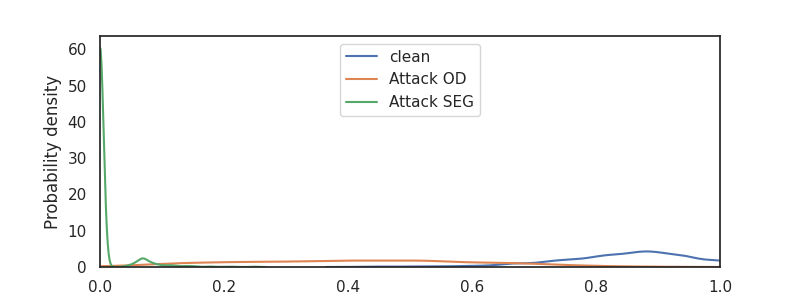}
    \caption{OD\_frcnn\_swint\_SEG\_mrcnn\_r101}
\end{figure}

\begin{figure}[!h]
    \centering
        \includegraphics[scale=0.4]{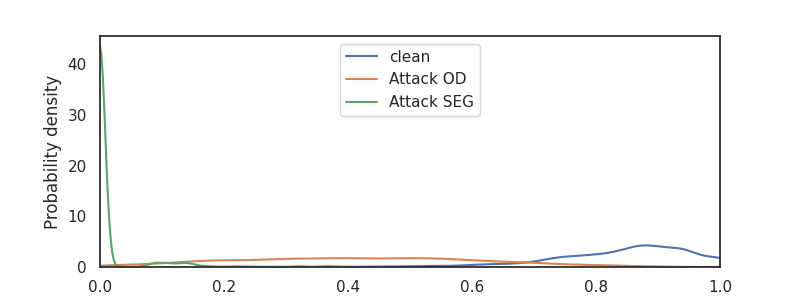}
    \caption{OD\_frcnn\_swint\_SEG\_gcnet\_r101}
\end{figure}

\begin{figure}[!h]
    \centering
        \includegraphics[scale=0.4]{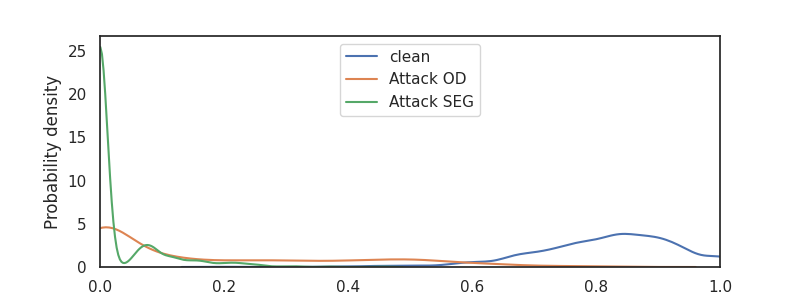}
    \caption{OD\_retinanet\_pvtv2\_SEG\_mrcnn\_r50}
\end{figure}

\begin{figure}[!h]
    \centering
        \includegraphics[scale=0.4]{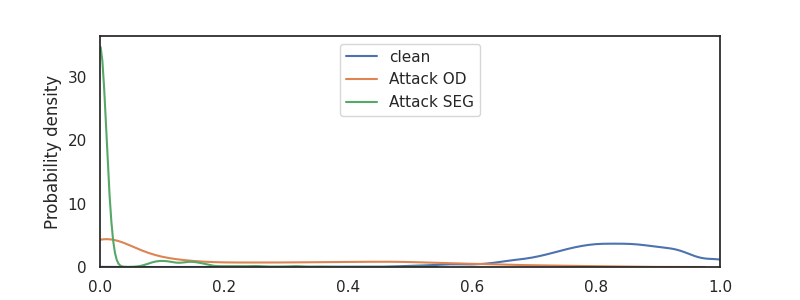}
    \caption{OD\_retinanet\_pvtv2\_SEG\_gcnet\_r50}
\end{figure}

\begin{figure}[!h]
    \centering
        \includegraphics[scale=0.4]{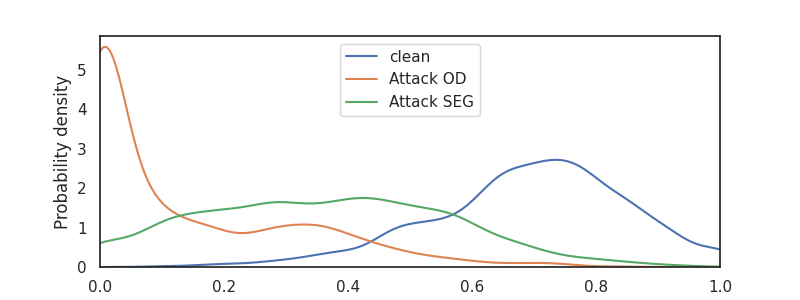}
    \caption{OD\_retinanet\_pvtv2\_SEG\_mask2former}
\end{figure}

\begin{figure}[!h]
    \centering
        \includegraphics[scale=0.4]{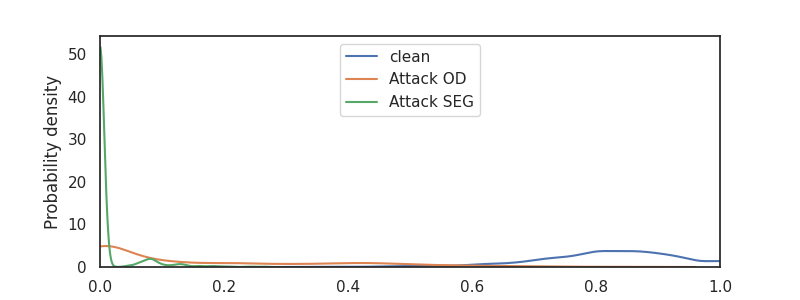}
    \caption{OD\_retinanet\_pvtv2\_SEG\_mrcnn\_r101}
\end{figure}

\begin{figure}[!h]
    \centering
        \includegraphics[scale=0.4]{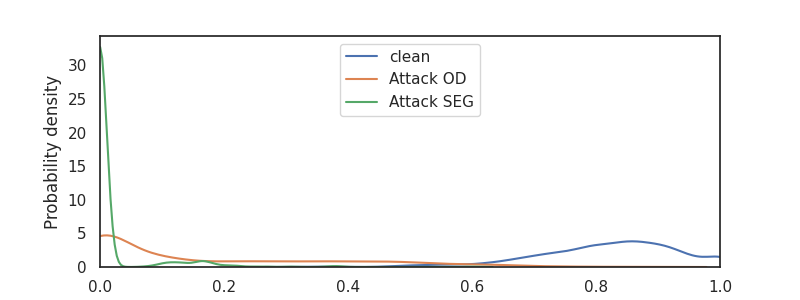}
    \caption{OD\_retinanet\_pvtv2\_SEG\_gcnet\_r101}
\end{figure}

\begin{figure}[!h]
    \centering
        \includegraphics[scale=0.4]{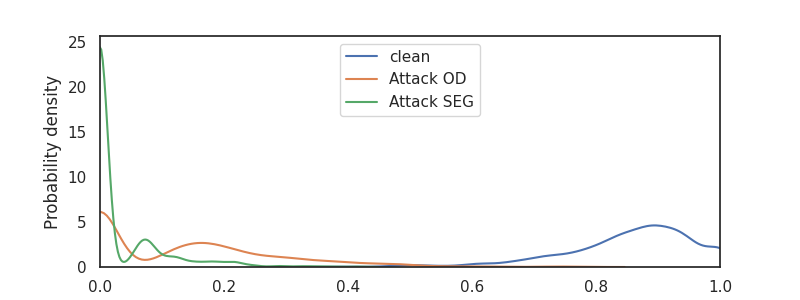}
    \caption{OD\_frcnn\_r101\_SEG\_mrcnn\_r50}
\end{figure}

\begin{figure}[!h]
    \centering
        \includegraphics[scale=0.4]{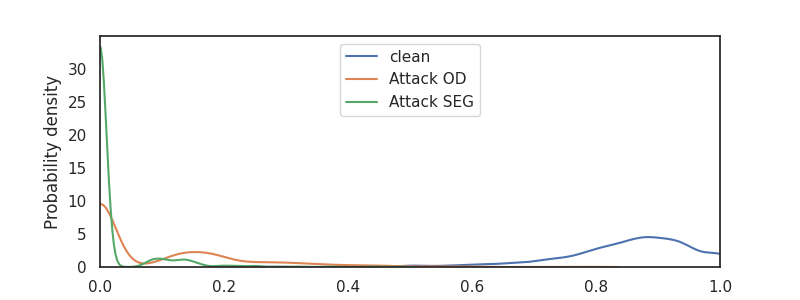}
    \caption{OD\_frcnn\_r101\_SEG\_gcnet\_r50}
\end{figure}

\begin{figure}[h!]
    \centering
    \begin{subfigure}{.475\linewidth}
        \centering
        \par\medskip
        \includegraphics[width=\linewidth]{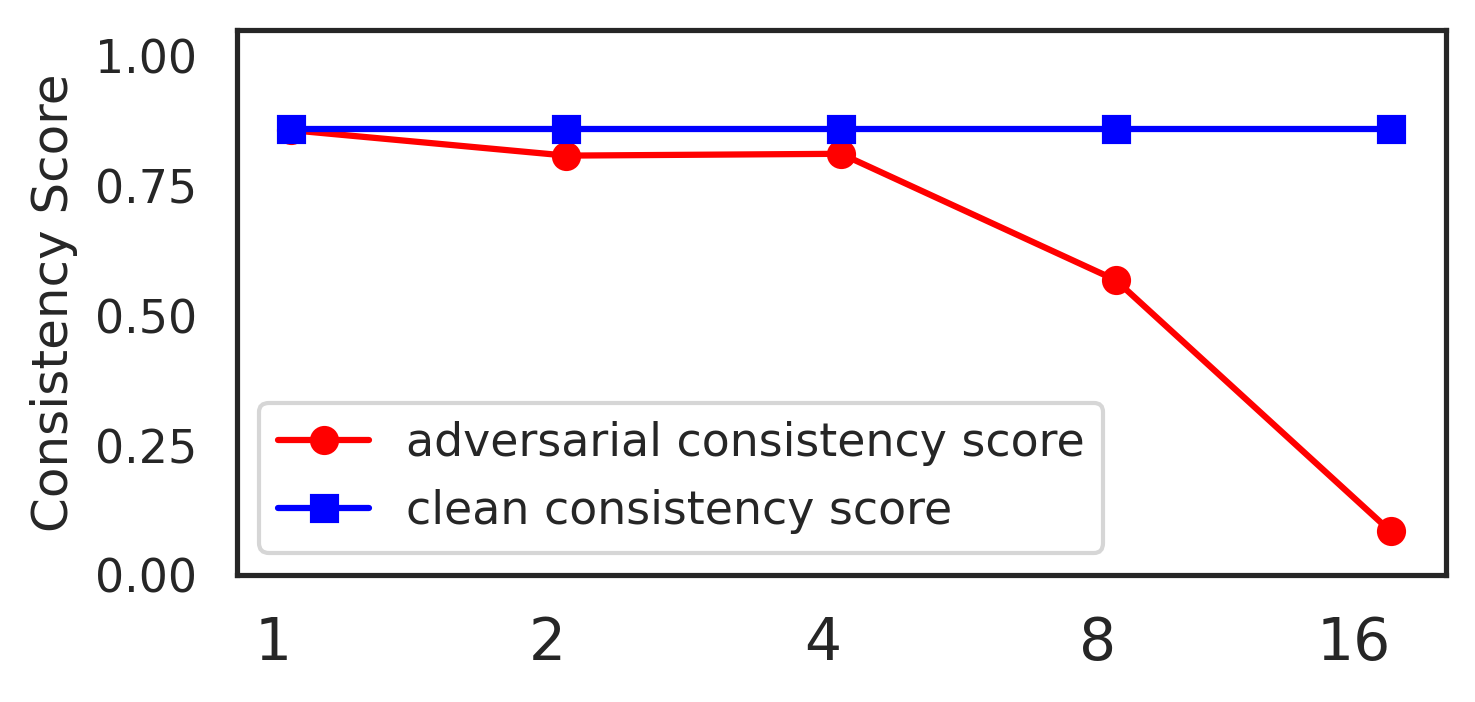}
        \caption{Consistency Score}
    \end{subfigure}
    \hfill
    \begin{subfigure}{.475\linewidth}
        \centering
        \par\medskip
        \includegraphics[width=\linewidth]{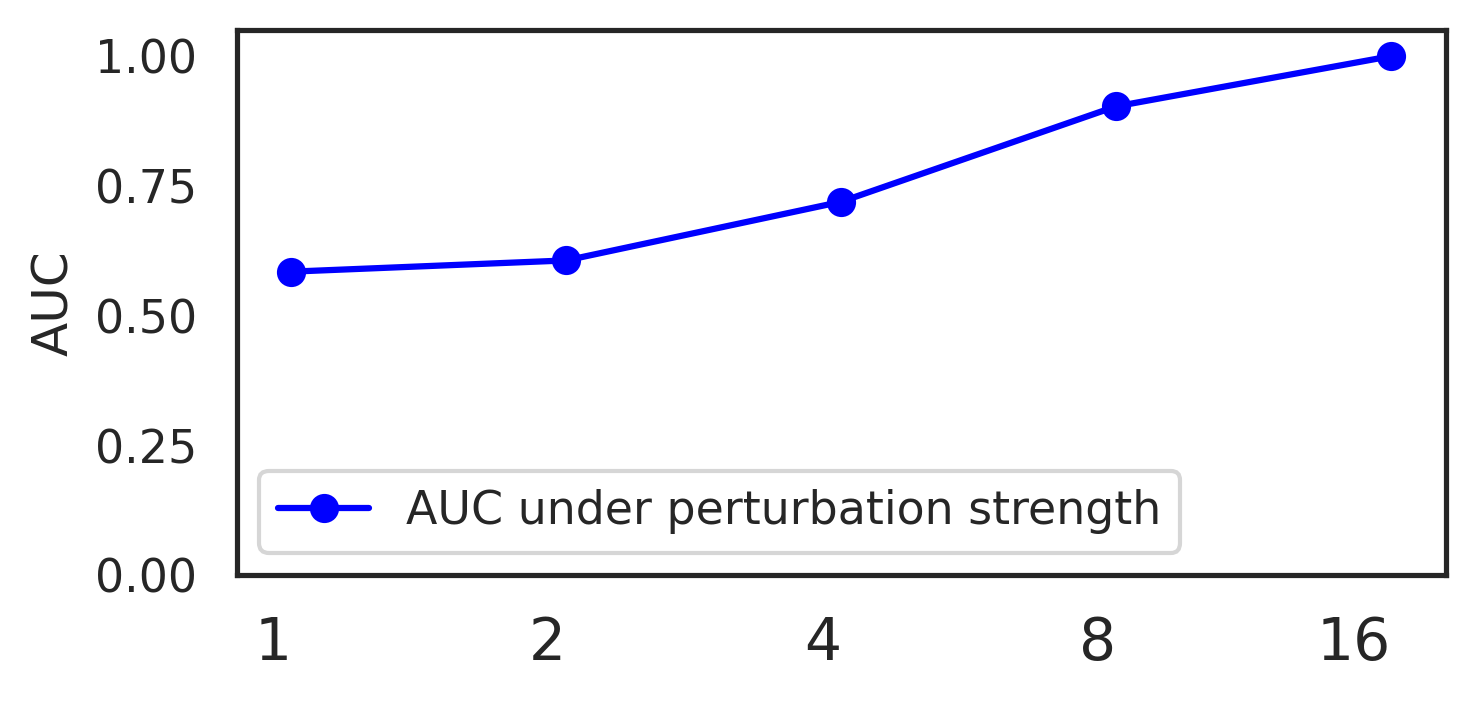}
        \caption{Consistency Score}
    \end{subfigure}
    \caption{Impact of perturbation strength for OD\_frcnn\_r50\_SEG\_mrcnn\_r50}
\end{figure}

\begin{figure}[h!]
    \centering
    \begin{subfigure}{.475\linewidth}
        \centering
        \par\medskip
        \includegraphics[width=\linewidth]{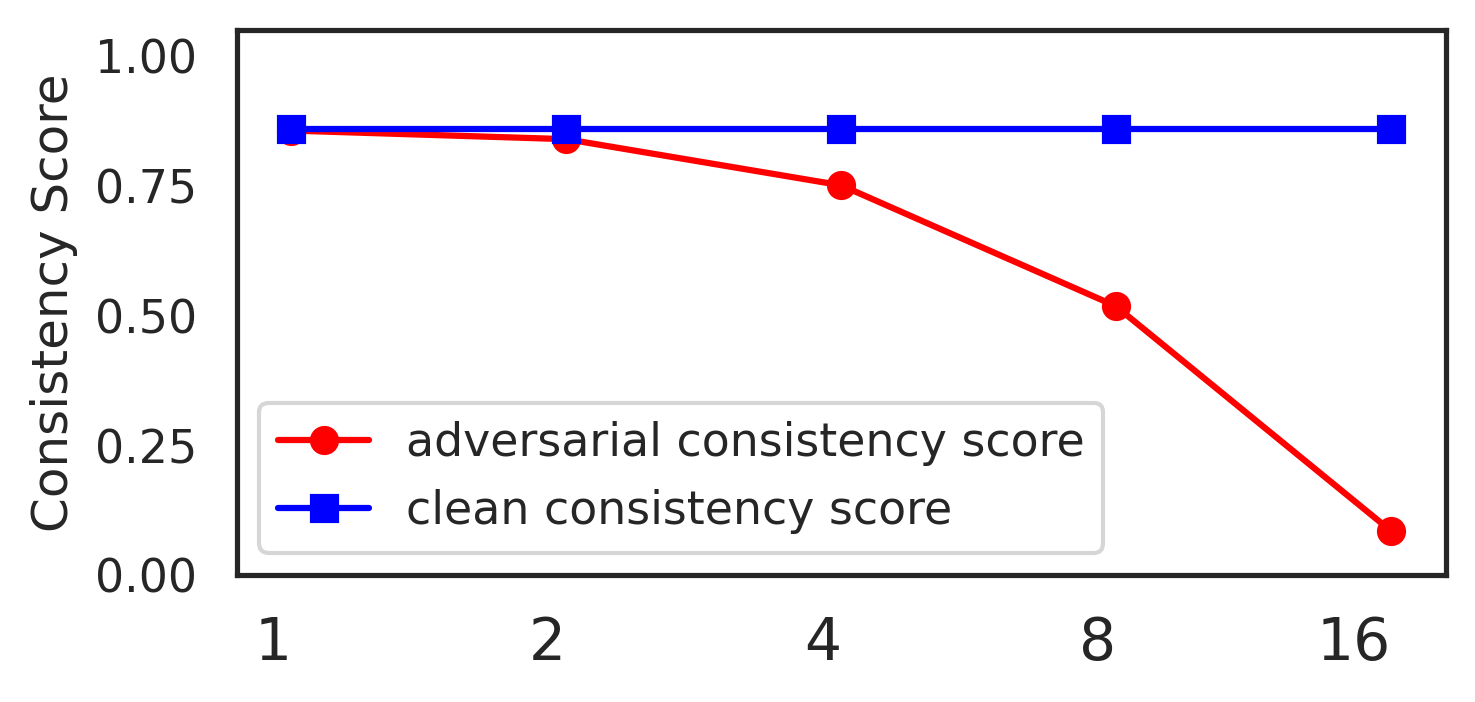}
        \caption{Consistency Score}
    \end{subfigure}
    \hfill
    \begin{subfigure}{.475\linewidth}
        \centering
        \par\medskip
        \includegraphics[width=\linewidth]{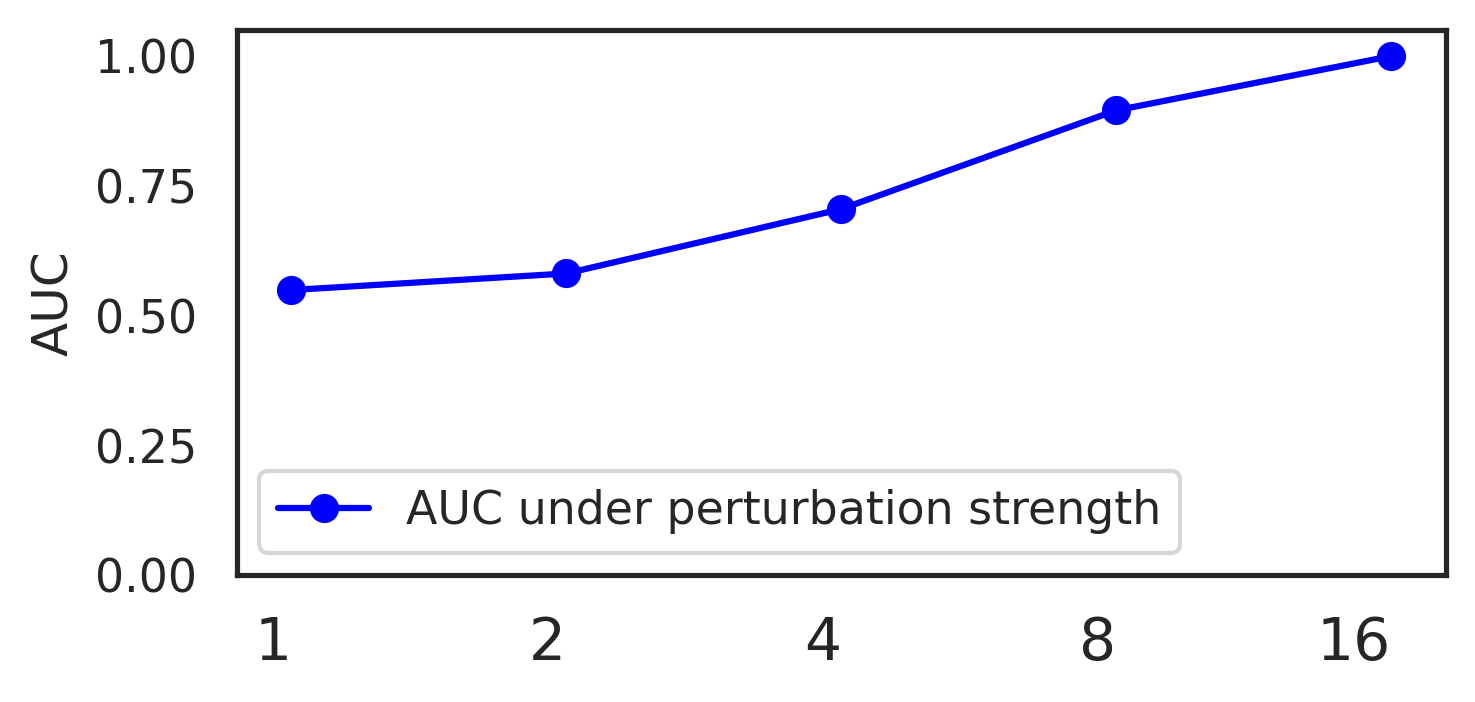}
        \caption{Consistency Score}
    \end{subfigure}
    \caption{Impact of perturbation strength for OD\_frcnn\_r50\_SEG\_gcnet\_r50}
\end{figure}

\begin{figure}[h!]
    \centering
    \begin{subfigure}{.475\linewidth}
        \centering
        \par\medskip
        \includegraphics[width=\linewidth]{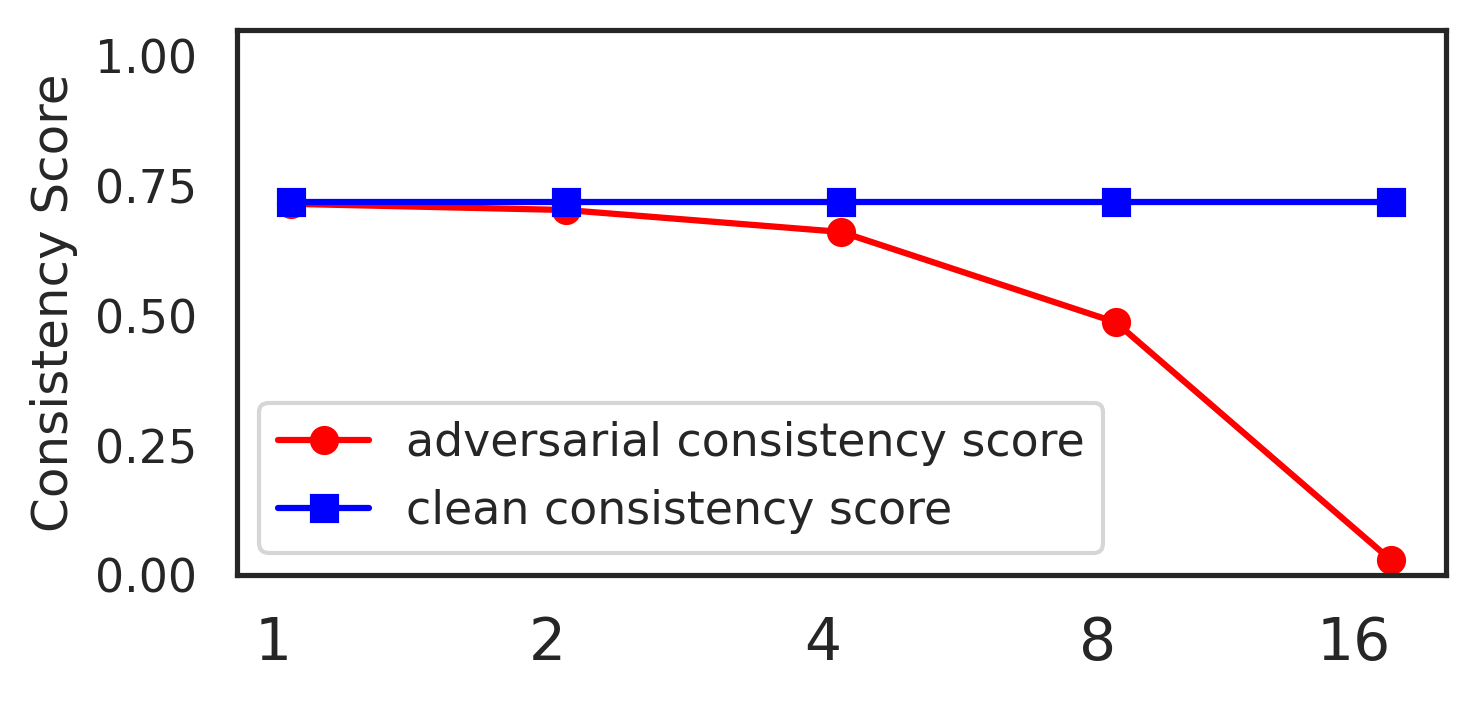}
        \caption{Consistency Score}
    \end{subfigure}
    \hfill
    \begin{subfigure}{.475\linewidth}
        \centering
        \par\medskip
        \includegraphics[width=\linewidth]{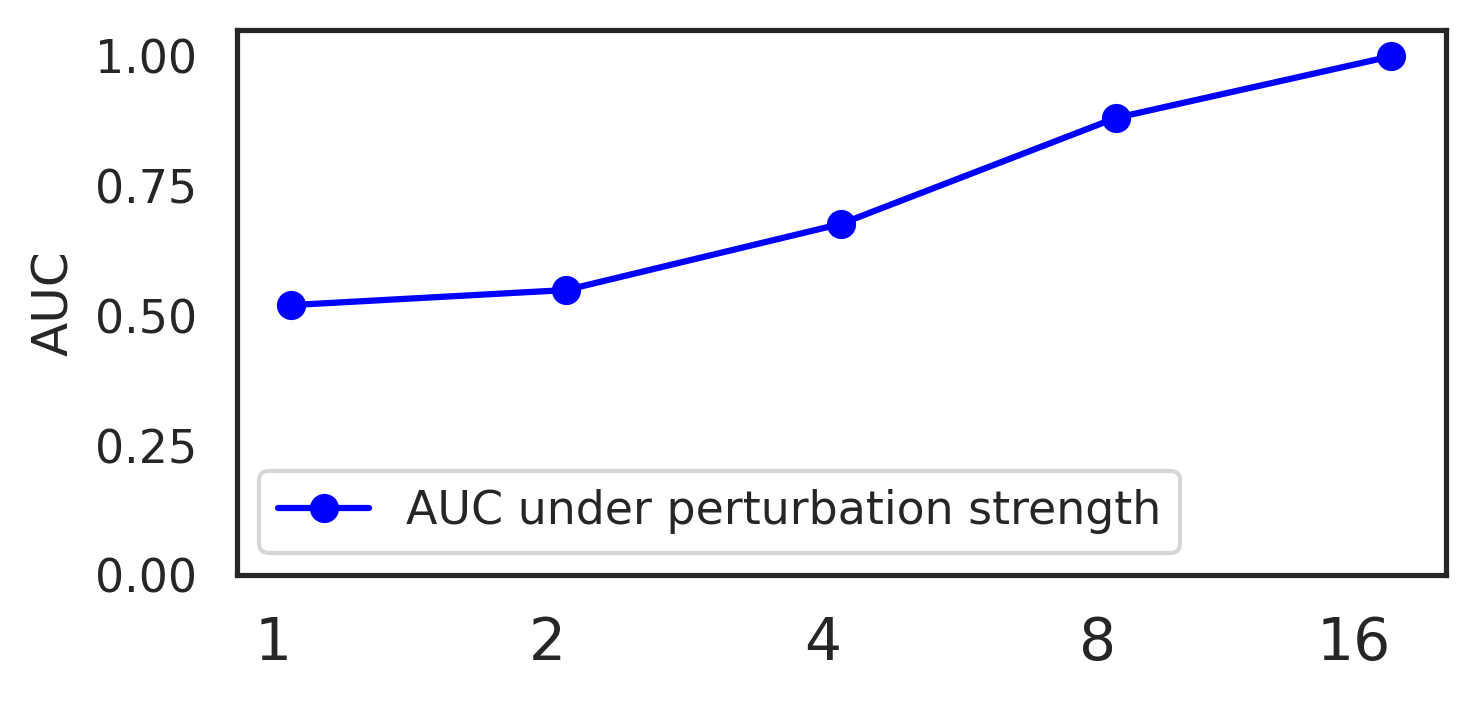}
        \caption{Consistency Score}
    \end{subfigure}
    \caption{Impact of perturbation strength for OD\_frcnn\_r50\_SEG\_mask2former}
\end{figure}

\begin{figure}[h!]
    \centering
    \begin{subfigure}{.475\linewidth}
        \centering
        \par\medskip
        \includegraphics[width=\linewidth]{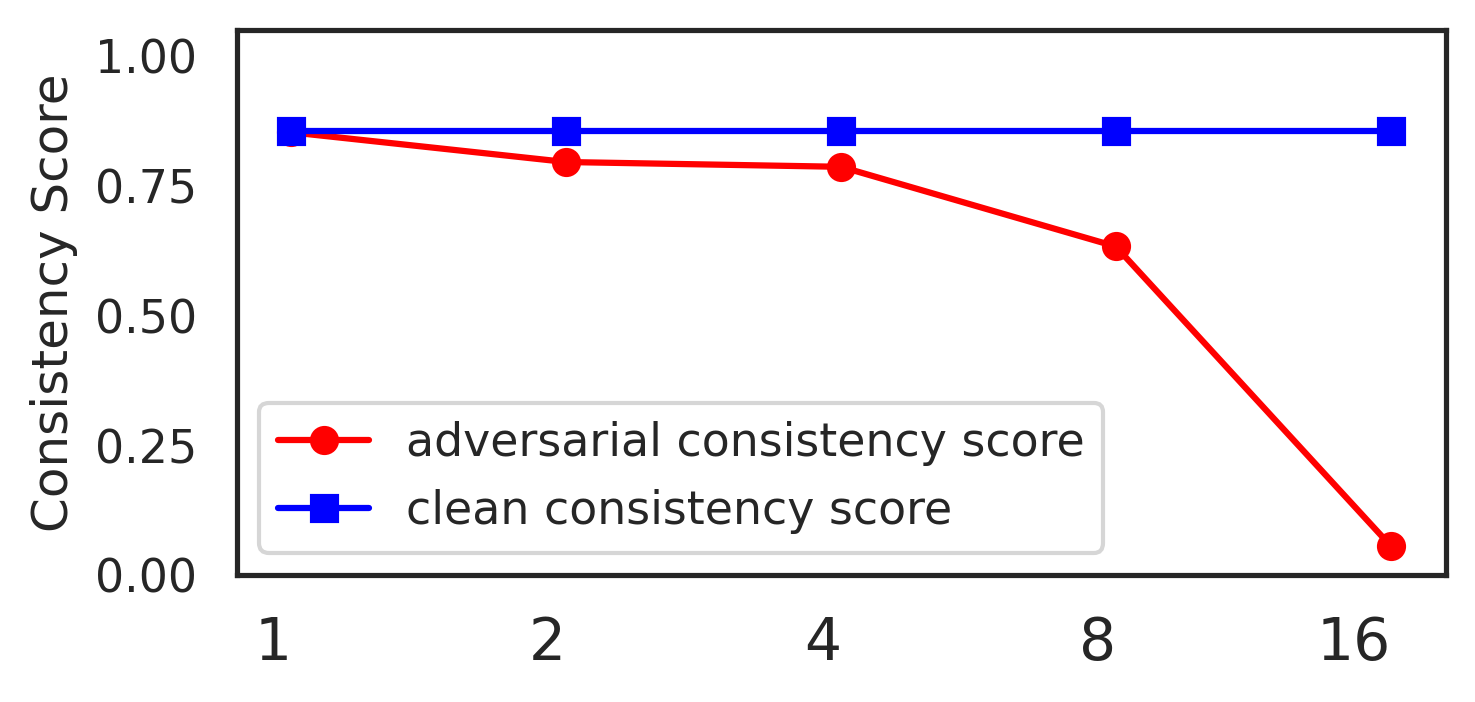}
        \caption{Consistency Score}
    \end{subfigure}
    \hfill
    \begin{subfigure}{.475\linewidth}
        \centering
        \par\medskip
        \includegraphics[width=\linewidth]{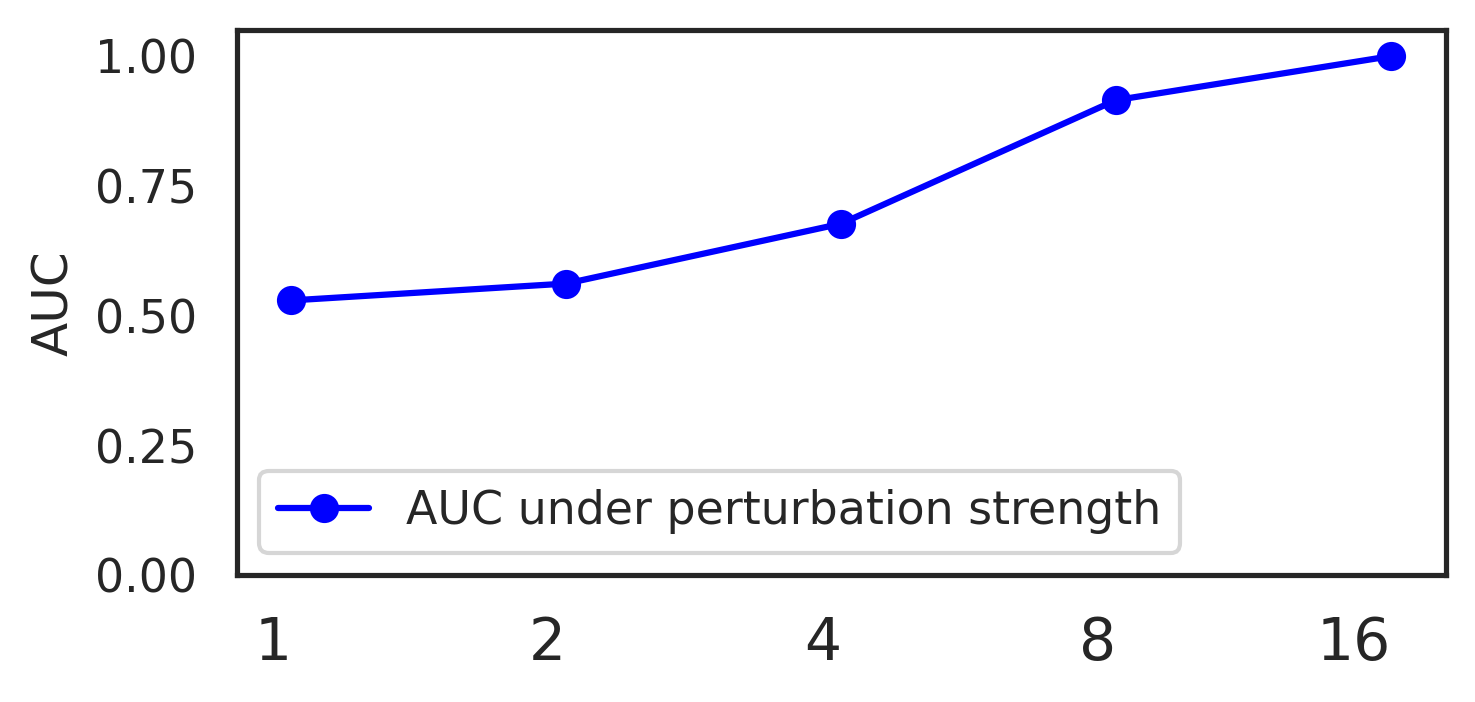}
        \caption{Consistency Score}
    \end{subfigure}
    \caption{Impact of perturbation strength for OD\_frcnn\_r50\_SEG\_mrcnn\_r101}
\end{figure}

\begin{figure}[h!]
    \centering
    \begin{subfigure}{.475\linewidth}
        \centering
        \par\medskip
        \includegraphics[width=\linewidth]{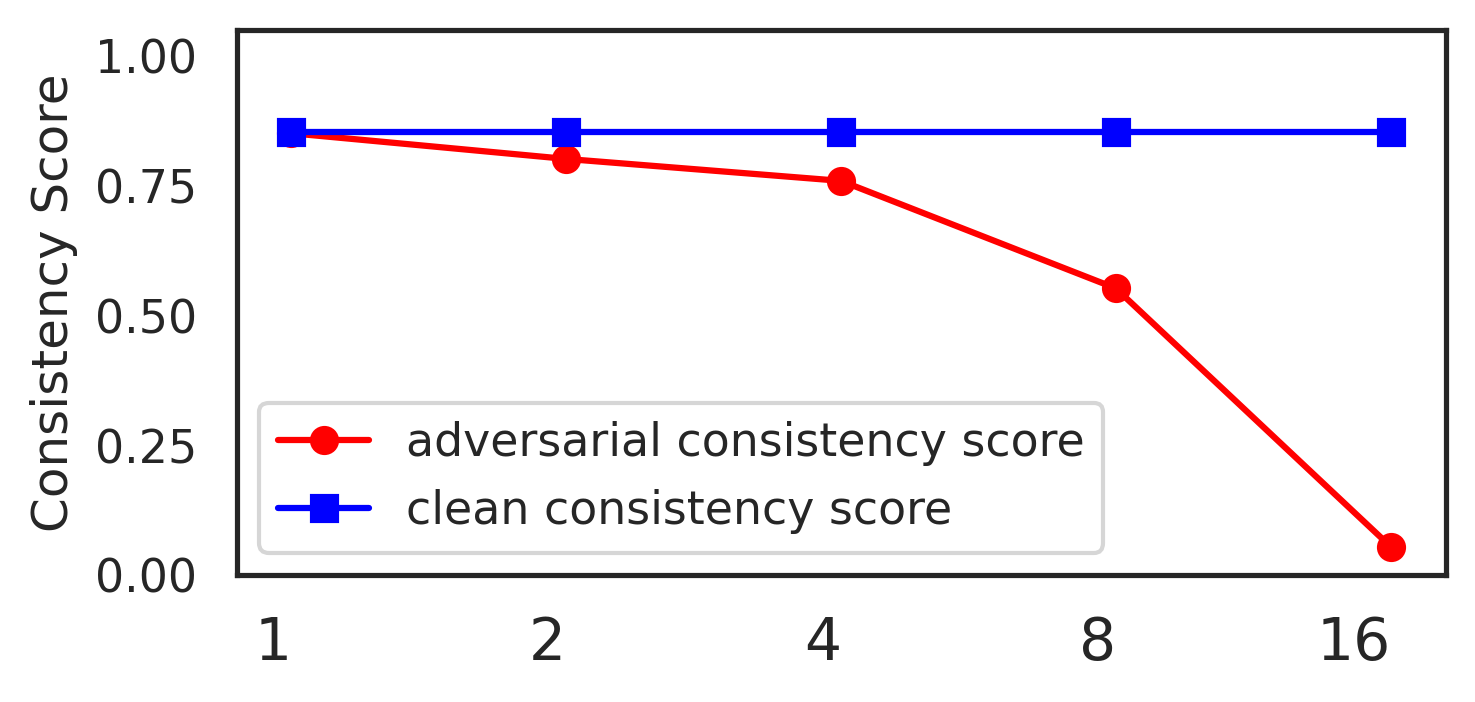}
        \caption{Consistency Score}
    \end{subfigure}
    \hfill
    \begin{subfigure}{.475\linewidth}
        \centering
        \par\medskip
        \includegraphics[width=\linewidth]{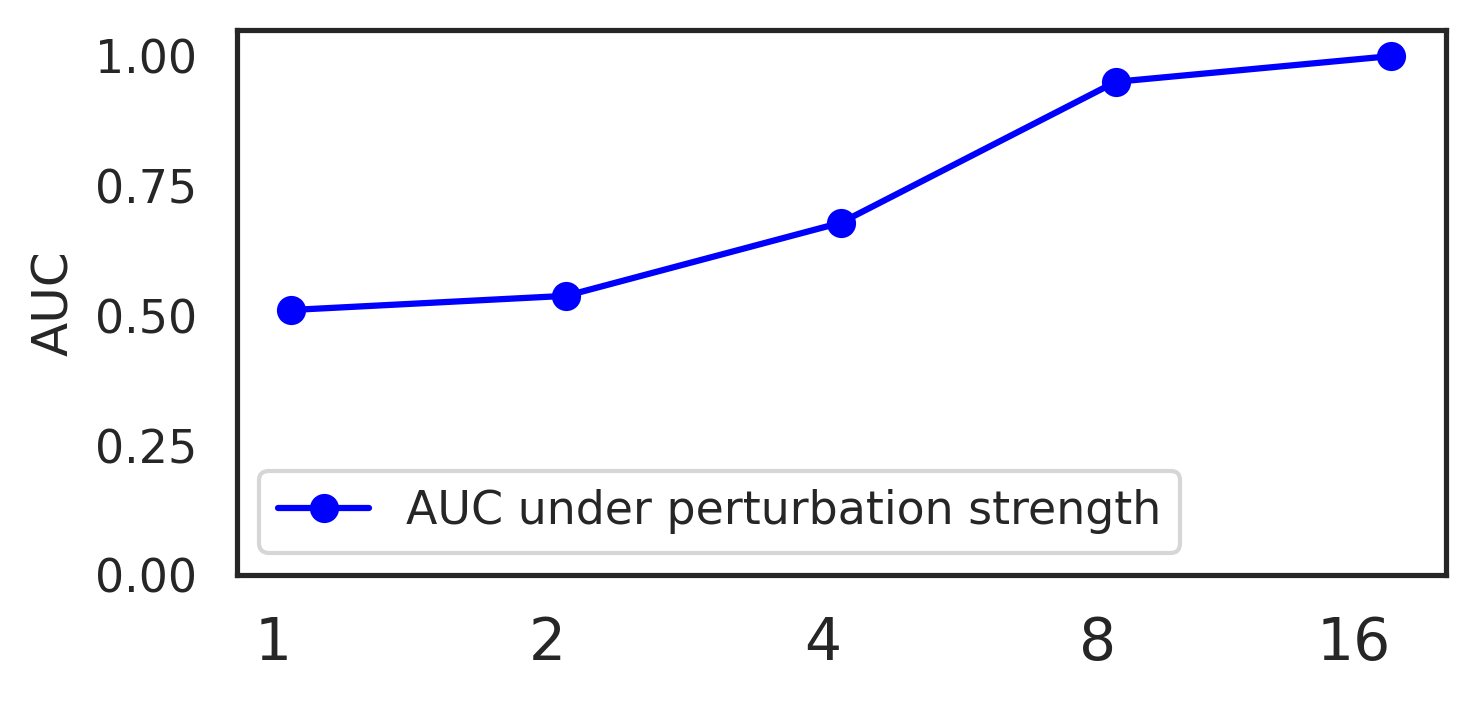}
        \caption{Consistency Score}
    \end{subfigure}
    \caption{Impact of perturbation strength for OD\_frcnn\_r50\_SEG\_gcnet\_r101}
\end{figure}

\begin{figure}[h!]
    \centering
    \begin{subfigure}{.475\linewidth}
        \centering
        \par\medskip
        \includegraphics[width=\linewidth]{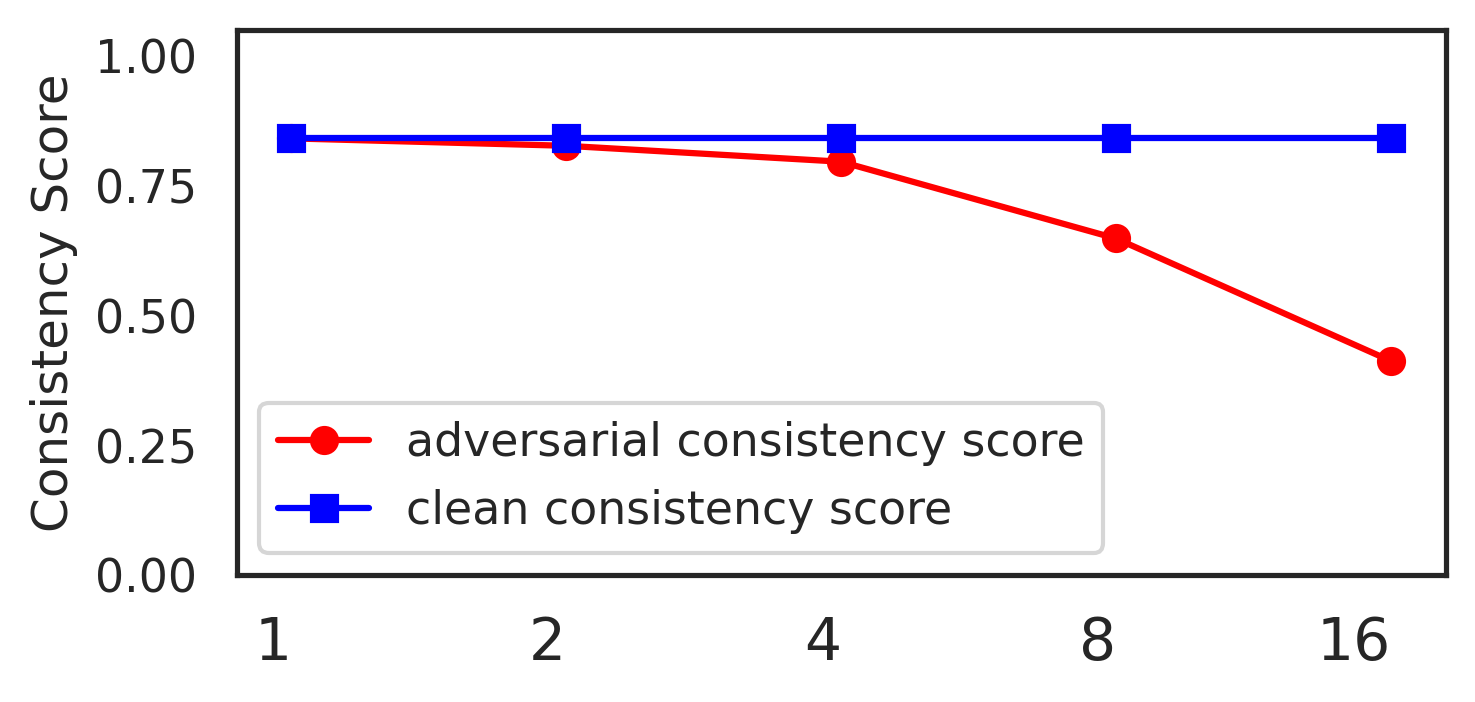}
        \caption{Consistency Score}
    \end{subfigure}
    \hfill
    \begin{subfigure}{.475\linewidth}
        \centering
        \par\medskip
        \includegraphics[width=\linewidth]{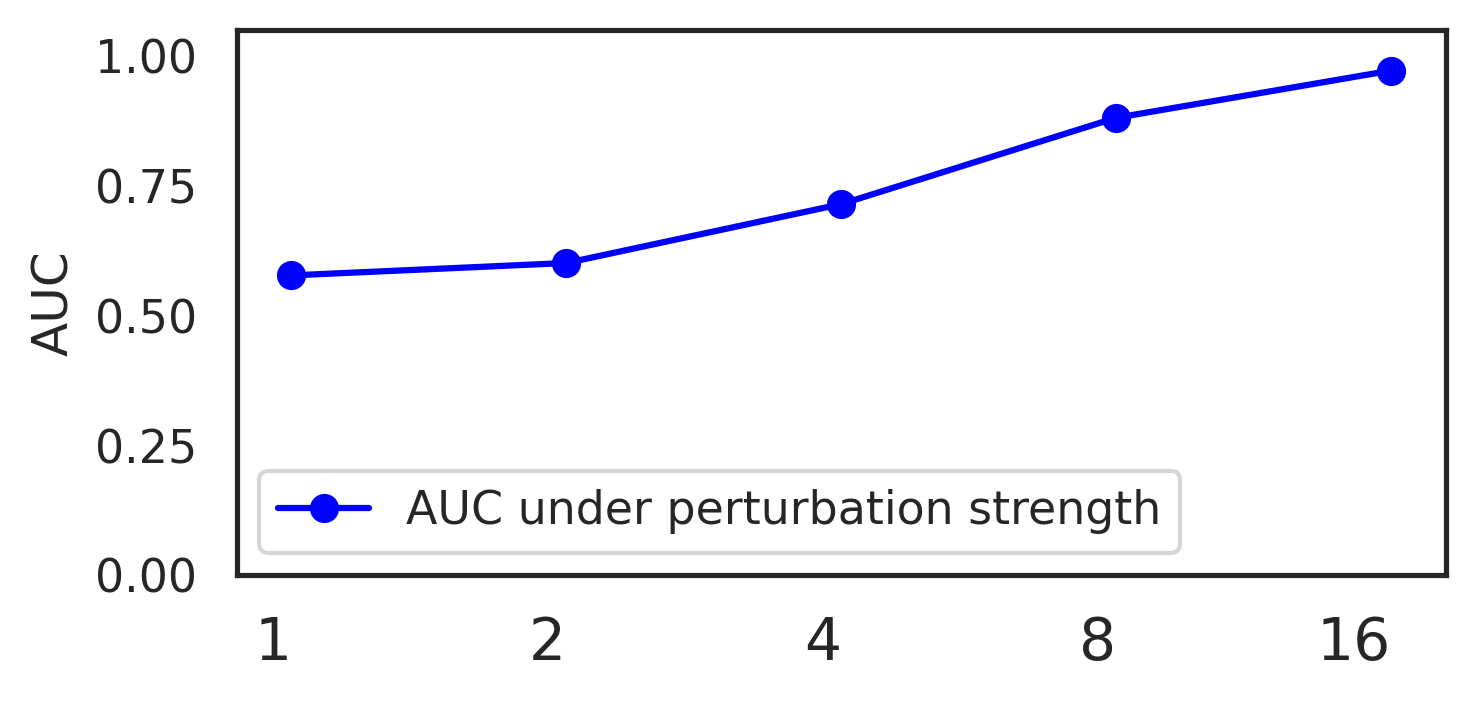}
        \caption{Consistency Score}
    \end{subfigure}
    \caption{Impact of perturbation strength for OD\_frcnn\_swint\_SEG\_mrcnn\_r50}
\end{figure}

\begin{figure}[h!]
    \centering
    \begin{subfigure}{.475\linewidth}
        \centering
        \par\medskip
        \includegraphics[width=\linewidth]{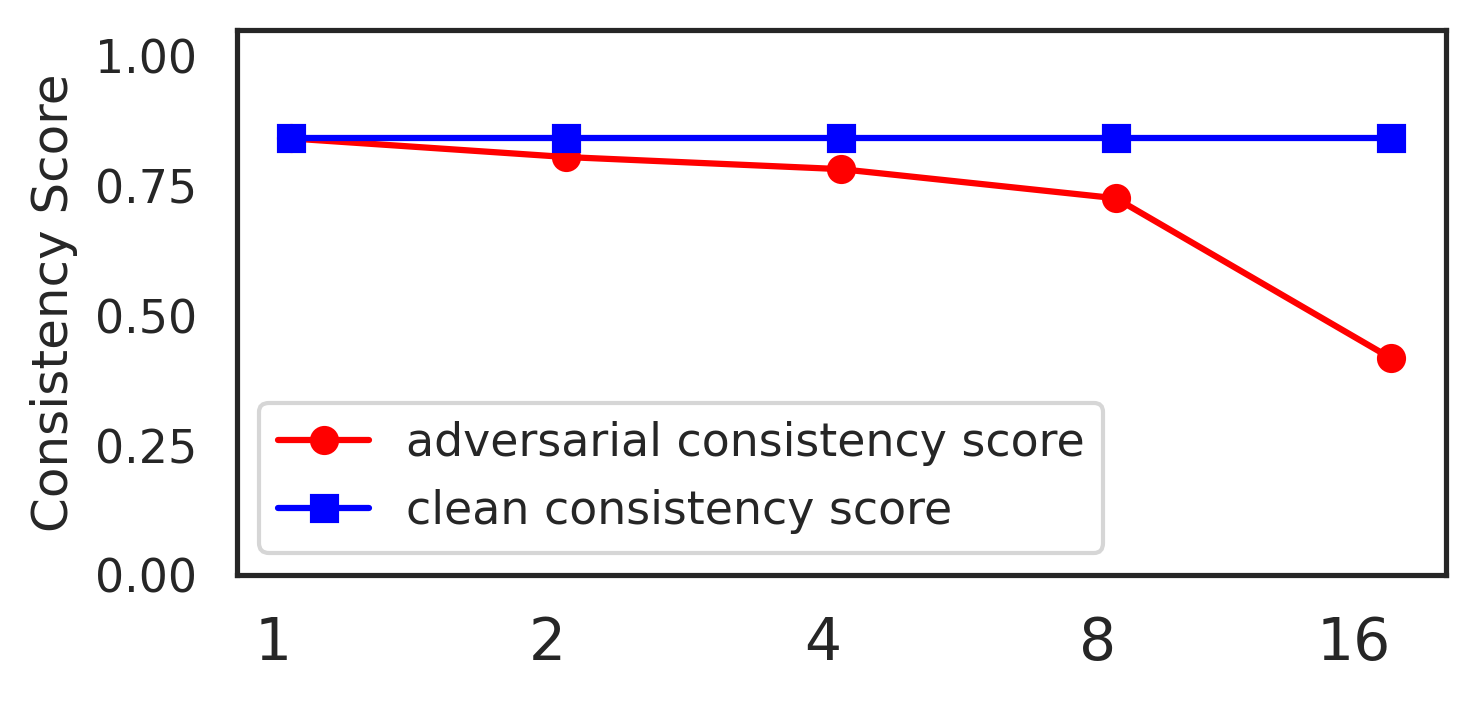}
        \caption{Consistency Score}
    \end{subfigure}
    \hfill
    \begin{subfigure}{.475\linewidth}
        \centering
        \par\medskip
        \includegraphics[width=\linewidth]{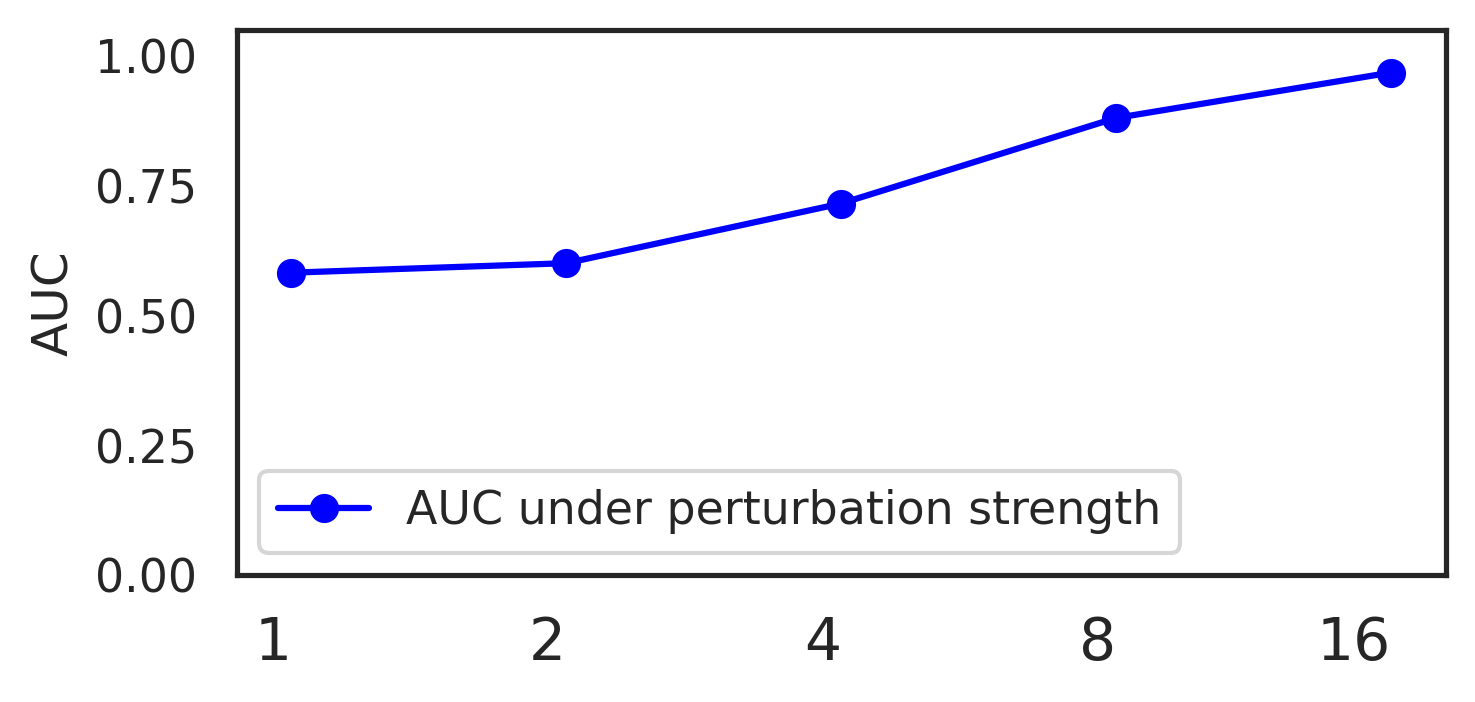}
        \caption{Consistency Score}
    \end{subfigure}
    \caption{Impact of perturbation strength for OD\_frcnn\_swint\_SEG\_gcnet\_r50}
\end{figure}

\begin{figure}[h!]
    \centering
    \begin{subfigure}{.475\linewidth}
        \centering
        \par\medskip
        \includegraphics[width=\linewidth]{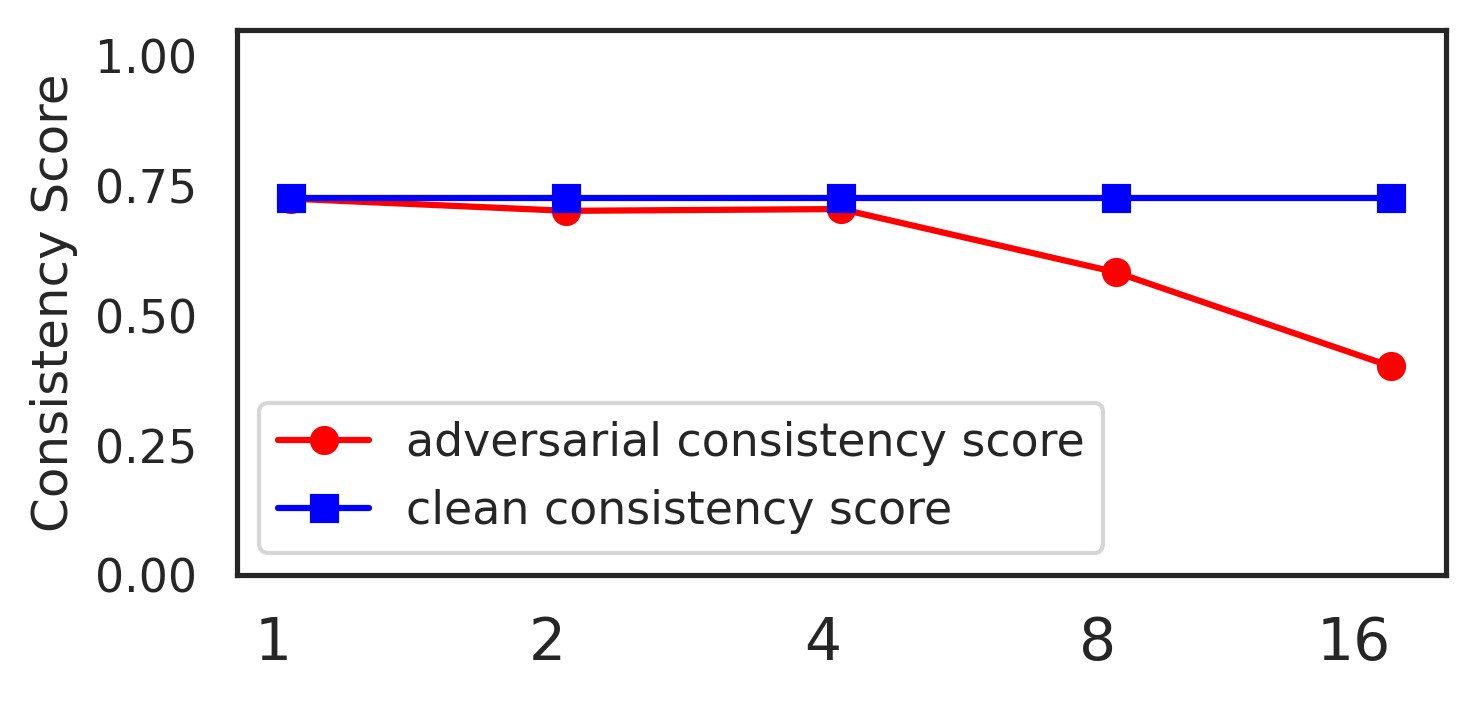}
        \caption{Consistency Score}
    \end{subfigure}
    \hfill
    \begin{subfigure}{.475\linewidth}
        \centering
        \par\medskip
        \includegraphics[width=\linewidth]{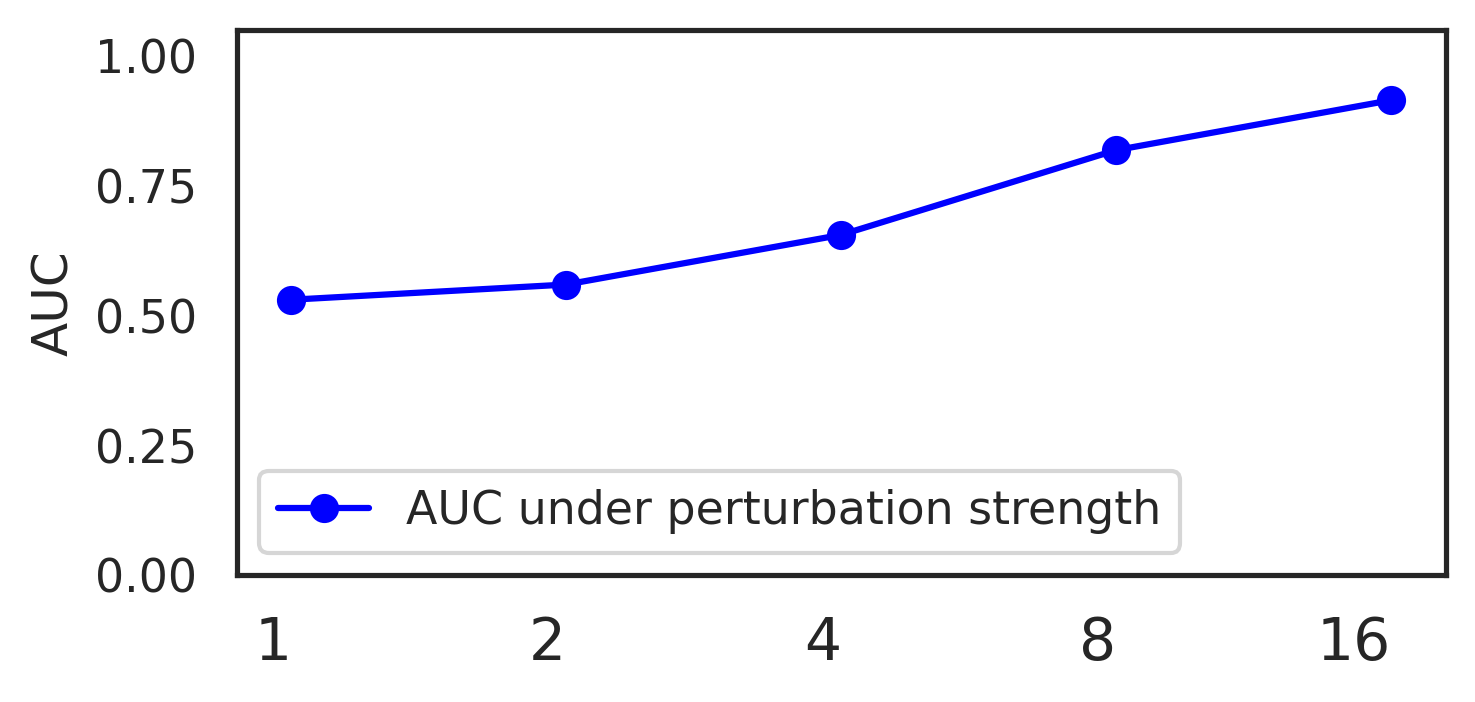}
        \caption{Consistency Score}
    \end{subfigure}
    \caption{Impact of perturbation strength for OD\_frcnn\_swint\_SEG\_mask2former}
\end{figure}

\begin{figure}[h!]
    \centering
    \begin{subfigure}{.475\linewidth}
        \centering
        \par\medskip
        \includegraphics[width=\linewidth]{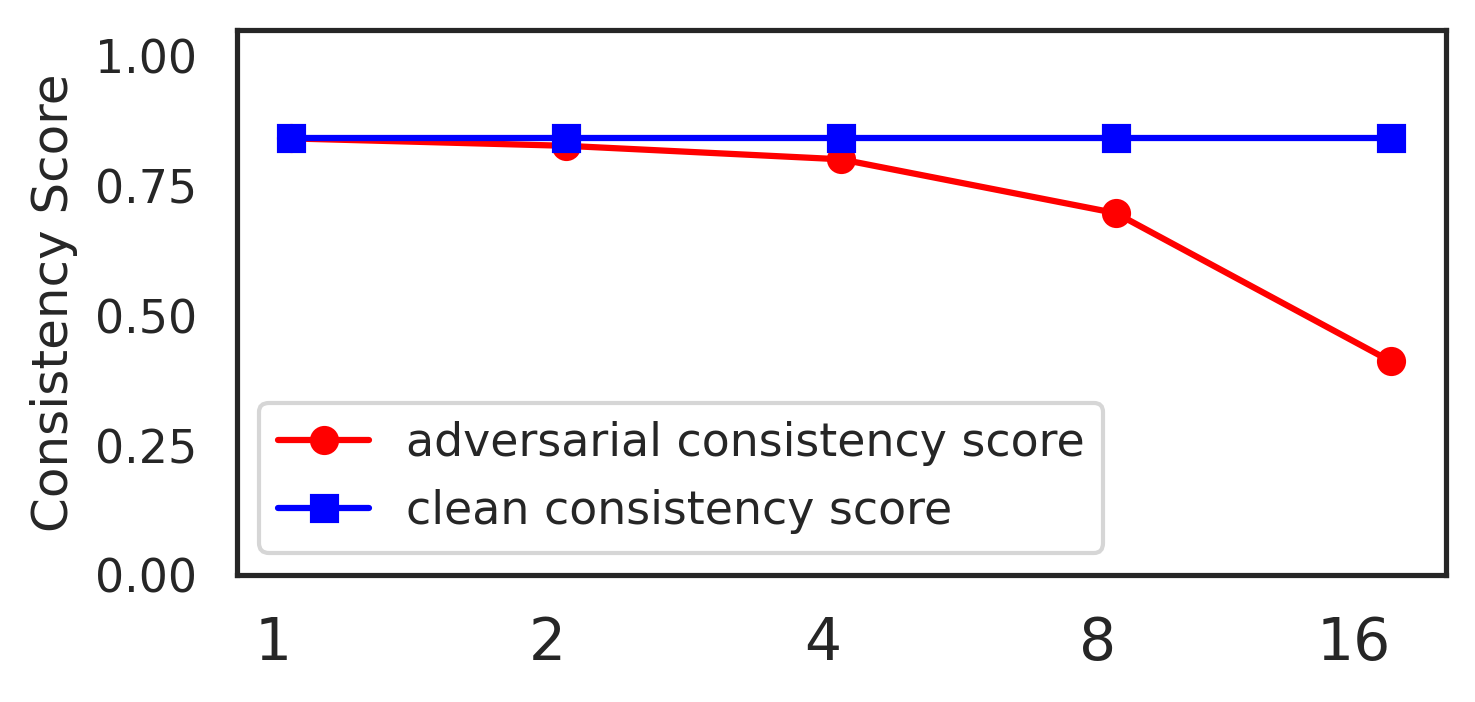}
        \caption{Consistency Score}
    \end{subfigure}
    \hfill
    \begin{subfigure}{.475\linewidth}
        \centering
        \par\medskip
        \includegraphics[width=\linewidth]{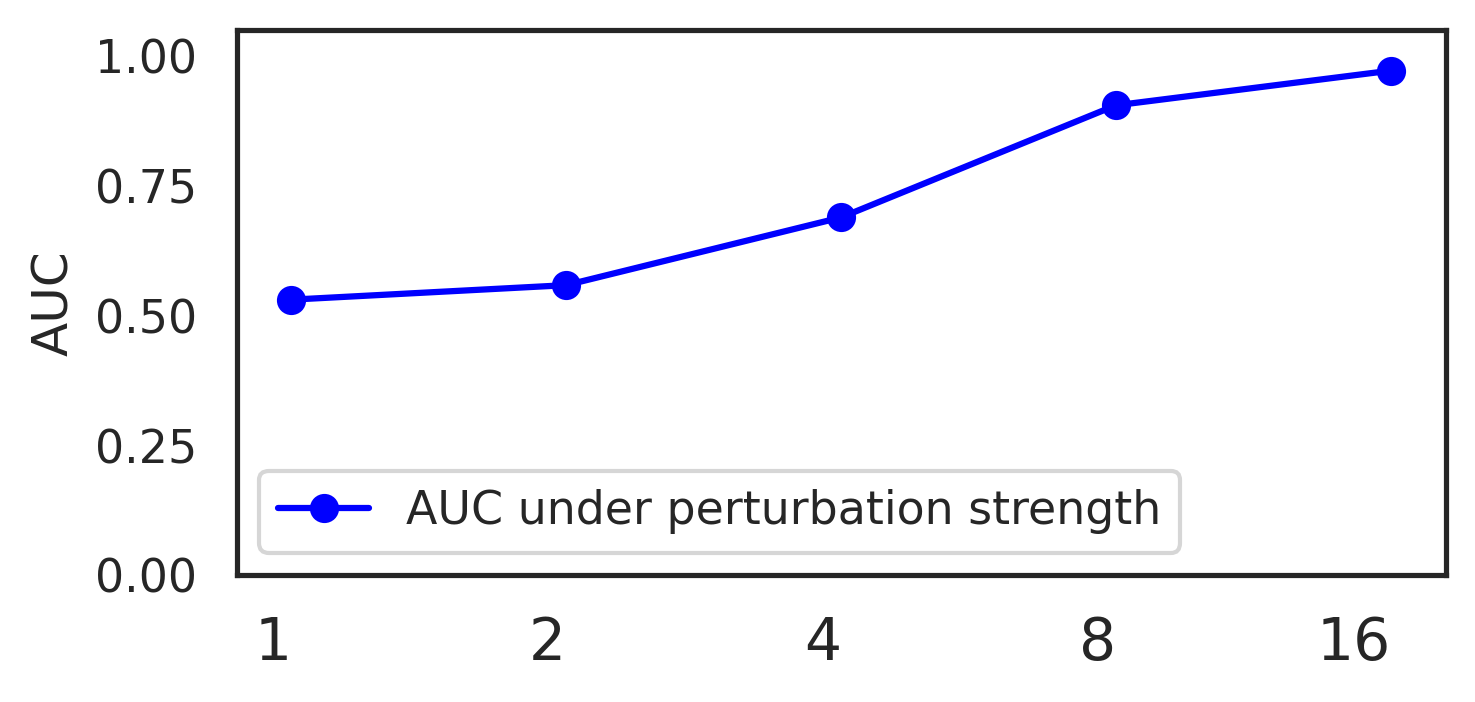}
        \caption{Consistency Score}
    \end{subfigure}
    \caption{Impact of perturbation strength for OD\_frcnn\_swint\_SEG\_mrcnn\_r101}
\end{figure}

\begin{figure}[h!]
    \centering
    \begin{subfigure}{.475\linewidth}
        \centering
        \par\medskip
        \includegraphics[width=\linewidth]{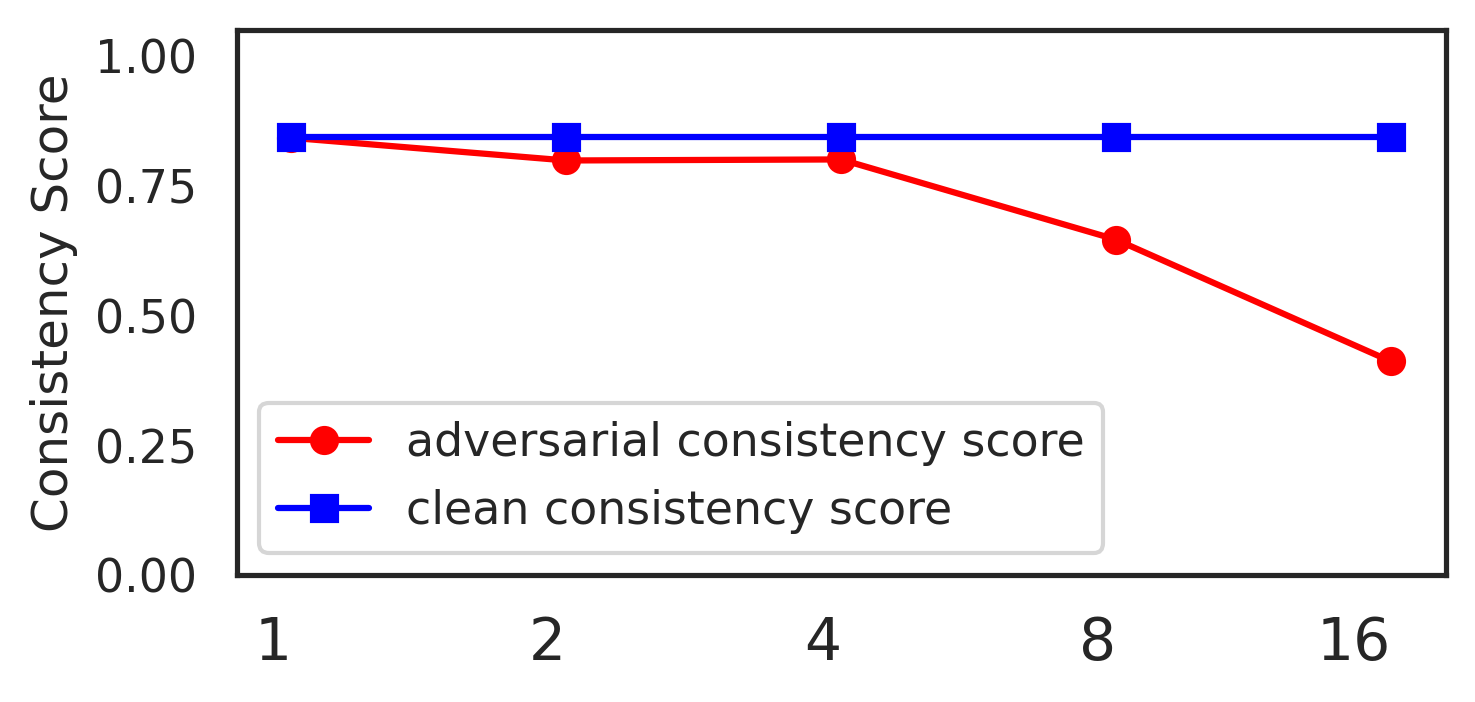}
        \caption{Consistency Score}
    \end{subfigure}
    \hfill
    \begin{subfigure}{.475\linewidth}
        \centering
        \par\medskip
        \includegraphics[width=\linewidth]{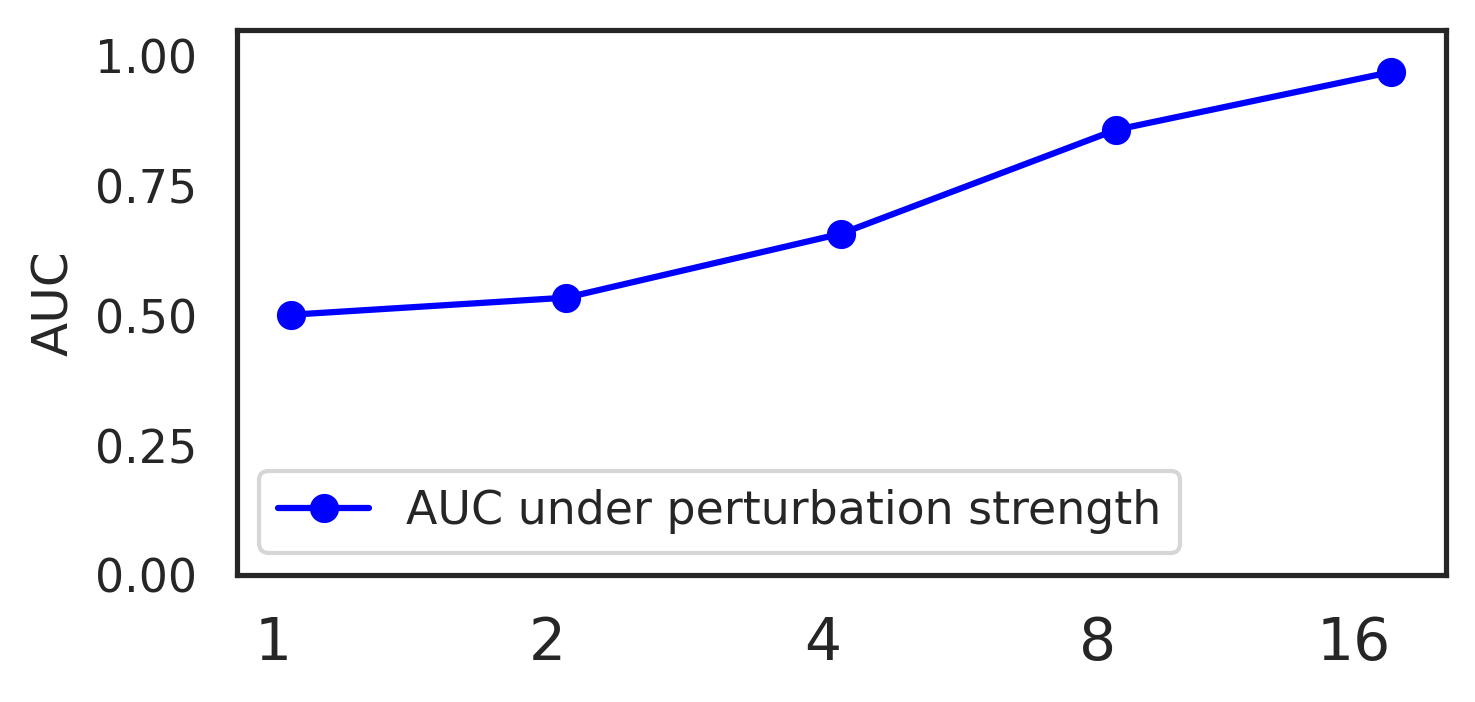}
        \caption{Consistency Score}
    \end{subfigure}
    \caption{Impact of perturbation strength for OD\_frcnn\_swint\_SEG\_gcnet\_r101}
\end{figure}

\begin{figure}[h!]
    \centering
    \begin{subfigure}{.475\linewidth}
        \centering
        \par\medskip
        \includegraphics[width=\linewidth]{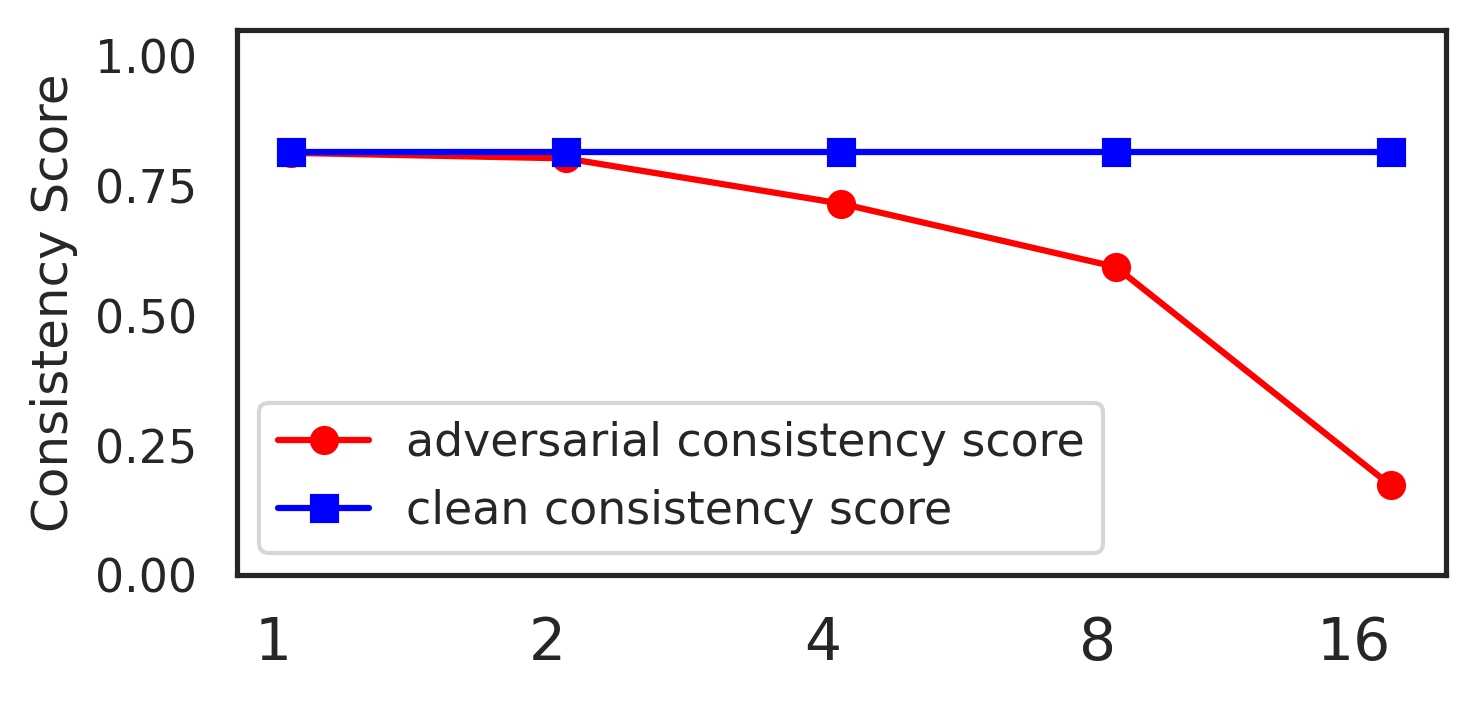}
        \caption{Consistency Score}
    \end{subfigure}
    \hfill
    \begin{subfigure}{.475\linewidth}
        \centering
        \par\medskip
        \includegraphics[width=\linewidth]{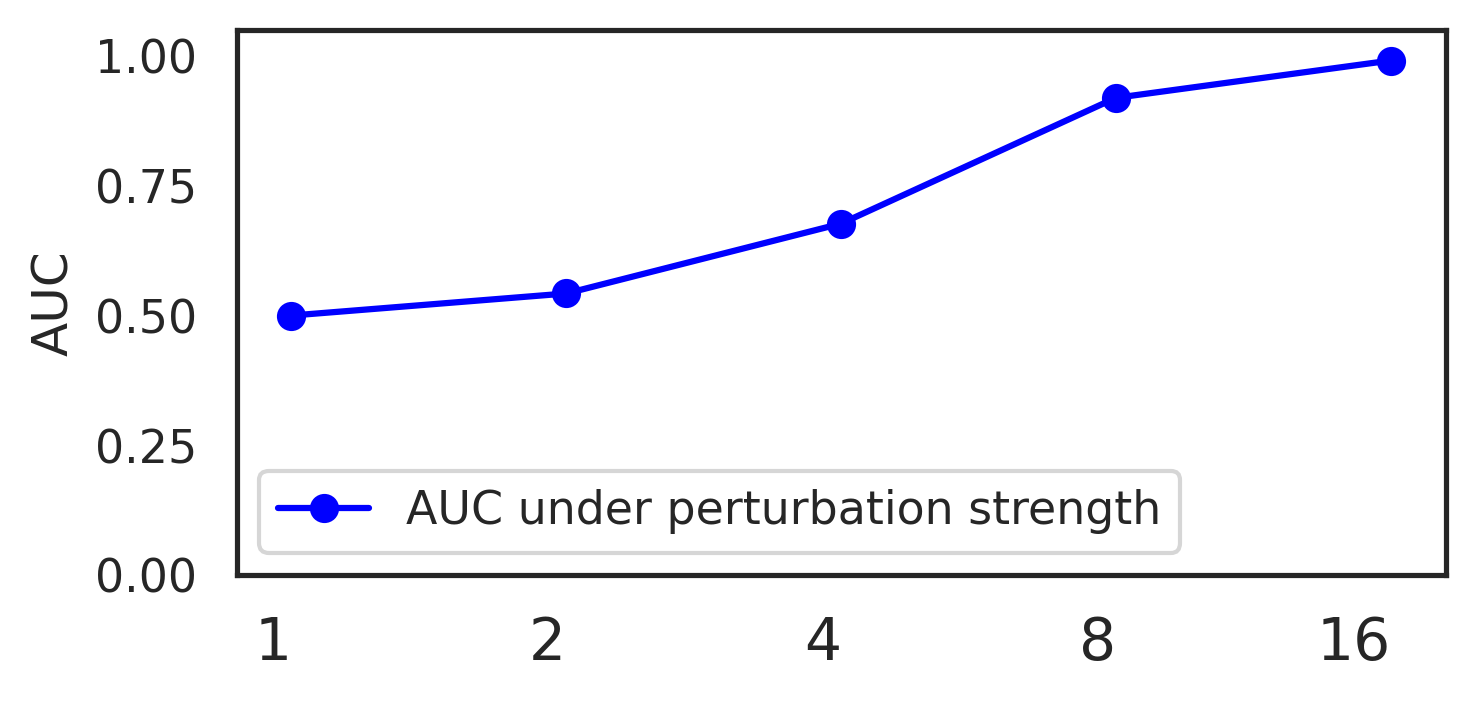}
        \caption{Consistency Score}
    \end{subfigure}
    \caption{Impact of perturbation strength for OD\_retinanet\_pvtv2\_SEG\_mrcnn\_r50}
\end{figure}

\begin{figure}[h!]
    \centering
    \begin{subfigure}{.475\linewidth}
        \centering
        \par\medskip
        \includegraphics[width=\linewidth]{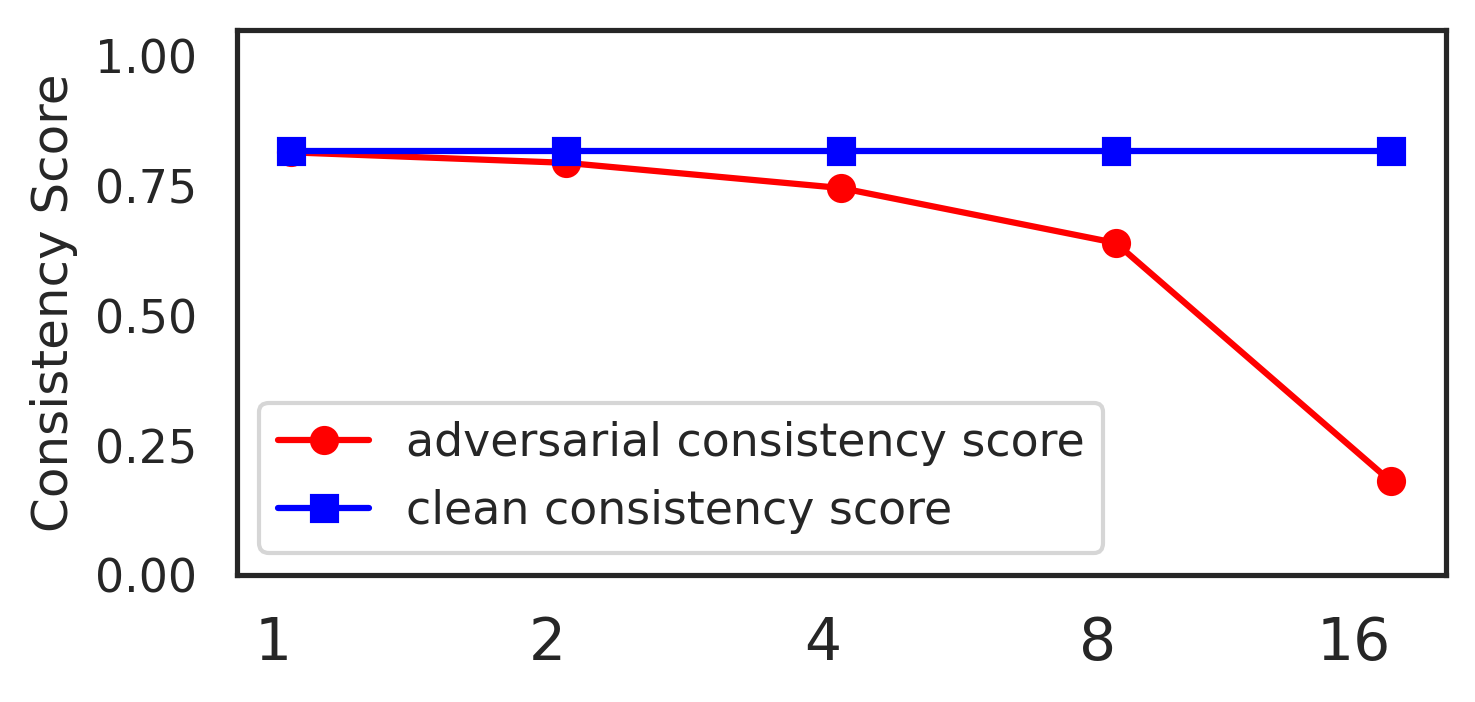}
        \caption{Consistency Score}
    \end{subfigure}
    \hfill
    \begin{subfigure}{.475\linewidth}
        \centering
        \par\medskip
        \includegraphics[width=\linewidth]{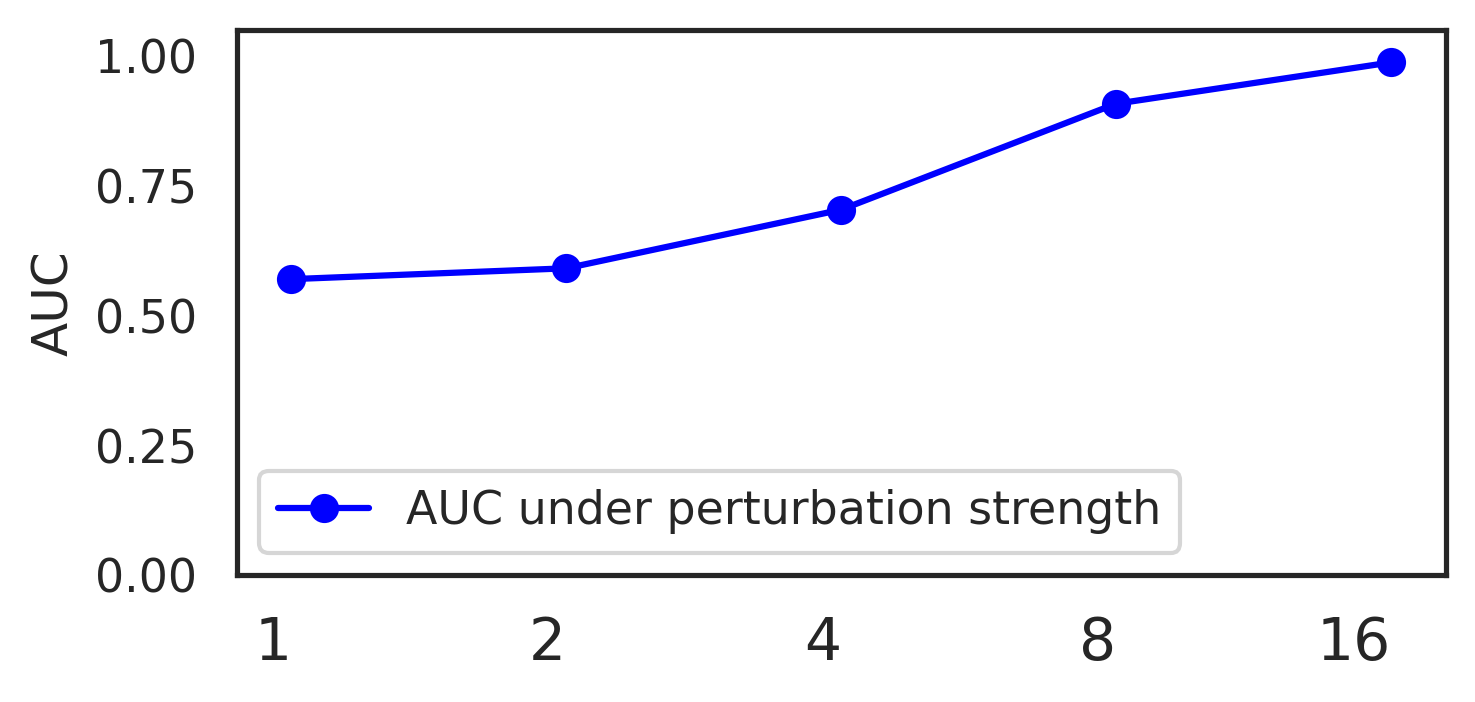}
        \caption{Consistency Score}
    \end{subfigure}
    \caption{Impact of perturbation strength for OD\_retinanet\_pvtv2\_SEG\_gcnet\_r50}
\end{figure}

\begin{figure}[h!]
    \centering
    \begin{subfigure}{.475\linewidth}
        \centering
        \par\medskip
        \includegraphics[width=\linewidth]{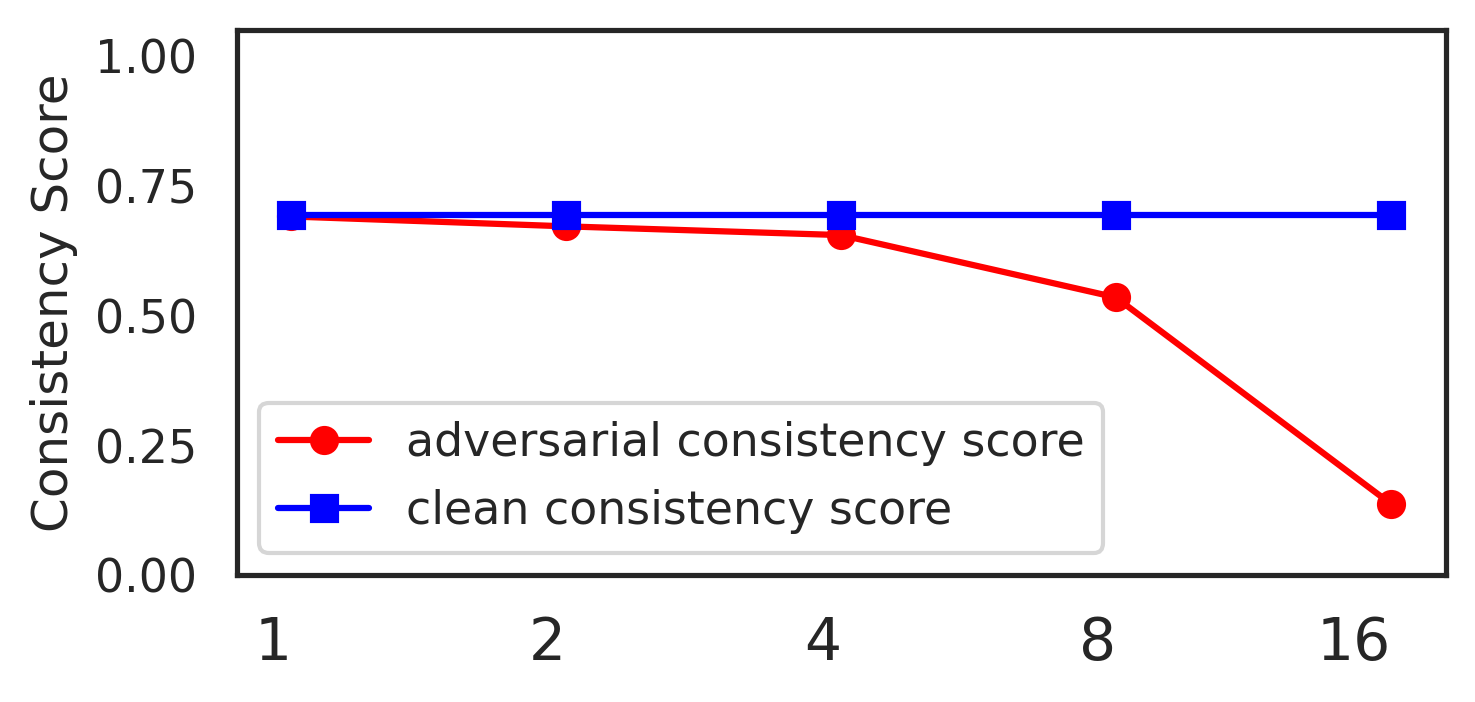}
        \caption{Consistency Score}
    \end{subfigure}
    \hfill
    \begin{subfigure}{.475\linewidth}
        \centering
        \par\medskip
        \includegraphics[width=\linewidth]{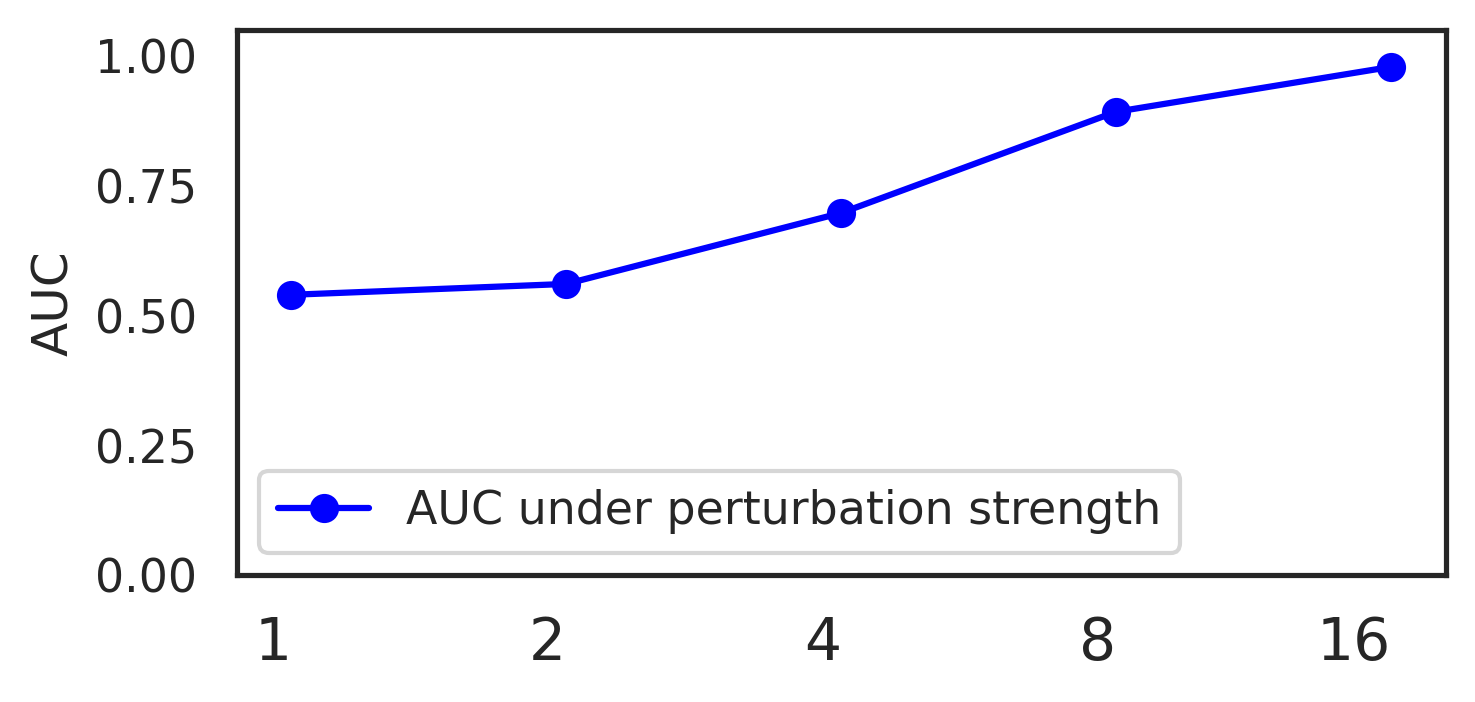}
        \caption{Consistency Score}
    \end{subfigure}
    \caption{Impact of perturbation strength for OD\_retinanet\_pvtv2\_SEG\_mask2former}
\end{figure}

\begin{figure}[h!]
    \centering
    \begin{subfigure}{.475\linewidth}
        \centering
        \par\medskip
        \includegraphics[width=\linewidth]{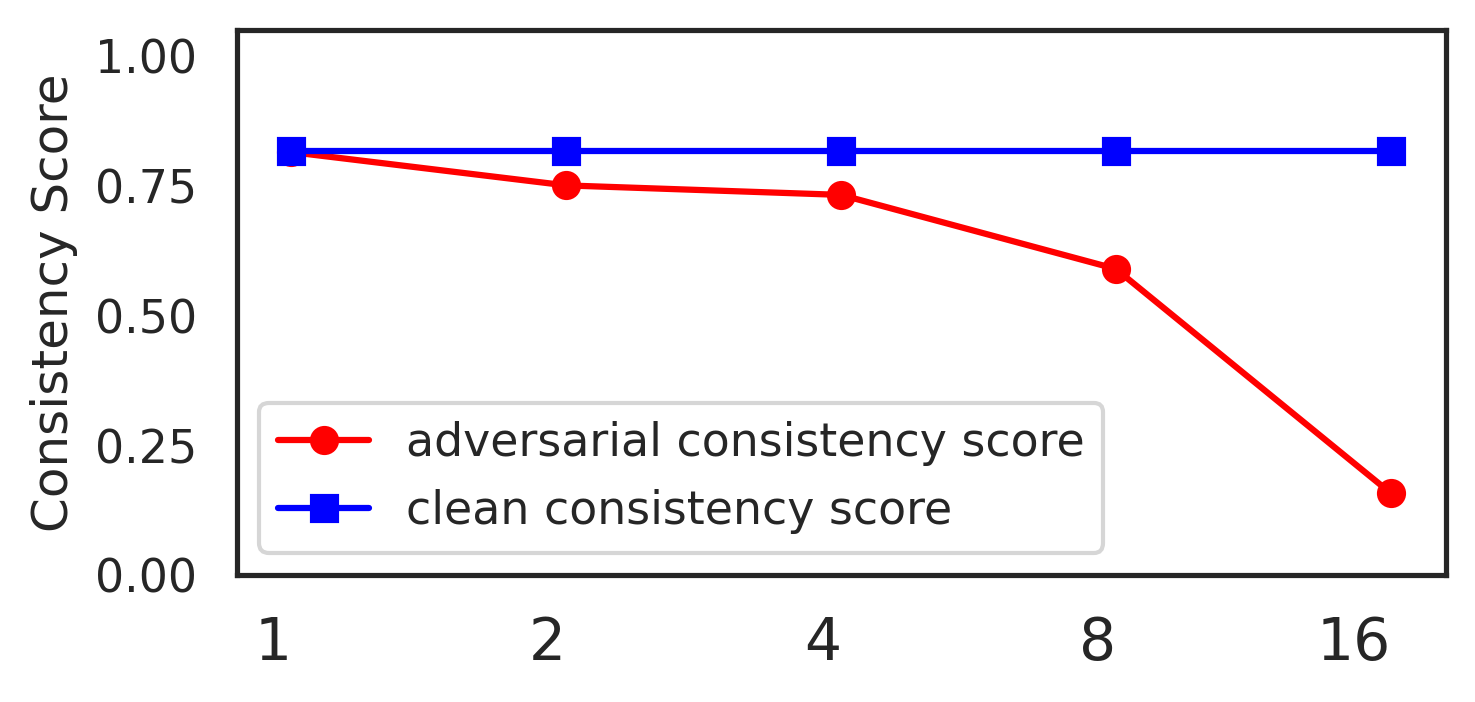}
        \caption{Consistency Score}
    \end{subfigure}
    \hfill
    \begin{subfigure}{.475\linewidth}
        \centering
        \par\medskip
        \includegraphics[width=\linewidth]{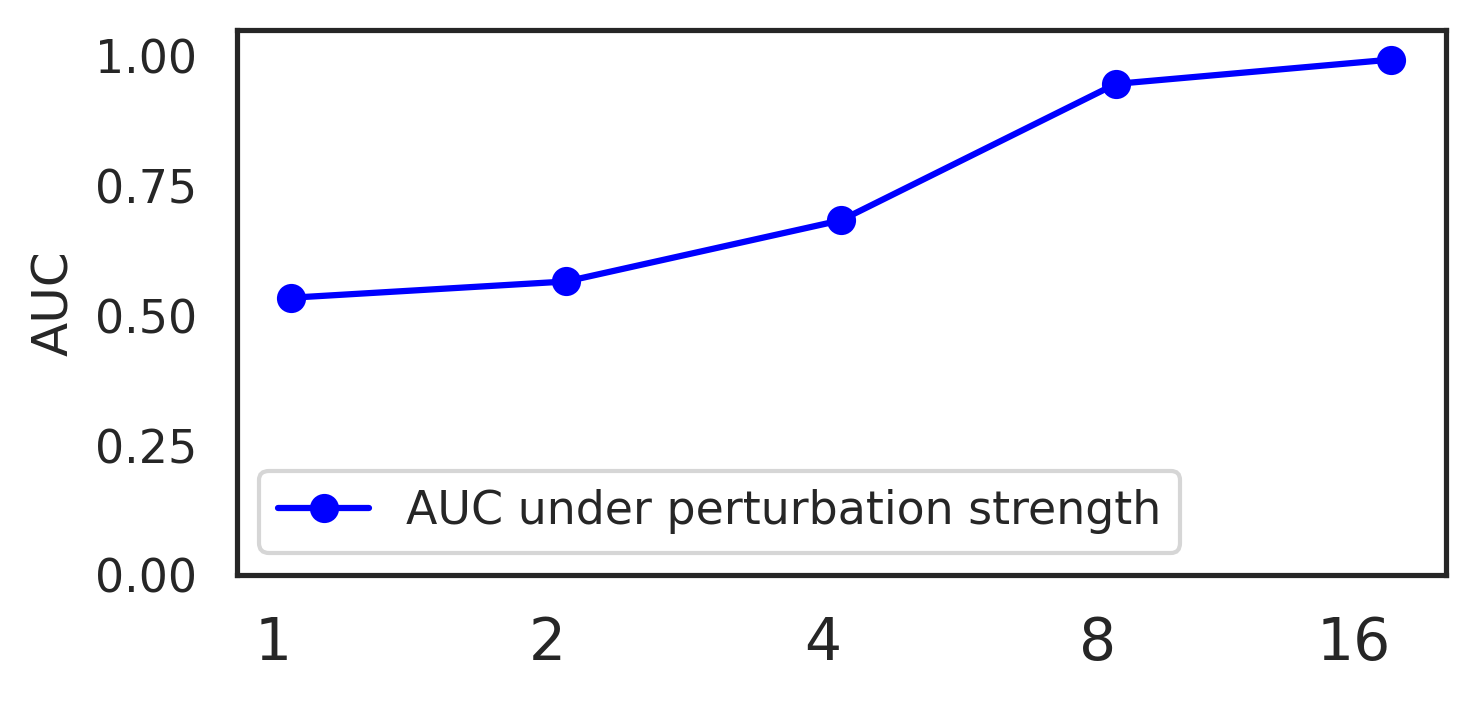}
        \caption{Consistency Score}
    \end{subfigure}
    \caption{Impact of perturbation strength for OD\_retinanet\_pvtv2\_SEG\_mrcnn\_r101}
\end{figure}

\begin{figure}[h!]
    \centering
    \begin{subfigure}{.475\linewidth}
        \centering
        \par\medskip
        \includegraphics[width=\linewidth]{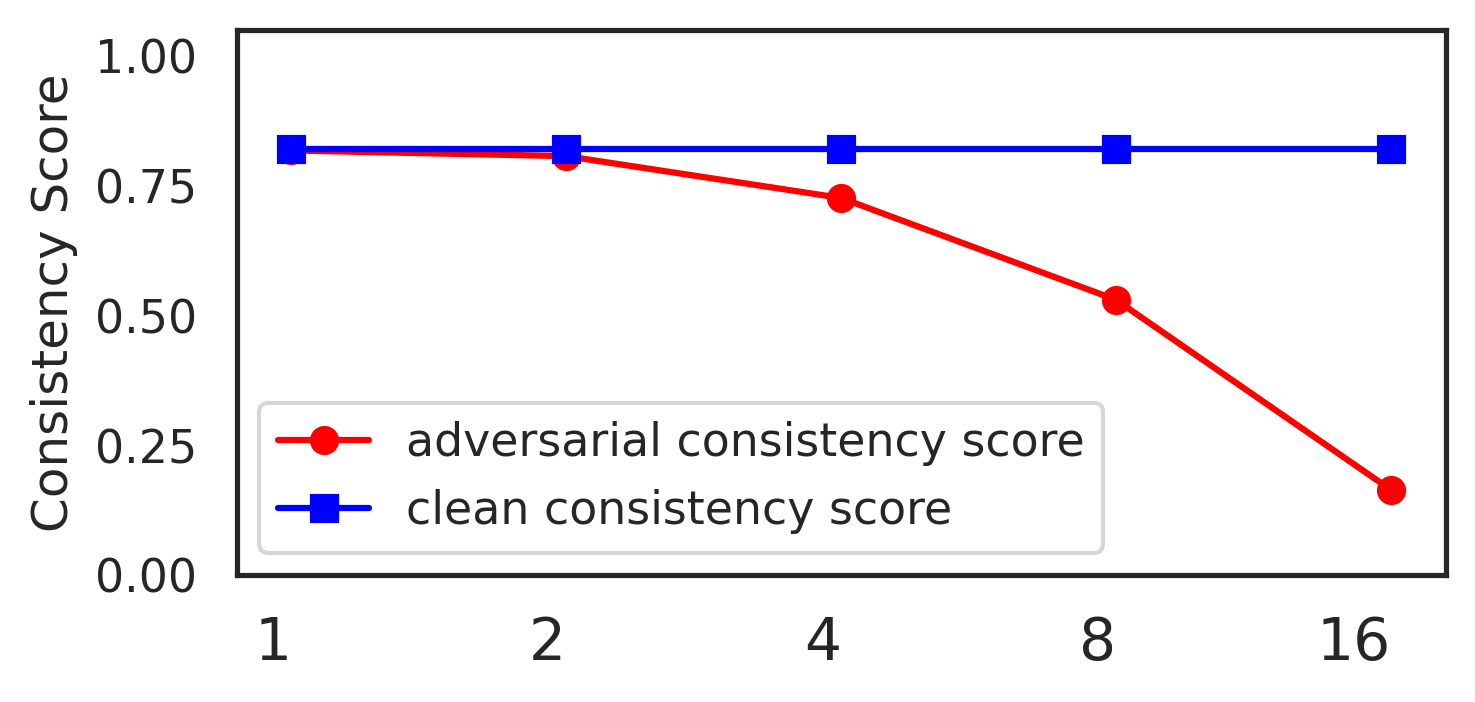}
        \caption{Consistency Score}
    \end{subfigure}
    \hfill
    \begin{subfigure}{.475\linewidth}
        \centering
        \par\medskip
        \includegraphics[width=\linewidth]{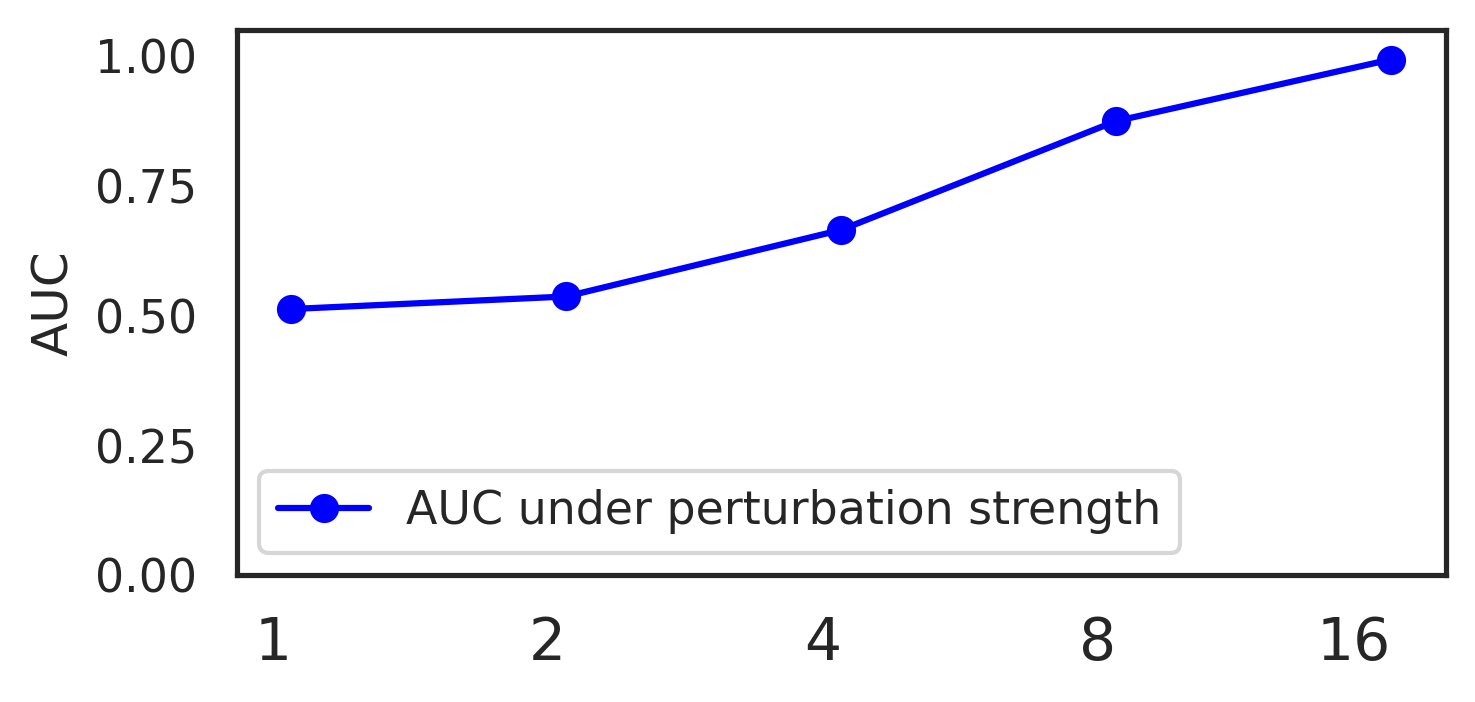}
        \caption{Consistency Score}
    \end{subfigure}
    \caption{Impact of perturbation strength for OD\_retinanet\_pvtv2\_SEG\_gcnet\_r101}
\end{figure}

\begin{figure}[h!]
    \centering
    \begin{subfigure}{.475\linewidth}
        \centering
        \par\medskip
        \includegraphics[width=\linewidth]{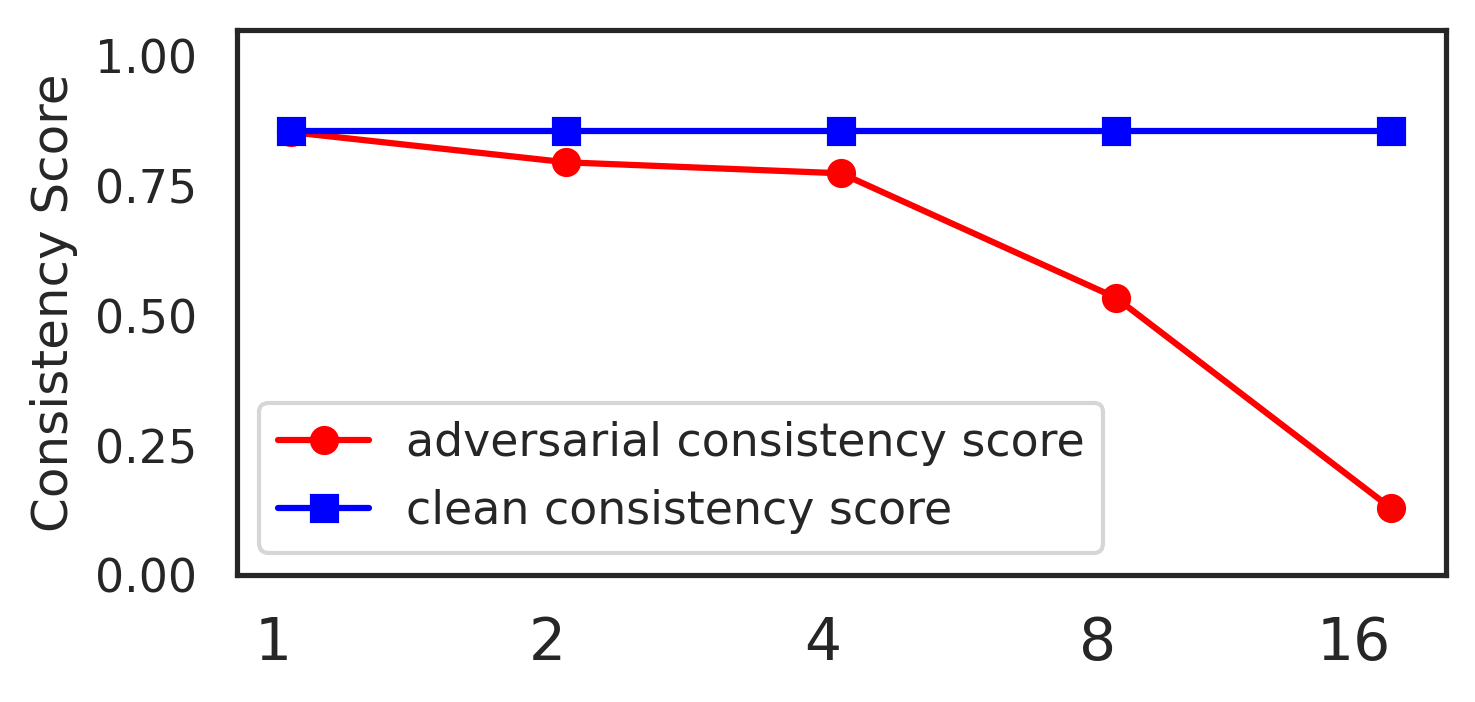}
        \caption{Consistency Score}
    \end{subfigure}
    \hfill
    \begin{subfigure}{.475\linewidth}
        \centering
        \par\medskip
        \includegraphics[width=\linewidth]{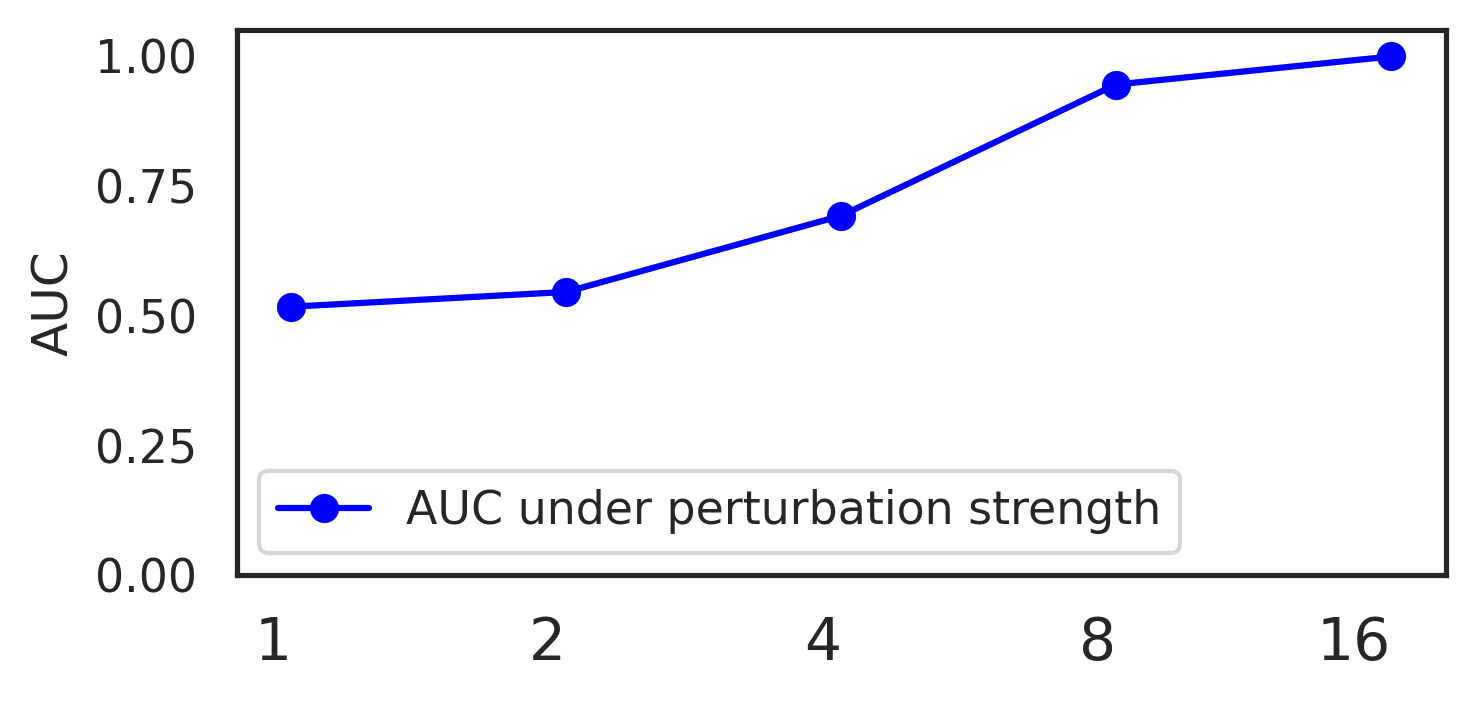}
        \caption{Consistency Score}
    \end{subfigure}
    \caption{Impact of perturbation strength for OD\_frcnn\_r101\_SEG\_mrcnn\_r50}
\end{figure}

\begin{figure}[h!]
    \centering
    \begin{subfigure}{.475\linewidth}
        \centering
        \par\medskip
        \includegraphics[width=\linewidth]{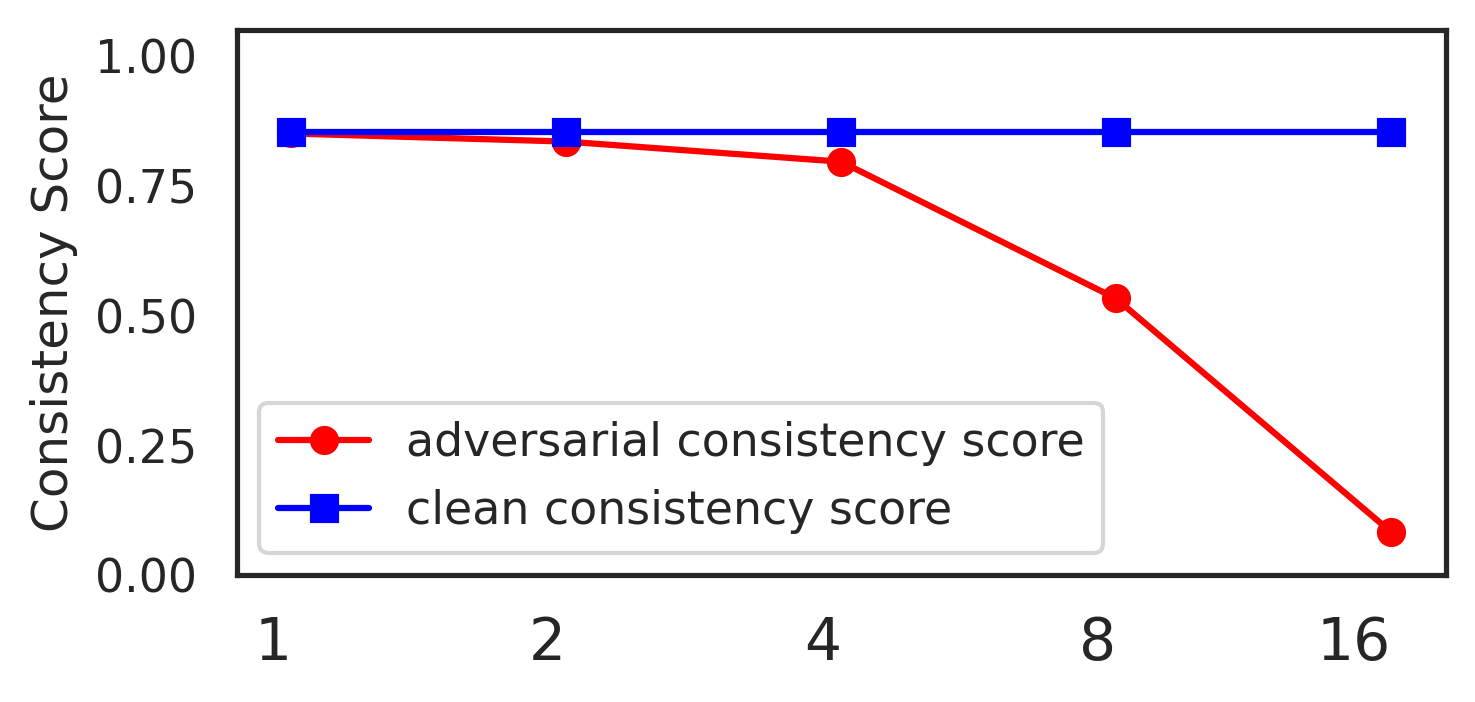}
        \caption{Consistency Score}
    \end{subfigure}
    \hfill
    \begin{subfigure}{.475\linewidth}
        \centering
        \par\medskip
        \includegraphics[width=\linewidth]{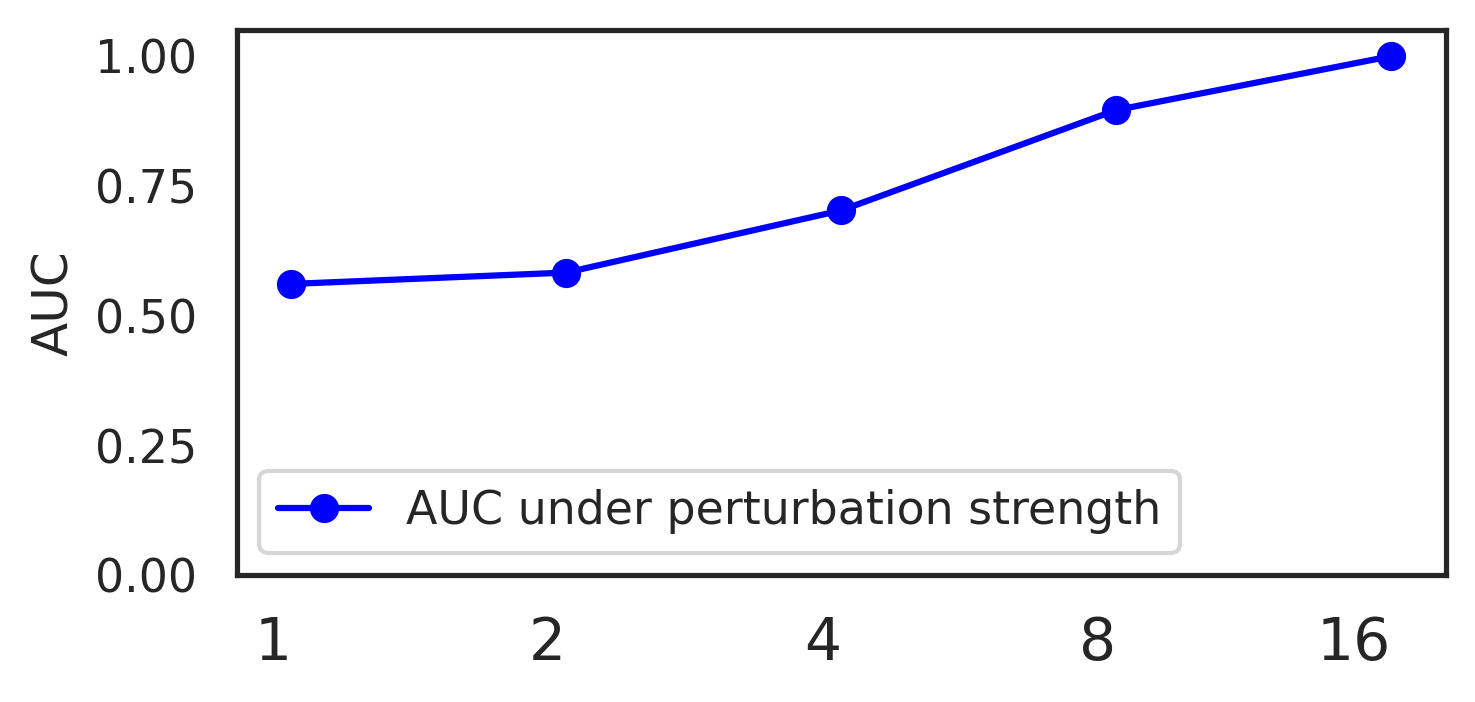}
        \caption{Consistency Score}
    \end{subfigure}
    \caption{Impact of perturbation strength for OD\_frcnn\_r101\_SEG\_gcnet\_r50}
\end{figure}

\begin{figure}[h!]
    \centering
    \begin{subfigure}{.475\linewidth}
        \centering
        \par\medskip
        \includegraphics[width=\linewidth]{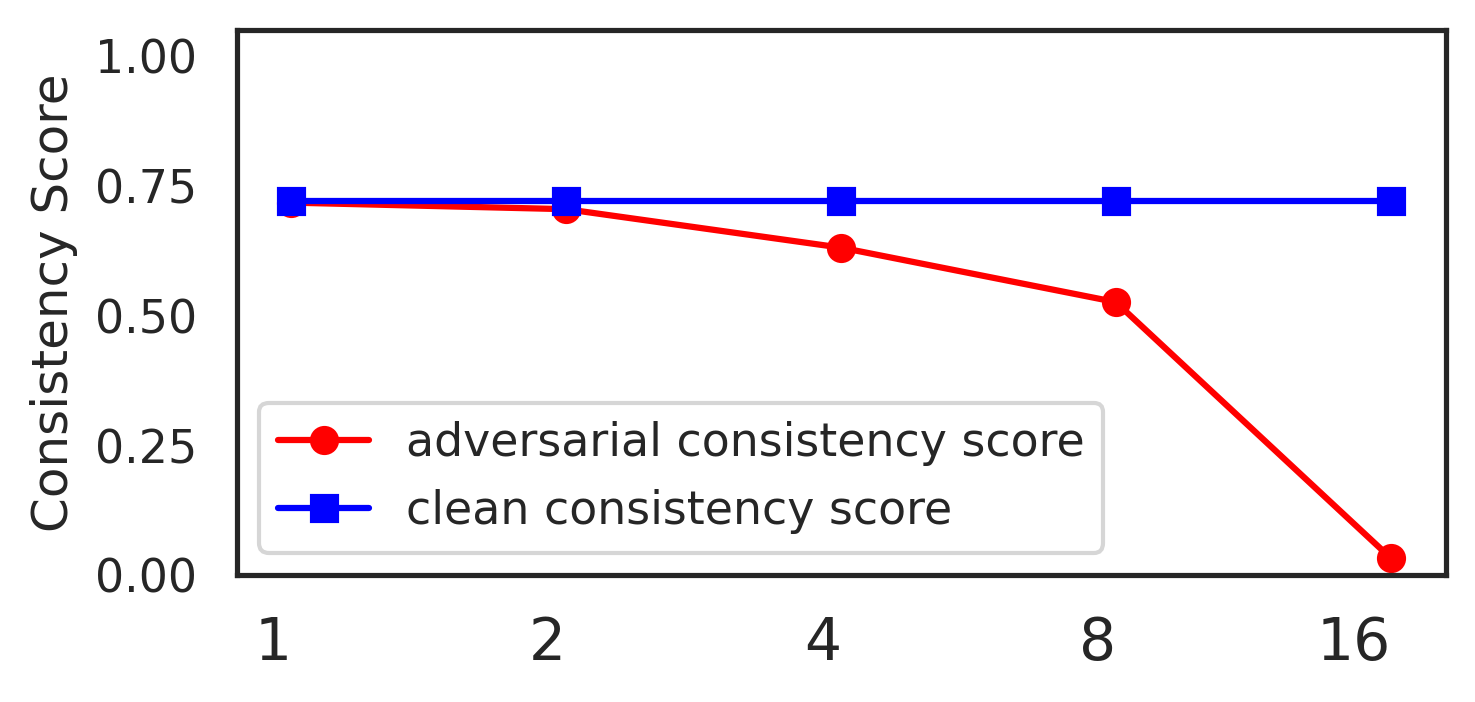}
        \caption{Consistency Score}
    \end{subfigure}
    \hfill
    \begin{subfigure}{.475\linewidth}
        \centering
        \par\medskip
        \includegraphics[width=\linewidth]{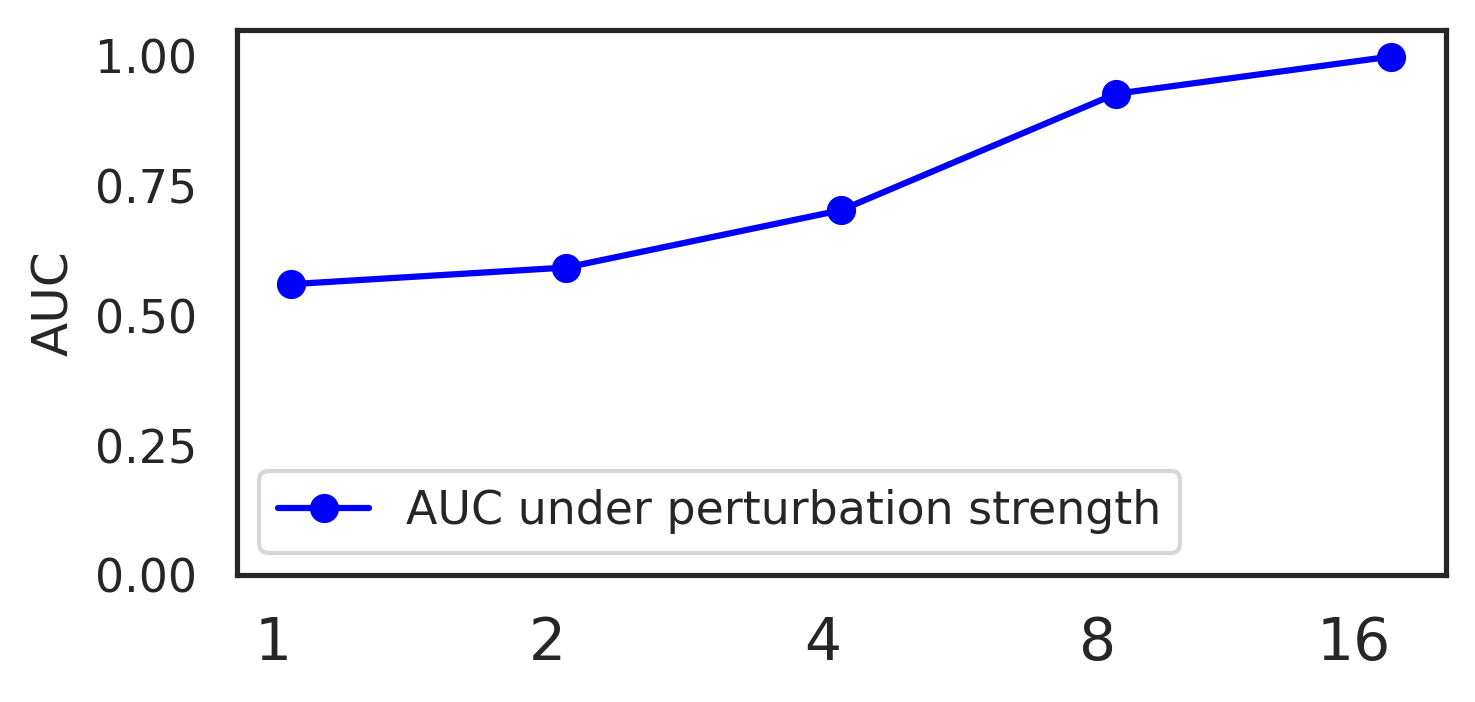}
        \caption{Consistency Score}
    \end{subfigure}
    \caption{Impact of perturbation strength for OD\_frcnn\_r101\_SEG\_mask2former}
\end{figure}

\begin{figure}[h!]
    \centering
    \begin{subfigure}{.475\linewidth}
        \centering
        \par\medskip
        \includegraphics[width=\linewidth]{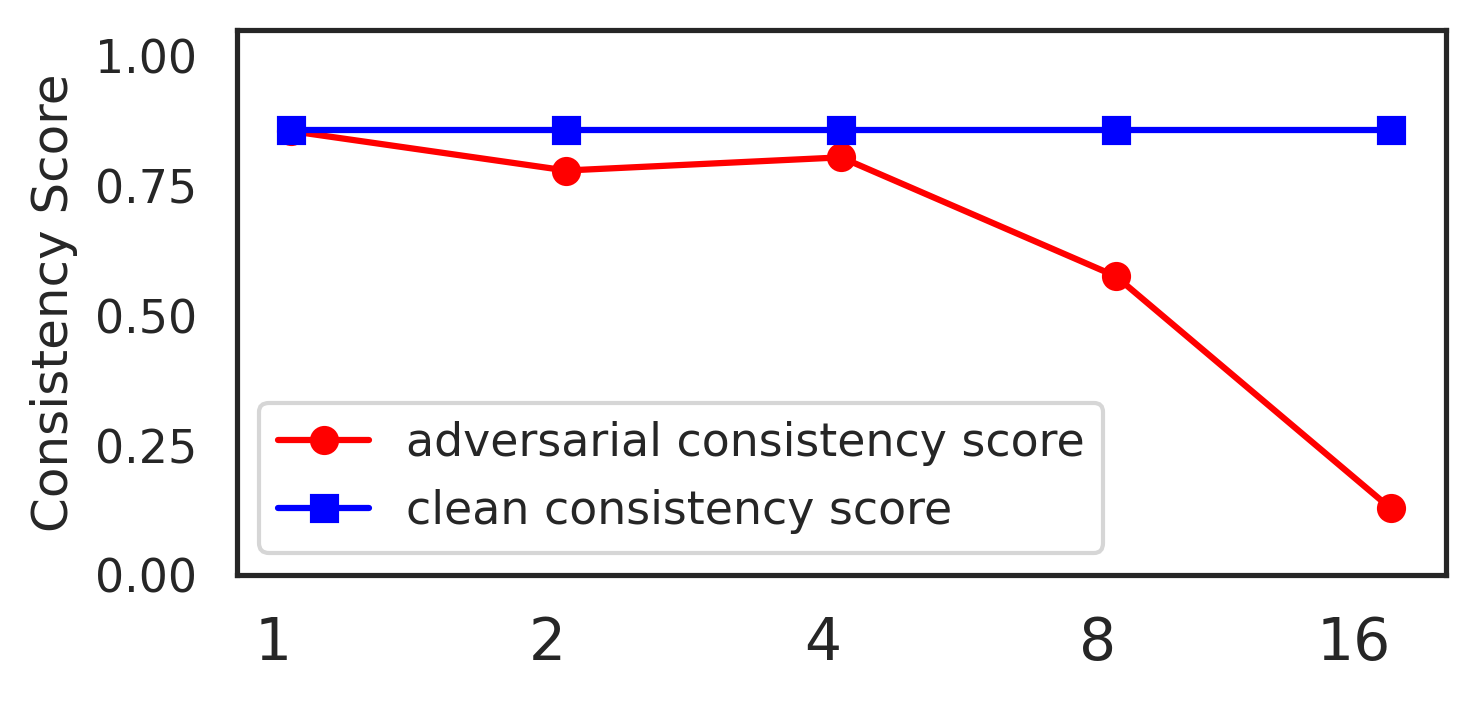}
        \caption{Consistency Score}
    \end{subfigure}
    \hfill
    \begin{subfigure}{.475\linewidth}
        \centering
        \par\medskip
        \includegraphics[width=\linewidth]{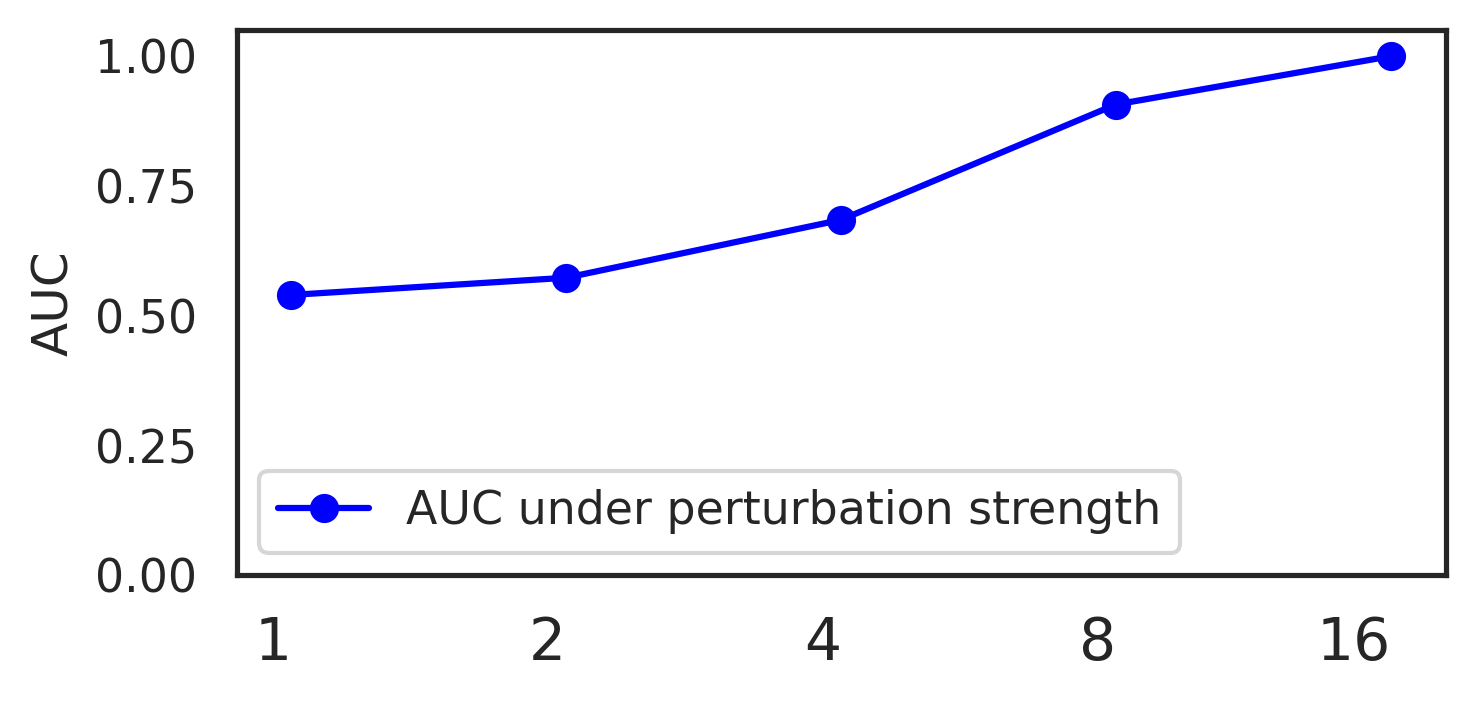}
        \caption{Consistency Score}
    \end{subfigure}
    \caption{Impact of perturbation strength for OD\_frcnn\_r101\_SEG\_mrcnn\_r101}
\end{figure}

\begin{figure}[h!]
    \centering
    \begin{subfigure}{.475\linewidth}
        \centering
        \par\medskip
        \includegraphics[width=\linewidth]{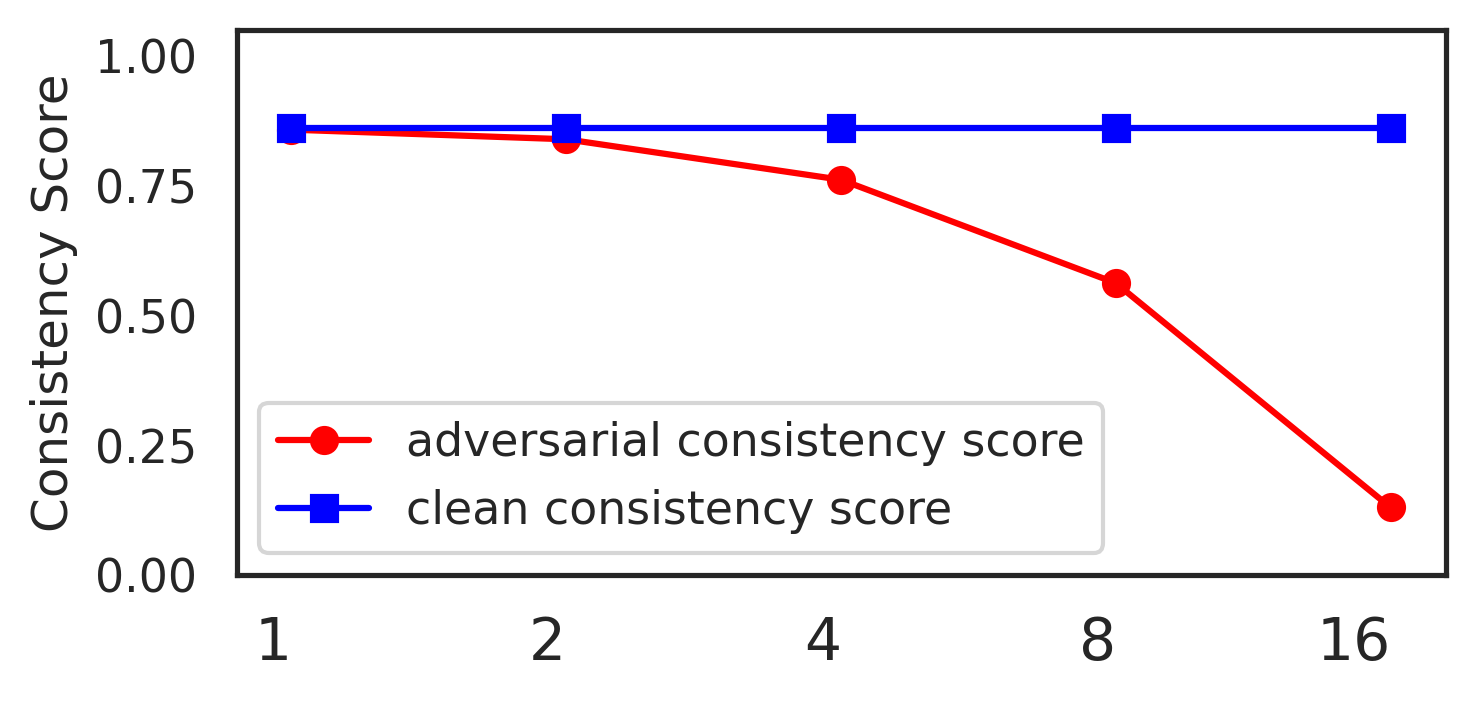}
        \caption{Consistency Score}
    \end{subfigure}
    \hfill
    \begin{subfigure}{.475\linewidth}
        \centering
        \par\medskip
        \includegraphics[width=\linewidth]{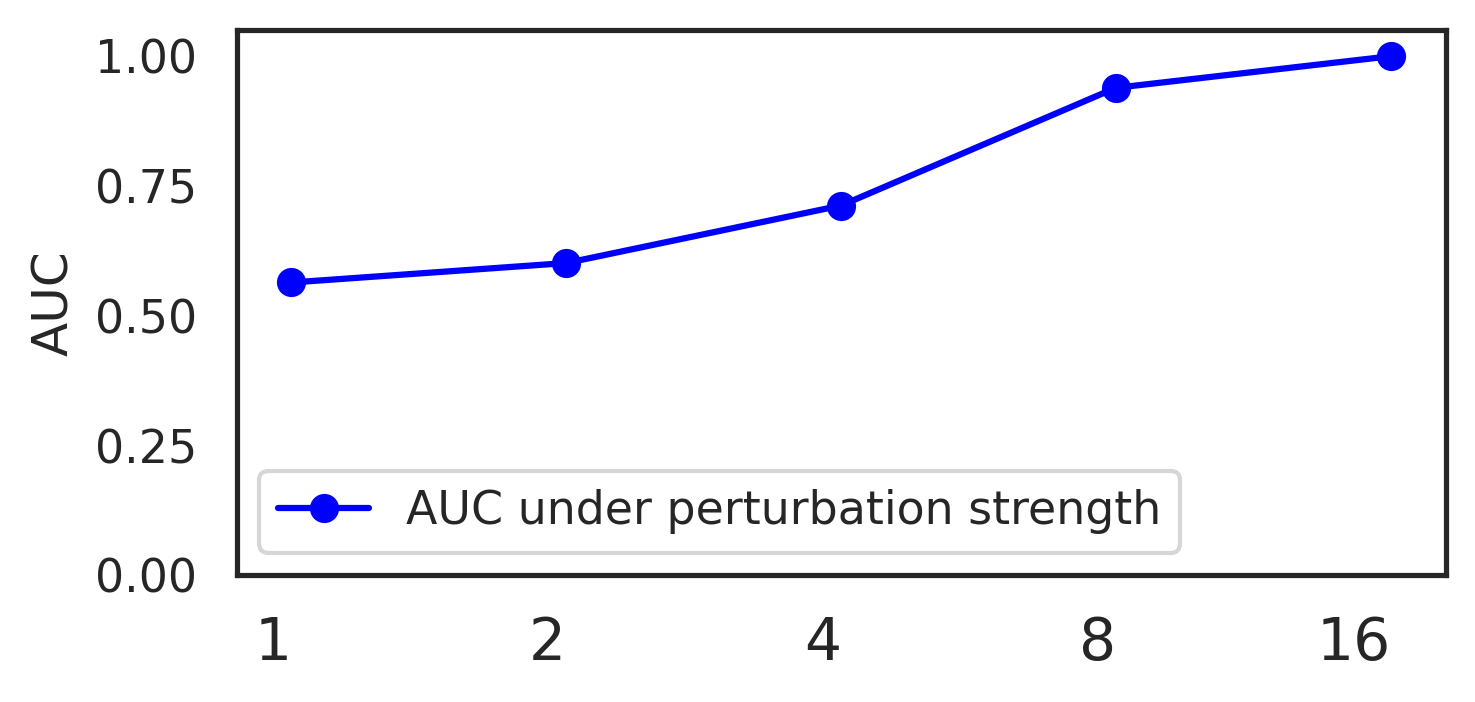}
        \caption{Consistency Score}
    \end{subfigure}
    \caption{Impact of perturbation strength for OD\_frcnn\_r101\_SEG\_gcnet\_r101}
\end{figure}

\begin{figure}[h!]
    \centering
    \begin{subfigure}{.475\linewidth}
        \centering
        \par\medskip
        \includegraphics[width=\linewidth]{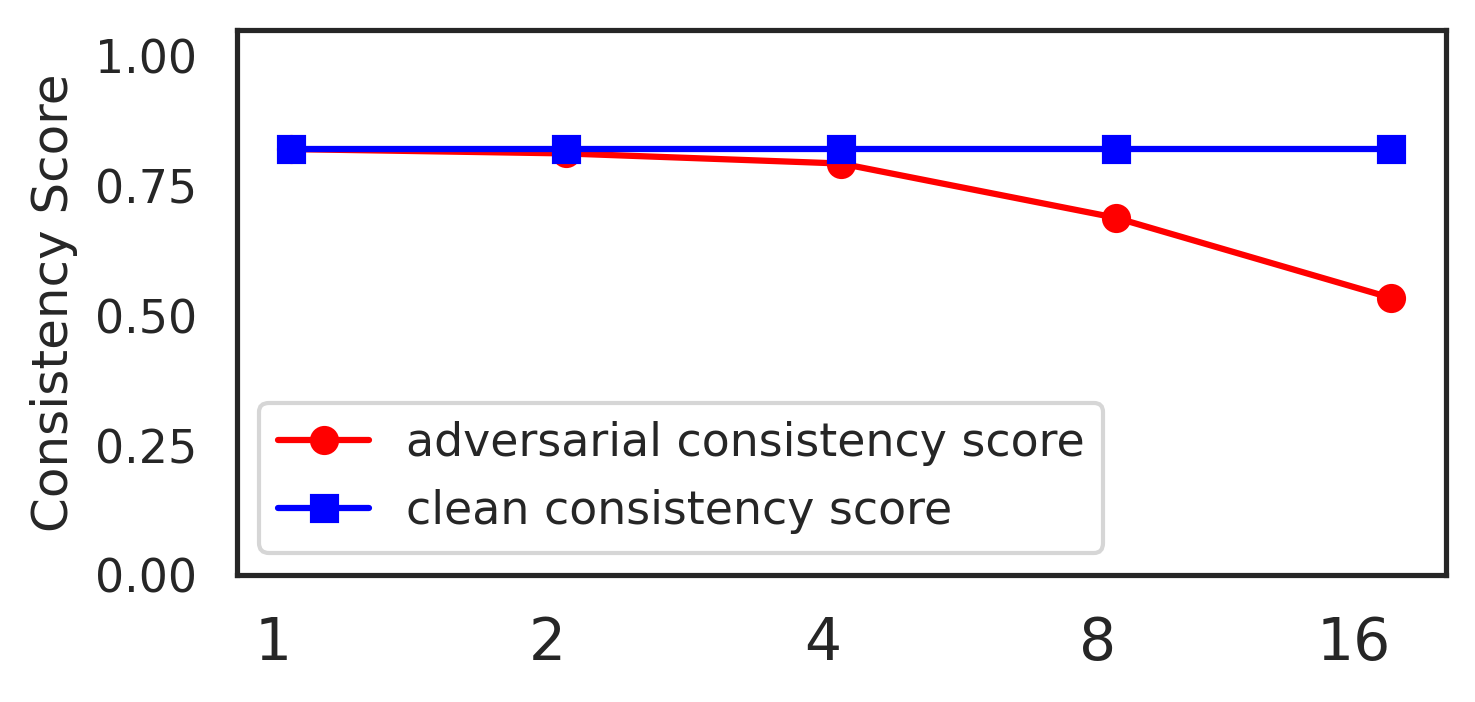}
        \caption{Consistency Score}
    \end{subfigure}
    \hfill
    \begin{subfigure}{.475\linewidth}
        \centering
        \par\medskip
        \includegraphics[width=\linewidth]{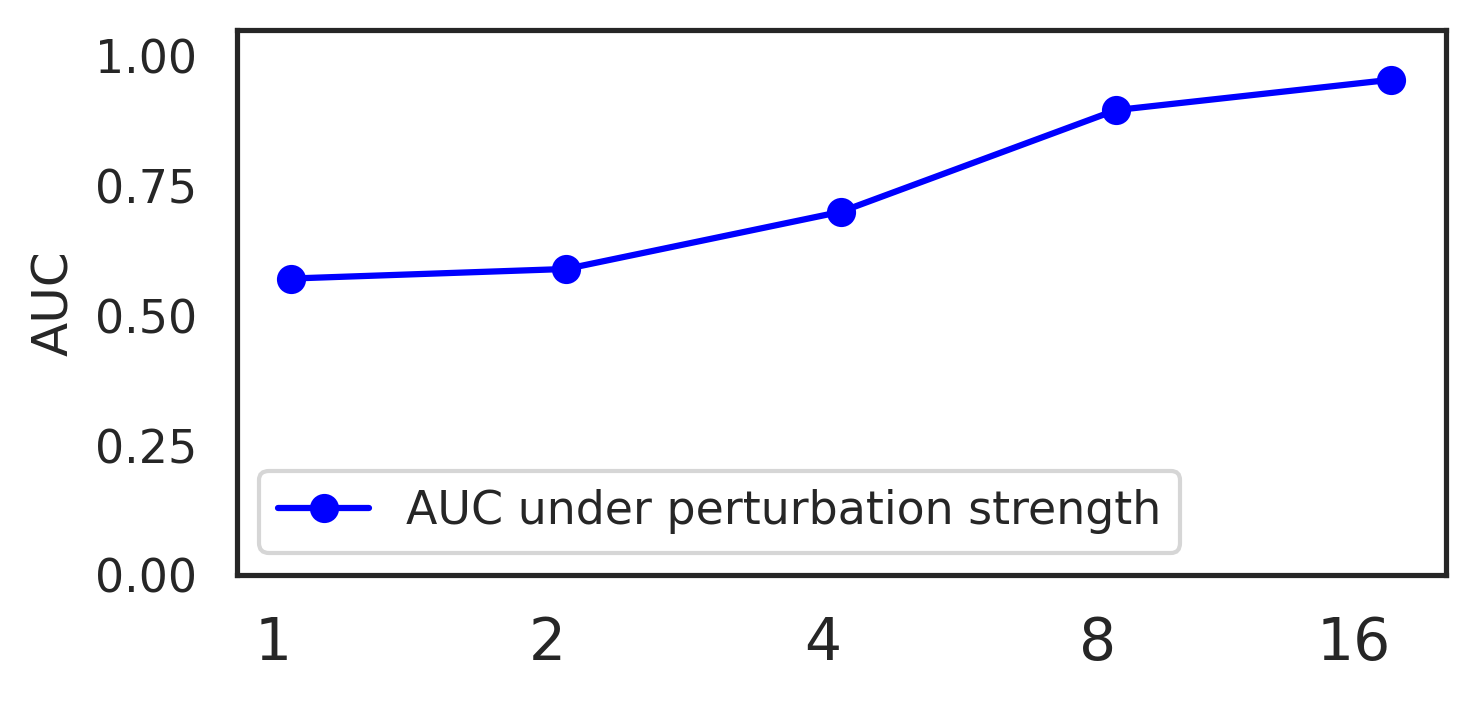}
        \caption{Consistency Score}
    \end{subfigure}
    \caption{Impact of perturbation strength for OD\_retinanet\_r50\_SEG\_mrcnn\_r50}
\end{figure}

\begin{figure}[h!]
    \centering
    \begin{subfigure}{.475\linewidth}
        \centering
        \par\medskip
        \includegraphics[width=\linewidth]{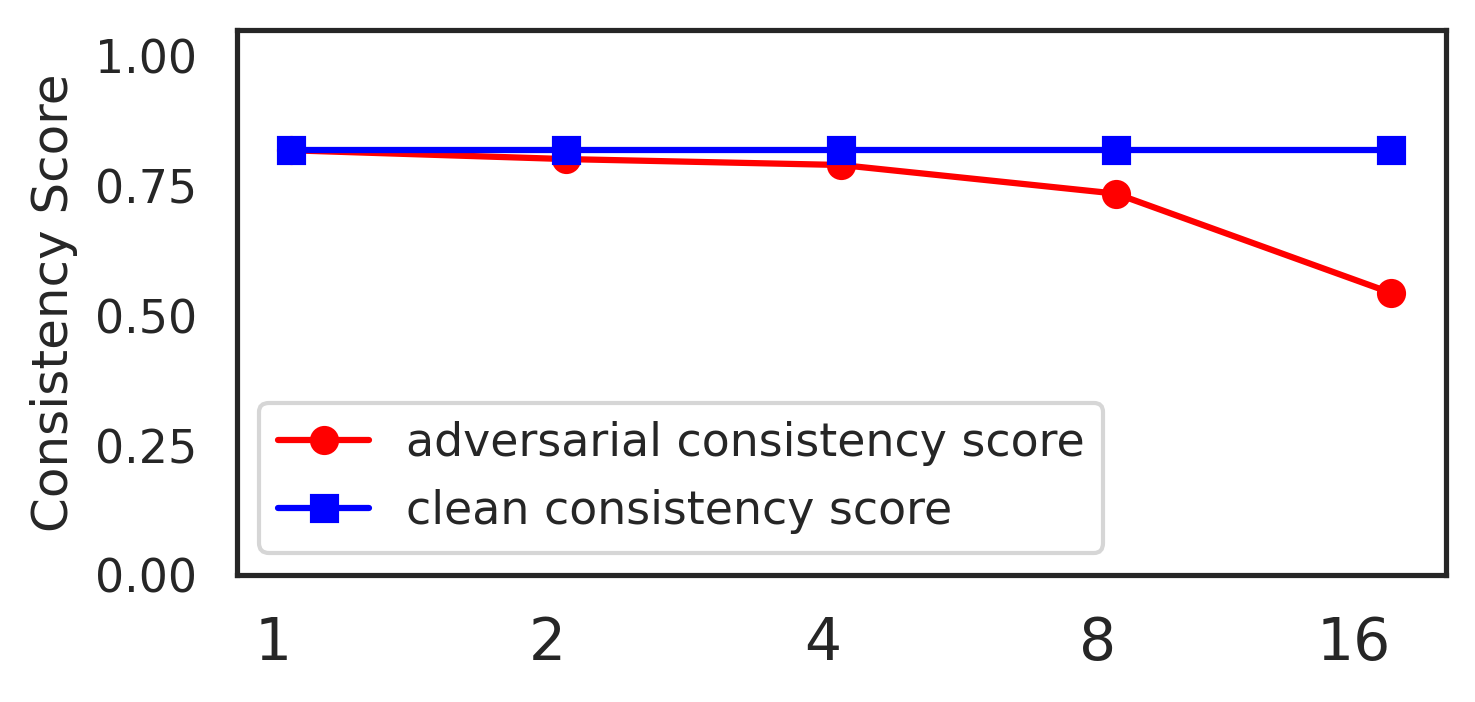}
        \caption{Consistency Score}
    \end{subfigure}
    \hfill
    \begin{subfigure}{.475\linewidth}
        \centering
        \par\medskip
        \includegraphics[width=\linewidth]{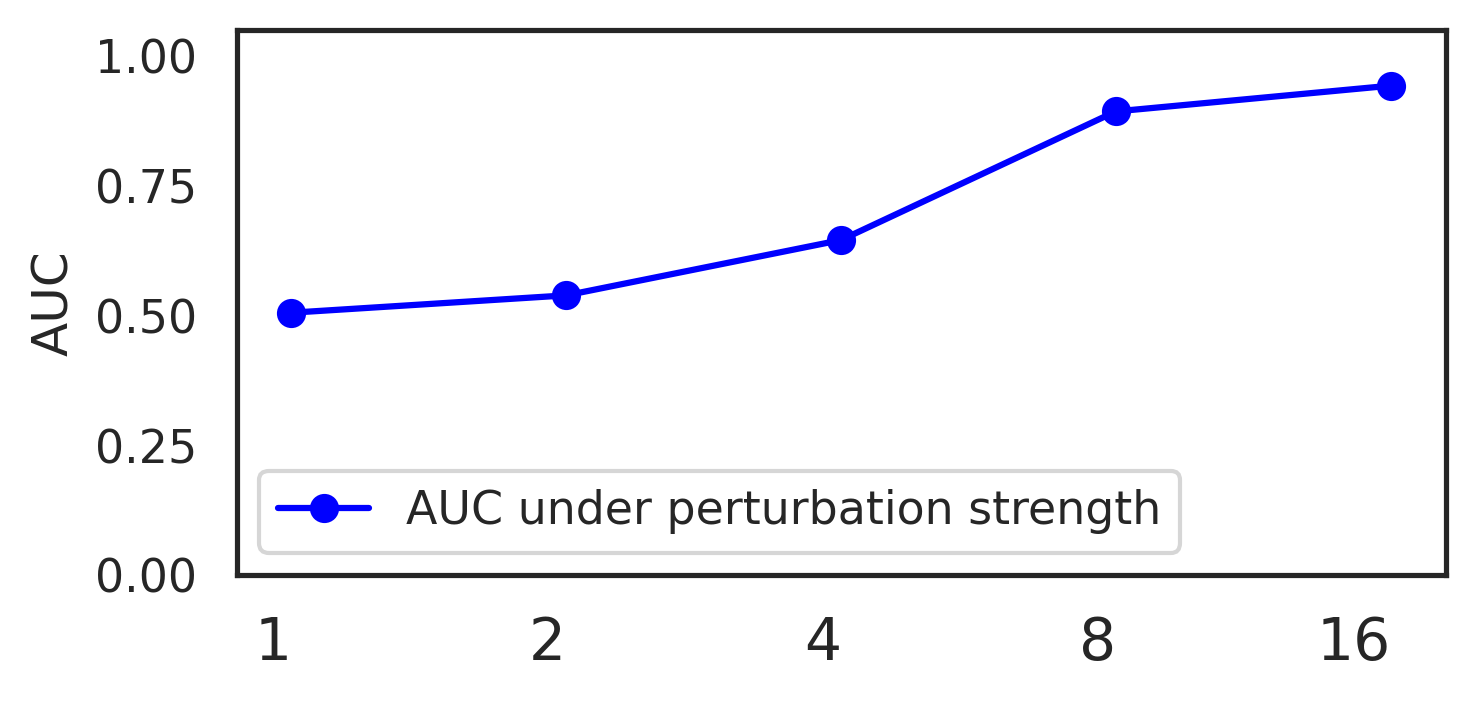}
        \caption{Consistency Score}
    \end{subfigure}
    \caption{Impact of perturbation strength for OD\_retinanet\_r50\_SEG\_gcnet\_r50}
\end{figure}

\begin{figure}[h!]
    \centering
    \begin{subfigure}{.475\linewidth}
        \centering
        \par\medskip
        \includegraphics[width=\linewidth]{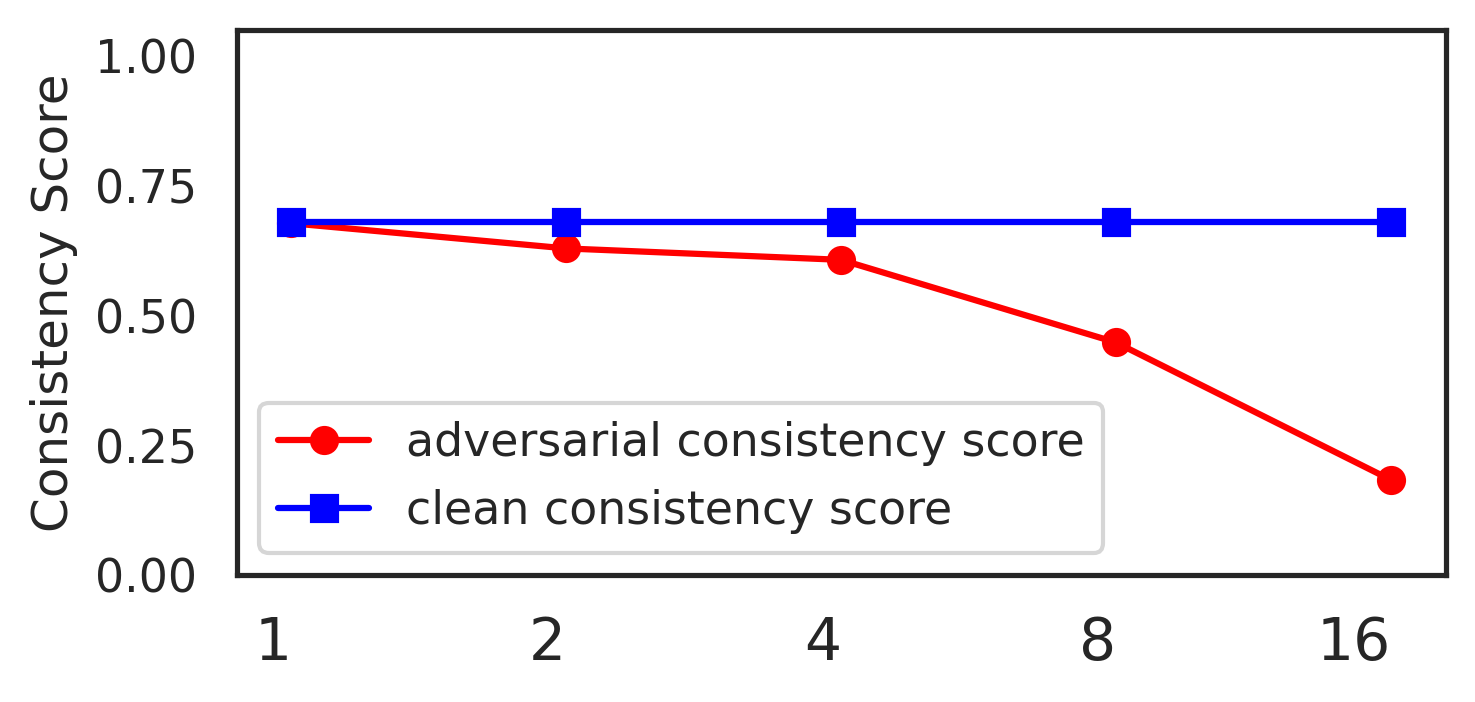}
        \caption{Consistency Score}
    \end{subfigure}
    \hfill
    \begin{subfigure}{.475\linewidth}
        \centering
        \par\medskip
        \includegraphics[width=\linewidth]{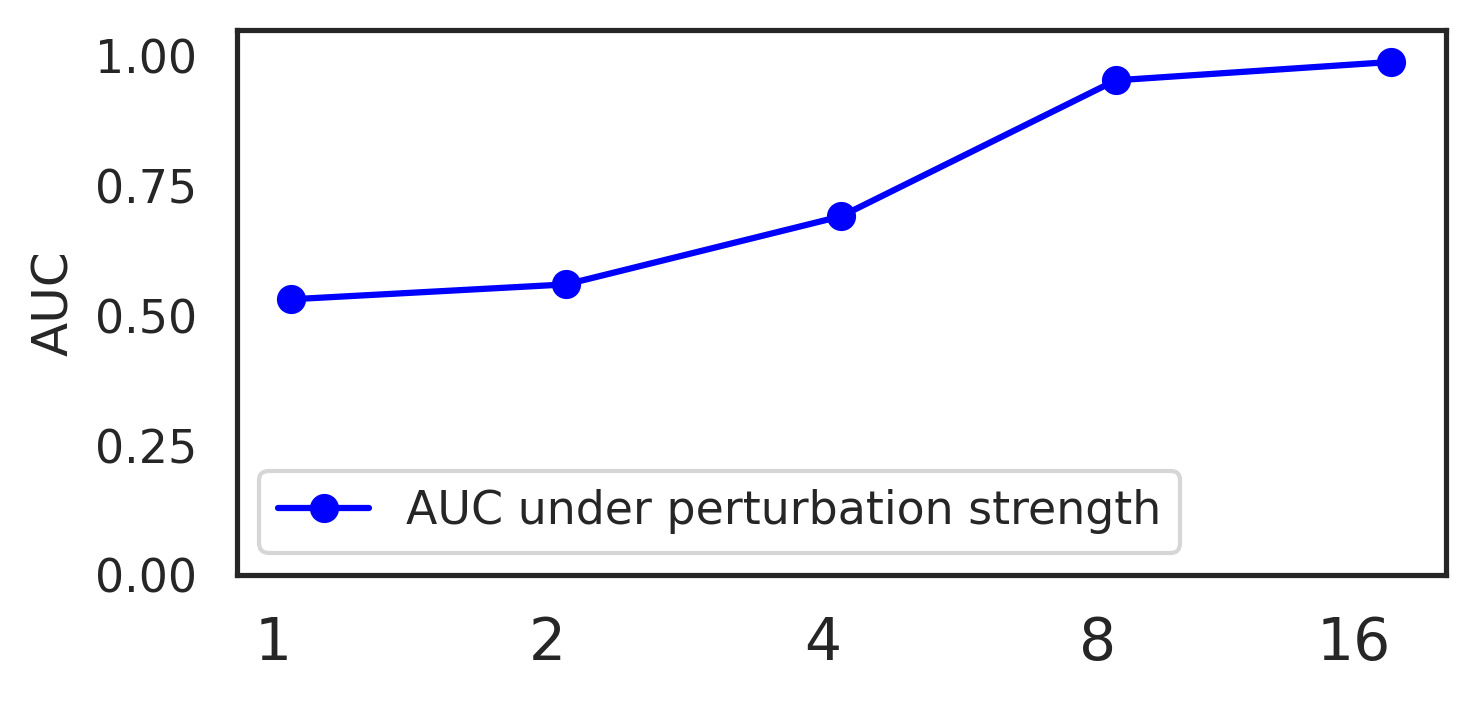}
        \caption{Consistency Score}
    \end{subfigure}
    \caption{Impact of perturbation strength for OD\_retinanet\_r50\_SEG\_mask2former}
\end{figure}

\begin{figure}[h!]
    \centering
    \begin{subfigure}{.475\linewidth}
        \centering
        \par\medskip
        \includegraphics[width=\linewidth]{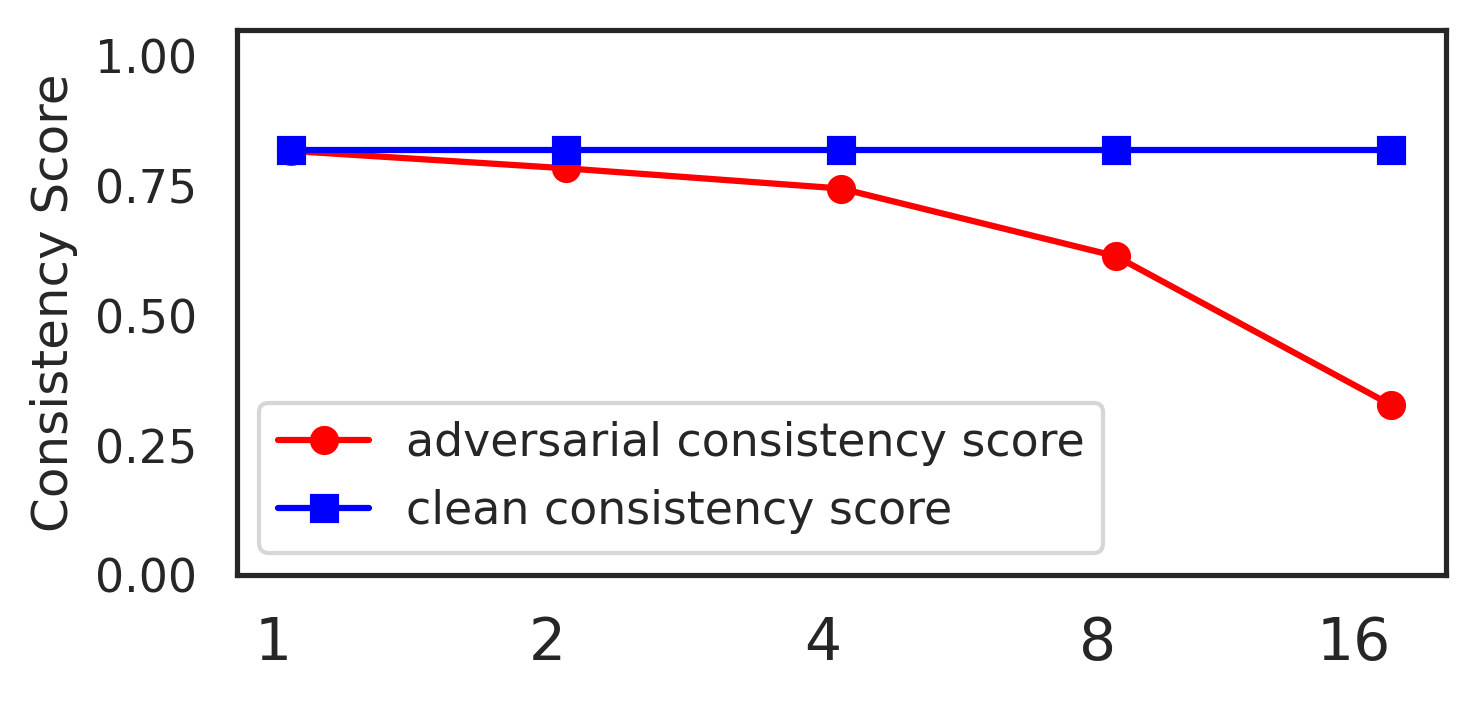}
        \caption{Consistency Score}
    \end{subfigure}
    \hfill
    \begin{subfigure}{.475\linewidth}
        \centering
        \par\medskip
        \includegraphics[width=\linewidth]{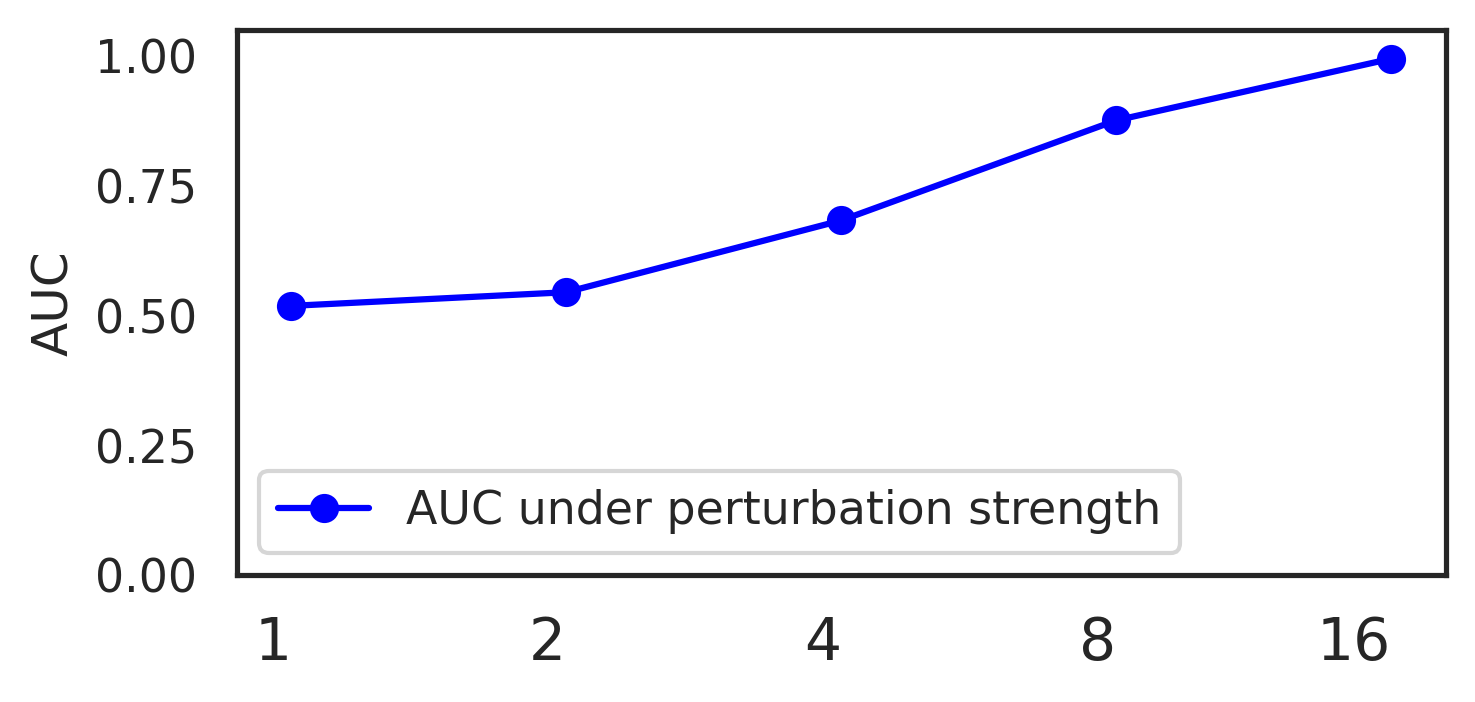}
        \caption{Consistency Score}
    \end{subfigure}
    \caption{Impact of perturbation strength for OD\_retinanet\_r50\_SEG\_mrcnn\_r101}
\end{figure}

\begin{figure}[h!]
    \centering
    \begin{subfigure}{.475\linewidth}
        \centering
        \par\medskip
        \includegraphics[width=\linewidth]{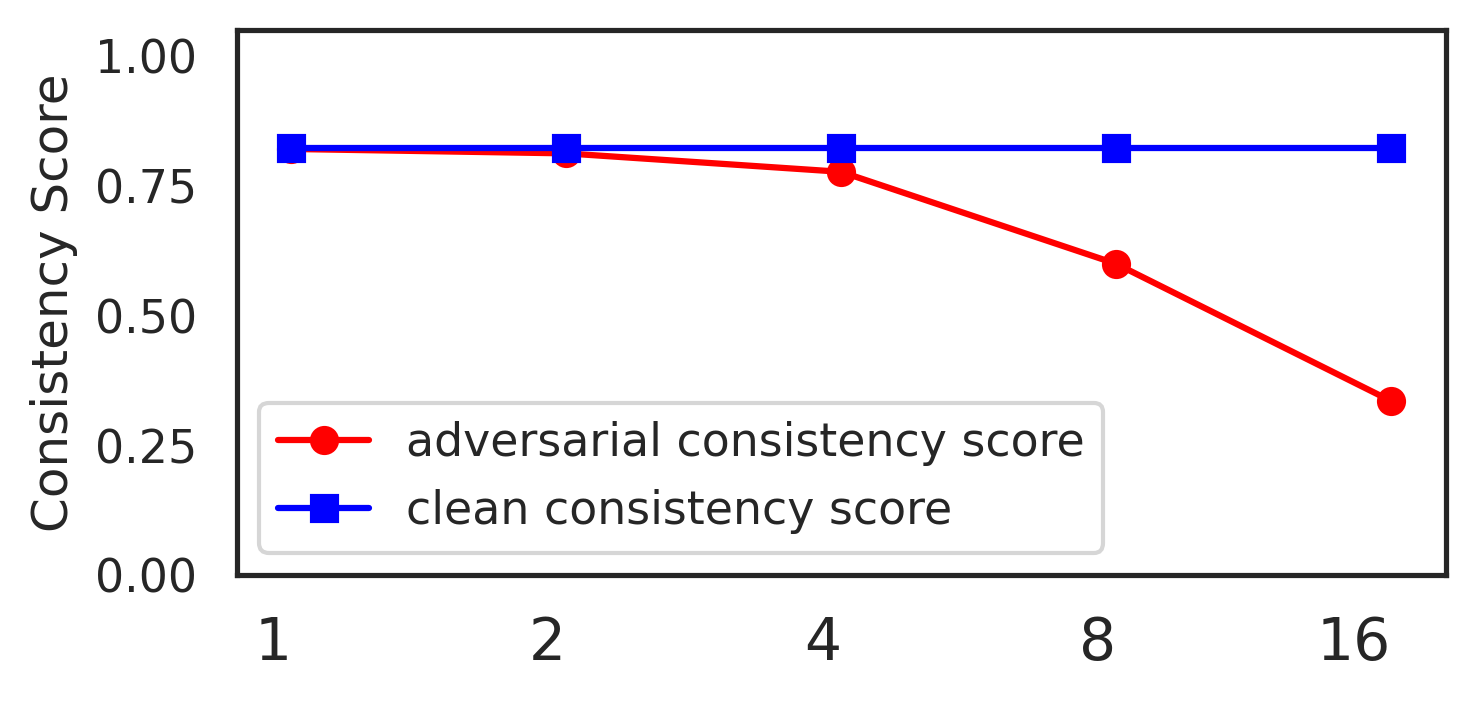}
        \caption{Consistency Score}
    \end{subfigure}
    \hfill
    \begin{subfigure}{.475\linewidth}
        \centering
        \par\medskip
        \includegraphics[width=\linewidth]{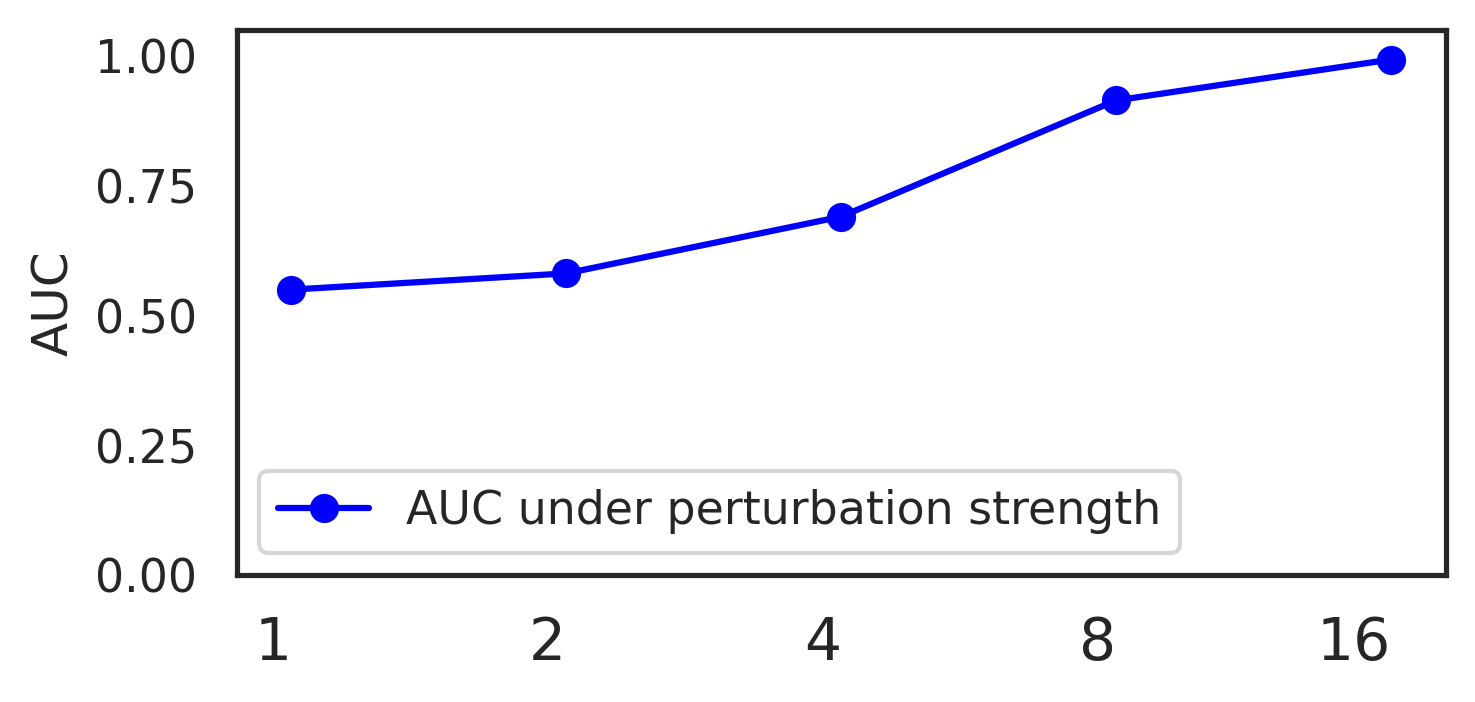}
        \caption{Consistency Score}
    \end{subfigure}
    \caption{Impact of perturbation strength for OD\_retinanet\_r50\_SEG\_gcnet\_r101}
\end{figure}

\begin{figure}[h!]
    \centering
    \begin{subfigure}{.475\linewidth}
        \centering
        \par\medskip
        \includegraphics[width=\linewidth]{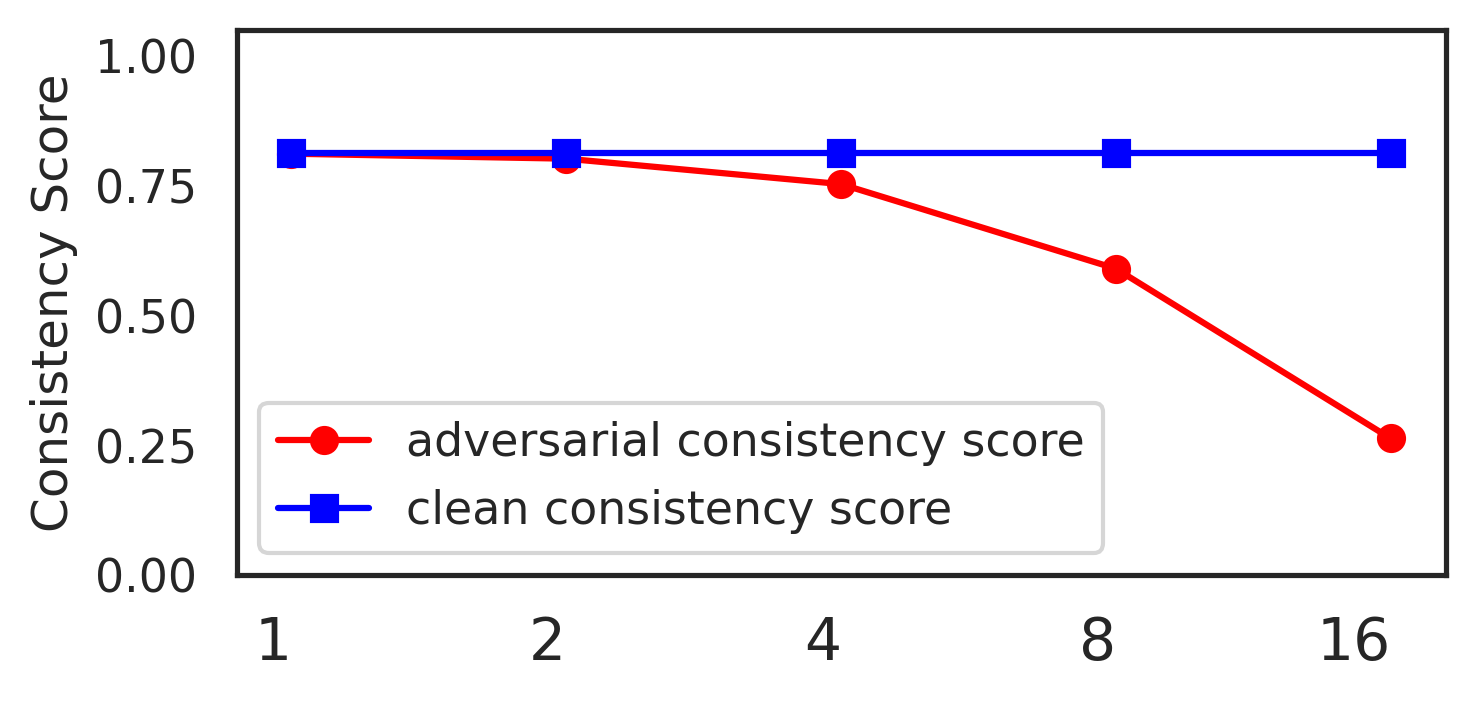}
        \caption{Consistency Score}
    \end{subfigure}
    \hfill
    \begin{subfigure}{.475\linewidth}
        \centering
        \par\medskip
        \includegraphics[width=\linewidth]{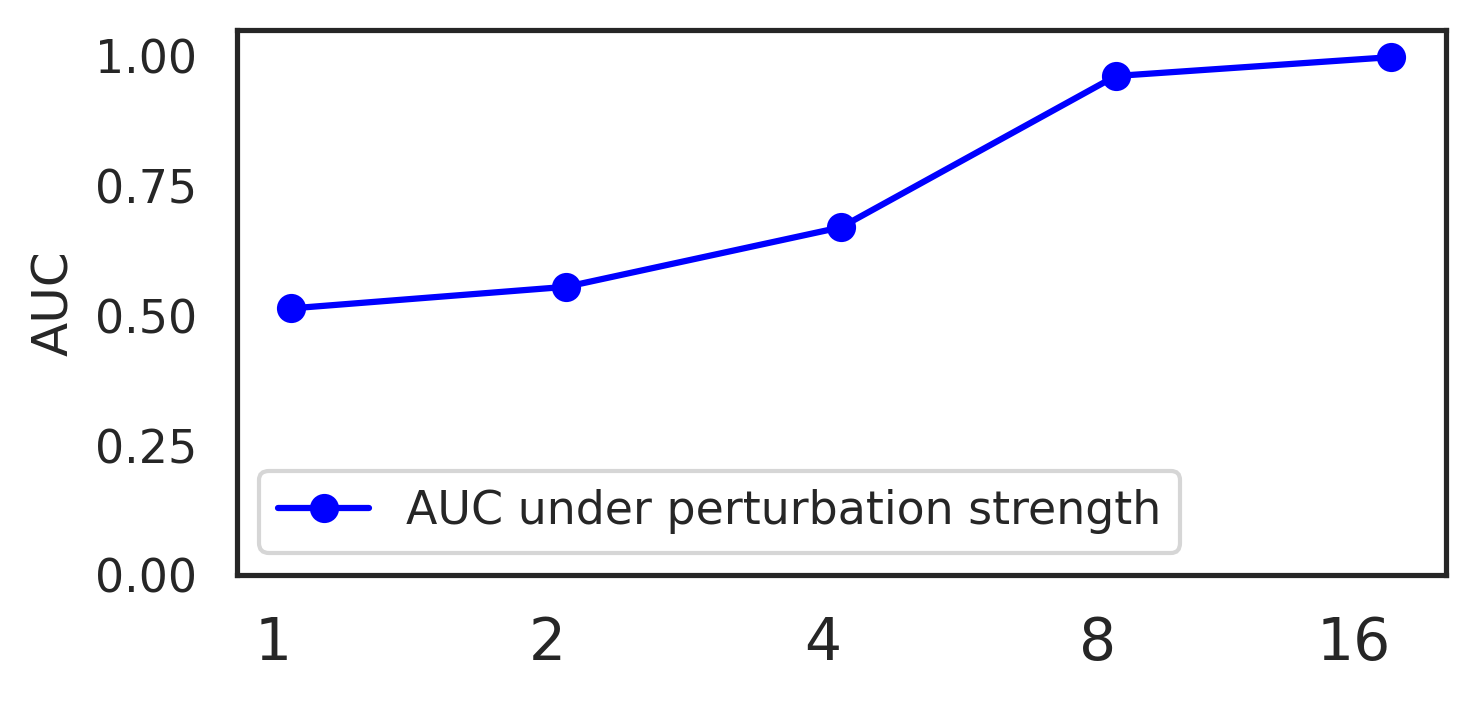}
        \caption{Consistency Score}
    \end{subfigure}
    \caption{Impact of perturbation strength for OD\_retinanet\_r101\_SEG\_mrcnn\_r50}
\end{figure}

\begin{figure}[h!]
    \centering
    \begin{subfigure}{.475\linewidth}
        \centering
        \par\medskip
        \includegraphics[width=\linewidth]{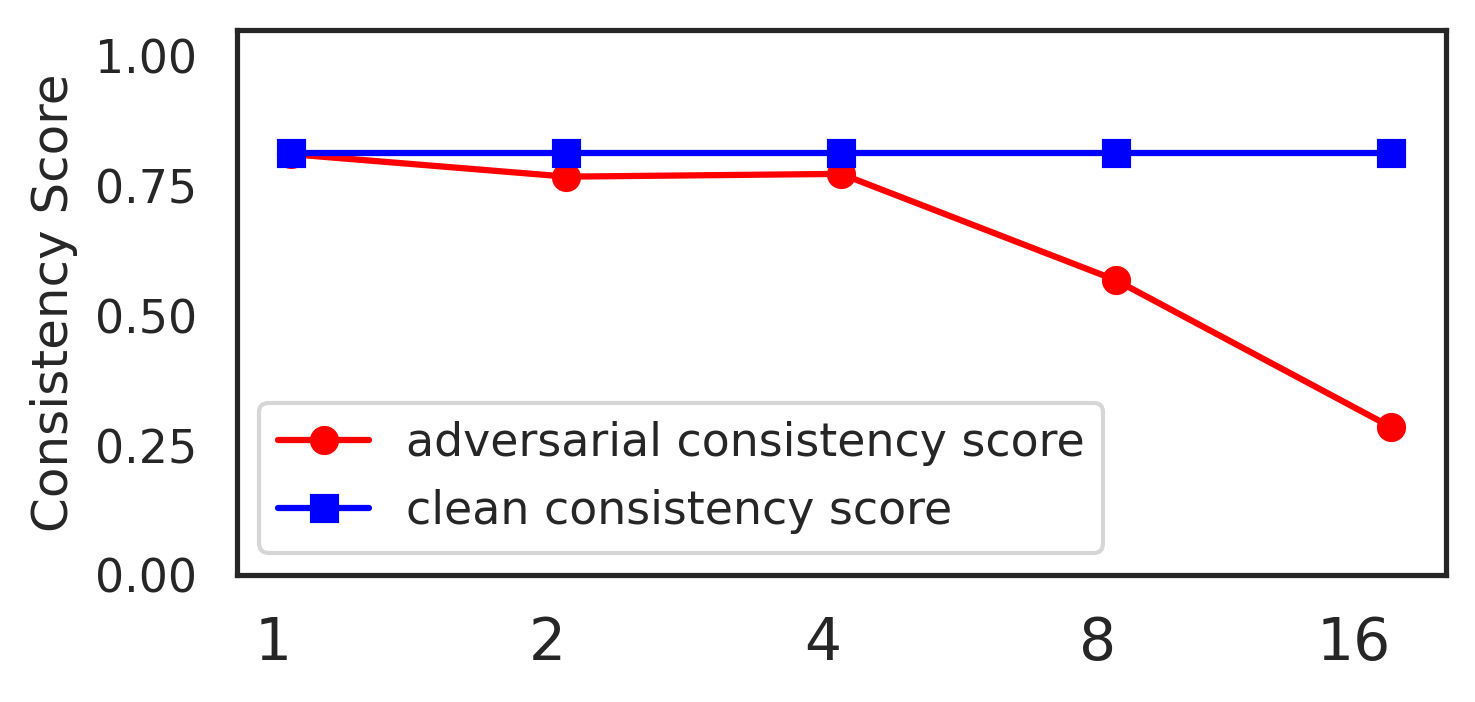}
        \caption{Consistency Score}
    \end{subfigure}
    \hfill
    \begin{subfigure}{.475\linewidth}
        \centering
        \par\medskip
        \includegraphics[width=\linewidth]{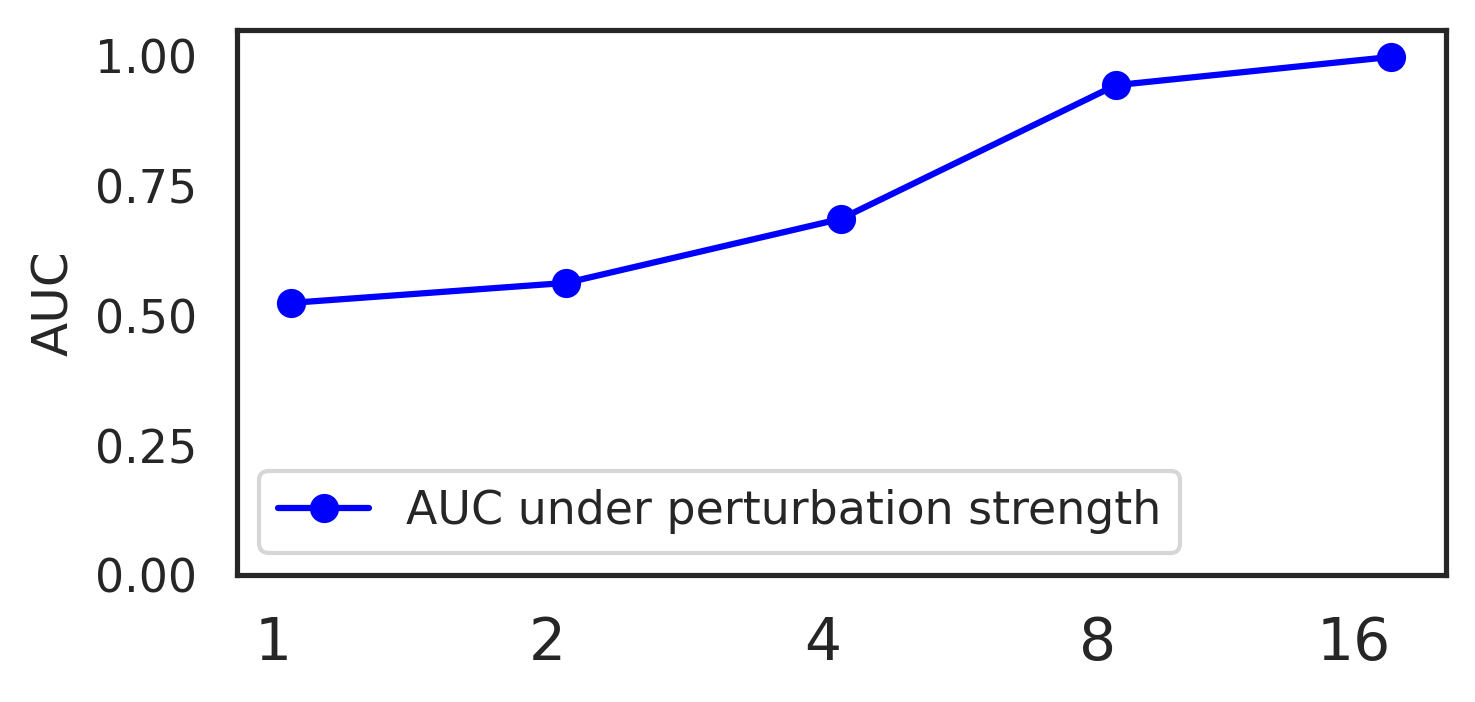}
        \caption{Consistency Score}
    \end{subfigure}
    \caption{Impact of perturbation strength for OD\_retinanet\_r101\_SEG\_gcnet\_r50}
\end{figure}

\begin{figure}[h!]
    \centering
    \begin{subfigure}{.475\linewidth}
        \centering
        \par\medskip
        \includegraphics[width=\linewidth]{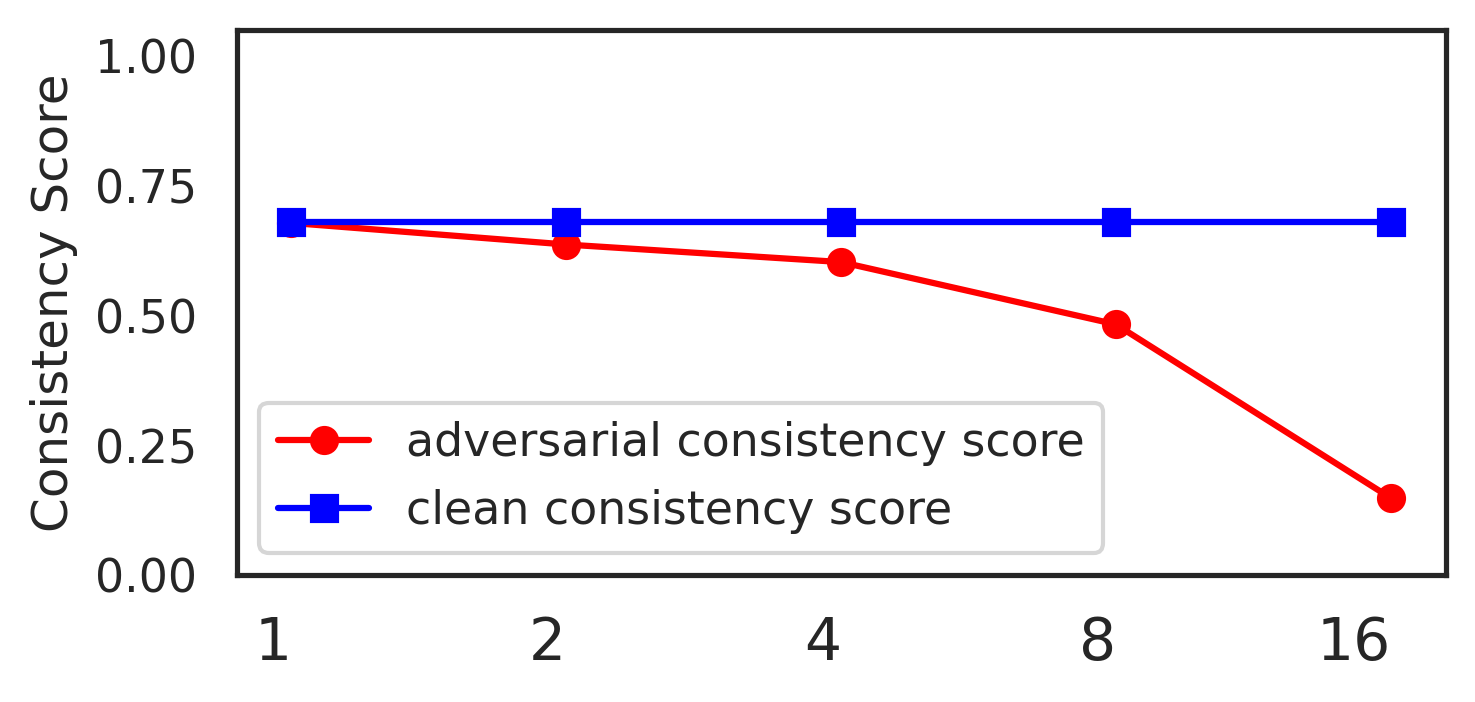}
        \caption{Consistency Score}
    \end{subfigure}
    \hfill
    \begin{subfigure}{.475\linewidth}
        \centering
        \par\medskip
        \includegraphics[width=\linewidth]{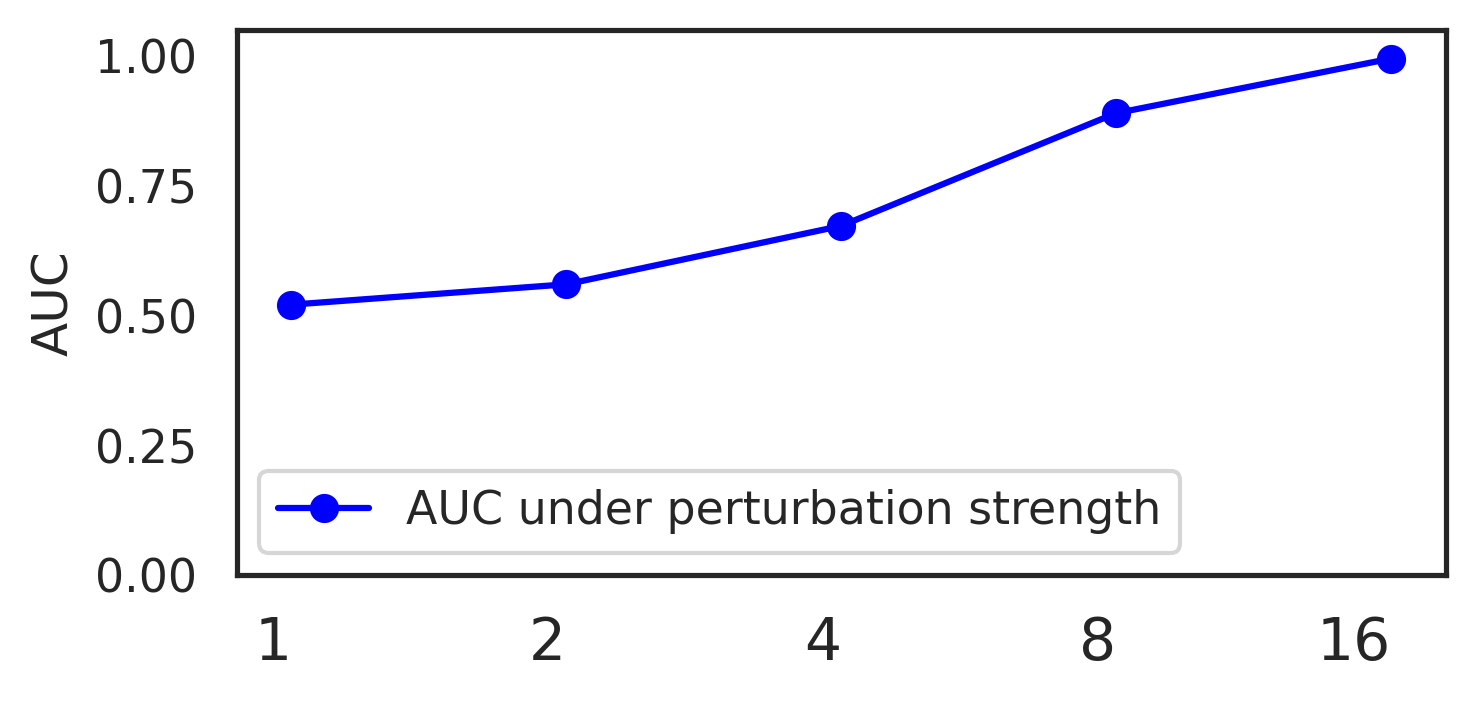}
        \caption{Consistency Score}
    \end{subfigure}
    \caption{Impact of perturbation strength for OD\_retinanet\_r101\_SEG\_mask2former}
\end{figure}



\begin{table}[b]
\centering
\caption{Effect of perturbation strength $\epsilon$ on mAP. Target model is FRCNN R50.}
\label{tab:pert_strength}
\small
\begin{tabular}{l|cccccc}
\hline
\multicolumn{1}{c|}{\textbf{Model}} & \multicolumn{6}{c}{\textbf{Perturbation Strength}} \\ 
 & clean & 1 & 2 & 4 & 8 & 16  \\ 
\hline
\textbf{FRCNN R50} & 30.2 & 28.4 & 25.4 & 18.7 & 4.2 & 0.18  \\ 
\hline
\textbf{MRCNN R50} & 19.8 & 18.9 & 17.0 & 13.4 & 6.2 & 1.5  \\ 
\hline
\end{tabular}
\end{table}

\end{document}